\documentclass[english,12pt]{article}
\include{macros}

\makeatletter
\newcommand\myCref[1]{\@ifundefined{r@#1}{\Cref*{sm:#1}}{\Cref{#1}}}
\newcommand\myref[1]{\@ifundefined{r@#1}{\ref*{sm:#1}}{\ref{#1}}}

\usepackage{setspace}
\makeatother

\begin{document}

\title{\Large The Nonstationarity-Complexity Tradeoff in Return Prediction\footnote{We are grateful to the seminar participants of the Tech 4 Finance conference, Fordham Quant 2026 Conference, Imperial College, and Voleon Research group. We acknowledge the constructive comments from Guillaume Coqueret, Marcin Kacperczyk, Serge Darolles, Blair Bilodeau, Yash Deshpande, Francesco Sangiorgi, Johannes Muhle-Karbe. Agostino Capponi's research was supported in part through grants of the Global Risk Institute and of Fi-Tek.
Chengpiao Huang and Kaizheng Wang's research is supported by National Science Foundation grants DMS-2210907 and DMS-2515679.}}
\blfootnote{Authors are listed in alphabetical order. Code is available at \url{https://github.com/ch3702/ATOMS}.}

\author{
Agostino Capponi\footnote{Department of IEOR and Columbia Business School, Columbia University. Email: \href{mailto:ac3827@columbia.edu}{ac3827@columbia.edu}.}
	\and Chengpiao Huang\footnote{Department of IEOR, Columbia University. Email: \href{mailto:chengpiao.huang@columbia.edu}{chengpiao.huang@columbia.edu}.}
	\and J.~Antonio Sidaoui\footnote{Department of IEOR, Columbia University. Email: \href{mailto:j.sidaoui@columbia.edu}{j.sidaoui@columbia.edu}.}
	\and Kaizheng Wang\footnote{Department of IEOR and Data Science Institute, Columbia University. Email: \href{mailto:kaizheng.wang@columbia.edu}{kaizheng.wang@columbia.edu}.}
	\and Jiacheng Zou\footnote{Email: \href{mailto:jiachengzou@alumni.stanford.edu}{jiachengzou@alumni.stanford.edu}.}
}

\date{This version: \today}

\maketitle

\begin{abstract}
Does more data improve return prediction? In non-stationary financial markets, longer training windows improve prediction of complex models but incorporate outdated economic regimes, whereas simpler models require less data and are less vulnerable to changes in economic conditions. We formally characterize this nonstationarity-complexity tradeoff, showing that model complexity and training window length must be jointly optimized. We propose an adaptive selection procedure with formal performance guarantees. Over three decades of U.S. equity markets, our method improves out-of-sample $R^2$ on industry portfolios by 14\% relative to fixed-window and regime-switching benchmarks, with large gains during recessions.
\end{abstract}

\noindent{\bf Keywords:} Non-stationarity, Model complexity, Return prediction, Model selection, Machine learning, Adaptive window selection

\newpage

\section{Introduction}\label{sec-intro}

Does more data improve return predictions? The machine learning models now standard in this literature \citep{GKX20,FNW20,KXi23,KMZ24} are heavily parameterized, and estimating them well requires ample samples, which in turn requires long periods of observation whenever the sampling frequency is fixed.\footnote{For instance, if the firm characteristics are constructed from quarterly disclosures of accounting information \citep{FNW20}, researchers cannot easily choose to increase the observation frequency.} Yet old data reflect economic conditions that may no longer hold: risk premia vary over the business cycle \citep{Ham89}, policy regimes change, and the returns to known predictors decline after those predictors become widely known \citep{MPo16}. 

Classical bias-variance tradeoff centers on model complexity and sample size under stationary conditions, so additional observations reduce estimation variance at no cost. Non-stationarity complicates this logic. A complex model has low misspecification error but high estimation variance. Controlling that variance and the overall prediction error requires training data drawn from economic conditions similar to those prevailing at the prediction date. Hence, a longer training window of observations help only if they come from the same regime rather than an outdated one. Otherwise, a simple model estimated on a short, recent window can outperform a complex model estimated on the full history when market conditions are most volatile. We refer to the resulting tension as the \emph{nonstationarity--complexity tradeoff}. Resolving this tension requires answering:
\begin{center}
\emph{How to choose model complexity and training window jointly under non-stationarity?}
\end{center}
Existing studies treat these as two separate decisions: a modeler first commits to a model class, then to a training window, typically an expanding window that uses all available history or a rolling window of fixed length. Empirically, we find that this separation could be costly when it matters economically. In doing so, we complement the machine learning in asset pricing literature, including the documented success of complex models and large factor zoos in return prediction \citep{Kelly2022Virtue,kelly2024large,DKK24}, by considering markets in which economic conditions shift over time in ways that cannot be known in advance.

To answer this, we first
derive a finite-sample bound that decomposes 
prediction error into model misspecification, 
statistical uncertainty, and a cost of 
non-stationarity, and use it to characterize 
the model class and training window that jointly minimize prediction error. We then develop an adaptive data-driven procedure that selects both without advance knowledge of when the economic environment will shift, and prove it performs nearly as well as the best choice one could have made in hindsight. Applied to 17 U.S.\ industry portfolios from 1990--2016, the procedure raises out-of-sample $R^2$ by 14\% and the mean-variance monthly Sharpe ratio from 0.30 to 0.33 relative to the best fixed-window alternative, with the largest gains concentrated in the three NBER recessions in our sample.

Our paper makes methodological and empirical contributions. We prove a finite-sample bound on prediction error that separates three sources of error: the misspecification of the model class, the statistical uncertainty from estimating the model on a finite training sample, and a cost of non-stationarity that reflects how far the economic conditions spanned by the training window have drifted from those prevailing at the time of prediction. The bound makes precise how model complexity and training-window length jointly affect the prediction error: a richer model class lowers misspecification error but raises statistical uncertainty, which can only be reduced with a longer window, and a longer window in turn raises exposure to outdated regimes. Jointly, these forces pin down the model class and window length that minimize prediction error, and the optimal choice depends on how severe the prevailing non-stationarity is.

We develop an adaptive procedure that selects the model class and training window jointly, without requiring prior knowledge of when or how the economic environment will shift. The procedure compares candidate models pairwise, choosing a separate look-back window for each comparison to balance stale data against too little data, and combines these comparisons through a sequential elimination tournament that stays fast even as more candidate models are added. We prove that the resulting procedure selects a model class and training window whose performance is nearly as good as the best choice one could have made in hindsight, without knowing in advance how market conditions would evolve. The procedure requires no parametric model of the return-generating process and imposes only mild regularity conditions on how that process drifts over time, so it accommodates any set of candidate models, whatever their model class or training horizon.

We evaluate this procedure on daily returns of 17 U.S.\ industry portfolios from Kenneth French's data library, with models estimated on daily data and predictive accuracy measured at the monthly frequency, over an out-of-sample period from January 1990 to November 2016. The procedure raises the average out-of-sample $R^2$, computed against a zero-forecast benchmark, to 0.049, a 14\% improvement over the strongest fixed-window alternative and more than double the performance of a benchmark validated only on recent data. In an implementable mean-variance strategy across the 17 portfolios, the same procedure raises the monthly Sharpe ratio from 0.30 for the best fixed-window alternative to 0.33 over the full sample, without higher turnover.

These gains concentrate exactly where the tradeoff predicts they should: in the three recessions identified by the National Bureau of Economic Research (NBER)\footnote{NBER Business Cycle Dating, \url{https://www.nber.org/research/business-cycle-dating}.} that fall within our sample, when non-stationarity is most severe. During the 1990 Gulf War recession, our procedure delivers the highest out-of-sample $R^2$ (0.027); the medium- and long-window benchmarks and the regime-switching model all turn negative, and even the shortest fixed window, restricted to only the most recent data, manages just 0.009. During the 2001 recession, which combined the dot-com repricing with the September 11 attacks, our procedure attains an $R^2$ of 0.125 --- more than five percentage points above cross-validation's 0.071, and narrowly ahead of the best long-window benchmark, which reaches 0.117. During the 2008 financial crisis, a longer-lived downturn that gave fixed-window methods more time to catch up, our procedure's advantage narrows but does not disappear: its $R^2$ of 0.041 is slightly above the best long-window benchmark's 0.039, and its mean-variance Sharpe ratio of 0.24 remains well above cross-validation's 0.05. This pattern holds across all 17 industries rather than concentrating in a few, and the resulting portfolio gains are not offset by higher trading costs.

The rest of the paper is organized as follows. \Cref{sec-setup} describes the problem setup. \Cref{sec-tradeoff} investigates the nonstationarity-complexity tradeoff. \Cref{sec-select} presents the adaptive model selection algorithm. \Cref{sec-experiments} illustrates our algorithm on real datasets. \Cref{sec-discussions} concludes the paper and discusses future directions. Mathematical proofs are deferred to the appendices. The code for implementing the algorithms in the paper is available at \url{https://github.com/ch3702/ATOMS}.

\subsection{Related Literature}

Our paper relates to two strands of research in asset pricing: one that documents the gains from model complexity in predicting stock returns, and one that studies how much historical data investors should use to estimate return models as economic conditions change.

The integration of machine learning into return prediction and asset pricing was initially driven by the ``multidimensional challenge'', that is, the need to identify which of the hundreds of proposed firm characteristics provide independent information for expected returns.

Early influential work by \cite{GKX20} demonstrated that nonlinear interactions missed by traditional regressions are a primary source of predictive gains, identifying trees and neural networks as superior methods. \cite{FNW20}  used adaptive group LASSO to show that only a small subset of characteristics provides incremental information when nonlinearities are properly accounted for. \cite{ChoiJiangZhang2025} applied machine learning to 32 international markets, concluding that market-specific neural networks achieve stronger results than global models by capturing local return-characteristic relationships. We refer to \cite{KXi23} for an excellent survey on financial machine learning.

Building on these empirical successes, a series of theoretical research have formalized the virtue of complexity, proving that out-of-sample forecast accuracy and portfolio Sharpe ratios can be strictly increasing in model complexity. This phenomenon occurs because high complexity induces ``implicit shrinkage'', which reduces prediction variance without the heavy bias costs associated with explicit shrinkage. This line of research advocates for the largest approximating model one can compute, because the gains from better approximation of the unknown truth dominate the statistical costs of heavy parameterization. Foundational works include \cite{Kelly2022Virtue, KellyMalamud2025Understanding, KMZ24}. They focus on time-series return prediction and market timing, resolving the ``double limit'' problem of growing parameters and observations to show that complexity captures unknown nonlinearities that improve  Sharpe ratios. A recent study by \cite{DKK24} extends these insights to the cross-section of returns, tackling a ``three infinities'' problem involving a simultaneously large number of assets, parameters, and observations. Their work proposes using random Fourier features to generate vast numbers of nonlinear factors, shifting the statistical objective from pure return prediction to minimizing pricing errors and constructing a high-complexity stochastic discount factor that reflects the true drivers of investors' marginal rates of substitution.

\cite{KellyMalamud2025Understanding} explicitly acknowledge the limitations of the large-$T$ asymptotic framework when applied to real-world financial data. In their discussion of finite-sample properties, they note that ``likely non-stationarities in financial markets data ... begs an agenda to understand how machine learning models can aid in the development of models that are robust to structural breaks''. Our work directly addresses this question by developing a framework that accounts for non-stationarity in the joint selection of model complexity and training windows. While \cite{KellyMalamud2025Understanding} establish that complexity can be virtuous under stationarity, we demonstrate that realizing this virtue in practice requires careful calibration to the local non-stationary environment, particularly during structural breaks such as recessions, in which, as we show below, simple models with a short training window can outperform complex models estimated over longer histories.

A separate literature studies how to detect changes in market conditions, such as structural breaks and regime changes \citep{BUr05}. Foundational works such as \cite{Cho60, And93, BPe98, Chi98} developed methods to identify these changes, which have been applied to financial time series including realized volatility \citep{LMa07} and speculative bubbles \citep{HBr12}. While these studies focus on identifying when breaks occur, our work addresses how to optimally use non-stationary data for estimation. Rather than pinpointing change points, we determine the optimal training window to minimize prediction error in the presence of non-stationarity.

A complementary literature examines optimal training window selection under non-stationarity but with fundamentally different assumptions than this paper. \cite{PTi07} showed that under structural breaks, optimal window selection should balance the bias from including pre-break data against the variance from using only post-break data. Subsequent work explored various selection criteria, including minimizing estimation loss functions \citep{PTi07, IJR17} and aggregating predictions across multiple windows \citep{PPi11}. However, these approaches typically assume linear models and specific non-stationary structures, such as single breaks or random walks. Our contribution differs along two dimensions. We take a model-free and distribution-free approach that imposes no parametric assumptions on either the prediction function or the non-stationary dynamics, allowing us to handle more general patterns; and we extend the bias-variance tradeoff identified by \cite{PTi07} to the modern machine learning context by jointly optimizing model complexity 
and training window length with formal finite sample error bound,
whereas prior work typically selects a window for a model that 
is fixed in advance.

\paragraph{Notation.}
We introduce the mathematical notation used throughout the paper. Let $\ZZ_+=\{1,2,...\}$ be the set of positive integers. For $n\in\ZZ_+$, define $[n]=\{1,2,...,n\}$. For $a,b\in\RR$, define $a \wedge b = \min \{ a, b \}$ and $a \vee b = \max \{ a, b \}$. For $x \in \RR$, let $x_+ = x \vee 0$. For non-negative sequences $\{a_n\}_{n=1}^{\infty}$ and $\{b_n\}_{n=1}^{\infty}$, we write $a_n=O(b_n)$ if there exists $C>0$ such that for all $n\in\ZZ_+$, $a_n\le Cb_n$. We write $a_n=\Theta(b_n)$ if $a_n=O(b_n)$ and $b_n=O(a_n)$. Unless otherwise stated, $a_n\lesssim b_n$ also represents $a_n=O(b_n)$, and $a_n\asymp b_n$ also represents $a_n=\Theta(b_n)$. For a finite set $S$, we use $|S|$ to denote its cardinality.

\section{Framework for Return Prediction under Non-Stationarity}\label{sec-setup}
We describe the return prediction problem common in asset pricing. At modeling time $t$, the goal is to estimate the conditional mean $\EE_t[y\mid\bx]$ of an asset return $y \in \RR$ given predictors $\bx \in \cX \subseteq \RR^d$. The model is trained on historical data from periods $\tau = 1, \ldots, t-1$, where for each past period $\tau$ we observe a batch $\batch_{\tau}$\footnote{We use a batch $\batch_{\tau}$ to refer to a general collection of observations at a time $\tau$. This could represent, for example, (i) multiple observations of the same asset from intraday intervals; (ii) a cross-sectional collection of returns from different assets. Our framework accommodates (i) and (ii) as well as any other sampling and observation scheme, so long as they satisfy the no-look-ahead-bias constraint of $\tau\le t-1$ at the given time $\tau$. With a slight abuse of notation, we also use $\batch_{\tau}$ to denote its own cardinality.} of observations $\dataset_{\tau} = \{(\bx_s, y_s)\}_{s \in \batch_{\tau}}$. The key challenge is non-stationarity: the joint distribution $\distP_{\tau}$ of $(\bx, y)$ varies over time, with $\distP_{\tau} \neq \distP_{\tau'}$ for $\tau \neq \tau'$ in a fully general sense.

The data points $\dataset_{\tau} = \{(\bx_s, y_s)\}_{s\in \batch_{\tau}}$ are independent and identically distributed (i.i.d.) following the period-$\tau$ distribution $\distP_{\tau}$, where $s\in B_{\tau}$ describes the period-$\tau$ batch of data. We use this notation to reflect that the distribution $\distP_{\tau}$  could change in unknown manners as its time index $\tau$ changes, but has the typical statistical similarity within a given period.  That is, throughout our paper, we assume that observations within each batch are i.i.d., whereas batches from different time periods are independent but may follow distinct distributions, hence allowing for non-stationarity. Formally,
\begin{assumption}[Independent data]\label{assumption-independence}
For each $\tau\le t-1$, the dataset $\dataset_{\tau}$ consists of independent and identically distributed samples. The datasets $\{\dataset_{\tau}\}_{\tau=1}^{\infty}$ are independent across time periods.
\end{assumption}

While returns and many other financial time series inherently exhibit temporal dependence, Assumption \ref{assumption-independence} is a standard simplification in the theoretical analysis of machine learning \citep{KMZ24, DKK24}.  Adopting this independence assumption allows us to isolate the effect of non-stationarity, without introducing additional technicalities from temporal dependence. More generally, our theoretical analysis extends to certain dependent settings. In particular, suppose there exists a latent state process $\{z_{\tau}\}_{\tau=1}^{\infty}$, which may be temporally dependent. In each period $\tau$, conditioned on $z_{\tau}$, the dataset $\dataset_{\tau}$ is drawn i.i.d.~from a distribution $Q_{\tau}(\cdot\mid z_{\tau})$, independent of the past states $\{z_{\tau'}\}_{\tau'\le\tau-1}$ and data $\{\dataset_{\tau'}\}_{\tau'\le\tau-1}$. In this case, the datasets $\{\dataset_{\tau}\}_{\tau=1}^{\infty}$ are independent conditioned on the latent states $\{z_{\tau}\}_{\tau=1}^{\infty}$, and our theoretical results should be interpreted as being conditioned on the realized latent states. Unconditionally, however, the datasets may remain dependent.

Our goal is to use the historical data $\{\dataset_{\tau}\}_{\tau=1}^{t-1}$ to construct a prediction model $f_t:\cX\to\RR$ that performs well on the current, unobserved distribution $\distP_t$.
The performance of a model $f:\cX\to\RR$ with respect to the data distribution $\distP_t$ is measured by the mean squared error (MSE):
\begin{equation}\label{eqn-MSE}
L_t(f) = \EE_{(\bx,y)\sim\distP_t} \left[ \left( f(\bx)-y \right)^2 \right].
\end{equation}
In line with the empirical finance literature, we also use the $R^2$ metric to evaluate the performance of a given modeling procedure or algorithm $\method$ that produces a prediction model $f_t^{\method}$ at each time $t$. The out-of-sample $R^2$ for the algorithm $\method$ over an evaluation period $[t_1,t_2]$ is computed as\footnote{We note that in the $R^2$ metric \eqref{eqn-R2-zero-cumulative}, the denominator is the sum of the squared responses $y_s^2$ \emph{without demeaning}. In other words, we are benchmarking against a forecast of zero rather than the historical mean as in the statistical $R^2$ metric. As noted by \cite{GKX20}, predicting future excess stock returns with historical averages can be problematic and is not assumption-free, because the historical mean is estimated with significant noise, often performing worse than a forecast of zero. }
\begin{equation}\label{eqn-R2-zero-cumulative}
R_{[t_1,t_2]}^2(\method) = 1 - \frac{\sum_{t=t_1}^{t_2}  \sum_{s\in \batch_t} \left(f_t^{\method}(x_{s}) - y_{s}\right)^2 }{\sum_{t=t_1}^{t_2} \sum_{s\in \batch_t} y_{s}^2}.
\end{equation}
For completeness, in the appendices we also present results using the statistical $R^2$ metric
\begin{equation}\label{eqn-R2-cumulative}
R_{[t_1,t_2],\texttt{s}}^2(\method) = 1 - \frac{\sum_{t=t_1}^{t_2} \sum_{s\in \batch_t} \left(f_t^{\method}(x_s) - y_s\right)^2}{\sum_{t=t_1}^{t_2} \sum_{s\in \batch_t} (y_{s}-\bar{y})^2},
\end{equation}
where $\bar{y}$ is the mean of the samples $\{y_{s}:s \in \batch_t ,\, t\in[t_1,t_2] \}$.

In a stationary environment where $\distP_{\tau} = \distP$ for all $\tau$, the standard approach for learning a model $f$ consists in choosing a model class $\modelclass$ (e.g., linear model, random forest) and then fitting a model $\widehat{f} \in \modelclass$ by minimizing the empirical loss over the training data. The choice of the model class $\modelclass$ involves a classic bias-variance trade-off. A simple class may exhibit high bias due to model misspecification, while a complex class may suffer from high estimation variance.

When the environment is non-stationary (that is, $\distP_{\tau}\neq\distP_{\tau'}$ for $\tau\neq\tau'$), the problem becomes significantly more complicated. One must now make two critical choices simultaneously: the model class $\modelclass$ and the amount of historical data used for training. Data from the distant past may no longer be representative of the current environment, and can be misleading for model training. This creates the core tension of our paper: complex models require more data to reduce estimation variance, but using more data may introduce stronger non-stationarity that increases bias. Thus, it is possible for simple models with less training data to outperform complex models trained on more data. Our goal is to develop an approach to jointly choose the model class and training window size.

\section{The Nonstationarity-Complexity Tradeoff}\label{sec-tradeoff}

In this section, we formally investigate the tension created by the joint choice of model class and training window. We begin by providing  empirical evidence for this tradeoff, and then formalize it through a prediction error bound as a new theoretical result of learning theory, which allows for a wide range of machine learning models and non-stationarity.

\subsection{Empirical Evidence}\label{sec-tradeoff-empirics}

We begin with an empirical illustration that highlights the challenges of jointly choosing a model class and a training window under non-stationarity. The task is to forecast the excess returns of $17$ industry portfolios from Kenneth French's data library using a set of covariates, with training data starting from September 1987 and ending in October 2016.\footnote{More details about the dataset are provided  in \myCref{sec-experiments}.} We highlight that our data spans several recessions documented in \href{https://www.nber.org/research/business-cycle-dating}{NBER Business Cycle Dating}: the 1990 Gulf War recession, the 2001 dot-com bubble bust and the 9/11 attack, and the 2008 Financial Crisis. To show that model performance is fundamentally linked to non-stationarity, we document a simple ranking of linear and nonlinear models in each period, across the different industries.

In each month $t$, for each industry, we fit three prediction models: (1) a linear model trained by ridge regression using the most recent $64$ months of data, (2) a random forest trained on the most recent $64$ months of data, and (3) a random forest trained on all historical data. More details of the experiments are given in \myCref{sec-tradeoff-empirics-details}. 
\begin{figure}[!t]
\centering
\caption{Number of Industries where Each Model Attains the Highest Annual Out-of-Sample $R^2$.} \label{fig-tradeoff-dominance}
\includegraphics[width=0.9\linewidth]{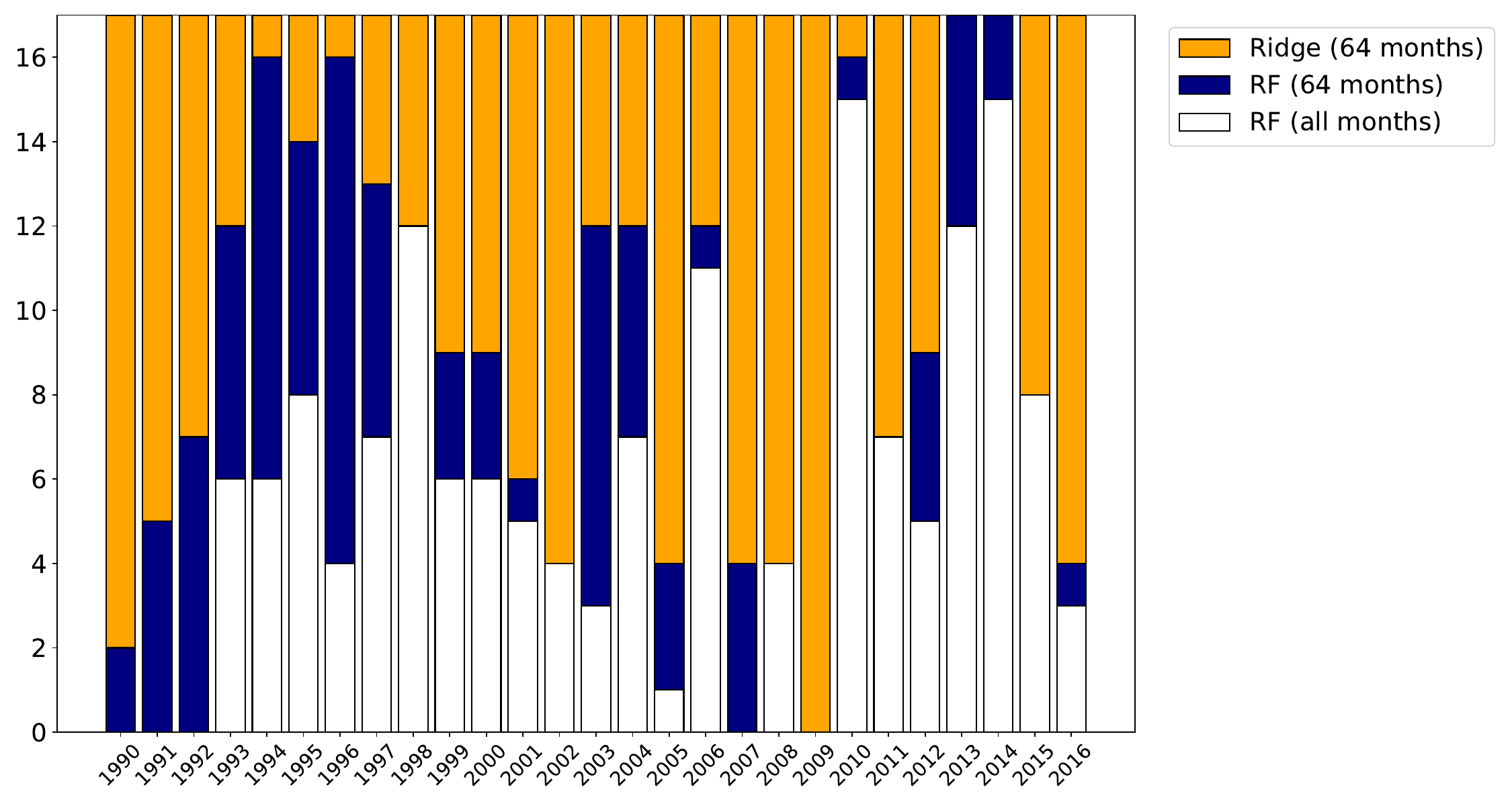}
\bnotefig{This figure reports the relative performance of three models in predicting the excess returns of the $17$ industry portfolios. The three models are: (1) a linear model trained by ridge regression on the most recent $64$ months of data (orange), (2) a random forest trained on the most recent $64$ months of data (blue), and (3) a random forest trained on all historical data (white). For each year from 1990 to 2016, we compute the models' annual out-of-sample $R^2$ in each of the $17$ industries, and then count, for each model, the number of industries in which it outperforms the other two models in terms of the annual out-of-sample $R^2$.}
\end{figure}

\begin{figure}[!t]
	\centering
	\caption{Annual Out-of-Sample $R^2$ of Random Forest and Ridge Models for $17$ Industry Portfolios.\label{fig-tradeoff-industry}}

    \begin{subfigure}{0.24\textwidth}
    	\centering
        \includegraphics[width=\linewidth]{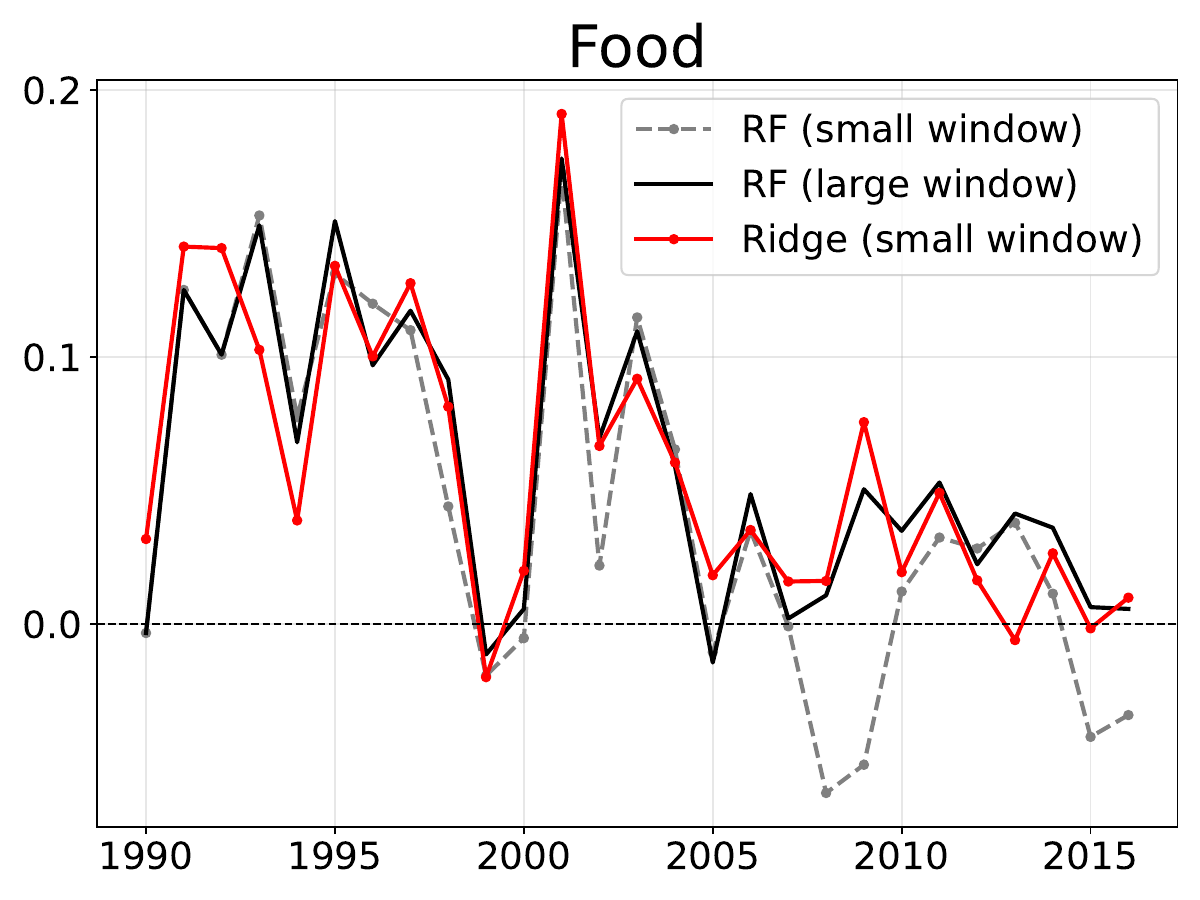}
	\end{subfigure}
    \begin{subfigure}{0.24\textwidth}
        \centering
        \includegraphics[width=\linewidth]{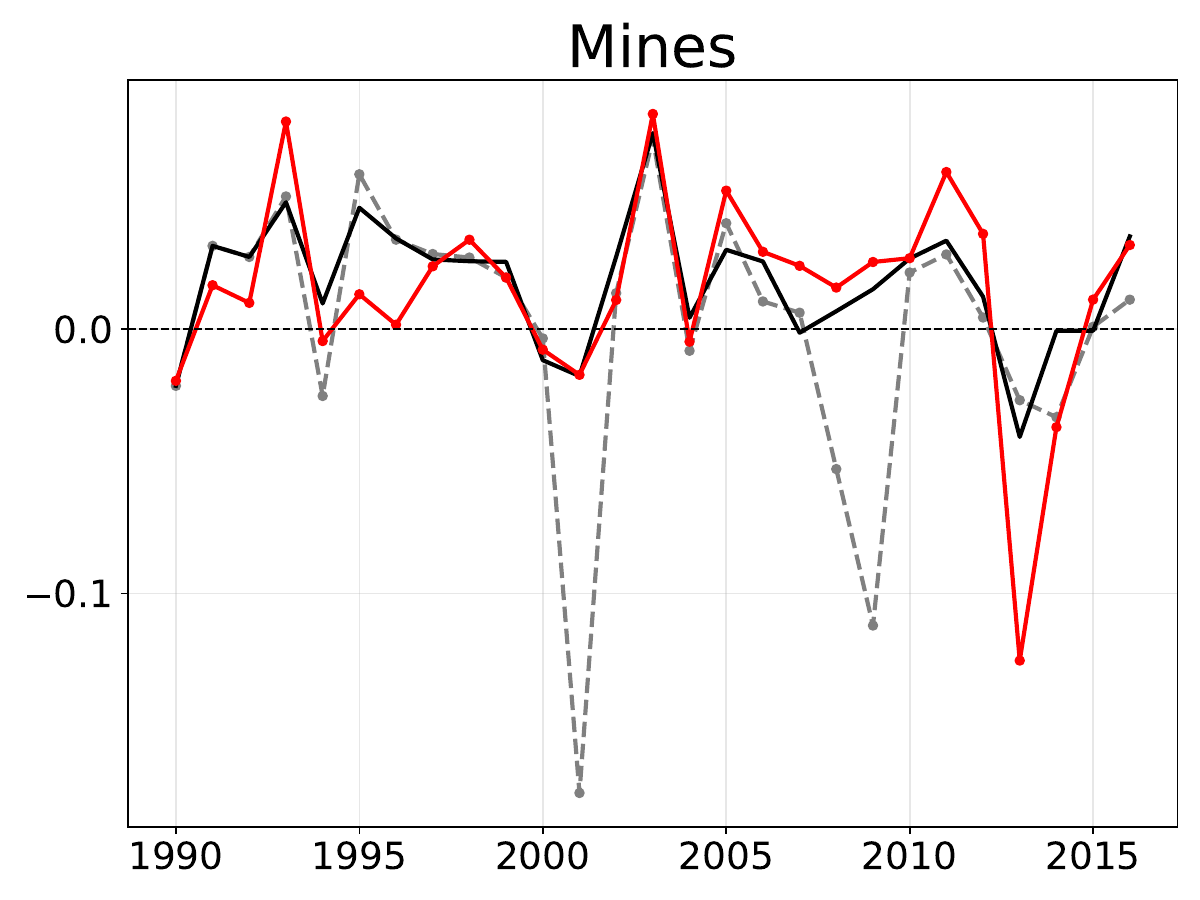}
	\end{subfigure}
    \begin{subfigure}{0.24\textwidth}
        \centering
        \includegraphics[width=\linewidth]{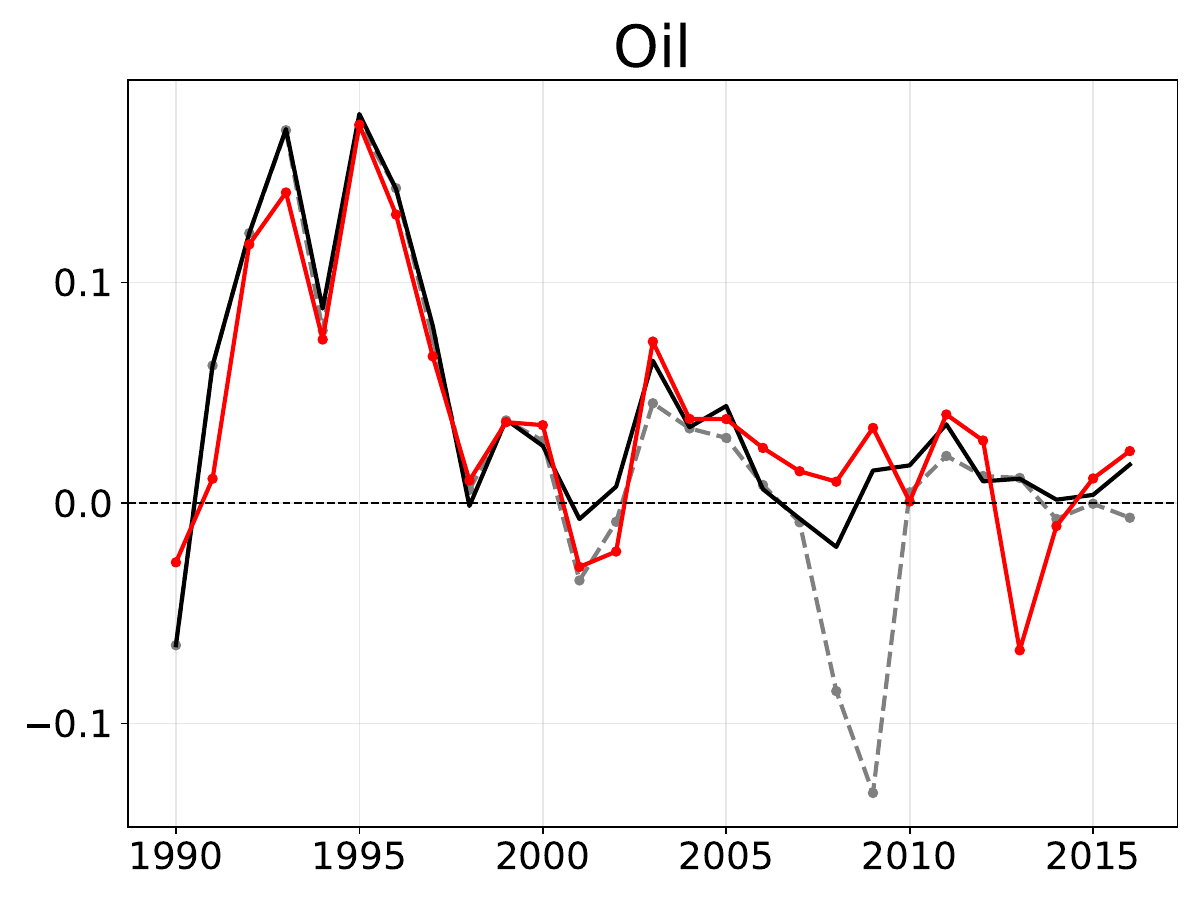}
	\end{subfigure}
    \begin{subfigure}{0.24\textwidth}
    	\centering
        \includegraphics[width=\linewidth]{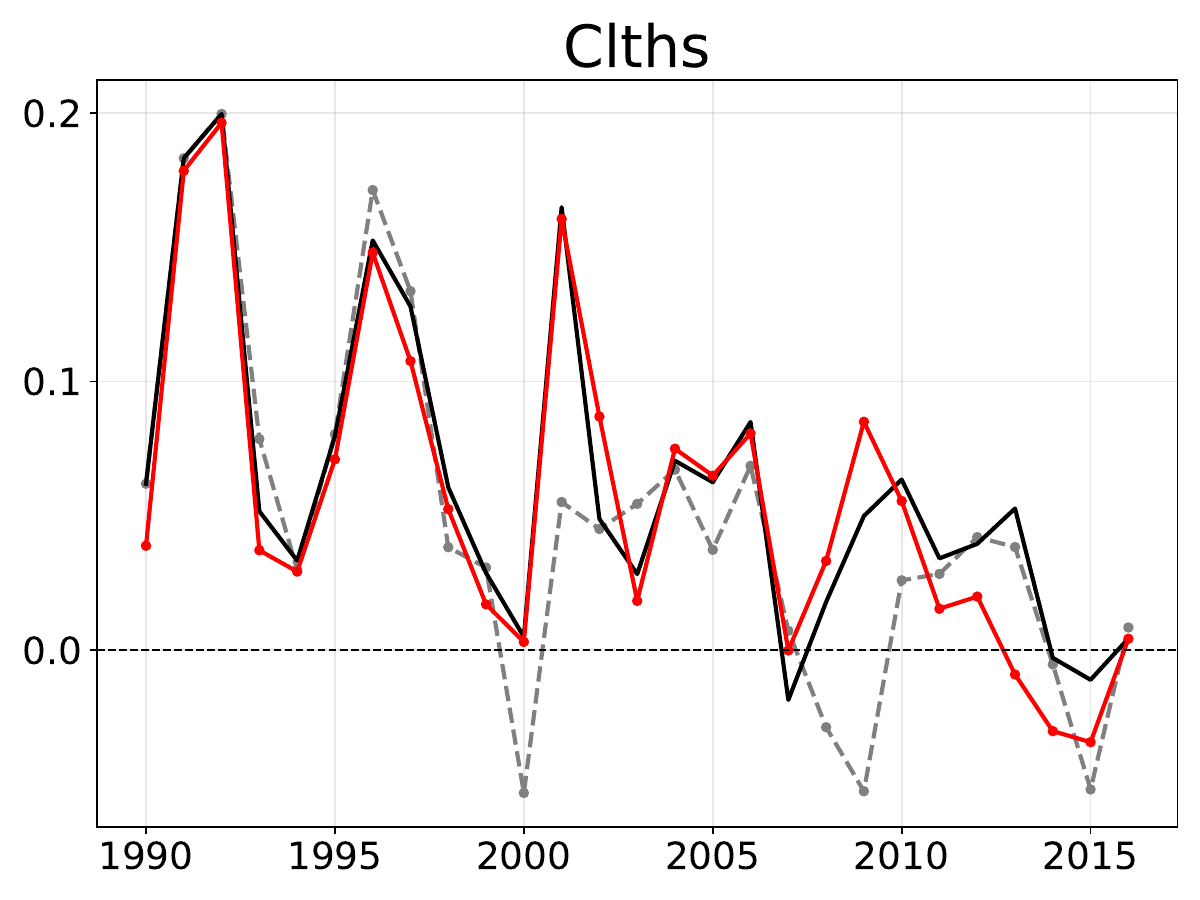}
	\end{subfigure}

    \begin{subfigure}{0.24\textwidth}
        \centering
        \includegraphics[width=\linewidth]{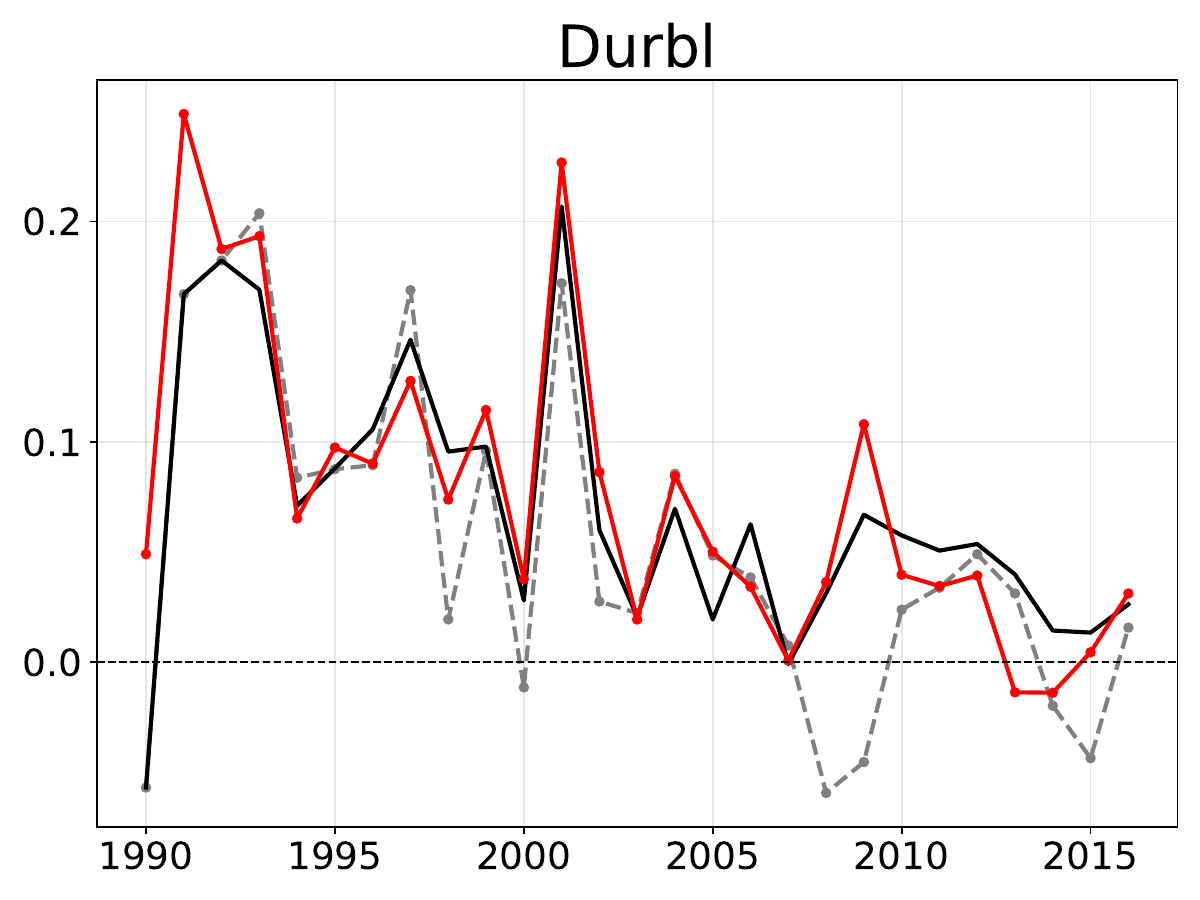}
	\end{subfigure}
    \begin{subfigure}{0.24\textwidth}
        \centering
        \includegraphics[width=\linewidth]{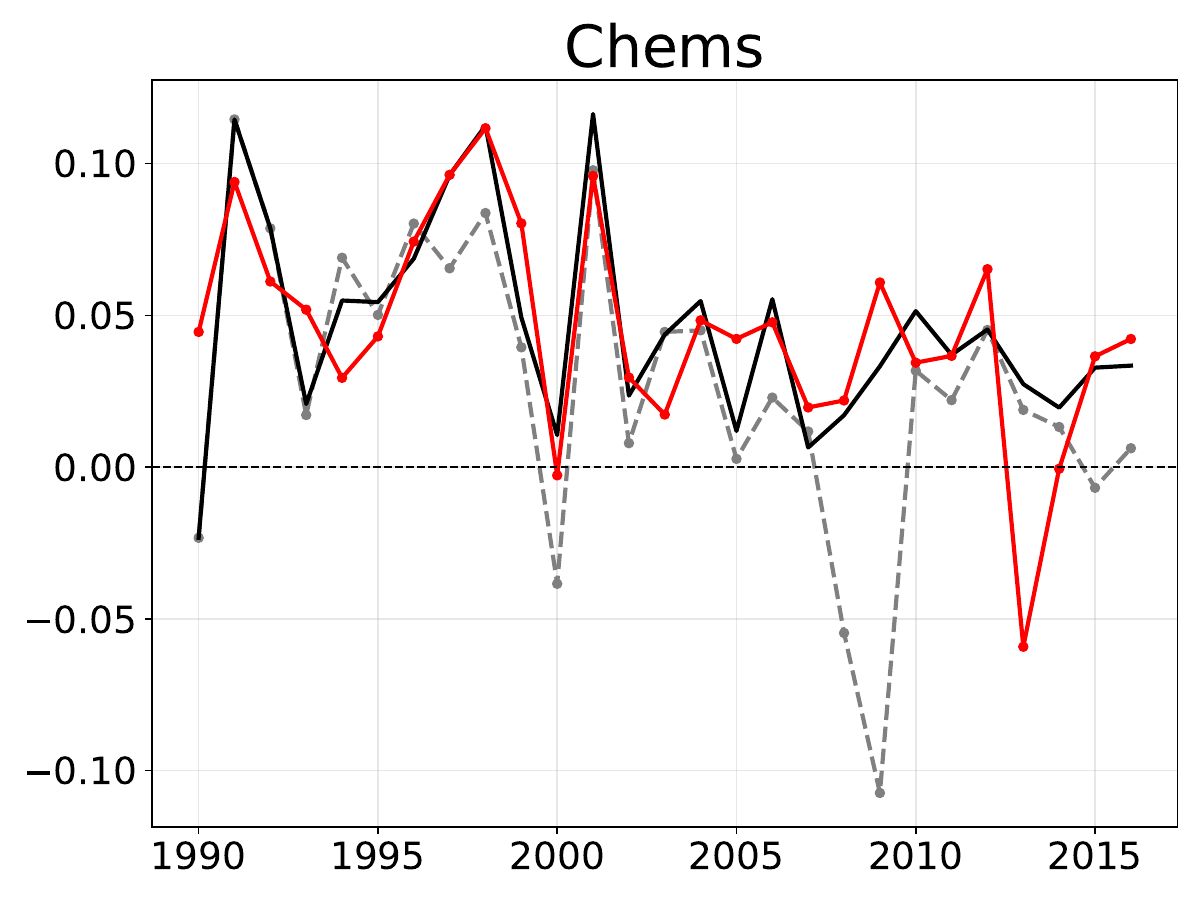}
	\end{subfigure}
    \begin{subfigure}{0.24\textwidth}
    	\centering
        \includegraphics[width=\linewidth]{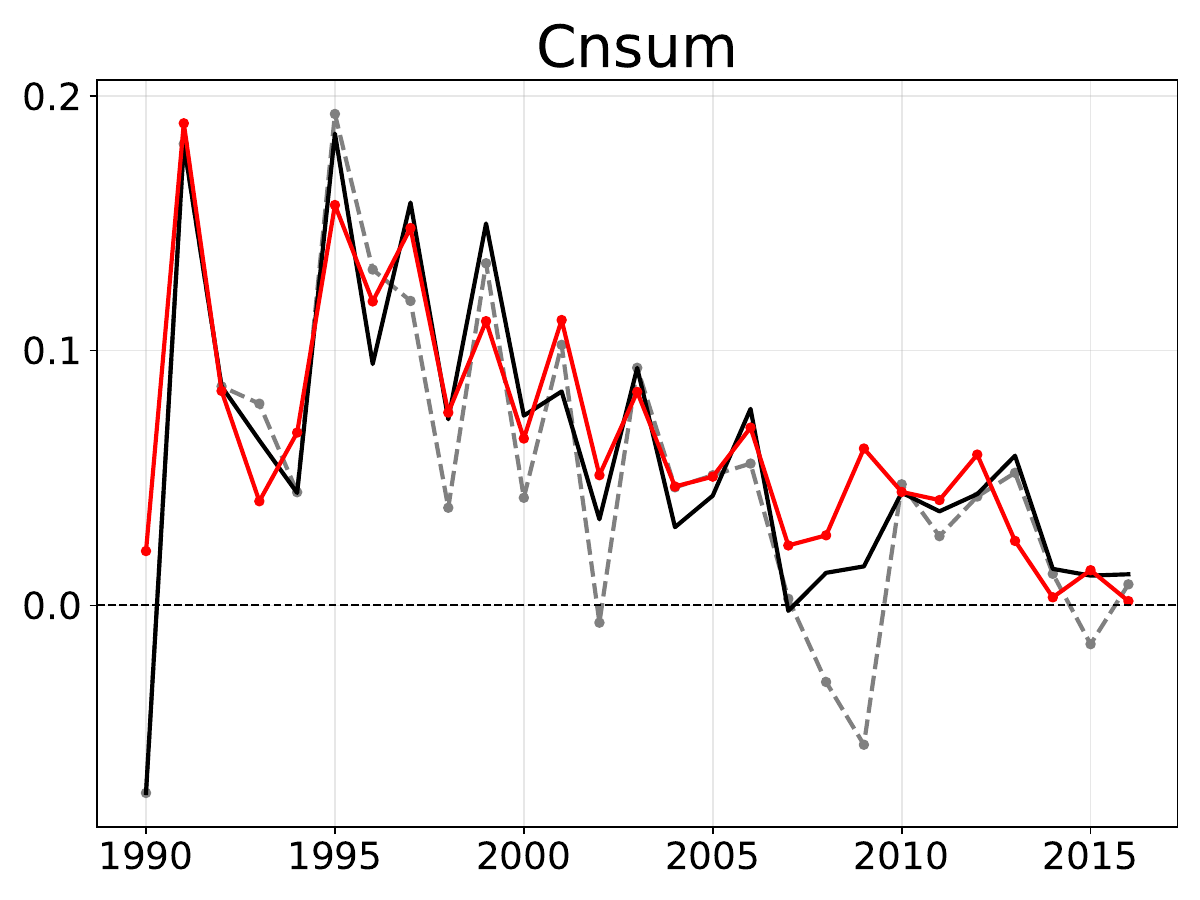}
	\end{subfigure}
    \begin{subfigure}{0.24\textwidth}
        \centering
        \includegraphics[width=\linewidth]{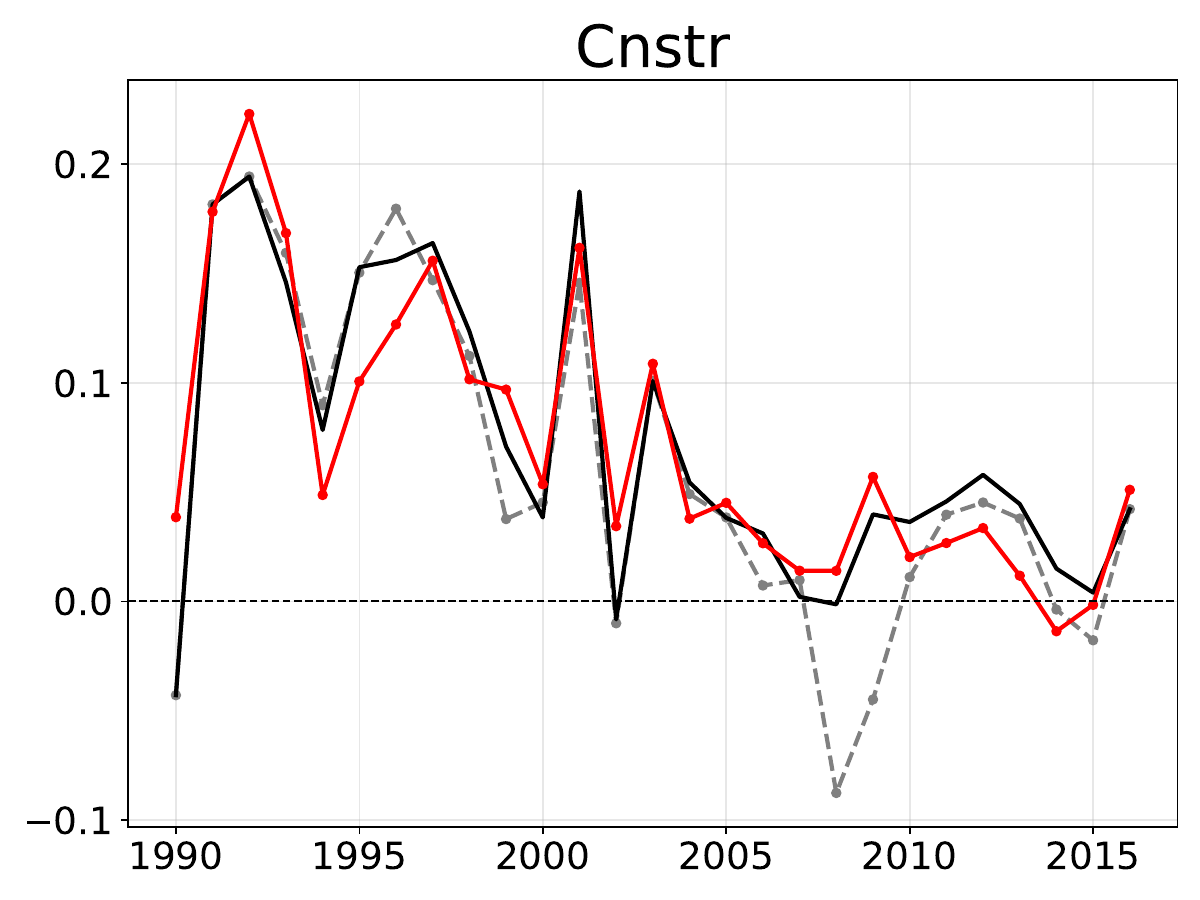}
	\end{subfigure}

    \begin{subfigure}{0.24\textwidth}
        \centering
        \includegraphics[width=\linewidth]{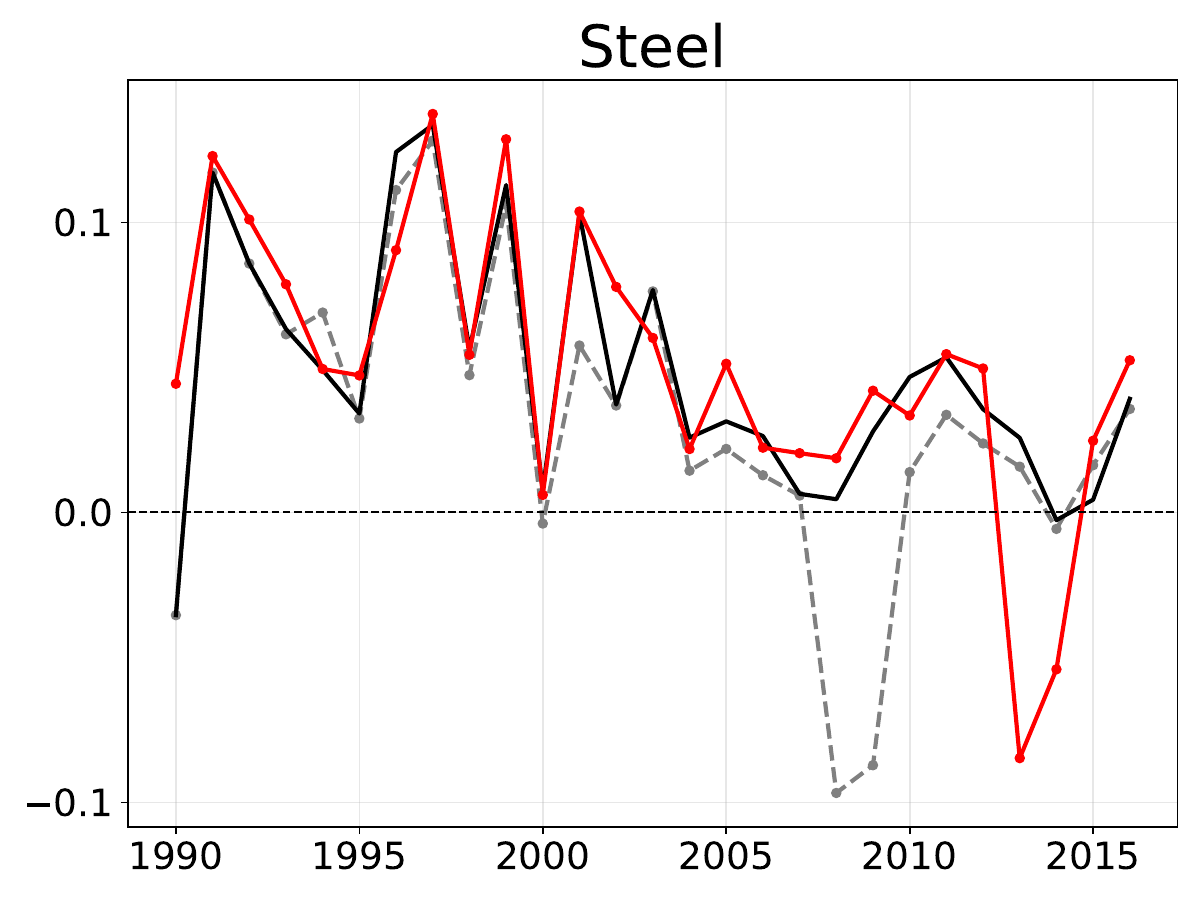}
	\end{subfigure}
    \begin{subfigure}{0.24\textwidth}
    	\centering
        \includegraphics[width=\linewidth]{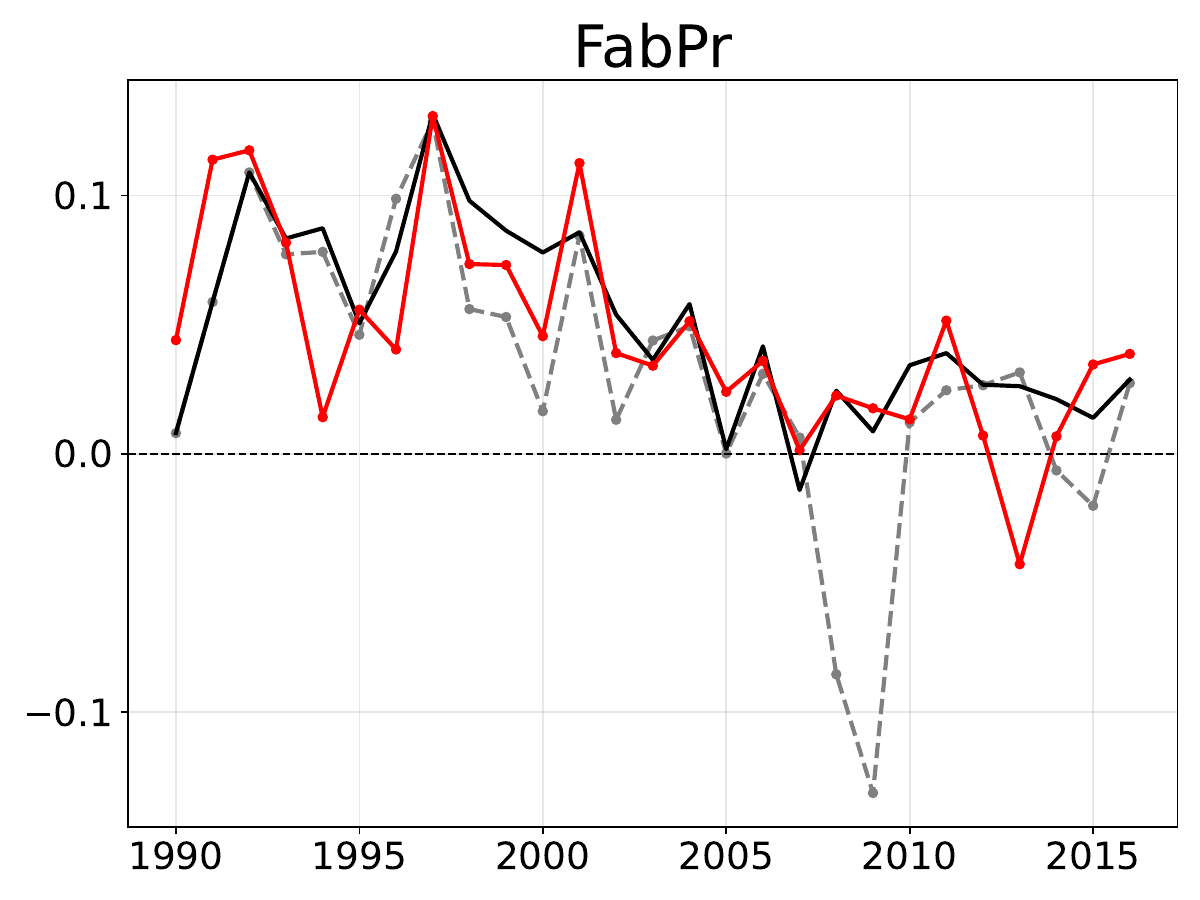}
	\end{subfigure}
    \begin{subfigure}{0.24\textwidth}
        \centering
        \includegraphics[width=\linewidth]{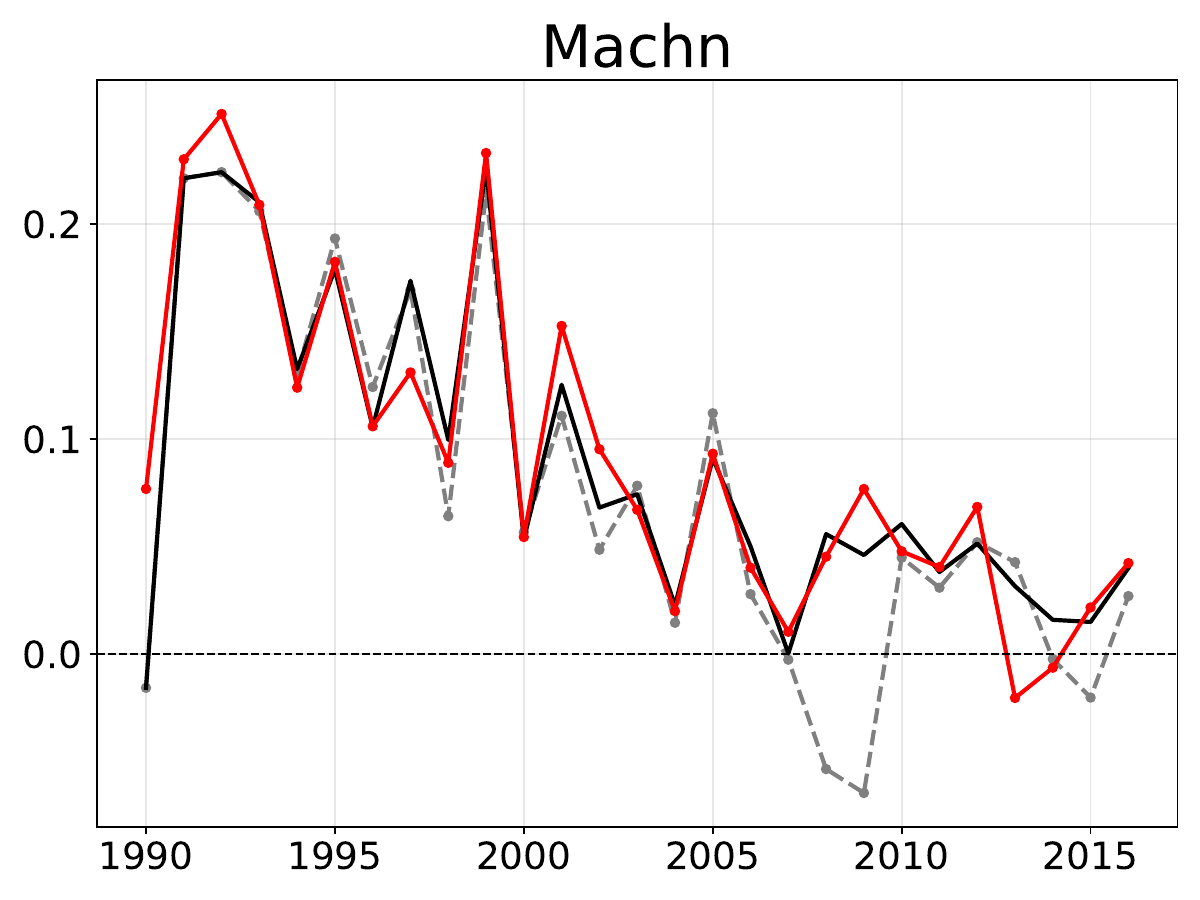}
	\end{subfigure}
    \hfill
    \begin{subfigure}{0.24\textwidth}
        \centering
        \includegraphics[width=\linewidth]{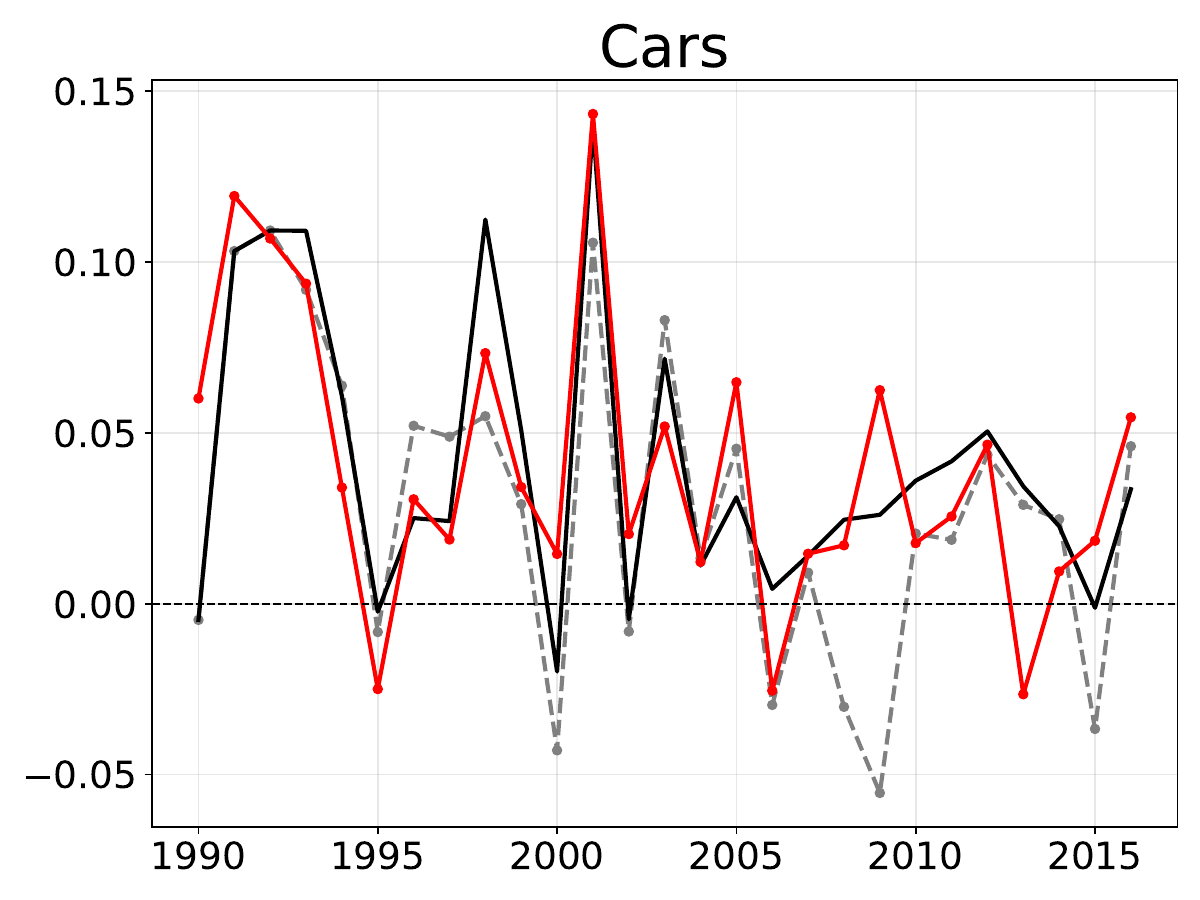}
	\end{subfigure}

    \begin{subfigure}{0.24\textwidth}
    	\centering
        \includegraphics[width=\linewidth]{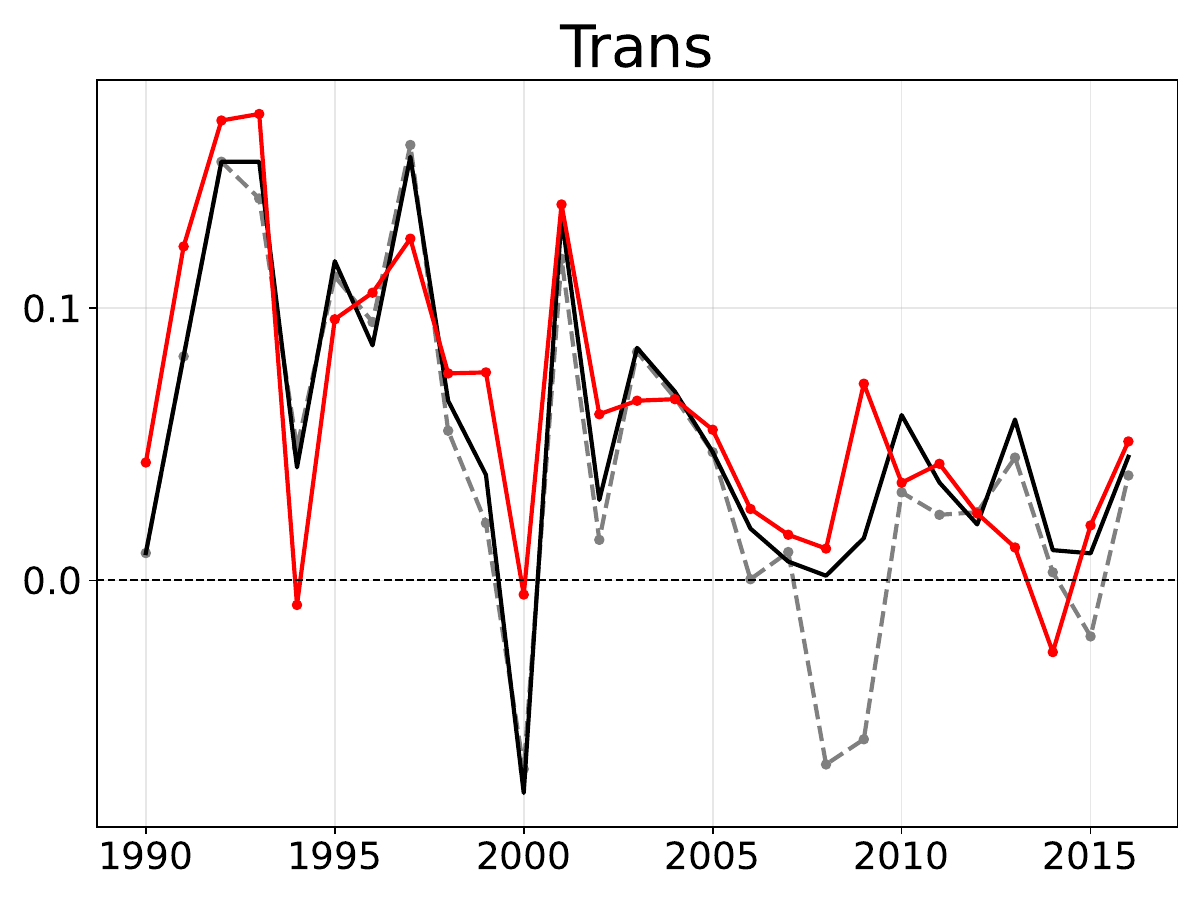}
	\end{subfigure}
    \begin{subfigure}{0.24\textwidth}
        \centering
        \includegraphics[width=\linewidth]{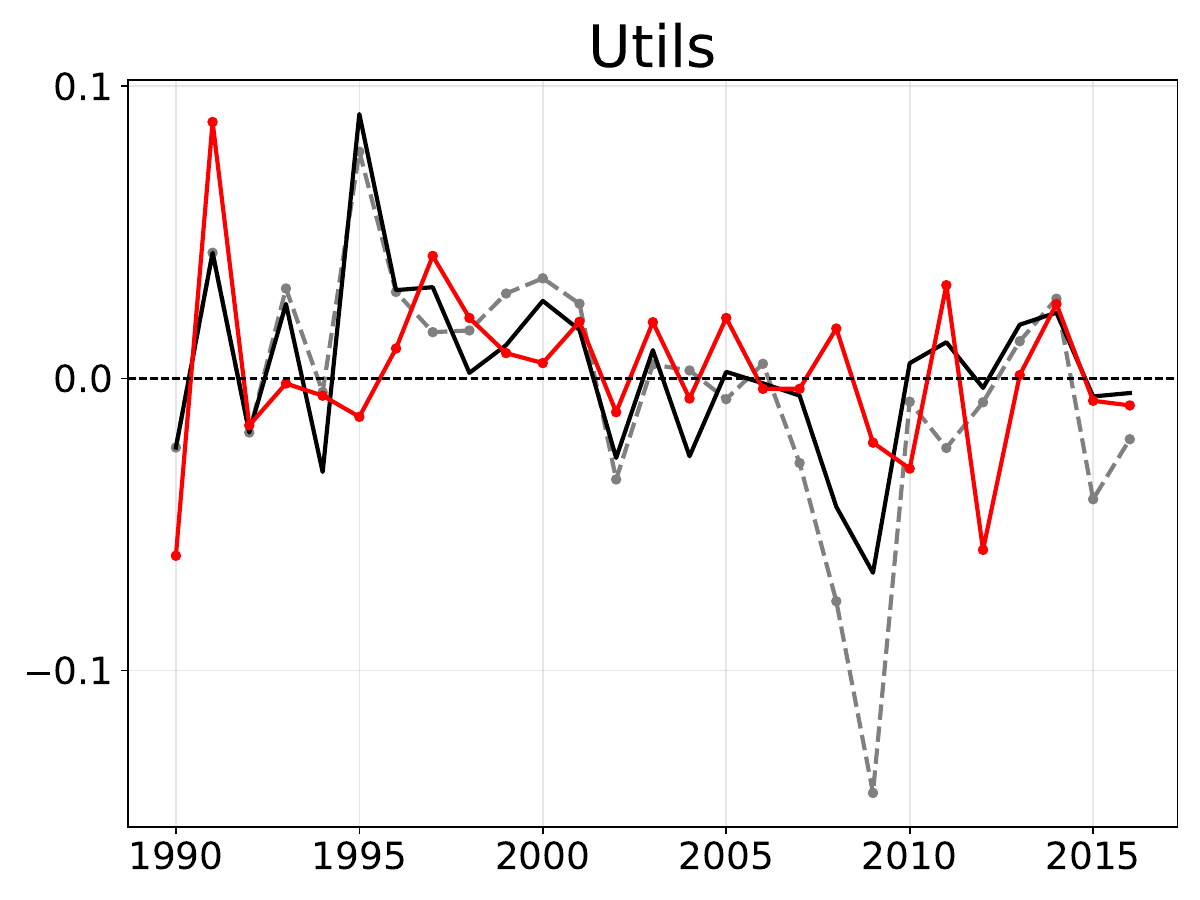}
	\end{subfigure}
    \begin{subfigure}{0.24\textwidth}
        \centering
        \includegraphics[width=\linewidth]{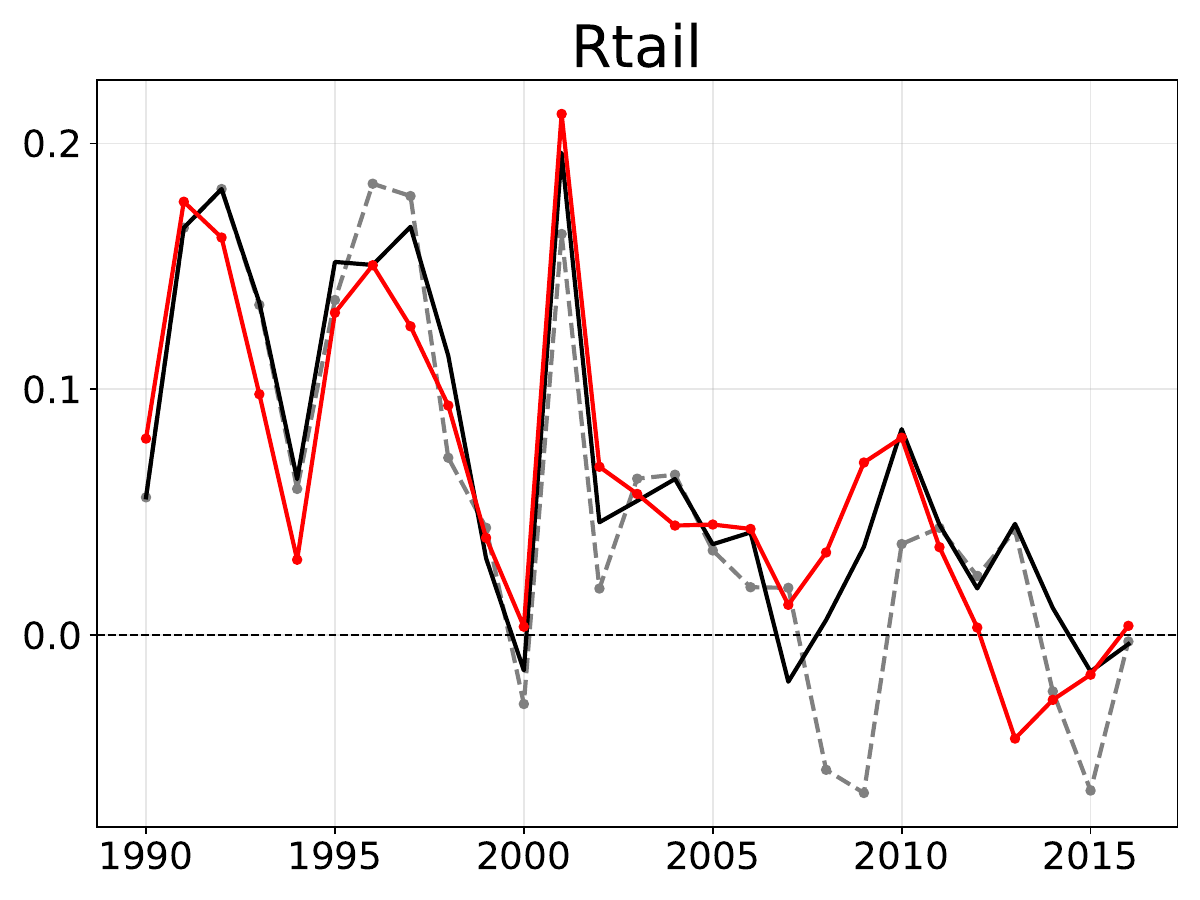}
	\end{subfigure}
    \begin{subfigure}{0.24\textwidth}
    	\centering
        \includegraphics[width=\linewidth]{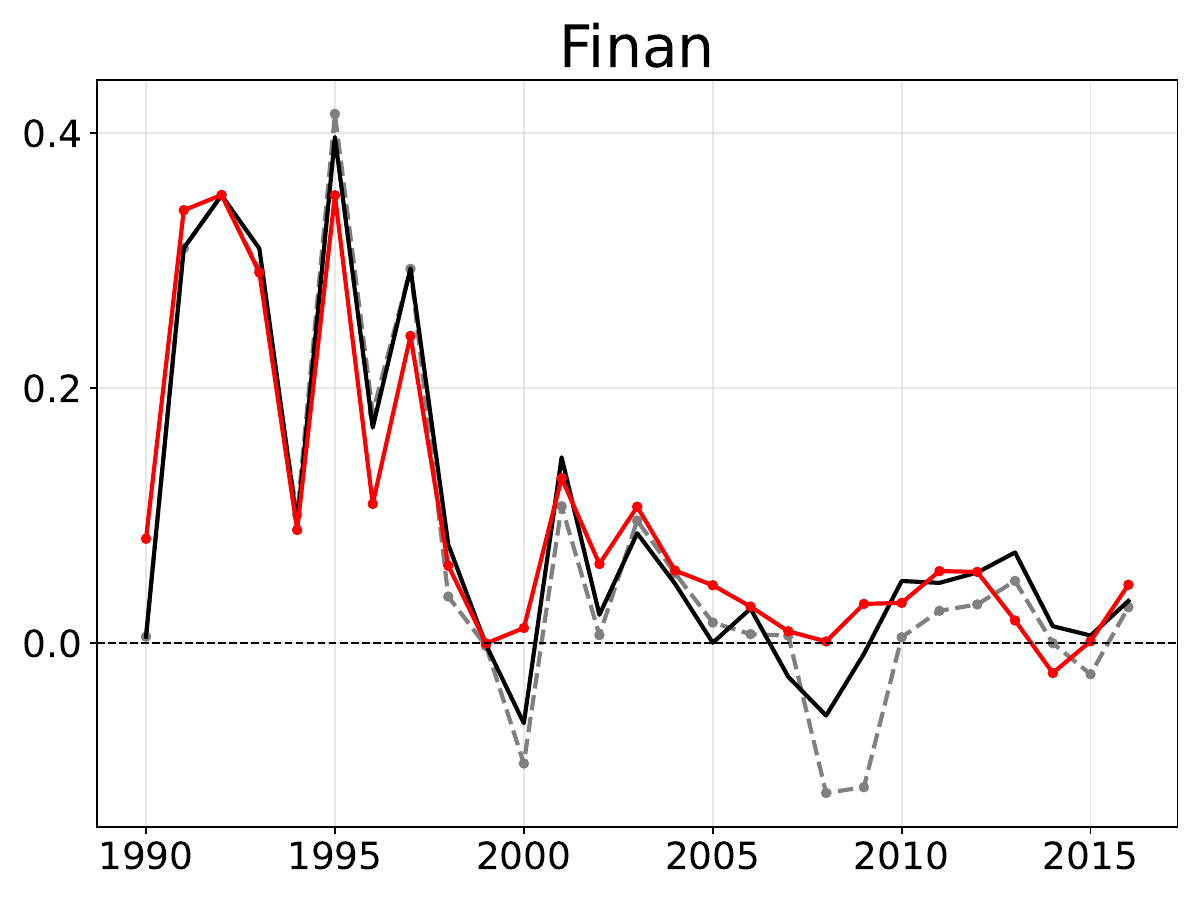}
	\end{subfigure}

    \begin{subfigure}{0.24\textwidth}
        \centering
        \includegraphics[width=\linewidth]{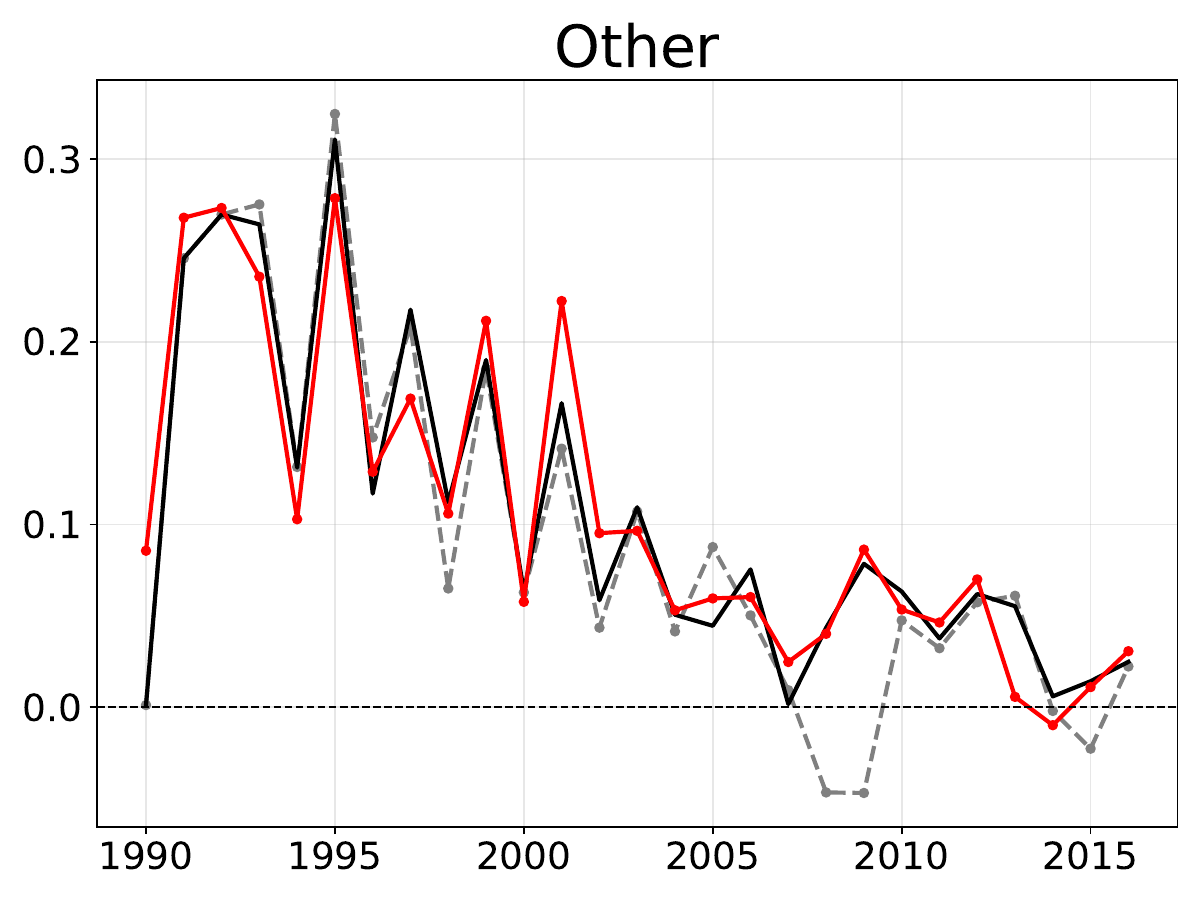}
	\end{subfigure}
    \bnotefig{This figure reports, for each of the $17$ industry portfolios, the annual out-of-sample $R^2$ from 1990 to 2016 for three models: (1) a linear model trained by ridge regression on the most recent $64$ months of data (red), (2) a random forest trained using the most recent $64$ months of data (gray), and (3) a random forest trained using all available historical data up to that year (black). In periods of strong non-stationarity, such as 1990-1991, 2001-2002 and 2008-2009, the  linear model trained on a small window constantly outperforms the more complex random forest trained on a large window. The labels in each figure is the Kenneth French acronym for the industries. For full names of these industries, please refer to Table \ref{tab-industry-name-mapping}.}
\end{figure}

We compute each model's annual out-of-sample $R^2$ for every industry. To visualize the models' relative performance across industries, we count the number of industries in which each model achieves the highest out-of-sample $R^2$ for a given year. \myCref{fig-tradeoff-dominance} summarizes the result. To provide a more granular understanding of the models' performance, \myCref{fig-tradeoff-industry} further plots the annual out-of-sample $R^2$ of the models for the $17$ industries to provide detailed accuracies in addition to ranking information.

We make two key observations. First, within the same model class, using less training data may lead to better performance. For example, in several years including $1994$, $1996$ and $2003$, the random forest trained on the most recent $64$ months of data outperforms the random forest trained on all historical data for at least half of the industries.

Second, and more strikingly, a simple model trained on a short window can outperform a complex model trained on a long window. In particular, during the three NBER-designated recessions, the simple linear model trained on $64$ months of data outperforms the more complex random forest model trained on all historical data for over half of the industries. This consistent pattern shows that the advantage of a more expressive model class can be completely negated by the non-stationarity in the training data. In \myCref{sec-select}, we propose a data-driven approache to select the best-performing model under non-stationarity.

These empirical findings highlight that in a non-stationary environment, the model complexity and training data size are intricately linked with each other. We call this phenomenon the \emph{nonstationarity-complexity tradeoff}. Crucially, the optimal choice of model class and training window size is not fixed; instead, it generally varies with the non-stationarity.

\subsection{Formal Characterization of the Tradeoff}\label{sec-tradeoff-theory}

We now provide theoretical support for the nonstationarity-complexity tradeoff, by deriving a finite-sample bound on a model's prediction error under non-stationarity. The bound decomposes the prediction error into three key components: model misspecification error, statistical uncertainty, and non-stationarity, and shows how each of them interacts with the choice of model class and training window size. We note that this bound is derived for theoretical analysis only, and does not explicitly characterize the optimal choice of the model class and training window. In the next section, we will develop a data-driven approach to adaptively select a near-optimal choice of model class and training window.

Consider a model $\widehat{f}$ trained from a model class $\modelclass$ by minimizing the empirical loss over training data from the last $k$ periods, denoted by $\big\{ \dataset^{\tr}_{\tau} \big\}_{\tau=t-k}^{t-1}$, where $\dataset^{\tr}_{\tau} = \{ (\bx_s,y_s) \}_{s\in\batchtr_{\tau}}$ is the training data in period $\tau$, with $\batchtr_{\tau}\subseteq\batch_{\tau}$. That is,
\begin{equation}\label{eqn-ERM}
\widehat{f} = \argmin_{f\in\modelclass} \frac{1}{\ntrain_{t,k}} \sum_{\tau=t-k}^{t-1} \sum_{s\in\batchtr_{\tau}} \left[ f(\bx_{s}) - y_{s} \right]^2,
\end{equation}
where $\ntrain_{t,k} = \sum_{{\tau}=t-k}^{t-1}\batchtr_{\tau}$ is the number of training data points in $\big\{ \dataset^{\tr}_{\tau} \big\}_{\tau=t-k}^{t-1}$. 

Define the Bayes optimal least squares estimator $f_t^*(\cdot) = \EE_{(\bx,y)\sim\distP_t}[y \mid \bx = \cdot]$, which minimizes the MSE $L_t(f)$ over all possible prediction models $f:\cX\to\RR$. Our bound will be stated in terms of the \emph{excess risk}
\[
\excessrisk_t(f) = L_t(f) - L_t(f_t^*),
\]
which compares the prediction error of a model $f$ against that of $f_t^*$. To facilitate analysis, we make the following boundedness assumption.

\begin{assumption}[Boundedness]\label{assumption-bounded}
There exists a constant $M > 0$ such that for all models $f$ in the class $\modelclass$, $(\bx,y)\sim\distP_{\tau}$ and $\tau\in\ZZ_+$, we have $|f(\bx)|\le M$ and $|y|\le M$. Without loss of generality, we assume $M\ge 1$.
\end{assumption}

To quantify the effective complexity of the model class $\modelclass$ relative to the training window size $k$, we employ a measure $r_{t,k}(\modelclass)$ derived from the well-known \emph{local Rademacher complexity} \citep{BBM05} commonly adopted in statistical learning theory.  Given the technical nature of this measure, we defer its formal definition to \myCref{sec-tradeoff-theory-appendix}. Intuitively, the local Rademacher complexity measures the ability of the near-optimal models in $\modelclass$ to fit random noise using data within the training window $k$. A higher local Rademacher complexity indicates a richer model class that is capable of approximating complex patterns, but also signals a higher estimation variance and thus a higher risk of overfitting. 

\paragraph{Examples of Local Rademacher Complexity in Finance.} To provide concrete illustrations, we now present the local Rademacher complexity measure $r_{t,k}(\modelclass)$ for several common model classes. These results are proved in \myCref{sec-proof-eg-classes}.


\begin{example}[Linear class]\label{eg-linear}
Recall $\featurespace \subseteq \RR^d$. For every $\btheta\in\RR^d$, define $f_{\btheta}:\featurespace\to\RR$ by $f_{\btheta}(\bx) = \langle\btheta , \bx\rangle$. Suppose $\modelclass \subseteq \{ f_{\btheta} : \btheta\in\RR^d \}$. Then,
$r_{t,k}(\modelclass) \le c d /\ntrain_{t,k}$ for some constant $c$. 
\end{example}

We note two relevant examples of linear class functionals in empirical asset pricing. Classical asset pricing fixed-dimension models such as \cite{fama1996multifactor,FFr15} fall clearly into this category. Ridge regression used in \cite{GKX20} corresponds to $\modelclass = \{ f_{\btheta} : \| \btheta \|_2 \le R \}$, which is also in the model class described by this example.

\begin{example}[Kernel class]\label{eg-kernel}
A kernel class is a nonparametric model class where we fit a linear model in a rich (possibly infinite-dimensional) feature space induced by a kernel. Let $\HH$ be a reproducing kernel Hilbert space \citep{Wah90} with inner product $\langle \cdot , \cdot \rangle$ and norm $\| \cdot \|_{\HH}$, and $\phi:\featurespace \to \HH$ be a feature mapping. For any $\btheta \in \HH$, define $f_{\btheta} :\featurespace \to \RR$ by $f_{\btheta}(\bx) = \langle \btheta, \phi(\bx) \rangle$. Consider the function class $\modelclass = \{ f_{\btheta}: \| \btheta \|_{\HH} \leq R \}$ for some constant $R > 0$. Model fitting in this class can be efficiently implemented through kernel ridge regression, which is a finite-dimensional convex program even if $\HH$ and $\modelclass$ are infinite-dimensional.

Suppose there exists a trace-class operator $\bS:\HH\to\HH$ such that for any $\tau \in \ZZ_+$ and $\bv \in \HH$, we have $\EE_{(\bx,y)\sim\distP_{\tau}} | \langle \phi (\bx) , \bv \rangle |^2 \leq \langle \bv , \bS \bv \rangle$. Let $\{ \mu_{j} \}_{j=1}^{\infty}$ be the eigenvalues of $\bS$ sorted in descending order. We have the following results:
\begin{itemize}
\item (Exponential  decay) If there are constants $c_1,c_2>0$ such that $\mu_{j} \leq c_1 e^{- c_2 j }$ holds for all $j$, then $r_{t,k}(\modelclass) \leq  (C \log \ntrain_{t, k} ) / \ntrain_{t, k}$ holds with some constant $C$. 

\item (Polynomial  decay) If there are constants $c>0$ and $\alpha \geq 1$ such that $\mu_j \leq c j^{-2 \alpha}$ holds for all $j$, then $r_{t,k}(\modelclass) \leq C  \big(\ntrain_{t, k}\big)^{-\frac{2 \alpha}{2 \alpha + 1}}$ holds with some constant $C$.
\end{itemize}

How does the kernel class relate to asset pricing? Fast (exponential) decay corresponds to a pricing relationship that is effectively spanned by a few dominant directions in the feature space, so the kernel class behaves like a low-complexity model despite its nominal richness. Slow (polynomial) decay corresponds to a pricing relationship in which many directions carry independent predictive content, so the model must be treated as genuinely high-dimensional.

\end{example}

Specifically, recent asset pricing literature has also considered machine learning models related to kernel classes.  \cite{DKK24} construct high-dimensional SDFs from random Fourier factors, which transform raw firm characteristics into many nonlinear sine and cosine features, and then estimate a linear SDF over these generated factors. This can be viewed as a finite-dimensional approximation to a kernel model, where the random features provide an explicit feature map for a rich nonlinear kernel class. The construction is also equivalent to a two-layer neural network with random first-layer weights and estimated output weights. More generally, wide neural networks are often connected to kernel classes through their neural tangent kernels \citep{JGH18}.

We now proceed to describe our prediction error bound under non-stationarity. In the classical setting where the training data $\{\dataset_{\tau}^{\tr}\}_{\tau=t-\wtrain}^{t-1}$ is i.i.d., the complexity measure $r_{t,\wtrain}(\modelclass)$ is a key component in bounding the excess risk of $\widehat{f}$. Standard results in statistical learning theory \citep{BBM05, Wai19} show that with high probability,
\begin{equation}\label{eqn-tradeoff-iid}
\excessrisk_t(\widehat{f})
\lesssim 
\min_{f\in\modelclass}\excessrisk_t(f)
+
\left(
r_{t,\wtrain}(\modelclass)
+
\frac{1}{\ntrain_{t,\wtrain}}
\right),
\end{equation}
where $\lesssim$ hides a multiplicative universal constant. In particular, the prediction error on the right-hand side of Eq. (\ref{eqn-tradeoff-iid}) is decomposed into two terms:
\begin{enumerate}
\item \emph{Model misspecification error} 
\[
\min_{f\in\modelclass}\excessrisk_t(f) = \min_{f\in\modelclass}L_t(f) - L_t(f_t^*),
\] 
which describes how well the model class $\modelclass$ can approximate the Bayes optimal least squares estimator $f_t^*$ at time $t$. A more complex model class tends to reduce the model misspecification error.
\item \emph{Statistical uncertainty} 
\[
r_{t,\wtrain}(\modelclass) + \frac{1}{\ntrain_{t,\wtrain}},
\] 
which quantifies the estimation variance of the model $\widehat{f}$. As is discussed above, using a more complex model class $\modelclass$ increases the statistical uncertainty of the fitted model. Consequently, a more complex model typically requires a longer training window $\wtrain$ to mitigate its estimation variance.
\end{enumerate}

The classical error bound \eqref{eqn-tradeoff-iid} shows that in the i.i.d.~case, increasing the training window size $\wtrain$ always reduces the statistical uncertainty, thereby lowering the total prediction error. However, we now present our theory to show that under non-stationarity, this logic is incomplete. As we increase the window size $\wtrain$ to reduce estimation variance, we inadvertently include older data distributions that differ from the target, introducing a third error component. We formalize this in the following theorem.

\begin{theorem}[Prediction error bound]\label{thm-tradeoff}
Let Assumptions \ref{assumption-independence} and \ref{assumption-bounded} hold, and fix $\delta\in(0,1)$. With probability at least $1-\delta$, the model $\widehat{f}$ defined by \eqref{eqn-ERM} satisfies
\[
\excessrisk_t(\widehat{f})
\lesssim 
\min_{f\in\modelclass}\excessrisk_t(f)
+
M^2\left(
r_{t,k}(\modelclass)
+
\frac{\log(1 / \delta)}{\ntrain_{t,k}}
\right)
+
M^2 \max_{t-k\le {\tau}\le t-1} \TV\left( \distP_{\tau},\distP_t \right).
\]
Here $\lesssim$ hides a universal constant, and $\TV(\distP_{\tau},\distP_t) = \max_{A}|\distP_{\tau}(A) - \distP_t(A)|$ is the total variation distance.
\end{theorem}

\begin{proof}[Proof of \myCref{thm-tradeoff}]
See \myCref{sec-tradeoff-theory-appendix}.
\end{proof}

Contrasting with the classical stationary case in \eqref{eqn-tradeoff-iid}, \myCref{thm-tradeoff} reveals that the non-stationarity adds a third dimension to the classical prediction error bound \eqref{eqn-tradeoff-iid}, namely, a \emph{non-stationarity} term
\[
\max_{t-k\le \tau\le t-1} \TV\left( \distP_{\tau},\distP_t \right),
\]
which quantifies the largest environment shift between the current modeling time $t$ and any period $\tau$ included in the training window. Mathematically, it measures the changes in the joint distribution of covariates and return in terms of the total variation distance.

In markets, non-stationarity naturally arises from shifts in risk premia, credit conditions, investor risk aversion, and the distribution of firm characteristics. These shifts as a high-dimensional event since any of these market indicators may have marginal or joint distributional variations. Whenever a shift happens, historical data becomes progressively less informative for model estimation. The term $\TV(\distP_{\tau},\distP_t)$ summarizes this mismatch between the past data distribution $\distP_{\tau}$ and the current distribution $\distP_t$.
The financial economics content of this bound is most transparent during episodes of sudden market stress.

The error decomposition in \myCref{thm-tradeoff} formalizes the empirical observations in \myCref{sec-tradeoff-empirics}. It shows that model complexity and training-window size affect different components of prediction error. Using a more complex model class $\modelclass$ can reduce misspecification error $\min_{f\in\modelclass} \excessrisk_t(f)$, but it also increases statistical uncertainty $r_{t,k}(\modelclass)$. Increasing the training window $k$ reduces statistical uncertainty $r_{t,k}(\modelclass) + \frac{\log(1 / \delta)}{\ntrain_{t,k}}$ by providing more samples, but is also more likely to include data generated under distributions $\distP_{\tau}$ far from the current distribution $\distP_t$. \myCref{tab-generalization-components} summarizes these effects.

\begin{table}[h!]
\centering
\caption{Impacts of Model Complexity and Training Window Size on Prediction Error
       \label{tab-generalization-components}}
\small
\begin{tabularx}{\textwidth}{%
  l
  >{\hsize=1.1\hsize\centering\arraybackslash}X
  >{\hsize=1\hsize\centering\arraybackslash}X
  >{\hsize=0.9\hsize\centering\arraybackslash}X}
\toprule
& Misspecification & Statistical Uncertainty & Non-stationarity \\
\midrule
Model complexity $\nearrow$   & $\searrow$ & $\nearrow$ & Intrinsic to window \\ [4pt]
Training window $\nearrow$ & Intrinsic to model class & $\searrow$ & $\nearrow$ \\
\bottomrule
\end{tabularx}
\bnotetab{The signs $\nearrow$ and $\searrow$ indicate the component increases or decreases as  the left-column factor increases. ``Intrinsic'' entries indicate the component is
governed by the other dimension and is not directly controlled by the factor shown.}
\end{table}

To interpret the tradeoff presented by \Cref{thm-tradeoff} from an investor's perspective, consider an asset manager who estimated a return-prediction model in mid-1990 using the prior five years of data. The August 1990 oil shock shifted the joint distribution of characteristics and returns in ways that were poorly represented in the pre-shock samples. In the notation of \Cref{thm-tradeoff}, the past distributions $\distP_{\tau}$ are far from the current distribution $\distP_t$, increasing $\TV(\distP_{\tau},\distP_t)$. At the same time, only a small number of post-shock samples were available, making very short windows statistically noisy. \Cref{thm-tradeoff} formalizes this dilemma: use more history to reduce estimation variance, or use less history to avoid outdated regimes.

A similar logic applies to the choice of model class. Parsimonious factor models, such as the Fama-French three- or five-factor model \citep{FFr93,FFr15}, may have a larger misspecification error $\min_{f\in\modelclass} \excessrisk_t(f)$, but its low complexity means that the estimation variance $r_{t,k}(\modelclass)$ can be controlled with relatively few samples. Lower complexity models can therefore be estimated on a shorter window, limiting exposure to outdated regimes. By contrast, a random forest or neural network using a large factor zoo \citep{GKX20} may better approximate nonlinear return relationships and has smaller misspecification error $\min_{f\in\modelclass} \excessrisk_t(f)$, but its larger complexity term $r_{t,k}(\modelclass)$ requires more data to drive down the estimation variance. This forces the model to reach further into the past, increasing the non-stationarity $\max_{t-k\le \tau\le t-1} \TV\left( \distP_{\tau},\distP_t \right)$. This is the mechanism behind the ``less can be more'' phenomenon observed in \Cref{sec-tradeoff-empirics}: complex models can face a larger non-stationarity cost when their variance reduction requires long historical windows.

Put together, neither greater model complexity nor more training data is uniformly beneficial under non-stationarity. Below we illustrate this phenomenon in a stylized setting.

\begin{example}[Selection of model class and window under non-stationarity]
Let $\eta,\gamma \in [0, 1]$ be two small constants. Suppose that at each time $t$, we collect one sample, and the covariate and response $(x,y)\sim\distP_t$ satisfy $x\sim\uniform [0,1]$, $y \sim N( f_t^*(x) , 1 )$ conditioned on $x$, and
    \[
    f_t^*(x) = c_t x + \gamma \sin (2 \pi x),
    \]
    where $\{c_t\}_{t=1}^{\infty}$ is a deterministic sequence in $[0,1]$ satisfying $|c_{t+1} - c_t| = \eta$. We observe a single sample per period. Consider two model classes: linear class and kernel class with a first-order Sobolev kernel (see, e.g., Example 12.16 in \cite{Wai19}). This kernel class contains smooth nonlinear functions represented through Fourier sine and cosine bases, which can capture the sinusoidal component in $f_t^*$, while the linear class cannot.
    \begin{itemize}
        \item If we train a linear model with a training window $k$, then the three components of the prediction error bound in \Cref{thm-tradeoff} satisfy
        \[
\min_{f\in\modelclass}\excessrisk_t(f) \asymp \gamma^2, \qquad r_{t,k}(\modelclass) \asymp k^{-1}, \qquad \max_{t-k\le \tau\le t-1} \TV\left( \distP_{\tau},\distP_t \right) \asymp k\eta.
        \]
        Optimizing their sum over $k$ yields the optimal window size $k^*\asymp \eta^{-1/2}$, which leads to an $O(\gamma^2 + \eta^{1/2})$ bound on the prediction error.
        \item If we use the kernel class, then $f_t^*$ is well-specified. For a training window $k$, we have
        \[
        \min_{f\in\modelclass}\excessrisk_t(f) = 0, \qquad r_{t,k}(\modelclass) \asymp k^{-2/3}, \qquad \max_{t-k\le \tau\le t-1} \TV\left( \distP_{\tau},\distP_t \right) \asymp k\eta.
        \]
        The optimal training window is $k^*\asymp \eta^{-3/5}$, which results in a prediction error of $O (\eta^{2/5})$.
    \end{itemize}
    
\end{example}
We observe that for both classes in the above example, the optimal window size depends on the severity of the drift $\eta$, and is in general not the full window size. If one na\"ively uses the kernel class with a large window size, then the resulting error scales as $O(k\eta)$, which is linear in $k$ and can be much worse than the above bounds.

As expected, the preferable model class depends on the interplay between misspecification $\gamma$ and drift $\eta$. The kernel class is more expressive but more sensitive to drift. When $\eta = O(\gamma^5)$, drift is relatively mild and the kernel is optimal,  consistent with the ``virtue of complexity'' \citep{KellyMalamud2025Understanding,Kelly2022Virtue, KMZ24}. However, when $\eta \gg \gamma^5$, severe non-stationarity requires shorter training windows, under which sample sizes are too limited for the kernel estimator to fully exploit its flexibility advantage. In this high-drift regime, the linear class achieves better performance with its shorter optimal window, explaining the ``less can be more'' phenomenon observed in our experiments.

\paragraph{Unavoidable Cost of Non-stationarity.} The non-stationarity term in \myCref{thm-tradeoff} is not an artifact of our proof technique: it reflects an intrinsic limitation of learning from non-stationary historical data. Formally, consider any class of non-stationary environments in which every training distribution is within total variation distance $V$ of the current distribution $\distP_t$; one can think of $V$ as the magnitude of the local non-stationarity. Over this class, no estimator, however cleverly designed, can achieve a worst-case prediction error better than the same three components identified in \myCref{thm-tradeoff}: the misspecification error of the model class, the statistical uncertainty from the finite training sample, and a term of order $V$ that is the irreducible cost of training on historical distributions that differ from the current one. Thus, even if the model class is correctly chosen and the estimator is optimal, non-stationarity alone can generate an error of order $V$, confirming that the upper bound in \myCref{thm-tradeoff} has the correct dependence on non-stationarity and is not simply a loose consequence of our analysis. We state and prove this lower bound formally as \Cref{prop-tv-lower-bound} in \Cref{sec-prop-tv-lower-bound-proof} of the appendix.

\paragraph{A Refined View of Non-stationarity in High Dimensions.} \Cref{thm-tradeoff} characterizes non-stationarity through a single total variation distance $\max_{t-k\le \tau\le t-1} \TV\left( \distP_{\tau},\distP_t \right)$ between the training and current distributions. This single distance bundles together two economically distinct sources of drift. The first is a shift in how characteristics map into expected returns: the same firm characteristic earns a different expected return in a different regime, i.e., a change in $f_{\tau}^*(\cdot) = \EE_{(\bx,y)\sim\distP_{\tau}}[y\mid\bx = \cdot]$, or more generally in the conditional distribution of $y$ given $\bx$. In the statistics and machine learning literature, this is called \emph{conditional shift} (or \emph{concept shift}). The second is a shift in the covariate distribution of $\bx$ itself \citep{Hec79, SKa12}, for instance, the cross-section of firms or the range of characteristic exposures observed in the market changes, even though the underlying pricing relationship is unchanged. This is known as \emph{covariate shift}. While shifts in the pricing relationship are the more classical concern, recent work has shown that covariate shift alone can substantially degrade performance in modern nonparametric and high-dimensional machine learning models \citep{KSM21, KMa21, MPW23, PDT24, Wan26}: models may be evaluated on combinations of characteristics that are poorly represented in the training data, leading to large out-of-sample error even when the pricing relationship itself has not changed. In \Cref{sec-thm-tradeoff-proof}, we prove a more refined prediction error bound (\Cref{thm-tradeoff-refined}) that separately decomposes the non-stationarity term into these two sources.

This refined view highlights the crucial role of non-stationarity in modern ML-based empirical asset pricing. Classical fixed-dimensional factor models such as the Fama-French three- and five-factor models \citep{FFr93,FFr15} admit small, fixed risk factor sets. For such low-dimensional linear models, the complexity term $r_{t,k}(\modelclass)$ is typically small and decreases quickly with the number of samples. This means using a shorter window to reduce non-stationarity does not come with a large estimation cost. If the covariate space is low-dimensional, moderate changes in factor realizations are less likely to shift into a region with little historical support. Therefore, fixed-dimension models can maintain reasonably stable performance under moderate distribution shifts.

However, the tradeoff becomes acute in the high-dimensional regime where the number of covariates $d$ grows with the sample size. This regime covers neural network and random-feature models with growing effective dimension \citep{KMZ24, DKK24}, LASSO-based methods that use a large cross-section of lagged returns \citep{CCY19}, regularized SDF estimators over many characteristic-sorted factors \citep{KNS20, FNW20}, and high-dimensional ML models using dozens or hundreds of firm characteristics \citep{GKX20}. These models can reduce misspecification error, but their complexity term $r_{t,k}(\modelclass)$ is larger and typically requires many more samples to control. They therefore benefit more from long windows in stable periods, but are also more heavily affected by non-stationarity when past data come from different regimes. Covariate shift is also more damaging in high dimensions because the current distribution of characteristics may be poorly represented in historical data. This is the setting in which the nonstationarity-complexity tradeoff is most relevant: the model's flexibility that improves approximation also makes it more reliant on careful and time-dependent choices of training window and hyperparameters.

\section{Adaptive Model Selection Algorithm \adaptive}\label{sec-select}

\myCref{sec-tradeoff} shows that the predictive performance of a model depends  jointly on its complexity and the size of the training data, and that the optimal choice often varies over time with the non-stationarity. As the non-stationarity is generally unknown \emph{a priori}, the selection of the model class and training window calls for a data-driven approach.

In this section, we develop a novel method that uses historical validation data to select the best model from a set of candidates. These candidate models can come from different model classes, be trained on different time windows, or use different hyperparameters. The main challenge is that the same non-stationarity that complicates model training also incapacitates standard model selection techniques such as holdout and cross validation.
Specifically, in a non-stationary environment, a model that performs well on a validation set from the distant past may not perform as well in the future. Our solution is to adaptively select the relevant validation data that best reflects the current environment, allowing for a more accurate comparison of the candidate models' future performance.

\begin{figure}[h]
    \centering
    \caption{Our Framework for Model Training and Selection under Non-stationarity. \label{fig-diagram-tr-va}}
    \includegraphics[scale=1]{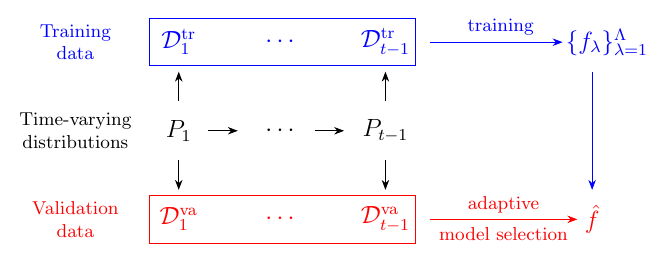}
\end{figure}

We now formally set up the framework, illustrated in \Cref{fig-diagram-tr-va}. In each period $t$, we split the available data $\{\dataset_{\tau}\}_{\tau=1}^{t-1}$ into a training dataset $\{\dataset_{\tau}^{\tr}\}_{\tau=1}^{t-1}$ and a validation dataset $\{\dataset_{\tau}^{\va}\}_{\tau=1}^{t-1}$. More precisely, each $\dataset_{\tau}=\{(\bx_s,y_s)\}_{s\in\batch_{\tau}}$ is stratified into $\dataset_{\tau}^{\tr}=\{(\bx_s,y_s)\}_{s\in\batchtr_{\tau}}$ and $\dataset_{\tau}^{\va}=\{(\bx_s,y_s)\}_{s\in\batchva_{\tau}}$. We use the training data $\{\dataset_{\tau}^{\tr}\}_{\tau=1}^{t-1}$ to produce a finite set of candidate models $\{f_{\modelidx}\}_{\modelidx=1}^{\nmodel}$.
These candidates can come from different model classes, be trained on different data horizons, or use different hyperparameters. We will use the validation data $\{\dataset_{\tau}^{\va}\}_{\tau=1}^{t-1}$ to select a good model $\widehat{f} = f_{\widehat{\modelidx}}$ that achieves near-optimal predictive performance on the current distribution $\distP_t$. The selected model $\widehat{f}$ is then tested on the out-of-sample data $\dataset_t$ at time $t$.

\subsection{Selecting a Model Through Pairwise Comparisons}

In this section, we describe our model selection approach, which compares candidate models through a sequential elimination tournament, much like a single-elimination sports bracket. The procedure relies on a pairwise comparison subroutine $\alg$, described below, that takes two candidate models $f$ and $f'$ and returns whichever one it judges to have better future performance, denoted $\alg(f,f')$. The tournament proceeds in rounds: in each round, one remaining model is drawn at random as the pivot and is compared, in turn, against every other remaining model. Any model that beats the pivot advances to the next round; if the pivot beats every other remaining model, it is declared the overall winner and the tournament stops. We record the full pseudocode in \myCref{sec-algorithmic-details} of the appendix, together with the proofs that rely on it.

This tournament structure has two attractive properties. It is computationally cheap:

\begin{lemma}[Computational complexity]\label{lem-complexity-tournament}
\myCref{alg-tournament} calls the subroutine $\alg$ for $\Theta(\nmodel)$ times in expectation.
\end{lemma}

\begin{proof}[Proof of \myCref{lem-complexity-tournament}]
See \myCref{sec-lem-complexity-tournament-proof}.
\end{proof}

It is also statistically efficient: we show later in this section that the tournament preserves the performance guarantee of whichever pairwise comparison subroutine $\alg$ it is built on, at the cost of only a logarithmic factor in the number of candidate models $\nmodel$.

\paragraph{How Much Validation History to Trust.} We now detail the pairwise comparison subroutine $\alg$. Comparing two models directly on the full non-stationary validation history $\{\dataset^{\va}_{\tau}\}_{\tau=1}^{t-1}$ can badly mislead an investor, because validation data from years ago may reflect a market regime that no longer applies. We instead build on the adaptive rolling-window approach of \cite{HHW24}, which chooses, separately for each comparison, how far back in the validation history to look.

The idea is as follows. To choose between two models, it suffices to determine the sign of their \emph{performance gap}
\[
\gap_t = L_t(f_1) - L_t(f_2).
\]
Indeed, $f_2$ is better than $f_1$ if and only if $\gap_t > 0$. To estimate $\gap_t$ from non-stationary validation data, a natural approach is to fix a look-back window $\wval\in[t-1]$ and average the performance gap over the validation data from the last $\wval$ periods, $\dataset^{\va}_{\tau} = \big\{(\bx_{s}, y_s)\big\}_{s\in\batchva_{\tau}}$, $\tau=t-\wval,...,t-1$, giving a rolling-window estimate
\begin{equation}\label{eqn-performance-gap-hat}
\widehat{\gap}_{t,\wval} = \frac{1}{\nval_{t,\wval}}\sum_{\tau=t-\wval}^{t-1}\sum_{s\in\batchva_{\tau}} u_s, \quad\text{where}\quad
u_s = \left[ f_1(\bx_s) - y_s \right]^2 - \left[ f_2(\bx_s) - y_s \right]^2,
\end{equation}
and $\nval_{t,\wval} = \sum_{\tau=t-\wval}^{t-1} \nval_{\tau}$. The accuracy of this estimate hinges on the look-back window $\wval$: a longer window pools in more validation observations, but risks pooling in data from a stale market regime. This is exactly the tension identified in \Cref{thm-tradeoff}, now applied to comparing two models' validation performance rather than to fitting a single model. Formally, with probability at least $1-\delta$,
\begin{equation}\label{eqn-bias-variance-decomp}
\big| \widehat{\gap}_{t,\wval} - \gap_t \big| \le \underbrace{\phi(t,\wval)}_{\text{cost of stale data}} + \underbrace{\psi(t,\wval,\delta)}_{\text{cost of limited data}}.
\end{equation}
Here, the cost of stale data,
\[
\phi (t , \wval ) = \max_{t-\wval\le \tau\le t-1} | \gap_{\tau} - \gap_t |,
\]
measures how much the performance gap has moved within the look-back window, and the cost of limited data, $\psi(t,\wval,\delta)$, is a statistical confidence bound that shrinks as more validation observations accumulate:
\[
\psi(t,\wval,\delta)
=
\begin{cases}
M',&\ \text{if } \nval_{t,\wval}=1 \\[6pt]
\displaystyle \sd_{t, \wval}  \sqrt{ \frac{2 \log( 2 / \delta) }{ \nval_{t, \wval} } } + \frac{ 2M' \log ( 2 / \delta) }{ 3 \nval_{t, \wval} } , &\ \text{if } \nval_{t,\wval}\ge 2
\end{cases},
\]
where
\[
\sd_{t,\wval}^2 = \frac{1}{\nval_{t,\wval}} \sum_{\tau=t-\wval}^{t-1} \sum_{s\in\batchva_{\tau}} \var( u_s )
\qquad\text{and}\qquad
M'=8M^2.
\]
These are exactly the non-stationarity and statistical uncertainty terms from the bound in \Cref{thm-tradeoff}, now measured for the performance gap between two models rather than for a single fitted model. The ideal look-back window would minimize the sum of the two costs,
\[
\wval^* = \argmin_{\wval\in[t-1]} \left\{ \phi(t,\wval) + \psi(t,\wval,\delta) \right\},
\]
but both costs depend on the unknown non-stationarity and cannot be computed directly.

To make this choice implementable, we construct data-driven proxies $\widehat{\psi}$ and $\widehat{\phi}$ for these two costs. The proxy for the cost of limited data is built by replacing the unknown variance $\sd_{t,\wval}^2$ with the sample variance
\begin{equation}\label{eqn-sigma-hat}
\totalsdhat_{t, \wval}^2 = \frac{1}{\nval_{t,\wval}-1} \sum_{\tau=t-\wval}^{t-1} \sum_{s\in\batchva_{\tau}} \big( u_s - \widehat{\gap}_{t, \wval}  \big)^2,
\end{equation}
which gives
\begin{equation}\label{eqn-psi-hat}
\widehat{\psi}(t, \wval, \delta)
=
\begin{cases}
M',&\ \text{if } \nval_{t,\wval}=1 \\[6pt]
\displaystyle \totalsdhat_{t, \wval}  \sqrt{ \frac{2 \log( 2 / \delta) }{ \nval_{t, \wval} } } + \frac{ 8M' \log ( 2 / \delta) }{ 3 ( \nval_{t, \wval} - 1 ) } , &\ \text{if } \nval_{t,\wval}\ge 2
\end{cases}.
\end{equation}
The proxy for the cost of stale data compares the window-$\wval$ estimate against every shorter window, after netting out each estimate's own statistical noise\footnote{This construction adapts the Goldenshluger--Lepski method from the nonparametric statistics literature on bandwidth selection \citep{GLe08} to the problem of choosing a look-back window.}:
\begin{equation}\label{eqn-phi-hat}
\widehat{\phi} (t, \wval, \delta)=
\max_{i \in [\wval]} \bigg( \big| \widehat{\gap}_{t, \wval} - \widehat{\gap}_{t, i} \big|
-  \big[ \widehat{\psi}\left(t, \wval , \delta \right) + \widehat{\psi}\left(t, i , \delta \right)  \big] \bigg)_+.
\end{equation}
Intuitively, in light of the decomposition \eqref{eqn-bias-variance-decomp}, the quantity
\begin{equation}\label{eqn-phi-hat-heart}
\left(| \widehat{\gap}_{t, \wval} - \widehat{\gap}_{t, i } | - \big[ \widehat{\psi}\left(t, \wval , \delta \right) + \widehat{\psi}\left(t, i , \delta \right) \big] \right)_+
\end{equation}
measures how much the window-$\wval$ and window-$i$ estimates disagree once their combined statistical noise has been subtracted off: any remaining disagreement must reflect a genuine shift in the performance gap over that stretch of history, rather than sampling noise. The estimated cost of stale data, $\widehat{\phi}(t,\wval,\delta)$, is the largest such disagreement over all shorter windows $i\in[\wval]$.

The procedure then selects a look-back window that minimizes the sum of the two estimated costs,
\begin{equation}
\widehat{\wval} = \argmin_{\wval\in[t-1]} \left\{ \widehat{\phi} (t , \wval ) + \widehat{\psi}(t, \wval, \delta) \right\},
\end{equation}
and compares the two models using $\widehat{\gap}_{t,\widehat{\wval}}$: the subroutine selects $f_1$ if and only if $\widehat{\gap}_{t,\widehat{\wval}} \le 0$. We record the full procedure as pseudocode (\myCref{alg-compare}) in \myCref{sec-algorithmic-details} of the appendix.

Combining this comparison subroutine with the tournament described above yields our model selection procedure, which we call \underline{A}daptive \underline{To}urnament \underline{M}odel \underline{S}election, or $\adaptive$ for short.

\subsection{Performance Guarantees for the Selection Procedure}

We now present theoretical guarantees for our model selection framework. \myCref{thm-model-comparison} establishes a performance bound for the pairwise model comparison subroutine described above.

\begin{theorem}[Near-optimal model comparison]\label{thm-model-comparison}
Let Assumptions \ref{assumption-independence} and \ref{assumption-bounded} hold. Choose $\delta \in (0, 1)$ and take $\delta' = \delta / (3t)$ in the comparison subroutine. With probability at least $1 - \delta $, the output $\widehat{f}$ satisfies
\begin{equation}\label{eqn-model-comparison-fast-squared}
\excessrisk_t(\widehat{f})
\lesssim
\min
\big\{
\excessrisk_t(f_{1}),
\excessrisk_t(f_{2})
\big\}
+
M^2 \log(t/\delta)  \cdot
\min_{\wval\in[t-1]}\bigg\{
\underbrace{\max_{t-\wval\le \tau\le t-1} \TV(\distP_{\tau}, \distP_t)}_{\text{cost of stale data}}
+
\underbrace{\ \frac{1}{\nval_{t,\wval}}\ }_{\text{cost of limited data}}
\bigg\}.
\end{equation}
Here $\lesssim$ hides a universal constant.
\end{theorem}

\begin{proof}[Proof of \myCref{thm-model-comparison}]
See \myCref{sec-thm-model-comparison-proof}.
\end{proof}

To interpret \Cref{thm-model-comparison}, consider an investor who knows, in hindsight, exactly how much historical data to use to choose a good prediction model. This theoretical result states that our comparison procedure performs nearly as well without that knowledge. It identifies on its own whether the current regime is better described by the last few months or the last several years of validation data. The procedure shortens its look-back window when conditions shift and widens it in stable periods, all without any prior model of how markets evolve.

From a learning theory perspective, \myCref{thm-model-comparison} gives a finite-sample oracle inequality \eqref{eqn-model-comparison-fast-squared}. It states that with high probability, the excess risk of $\widehat{f}$ does not exceed that of the better model between $f_1$ and $f_2$, up to an additive overhead term that we call the \emph{cost of adaptivity},
\[
\min_{\wval\in[t-1]}\left\{
\max_{t-\wval\le \tau\le t-1} \TV(\distP_{\tau}, \distP_t)
+
\frac{1}{\nval_{t,\wval}}
\right\}.
\]
This overhead reflects the same two forces as \Cref{thm-tradeoff}: the cost of stale data, $\max_{t-\wval\le \tau\le t-1} \TV(\distP_{\tau}, \distP_t)$, and the cost of limited data, $1/\nval_{t,\wval}$, associated with the $\nval_{t,\wval}$ validation samples used to make the comparison. The bound takes the minimum over all look-back windows $\wval$, meaning that our procedure performs almost as well as an oracle that knows in hindsight which window would lead to the most accurate comparison. This shows that the procedure adaptively chooses a near-optimal look-back window tailored to the local non-stationarity.

Building on this pairwise guarantee, \myCref{thm-select-tournament} below shows that our model selection algorithm $\adaptive$ inherits the same oracle property when selecting from multiple candidate models $\{f_{\modelidx}\}_{\modelidx=1}^{\nmodel}$.

\begin{theorem}[Near-optimal model selection]\label{thm-select-tournament}
Let Assumptions \ref{assumption-independence} and \ref{assumption-bounded} hold. Choose $\delta\in(0,1)$ and take $\delta' = \delta / (3\nmodel^2 t)$ in $\adaptive$. With probability at least $1-\delta$, $\adaptive$ outputs a model $\widehat{f}$ satisfying
\[
\excessrisk_t(\widehat{f})
\lesssim
\min_{\modelidx\in[\nmodel]} \excessrisk_t(f_{\modelidx})
+
M^2 \log(\nmodel t/\delta)  \cdot
\min_{\wval\in[t-1]}\bigg\{
\underbrace{\max_{t-\wval\le \tau\le t-1} \TV(\distP_{\tau}, \distP_t)}_{\text{cost of stale data}}
+
\underbrace{\ \frac{1}{\nval_{t,\wval}} \ }_{\text{cost of limited data}}
\bigg\}.
\]
Here $\lesssim$ hides a universal constant.
\end{theorem}

\begin{proof}[Proof of \myCref{thm-select-tournament}]
See \myCref{sec-thm-select-tournament-proof}.
\end{proof}

\myCref{thm-select-tournament} states that with high probability, the performance of the selected model $\widehat{f}$ is at most the performance of the best model in $\{f_{\modelidx}\}_{\modelidx=1}^{\nmodel}$, up to the same cost-of-adaptivity term and an extra $O(\log\nmodel)$ factor that represents the price of searching simultaneously over all candidate models rather than committing to one in advance. In other words, $\adaptive$ identifies a model whose performance is nearly as good as the best candidate one could have selected in hindsight using the non-stationary validation data. In practice, this resolves the investor's question of whether to use a linear or nonlinear specification this month, and over what history, without any prior knowledge of how market conditions will evolve.

We remark that our model selection framework (\myCref{alg-tournament}) is general and can be combined with any model comparison subroutine $\alg$. In particular, in \myCref{sec-thm-select-tournament-proof}, we prove a general reduction lemma (\myCref{lem-select-tournament-reduction}) that converts any theoretical guarantee of the subroutine $\alg$ to a guarantee of \myCref{alg-tournament}. In \myCref{sec-select-R2}, we further develop a $R^2$-based pairwise comparison subroutine that targets the $R^2$ metric. When equipped with this $R^2$-based subroutine, \myCref{alg-tournament} enjoys a guarantee with respect to the $R^2$ metric.

\begin{remark}[Benefits of pairwise comparison]
As a natural alternative, one may consider estimating the individual performance of the candidate models over a common window, then using the estimated performance for model selection or model aggregation. In a non-stationary environment, estimating the absolute performance of a model can be challenging. In contrast, our procedure makes use of pairwise comparisons and estimates the loss differences between two models. This is statistically advantageous because the two losses are evaluated on the same validation data, and taking their difference may cancel out common fluctuations and non-stationarity. As a consequence, it typically yields a lower-variance estimate of the performance gap than separate estimation of individual model performance. It is also operationally more flexible, since each model pair can be evaluated on its own window rather than forcing all models to share a single window.
\end{remark}

\begin{remark}[Comparison with prior work]
Our model selection framework builds upon the model comparison method of \cite{HHW24}. Below we discuss the main differences between our work and theirs. First, their analysis of the model comparison procedure assumes that the distribution of the covariates $\bx$ remains fixed across time. Our theory removes this assumption entirely, and covers the general non-stationary setting where the joint distribution of $(\bx,y)$ can change arbitrarily. Second, for model selection, they propose a single-elimination procedure which performs $\nmodel-1$ model comparisons, but incurs an additional $O(\log \nmodel)$ factor in the performance bound. In contrast, our approach maintains a linear complexity in $\nmodel$ in expectation while achieving a sharper bound.
\end{remark}

\subsection{Jointly Choosing Model Class and Training Window}

Finally, we apply \myCref{thm-select-tournament} to the joint selection of model class and training sample size.
Let $\metamodelclass$ be a finite collection of model classes, e.g., $\metamodelclass=\{\text{linear model}, \text{random forest of a certain size}\}$. For each model class $\modelclass\in\metamodelclass$, we train models on different windows $\wtrain\in[t-1]$ of the training data $\{\dataset_{\tau}^{\tr}\}_{\tau=1}^{t-1}$. Let $\widehat{h}(\modelclass,\wtrain)$ denote the model from $\modelclass$ trained on $\{\dataset_{\tau}^{\tr}\}_{\tau=t-\wtrain}^{t-1}$. Then, the set of candidate models is given by
\begin{equation}\label{eqn-joint-model-window-candidates}
\{f_{\modelidx}\}_{\modelidx=1}^{\nmodel}
=
\left\{ \widehat{h}(\modelclass,\wtrain) : \modelclass\in\metamodelclass,\, k\in[t-1] \right\}.
\end{equation}
Applying \myCref{thm-select-tournament} to this set of candidate models yields the following guarantee. For simplicity, we assume that training-validation data splitting ratio is fixed across time.

\begin{assumption}[Balanced training-validation split]\label{assumption-data-split}
There exists $c>0$ such that $|\dataset_{\tau}^{\tr}| / |\dataset_{\tau}^{\va}| = c$ for all $\tau\in\ZZ_+$.
\end{assumption}

\begin{theorem}[Near-optimal model-and-data selection]\label{thm-joint-selection}
Let Assumptions \ref{assumption-independence}, \ref{assumption-bounded} and \ref{assumption-data-split} hold. Suppose the set of candidate models is given by \eqref{eqn-joint-model-window-candidates}. Choose $\delta\in(0,1)$ and take $\delta' = \delta / (6|\metamodelclass|^2t^3)$ in $\adaptive$. Then, with probability at least $1-\delta$, the output $\widehat{f}$ of $\adaptive$ satisfies
\begin{equation}\label{eqn-oracle-joint-selection}
\excessrisk_t(\widehat{f})
\lesssim
\min_{\modelclass\in\metamodelclass,\,k\in[t-1]} \bigg\{ \underbrace{\min_{f\in\modelclass} \excessrisk_t(f)}_{\text{misspecification}} + \underbrace{\left( r_{t,k}(\modelclass) + \frac{1}{\ntrain_{t,k}} \right)}_{\text{cost of limited data}} + \underbrace{\max_{t-k\le \tau\le t-1} \TV(\distP_{\tau},\distP_t)}_{\text{cost of stale data}} \bigg\},
\end{equation}
where $\lesssim$ hides the constants $M$ and $c$ and logarithmic factors of $t$, $\delta^{-1}$ and $|\metamodelclass|$.
\end{theorem}

\begin{proof}[Proof of \myCref{thm-joint-selection}]
See \myCref{sec-thm-joint-selection-proof}.
\end{proof}

The bound on the right-hand side of \eqref{eqn-oracle-joint-selection} is exactly the model performance bound of the model $h(\modelclass,k)$ in \myCref{thm-tradeoff}: the same misspecification, cost of limited data, and cost of stale data that govern any single model's prediction error also govern the performance of the model that $\adaptive$ selects. Thus, our algorithm selects a near-optimal pair of model class and training window size, up to logarithmic factors, without ever being told which of the two costs dominates at a given point in time.

\section{Empirical Evidence: Industry Portfolios Across Economic Regimes}\label{sec-experiments}

In this section, we evaluate the predictive performance of our method on industry portfolios. The results below are the empirical counterpart to the near-optimality guarantee established in \myCref{sec-select}: without advance knowledge of when the economic environment will shift, $\adaptive$ should track the performance of the best fixed-window choice in hindsight, a pattern we now test directly. First, we describe our comprehensive dataset of firm-characteristic managed portfolios and industry portfolios. Then, we report our findings that our adaptive method $\adaptive$ achieves superior out-of-sample performance in predicting expected returns compared with fixed-window and regime-switching benchmarks across different economic regimes.

\subsection{Data}\label{subsection:data}
We examine the pricing of 17 industry portfolio returns from Kenneth French's data library, covering the period from September 1987 to November 2016. Our predictor set combines macroeconomic factors, risk premia from characteristic-sorted portfolios, and lagged returns to capture the complex dynamics driving industry returns, sourced for widely cited public datasets.\footnote{More specifically, our data combines daily and monthly sources to construct a comprehensive time series of covariates combining macroeconomic and cross-sectional signals. The final sample spans the time period from September 1987 to November 2016. We merge daily CRSP excess returns with monthly characteristics from \citet{GKX20}, which provides 94 standardized characteristics for U.S. equities encompassing valuation ratios, profitability measures, investment activity, liquidity, and past return dynamics. These characteristics have become the canonical set of firm-level predictors in modern empirical asset pricing. We construct daily long-short portfolios by sorting firms into deciles based on each characteristic and taking the difference between the top and bottom decile returns, following the methodology of \citet{GKX20}.} We provide full details of the dataset construction in \myCref{sec-dataapdx}, including data preprocessing and long-short portfolio constructions.

\paragraph{Macroeconomic and Systematic Factors.} We incorporate 15 factors from \citet{CPZ24}, who estimate a SDF using deep learning while imposing no-arbitrage restrictions. These factors include: (i) the estimated SDF representing the aggregate price of risk; (ii) ten beta-sorted decile portfolios based on firms' SDF exposure; and (iii) four macroeconomic hidden states extracted from 178 macro time series via a generative adversarial network. These monthly observations are assigned to all trading days within each month. We also include the daily Fama-French three factors (market, size, and value) from \citet{FFr93} as benchmark risk factors.

\paragraph{Characteristic-Sorted Portfolios.} Following \citet{GKX20}, we construct 94 long-short portfolios sorted on firm characteristics that capture price trends, liquidity, size, and risk measures. For each characteristic, we form decile portfolios using all CRSP-listed stocks and create a long-short strategy that buys the top decile and shorts the bottom decile. This approach transforms firm-level characteristics into interpretable factor returns that isolate the pricing implications of each characteristic.

\paragraph{Predictor Set.} Our final predictor set comprises: (i) $15$ macroeconomic factors from \citet{CPZ24}; (ii) $3$ Fama-French factors of \citet{FFr93}; (iii) $94$ characteristic-sorted long-short portfolio returns of \citet{GKX20} described above; and (iv) the $17$ lagged industry returns. This comprehensive set of $129$ predictors combines traditional risk factors with modern high-dimensional representations, allowing us to test whether our adaptive method can effectively navigate the complex, time-varying relationships between these predictors and industry returns.

\subsection{Estimation Framework and Benchmarks}\label{sec-experiment-setup}

We evaluate our adaptive method using candidate models from different specifications with varying parameters and training windows. We take one month as a period, where the data $\dataset_t=\{(\bx_s,y_s)\}_{s\in\batch_t}$ in month $t$ consists of daily covariate-return pairs in that month.

\paragraph{Model Specifications.} We consider the following specifications that approximate stochastic discount factors. For a vector of covariates $\bx\in\RR^d$, we write $\widetilde{\bx}=(\bx^\top,1)^\top\in\RR^{d+1}$.
\begin{enumerate}
\item Non-linear specification using random forests ($\rf$). Given training data $\{(\bx_i,y_i)\}_{i=1}^n$, and two parameters, namely the number of trees $n_{\texttt{tree}}$ and the maximum tree depth $d_{\max}$, $\rf$ estimates a random forest model.
\item Linear specification estimated with ridge regularization ($\ridge$). Given training data $\{(\bx_i,y_i)\}_{i=1}^n$ and regularization parameter $\alpha>0$, $\ridge$ estimates a linear model $f(\bx)= \inner{\widehat{\btheta}}{\widetilde{\bx}}$ by
\[
\widehat{\btheta} = \argmin_{\btheta\in\RR^{d+1}} \left\{ \frac{1}{n} \sum_{i=1}^{n} \left( \inner{\btheta}{\widetilde{\bx}_{i}} - y_{i} \right)^2 + \alpha \| \btheta \|_2^2 \right\}.
\]
\item Linear specification with LASSO regularization ($\lasso$). Given training data $\{(\bx_i,y_i)\}_{i=1}^n$ and regularization parameter $\alpha>0$, $\lasso$ estimates a linear model $f(\bx)= \inner{\widehat{\btheta}}{\widetilde{\bx}}$ by
\[
\widehat{\btheta} = \argmin_{\btheta\in\RR^{d+1}} \left\{ \frac{1}{2} \cdot \frac{1}{n} \sum_{i=1}^{n} \left( \inner{\btheta}{\widetilde{\bx}_{i}} - y_{i} \right)^2 + \alpha \| \btheta \|_1 \right\}.
\]
\item Linear specification with elastic net regularization ($\enet$). Given training data $\{(\bx_i,y_i)\}_{i=1}^n$ and regularization parameters $\alpha>0$ and $r\in(0,1)$, $\enet$ estimates a linear model $f(\bx)= \inner{\widehat{\btheta}}{\widetilde{\bx}}$ by
\[
\widehat{\btheta} = \argmin_{\btheta\in\RR^{d+1}} \left\{ \frac{1}{2} \cdot \frac{1}{n} \sum_{i=1}^{n} \left( \inner{\btheta}{\widetilde{\bx}_{i}} - y_{i} \right)^2 + \alpha r \| \btheta \|_1 + \frac{\alpha}{2}(1-r) \| \btheta \|_2^2 \right\}.
\]
\end{enumerate}
For each specification, we consider a grid of hyperparameter values, detailed in \myCref{sec-hyperparameters}. In each month $t$, we fit candidate models from these specifications on training windows of $4^k\wedge (t-1)$ months, where $0\le k\le 5$; more details are provided in \Cref{sec-window-details}. This creates a factorial candidate set formed by crossing model specifications, their hyperparameters, and training window lengths.

\begin{table}[t!]
\centering
\caption{Summary of methods\label{tab-fh-benchmarks}}
\resizebox{\textwidth}{!}{%
\begin{tabular}{@{}llccp{6cm}@{}}
\toprule
Benchmark & Notation & $\wval$ (months) & Approx.\ duration & Economic characterization \\
\midrule
Short-horizon  & $\fwshort$   & $32$         & $\approx 3$ yrs
  & Captures only the most recent regime; low non-stationarity cost, high estimation variance. \\[4pt]
Medium-horizon & $\fwmed$     & $128$        & $\approx 1$ business cycle
  & Spans a full expansion-recession cycle. \\[4pt]
Long-horizon   & $\fwlong$    & $\leq 512$   & All available history
  & Uses the full in-sample record; implicitly assumes stationarity across all historical regimes. \\[4pt]
Cross-validation & $\fixedwindowCV$ & $36$ (CV) & $\approx 3$ yrs
  & Selects models via 5-fold cross-validation over the most recent 36 months. \\[4pt]
Regime-switching & $\RS$ & N/A & N/A
  & Fits a 2-state Markov-switching linear model \citep{Ham89}; explicitly models non-stationarity but restricts each regime to a fixed linear specification. \\[4pt]
\midrule
Adaptive & $\adaptive$ & Data-driven & Time-varying
  & Selects window $\wval$ to balance non-stationarity and variance, without prior knowledge of how market conditions evolve. \\
\bottomrule
\end{tabular}}
\bnotetab{This table summarizes the six methods considered in \Cref{sec-experiments}.
$\fh$ stands for Fixed Horizon; the suffix indicates the validation window length.
The three $\fh$ benchmarks are the fixed-window counterparts of $\adaptive$
(\Cref{alg-fixed}) with window sizes $\wval = 32$, $128$, and up to $512$ months.
$\fixedwindowCV$ uses 5-fold cross-validation over the most recent 36 months.
$\RS$ is a 2-state Markov regime-switching linear model re-estimated monthly
using all available history.
$\adaptive$ (\Cref{alg-tournament}) jointly selects the model class
and validation window at each month $t$ without prior knowledge of the
non-stationarity structure.}
\end{table}

\paragraph{Benchmark Approaches.} To verify the adaptivity of our framework, we compare it with the following non-adaptive benchmarks that use a fixed window for training and/or validation. We organize the notations in Table \ref{tab-fh-benchmarks} for quick reference as well.
\begin{enumerate}
\item Fixed-horizon validation window for model selection ($\fwshort$, $\fwmed$, and $\fwlong$). This is the non-adaptive fixed-window counterpart of $\adaptive$. In each month $t$, we train the candidate models above, and then use validation data from the last $\wval$ periods $\{\dataset_{\tau}^{\va}\}_{\tau=t-\ell}^{t-1}$ to perform model selection. The detailed description for each of the fixed window length is given in \myCref{alg-fixed}. We consider three choices of the validation window $\wval$, which leads to three benchmarks as $\fwshort$, $\fwmed$, and $\fwlong$, where $\fh$ stands for Fixed-Horizon and the suffix indicates the validation window length. $\fwshort$ uses $\wval=32$ months of validation data (approximately 3 years), capturing only the recent regime. $\fwmed$ uses $128$ months (approximately one full business cycle), and $\fwlong$ uses all available history. The contrast across these three captures how much history a fixed-horizon practitioner would choose to use. $\adaptive$ is designed to adapt to the optimal choice of validation window size $\wval$ in hindsight at each time.

\item Fixed-horizon cross-validation ($\fixedwindowCV$). In each month $t$, we use data from the last $36$ months $\{\dataset_{\tau}\}_{\tau=t-36}^{t-1}$ to perform $5$-fold cross-validation to estimate and select a model out of the candidate models with the same sets of parameters.

\item Regime-switching model ($\RS$). We also fit a regime-switching model \citep{Ham89}, which explicitly models the non-stationary dynamics. 
We consider a $2$-state model, each day $s$ is associated with an underlying state $z_s\in\{1,2\}$, which evolves according to a time-homogeneous Markov chain. Conditioned on $z_s=z$, the covariates $\bx_s$ and response $y_s$ follow a linear model $y_s = \alpha_z + \bbeta_z^\top\bx_s + \varepsilon_s$, where $\varepsilon_s\sim N(0,\sigma_z^2)$. We re-estimate the model at a monthly frequency, using all available historical data. 
\end{enumerate}

For all the methods considered in Table \ref{tab-fh-benchmarks}, we use the same $129$-dimensional covariates set of macroeconomic factors, Fama-French factors, characteristic-sorted long-short portfolio returns and lagged industry returns, described in \Cref{subsection:data}.

As the methods $\adaptive$, $\fwshort$, $\fwmed$, $\fwlong$ and $\fixedwindowCV$ involve training-validation data splitting, we will evaluate each of them using $20$ independent repetitions of a stratified $80\%$-$20\%$ estimation-validation split. Specifically, fixing a random seed, we construct stratified samples in which, for each month $t$, $80\%$ of trading days are assigned to the training set and the remaining $20\%$ to the validation set. The validation observations are then pooled and ordered chronologically, and the method is applied to this validation set to select a model. Using the selected model, we compute a realized performance metric (e.g., out-of-sample $R^2$). We repeat this procedure $20$ times and report the average performance metric across repetitions. This evaluation design disciplines model selection and reduces the scope for data mining, while respecting the sequential arrival of information faced by investors and preserving the time-series structure of the data, thereby avoiding look-ahead bias. More details can be found in \myCref{sec-hyperparameters}.

\begin{algorithm}[h]
	\begin{algorithmic}
	\STATE {\bf Input:} Candidate models $\{f_{\modelidx}\}_{\modelidx=1}^{\nmodel}$, validation data $\{ \dataset^{\va}_{\tau} \}_{\tau=1}^{t-1}$, validation window size $\wval$.
	\STATE Select
	\[
	\widehat{\modelidx} = 
	\argmin_{\modelidx\in[\nmodel]}  \sum_{\tau=(t-\wval)\vee 1}^{t-1} \sum_{s\in\batchva_{\tau}} \left[ f_{\modelidx}(\bx_s) - y_s \right]^2.
	\]
	\RETURN $\widehat{f} = f_{\widehat{\modelidx}}$.
	\caption{Fixed Validation Window for Model Selection}
	\label{alg-fixed}
	\end{algorithmic}
\end{algorithm}
 
\paragraph{Performance Metrics.} We measure the performance of each approach using the out-of-sample $R^2$ metric \eqref{eqn-R2-zero-cumulative} that benchmarks against a zero forecast. We compute both the overall out-of-sample $R^2$ from January 1990 to November 2016, and the annual out-of-sample $R^2$. The latter provides a more granular understanding of the approaches' performance over time. In \myCref{sec-experiments-standard-R2}, we also report results for the standard $R^2$ metric \eqref{eqn-R2-cumulative}.

\subsection{Return Predictability Across Economic Regimes}
We next turn to the empirical analysis of pricing the 17 industry portfolios. The fundamental premise of our adaptive framework is that asset pricing relationships exhibit time-varying dynamics rather than remaining stationary across economic conditions. This non-stationarity is particularly pronounced during economic recessions, when structural breaks in risk premia, sudden shifts in investor risk aversion, and disruptions to market liquidity mechanisms create environments where long-term historical pricing relationships provide poor guidance for future returns. Recessions therefore serve as a natural laboratory for testing the adaptivity of our framework: if our approach can successfully navigate these turbulent periods when non-stationarity is most severe, it provides compelling evidence for the value of adaptive model selection in asset pricing more generally.

\paragraph{Performance During Recessions.}
Our adaptive method $\adaptive$ improves return predictability across the full 1990--2016 out-of-sample period, with significant gains during the three NBER-dated recessions. \myCref{tab:oos_r2_industry_time} presents out-of-sample $R^2$ values across distinct economic regimes. $\adaptive$ achieves an out-of-sample $R^2$ of 0.049 across the full sample period, representing a 14.0\% improvement over the best fixed-window benchmark $\fwlong$ which has $R^2 = 0.043$. The regime-switching model performs poorly in comparison, achieving a negative $R^2$ of $-0.026$.

\begin{table}[t!]
    \centering
    \caption{Out-of-Sample $R^2$ Averages Across Industries by Time Period}
    \label{tab:oos_r2_industry_time}
    \begin{tabular}{@{}lcccccc@{}}  
        \toprule
        \multirow{2}{*}{Method} & \multirow{2}{*}{Full OOS Period} & \multicolumn{3}{c}{Recessions}  \\
        \cmidrule{3-5}
        & & Gulf War & 2001 Recession & Financial Crisis  \\
        \midrule
        $\adaptive$               & $0.049$  & $0.027$  & $0.125$  & $0.041$    \\
        $\fwshort$ & $0.022$  & $0.009$ & $0.096$  & $-0.001$  \\
        $\fwmed$ & $0.041$  & $-0.031$ & $0.123$  & $0.035$  \\
        $\fwlong$ & $0.043$  & $-0.031$ & $0.117$  & $0.039$  \\
        $\fixedwindowCV$       &  $0.035$ &  $-0.009$ & $0.071$ & $0.014$   \\
        $\RS$ &  $-0.026$ & $-0.441$ & $0.081$ & $-0.008$ \\
        \bottomrule
    \end{tabular}\bnotetab{
    This table reports out-of-sample (OOS) $R^2$ averages for return prediction models across all 17 industry portfolios. Full OOS Period refers to our largest available OOS period covering 01/1990$\sim$11/2016. Columns report OOS $R^2$ averages across all industries and highlight this metric during three recessions, as documented in \href{https://www.nber.org/research/business-cycle-dating}{NBER Business Cycle Dating}: 
    \begin{itemize}
        \item the 1990 Gulf War recession (06/1990$\sim$10/1990);
        \item the 2001 Recession of dot-com bubble burst and the 9/11 attack (05/2001$\sim$10/2001);
        \item the Financial Crisis led by defaults of subprime mortgages (11/2007$\sim$06/2009). 
    \end{itemize}
    That is, the OOS performance in Gulf War column focuses on model performance comparisons exclusively in the out-of-sample period of 06/1990$\sim$10/1990. All values are calculated using monthly return data. The $\fh$ benchmarked methods use fixed windows as described in  Table \ref{tab-fh-benchmarks}.}
\end{table}

The 2001 recession consists of two distinct shocks. The dot-com collapse repriced technology-sector earnings, shifting growth-versus-value factor loadings rapidly and persistently. The September 11 attacks then introduced a macro shock with no historical analog. Data in the 1990s reflected an era of technology optimism and subdued risk aversion; neither property held after March 2000. Our method's validation window contracted to focus on recent data, recognizing that the bull-market decade was uninformative for the new regime. During the 2001 recession, $\adaptive$ achieves an impressive $R^2$ of 0.125, outperforming the long window $\fwlong$ by 6.8\% (0.117) and substantially exceeding the fixed window $\fixedwindowCV$'s 0.071 and regime-switching $\RS$'s $0.081$.

The 1990 Gulf War recession is the starkest finding of the benefits of adaptivity. Iraq's invasion of Kuwait in August 1990 triggered an oil supply shock with no precedent in the post-1987 sample observations. \citet{JonKau96} show that oil shocks transmit to U.S.\ equity returns through real cash flow effects. Data during 1987--1990 reflected post-crash recovery dynamics and had no exposure to oil-shock co-movements. In terms of Theorem~\ref{thm-tradeoff}, $\TV(\distP_{\tau},\distP_t)$ spiked sharply for times $\tau$ before August 1990, and our method's short validation window appropriately discounted that stale history. While $\adaptive$ maintains a positive $R^2$ of $0.027$ during this sharp but brief contraction, the fixed-window benchmark $\fwlong$ produces a negative $R^2$ of $-0.031$, and the regime switching model yields a negative $R^2$ of $-0.441$, indicating worse performance than a simple forecast of zero.\footnote{One possible reason for the significant underperformance of the regime-switching model during the Gulf War is the limited initial training samples. This is a common challenge in latent-state models, which require the unobserved latent states, transition probabilities, and state-dependent parameters to be inferred jointly from the data. With a large $129$-dimensional covariates set, our in-sample period, which starts from September 1987, is too short to produce stable and reliable parameter estimates for the out-of-sample period beginning in June 1990.}

The Global Financial Crisis showed a different non-stationarity structure. Systemic credit risk drove cross-industry return correlations toward 1, eliminating the return dispersion that prediction models depend on. For instance, liquidity risk reached historically extreme levels as market stresses occur and empirically cause the liquidity risk premium to drive significant commonality in stock returns \citep{PasSta03}. Leverage and funding costs also shifted to values outside the training distribution \citep{AdrShi14}. We found that during the Global Financial Crisis, the adaptive method $\adaptive$ achieves an $R^2$ of 0.041, marginally outperforming the long window $\fwlong$ (0.039) and substantially exceeding cross-validated $\fixedwindowCV$ (0.014) and regime switching $\RS$ ($-0.008$). The smaller performance gap here reflects the recession's 20-month duration. By the time $\adaptive$ is evaluated in late 2008 and 2009, even fixed-window methods have accumulated enough crisis-period observations to reflect the new economic conditions. The long time horizon of the financial crisis out-of-sample period diminishes the relative advantage of our method, as opposed to other shorter period shocks. We consider it as not a failure of our method's adaptivity but rather a natural consequence of a long-lived regime. Nevertheless, the adaptive $\adaptive$ maintains its competitiveness throughout this long period, exhibiting robustness across different types of economic contractions.

\paragraph{Why Adaptivity Matters During Downturns.}
The superior recession performance of $\adaptive$ has important implications for asset pricing theory and practice. Traditional asset pricing models assume stationary risk-return relationships, an assumption that becomes particularly problematic during economic downturns when risk aversion typically increases and market liquidity conditions deteriorate. Our adaptive framework explicitly recognizes this non-stationarity by allowing the validation window to expand or contract based on recent predictive performance.

The empirical evidence suggests that during recessions, the optimal window for model selection varies significantly, reflecting the rapid evolution of risk premia. The 1990 Gulf War recession provides the clearest example: its sudden onset and brief duration created an environment where only very recent validation data could accurately capture the new pricing dynamics. Conversely, during more prolonged downturns like the Global Financial Crisis, the optimal window likely expanded gradually as new market conditions became established.

From a theoretical perspective, these findings support the view that stochastic discount factors exhibit time-varying dynamics that are particularly pronounced during economic stress. The adaptive framework's ability to track these dynamics more effectively than fixed-window approaches suggests that the non-stationarity of asset pricing relationships is not merely a statistical artifact but reflects fundamental economic mechanisms that vary with the business cycle.

\paragraph{Why Regime-Switching Benchmark Appears Worse.}
The relatively weak performance of the regime-switching model is multi-fold. Firstly, we investigate whether our fitted  Markovian two-state regression captures economically meaningful states. In \Cref{sec-RS-addition-results}, we show that the estimated latent states are significantly correlated with the CBOE Volatility Index (VIX), suggesting that the model does capture economically meaningful variations in market conditions that increases as the ``fear index'' moves upwards. We believe its performance is not simply an artifact of using only two latent states, nor does it reflect a weakened implementation.\footnote{To the best of our knowledge, we use the most established publicly available implementation for Markov regime-switching regression, with details discussed in \Cref{sec-RS-addition-results}.} The limitation is instead the restricted form of conditional linearity: after conditioning on the latent state, the mapping from covariates to return is a fixed linear specification. This restriction arises largely from the need to maintain a tractable likelihood when the transition dynamics of the latent states are modeled explicitly. Our method $\adaptive$ avoids this rigid restriction by adaptively selecting over various model classes and training horizons, which allows it to flexibly handle both time-varying economic regimes and model specifications.

\begin{figure}[h]
\centering
\caption{Box Plot of Out-of-Sample $R^2$ of $\adaptive$ and Benchmarks for $17$ Industry Portfolios.\label{fig-boxplot}}
\includegraphics[scale=0.45]{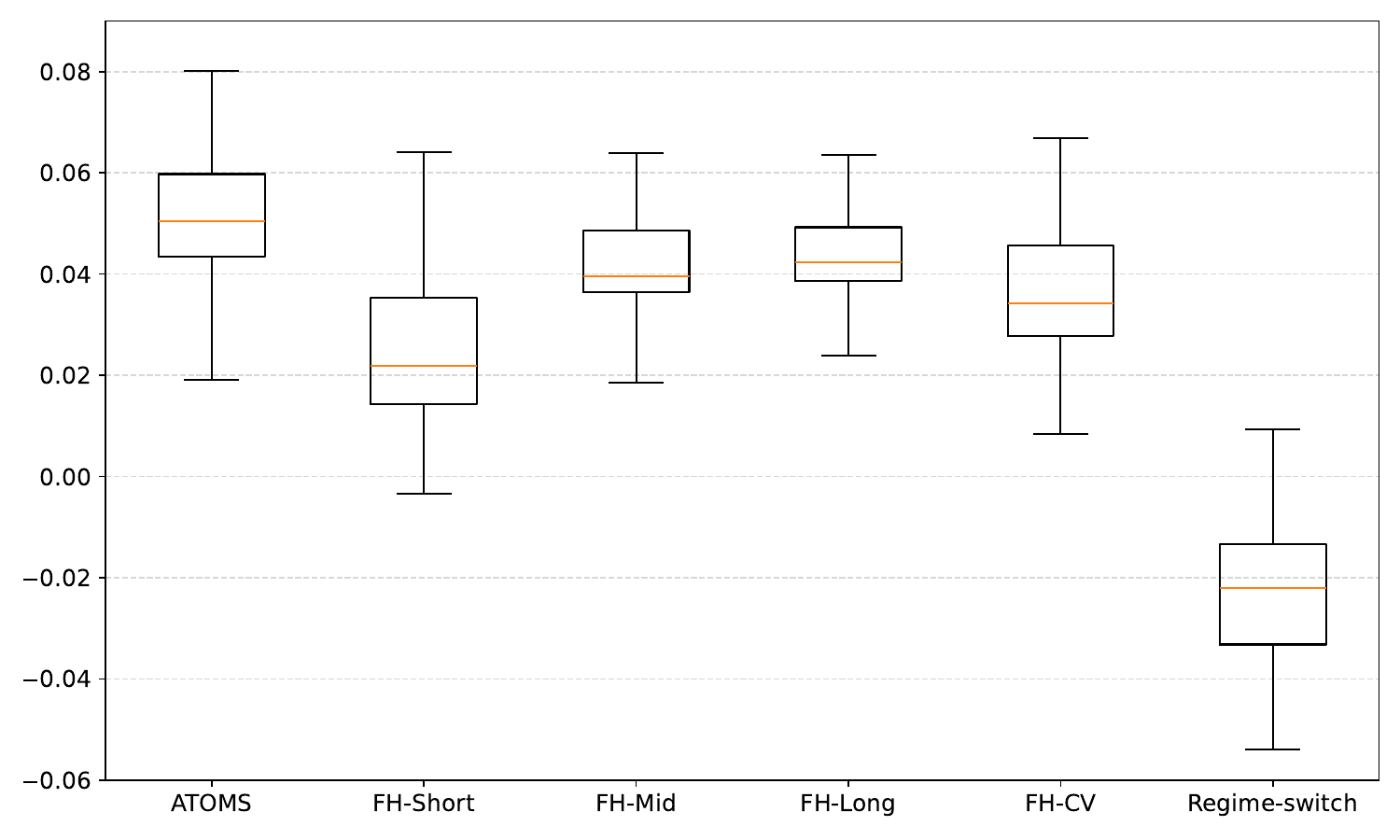}
\bnotefig{This figure describes the distribution of each method's OOS $R^2$. Each box represents one method and is constructed from $17$ industry-level $R^2$ across all years in the OOS horizon. The $\fh$ benchmarked methods use fixed windows as described in Table \ref{tab-fh-benchmarks}.}
\end{figure}

\paragraph{Robustness Across Industries.}
We conduct industry-by-industry robustness check to confirm that the recession outperformance of $\adaptive$ is not driven by a subset of industries but represents a broad-based phenomenon. First, Figure \ref{fig-boxplot} shows a box plot of performance across $17$ industries. Each box represents one method and is constructed from $17$ industry-level $R^2$ across all years in the OOS horizon. We observe that $\adaptive$ overall has better $R^2$ across all years, as its median, and level position of the box is higher than those of the other methods.

\begin{figure}[h!]
	\centering
	\caption{Annual Out-of-Sample $R^2$ of $\adaptive$ and Benchmarks for $17$ Industry Portfolios. \label{fig-industry-yearly}}

    \begin{subfigure}{0.24\textwidth}
    	\centering
        \includegraphics[width=\linewidth]{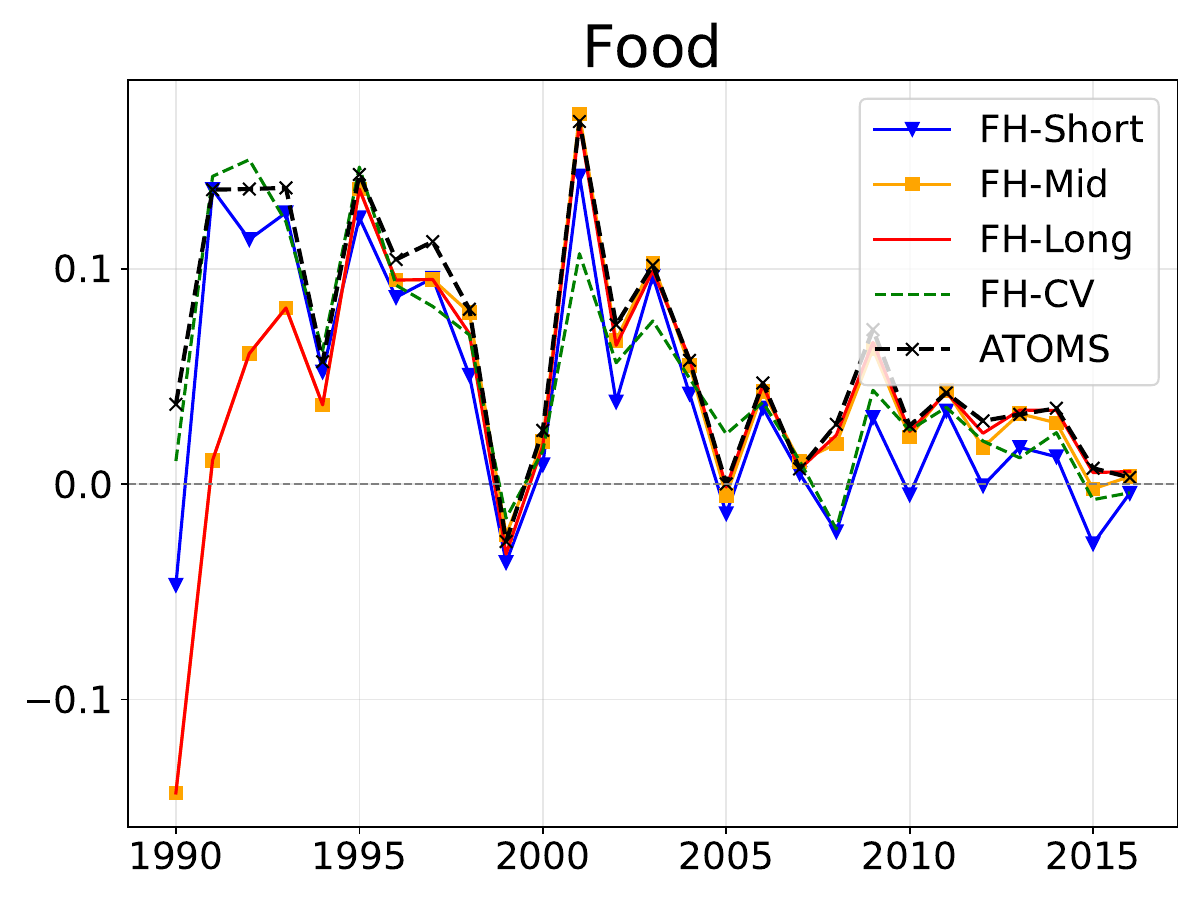}
	\end{subfigure}
    \begin{subfigure}{0.24\textwidth}
        \centering
        \includegraphics[width=\linewidth]{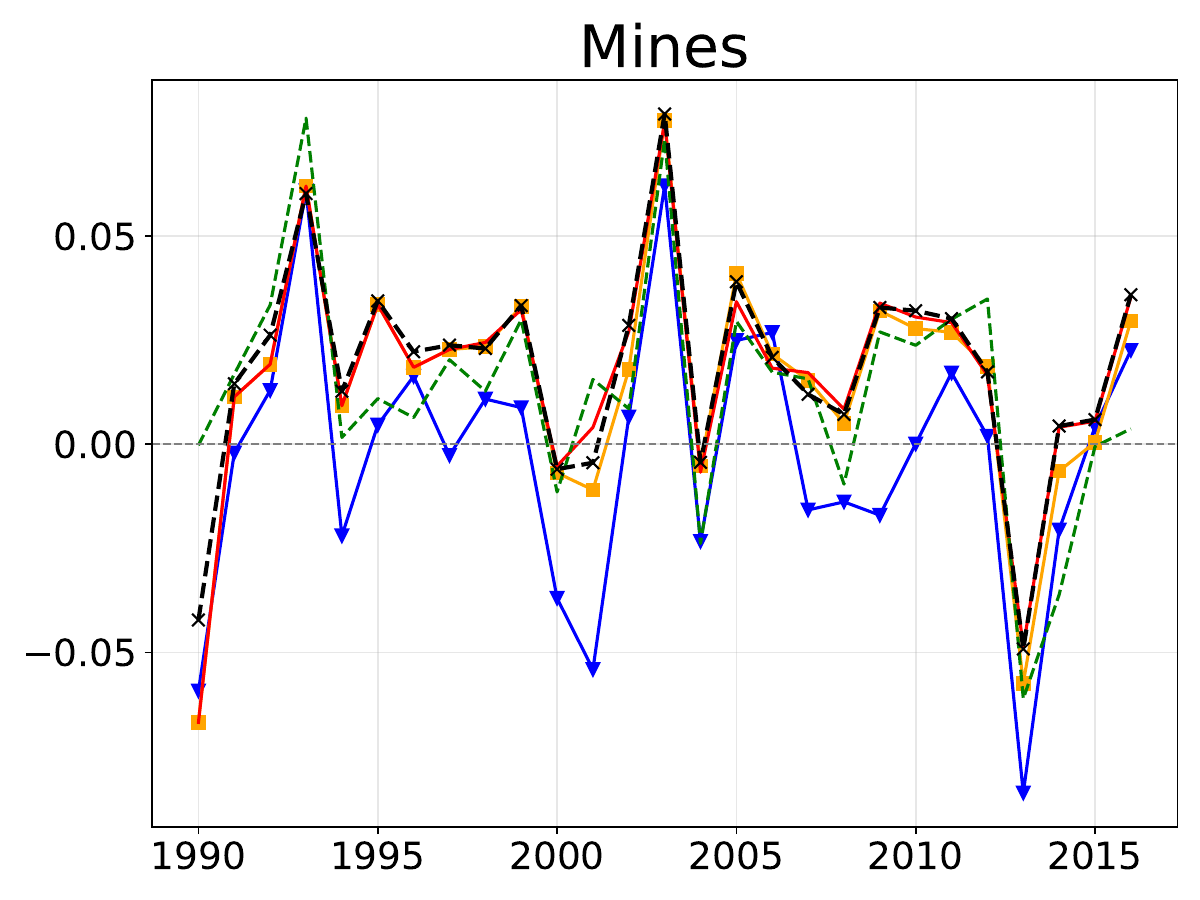}
	\end{subfigure}
    \begin{subfigure}{0.24\textwidth}
        \centering
        \includegraphics[width=\linewidth]{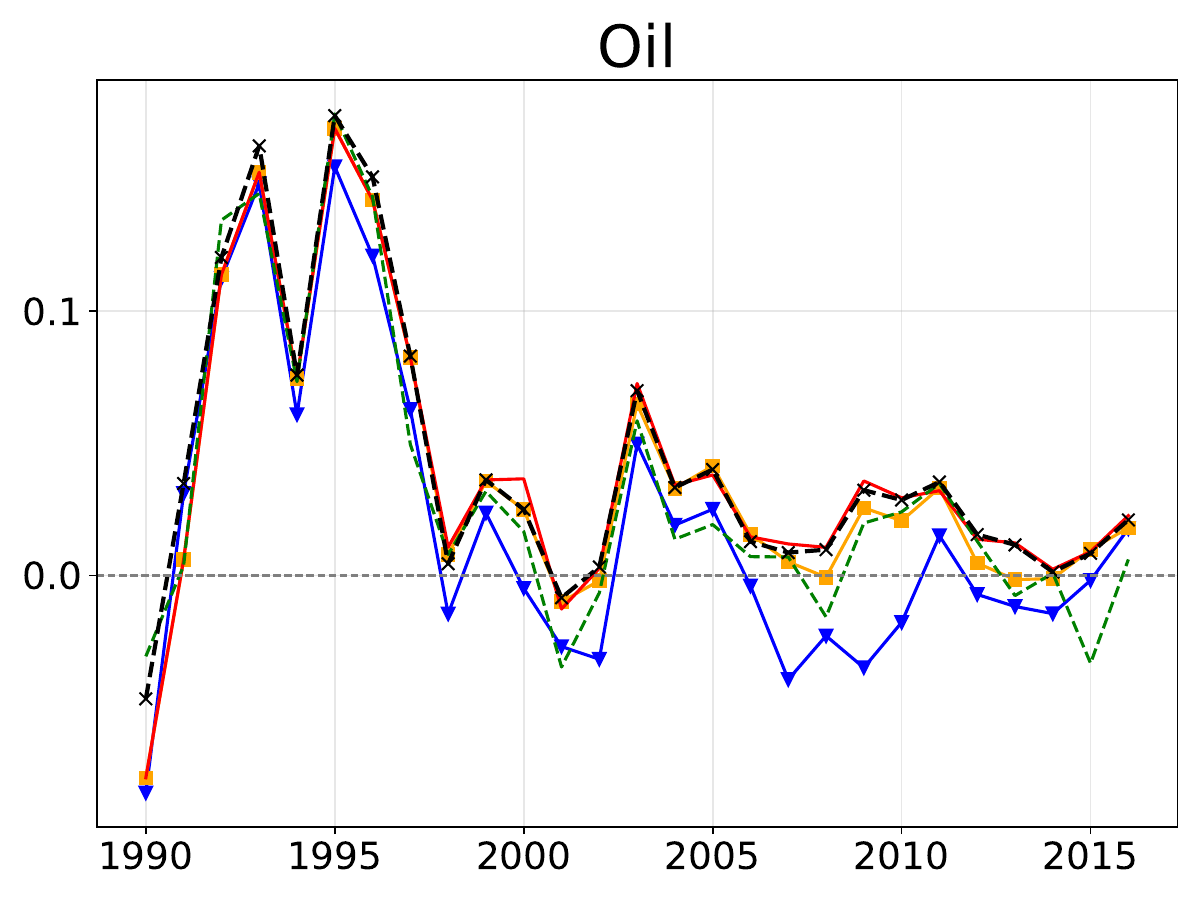}
	\end{subfigure}
    \begin{subfigure}{0.24\textwidth}
    	\centering
        \includegraphics[width=\linewidth]{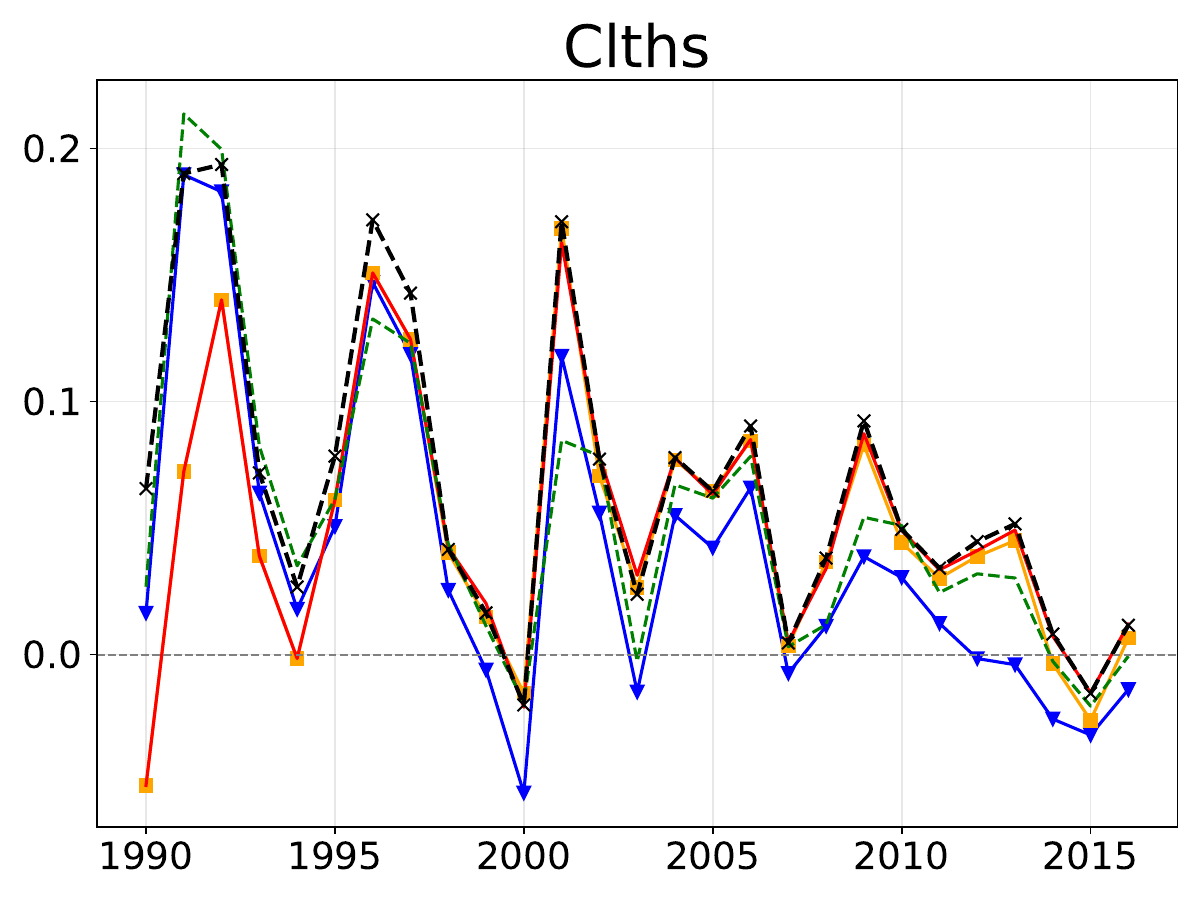}
	\end{subfigure}

    \begin{subfigure}{0.24\textwidth}
        \centering
        \includegraphics[width=\linewidth]{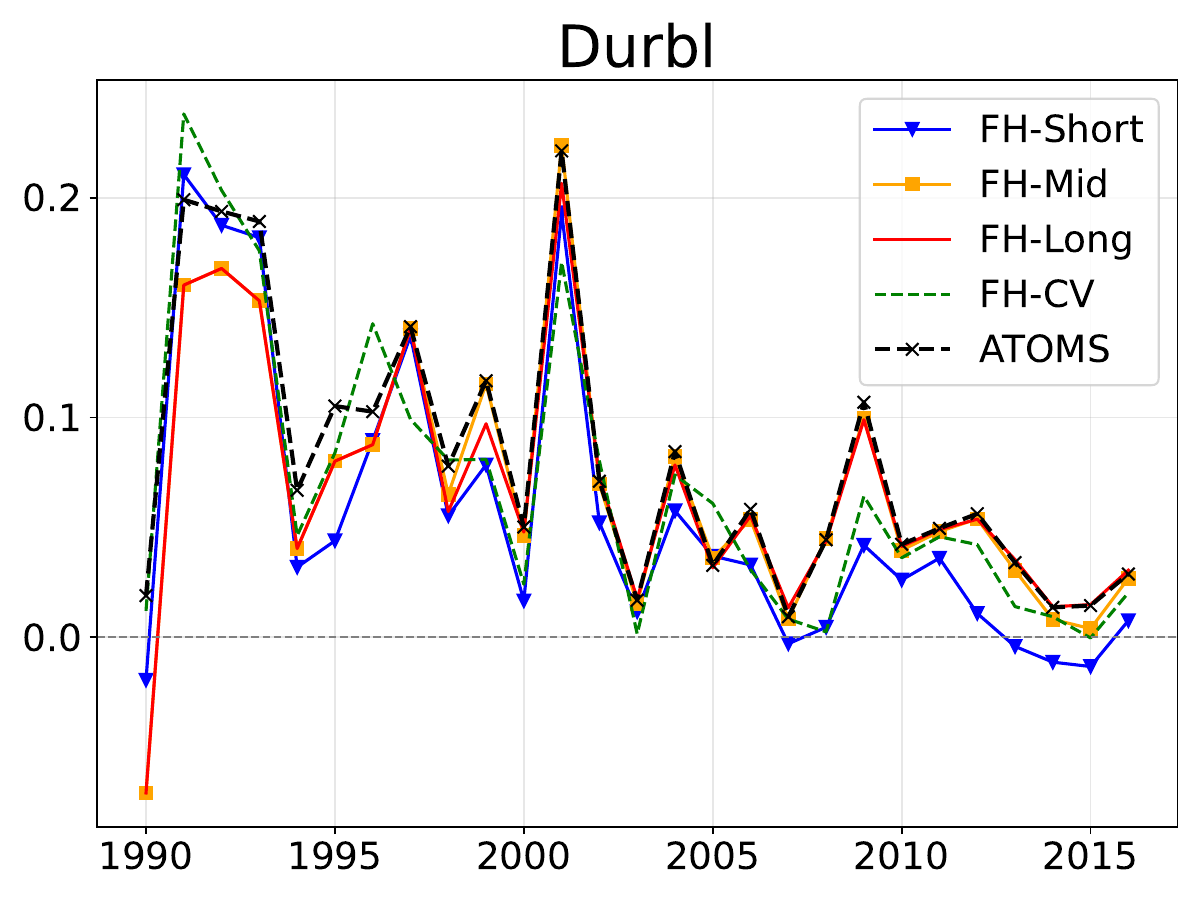}
	\end{subfigure}
    \begin{subfigure}{0.24\textwidth}
        \centering
        \includegraphics[width=\linewidth]{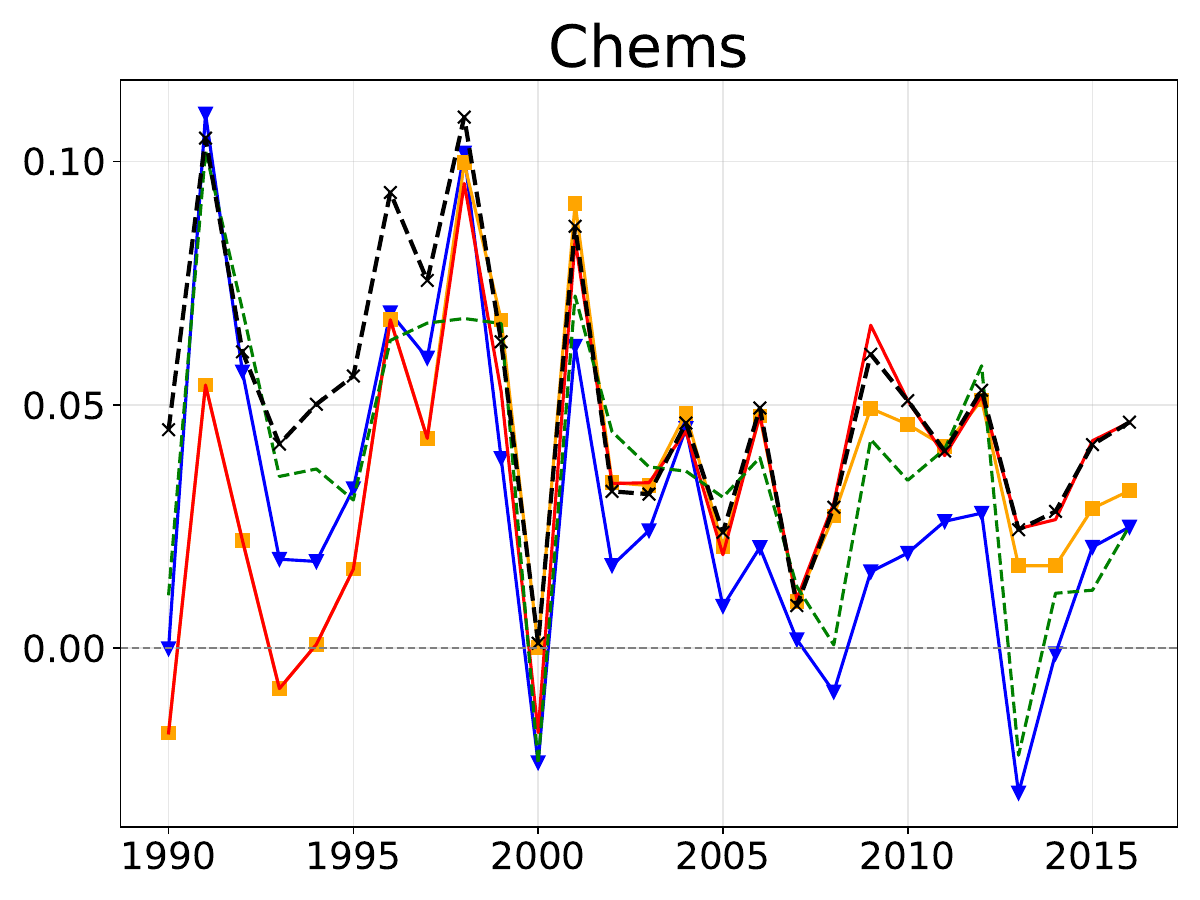}
	\end{subfigure}
    \begin{subfigure}{0.24\textwidth}
    	\centering
        \includegraphics[width=\linewidth]{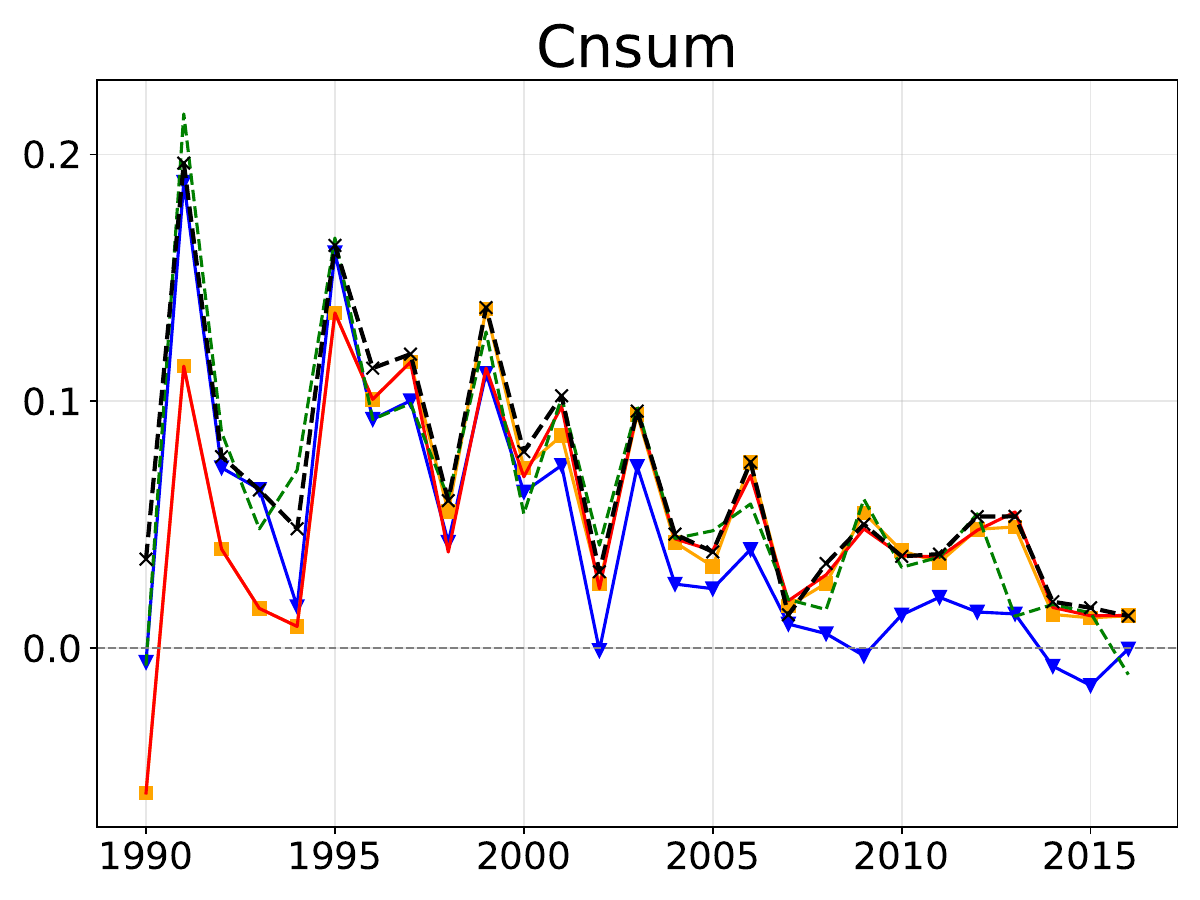}
	\end{subfigure}
    \begin{subfigure}{0.24\textwidth}
        \centering
        \includegraphics[width=\linewidth]{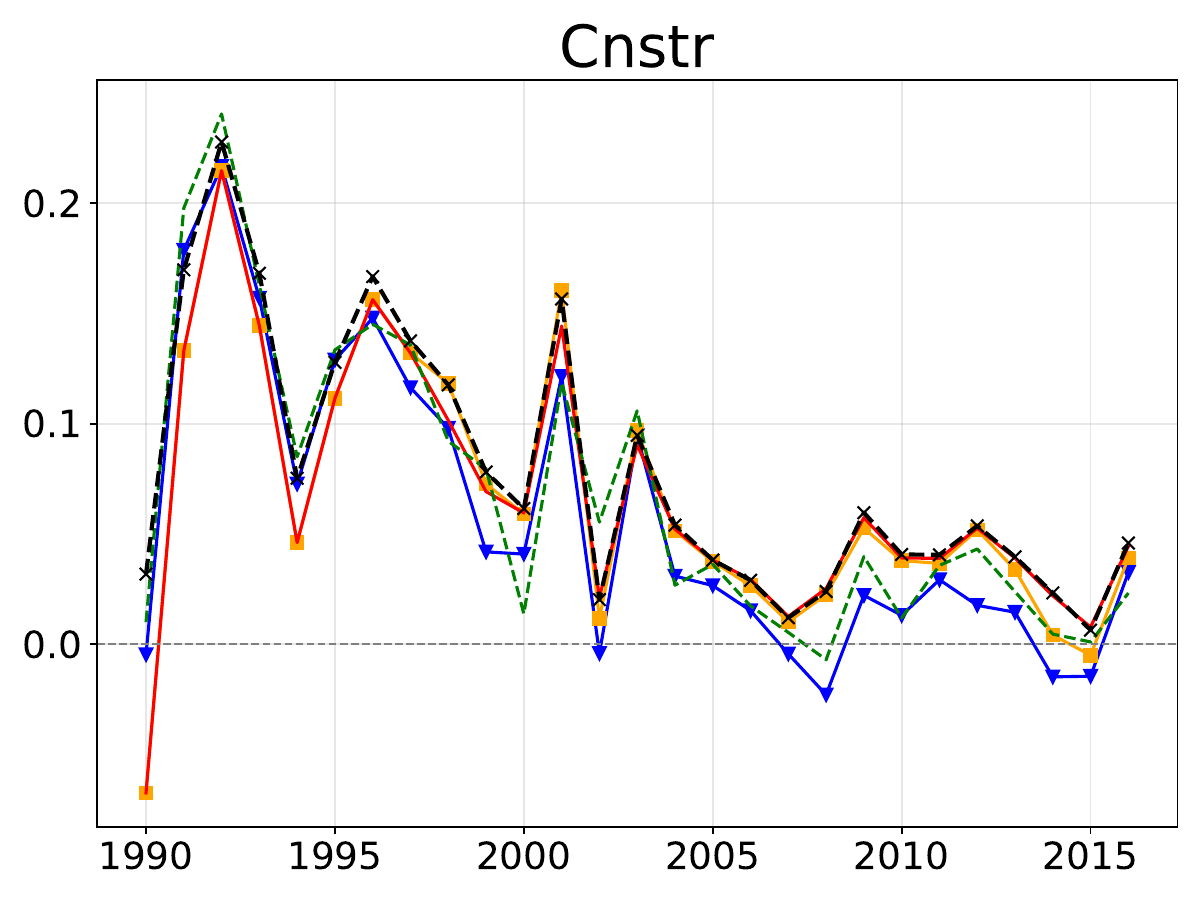}
	\end{subfigure}

    \begin{subfigure}{0.24\textwidth}
        \centering
        \includegraphics[width=\linewidth]{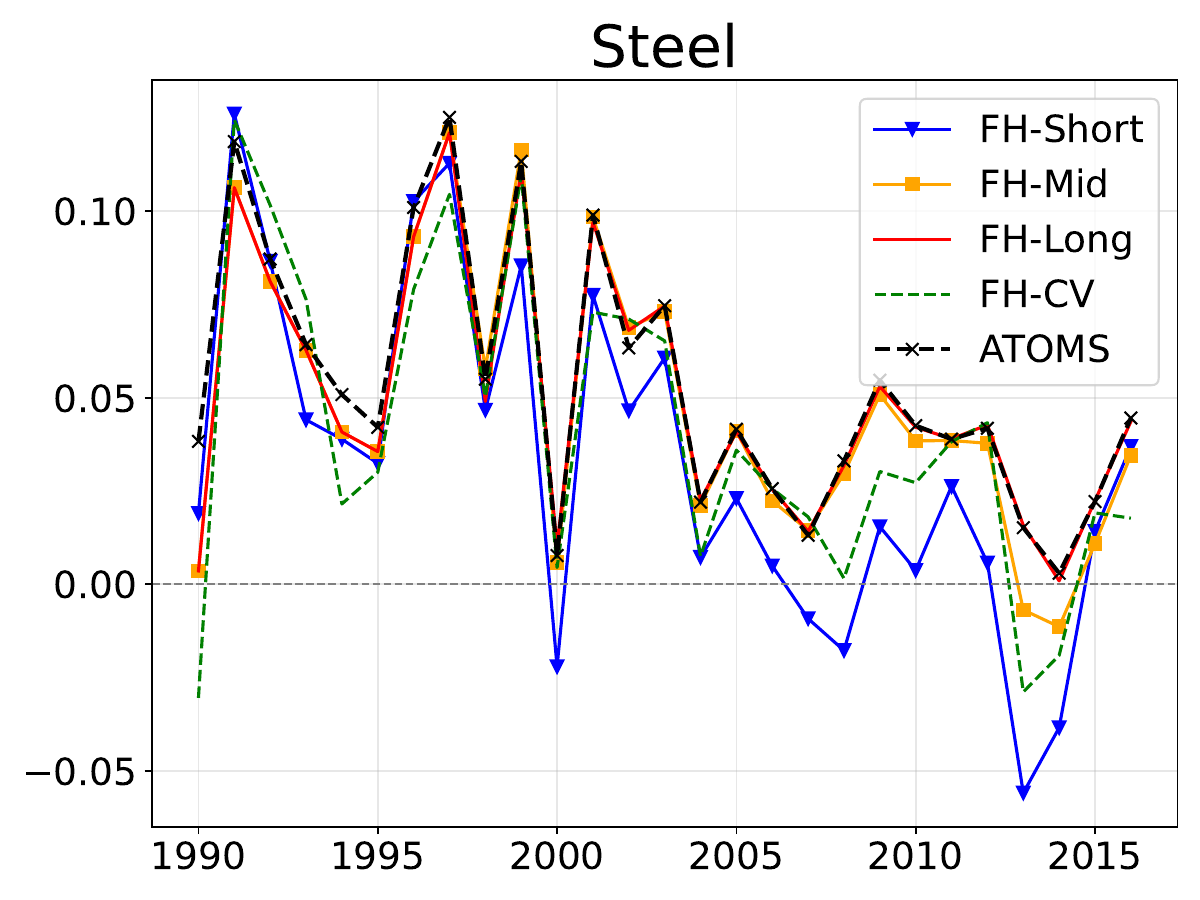}
	\end{subfigure}
    \begin{subfigure}{0.24\textwidth}
    	\centering
        \includegraphics[width=\linewidth]{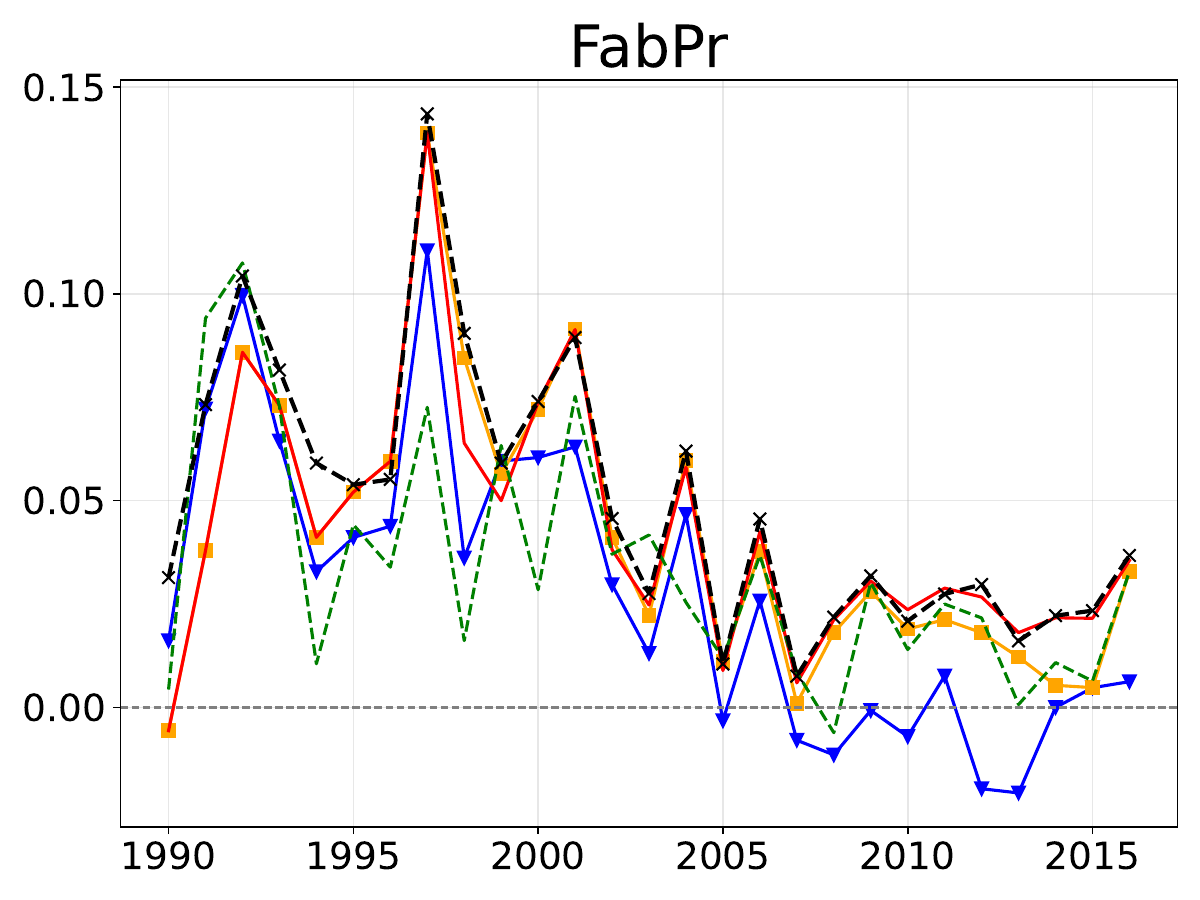}
	\end{subfigure}
    \begin{subfigure}{0.24\textwidth}
        \centering
        \includegraphics[width=\linewidth]{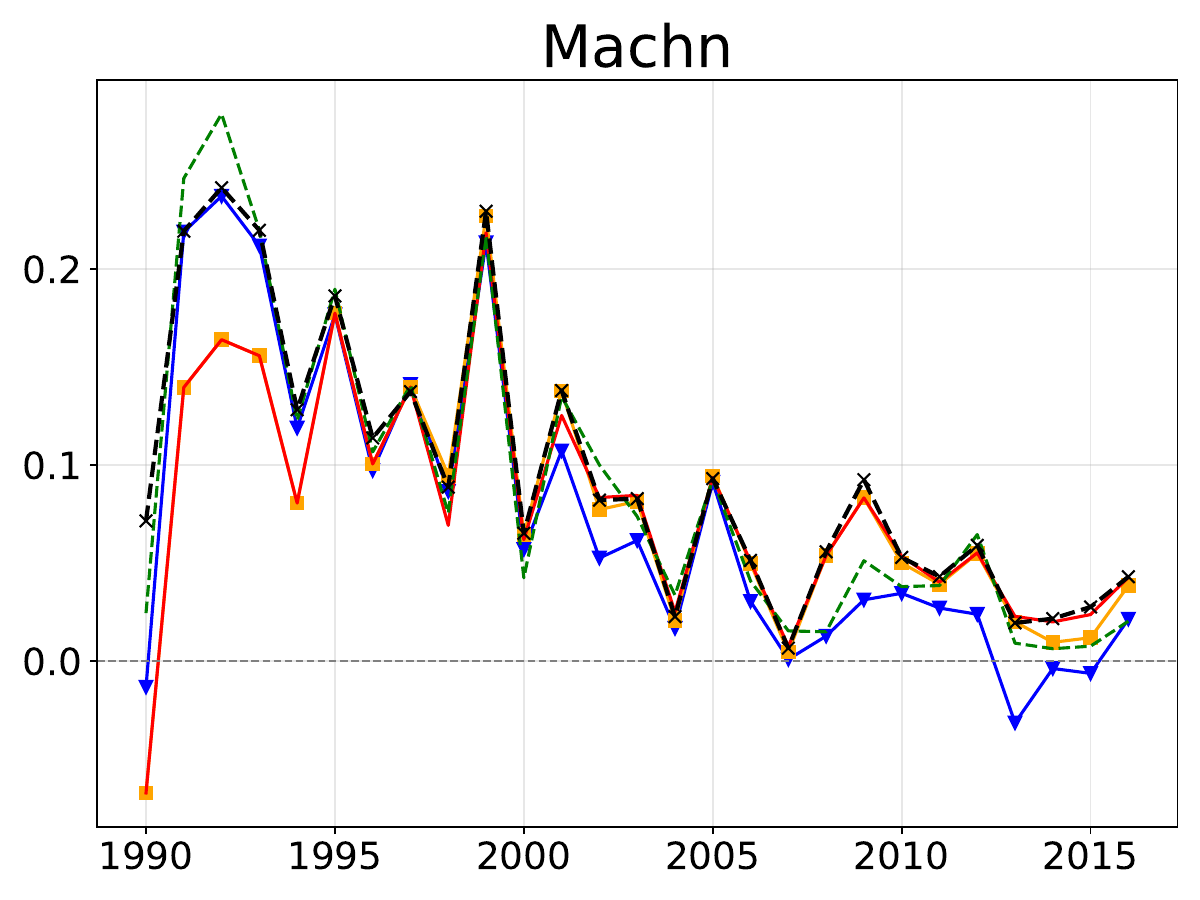}
	\end{subfigure}
    \hfill
    \begin{subfigure}{0.24\textwidth}
        \centering
        \includegraphics[width=\linewidth]{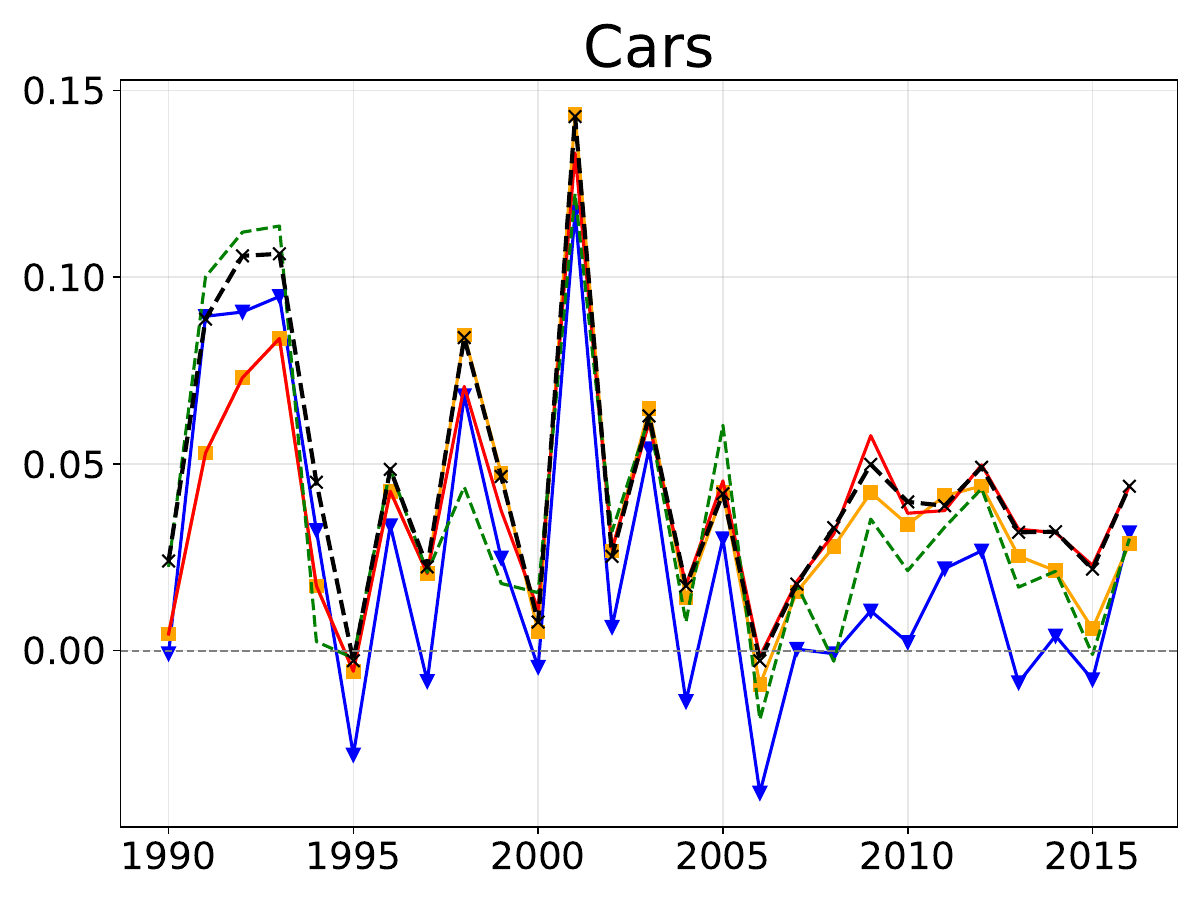}
	\end{subfigure}

    \begin{subfigure}{0.24\textwidth}
    	\centering
        \includegraphics[width=\linewidth]{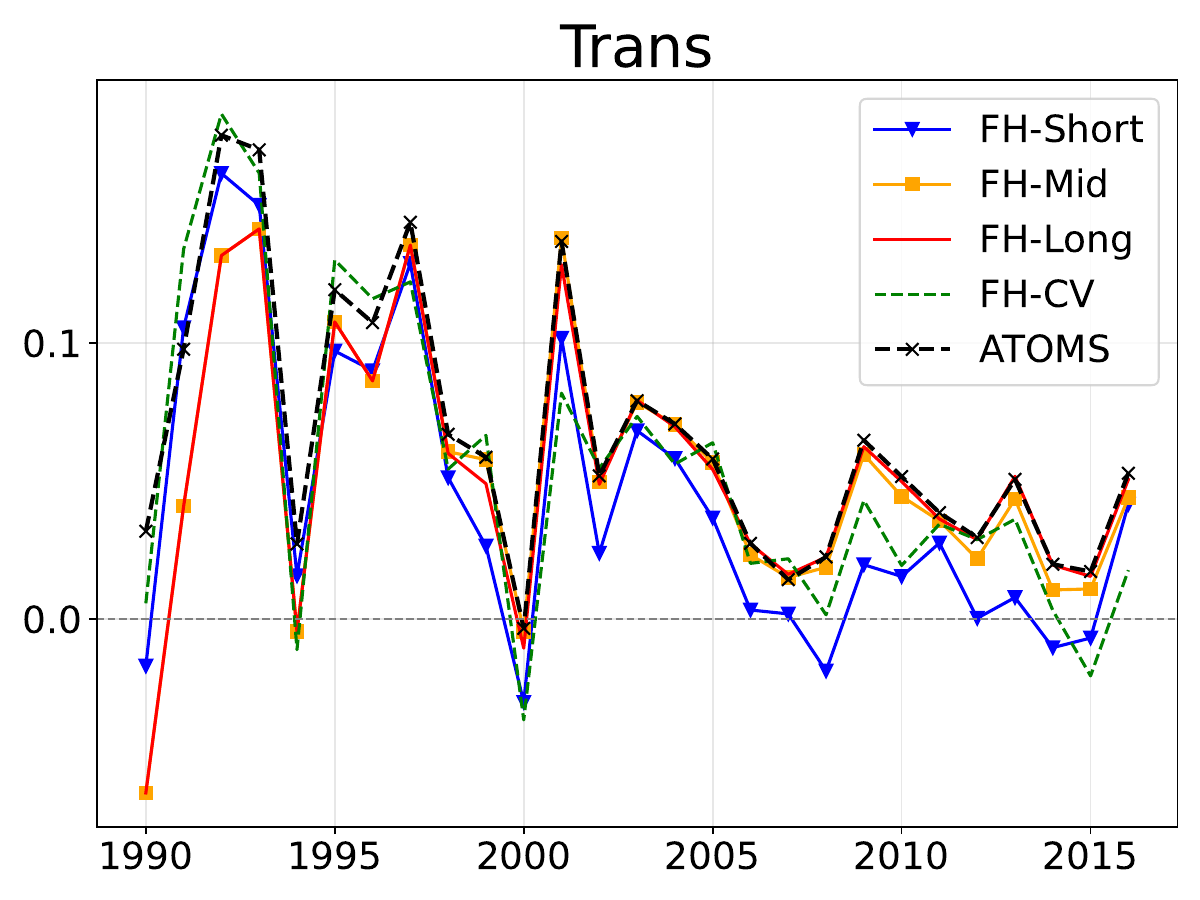}
	\end{subfigure}
    \begin{subfigure}{0.24\textwidth}
        \centering
        \includegraphics[width=\linewidth]{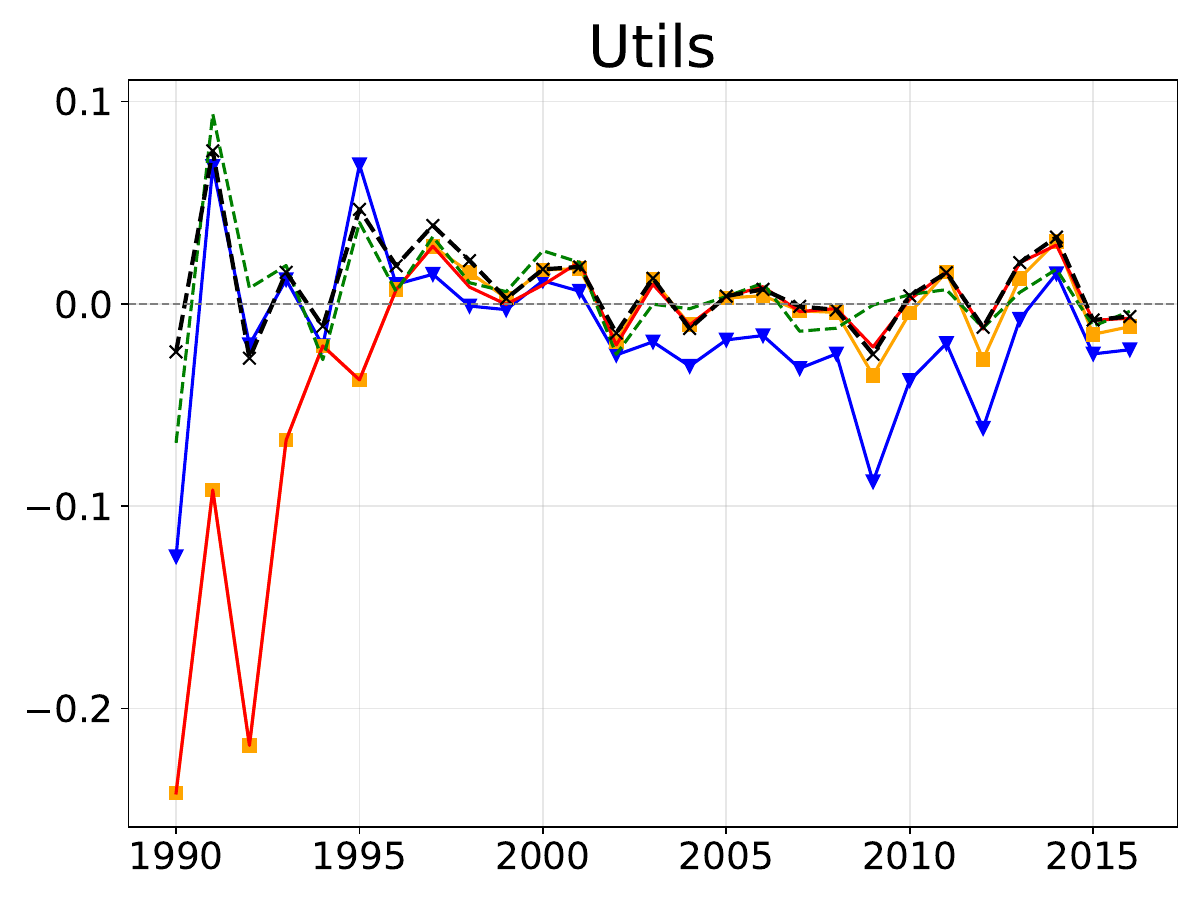}
	\end{subfigure}
    \begin{subfigure}{0.24\textwidth}
        \centering
        \includegraphics[width=\linewidth]{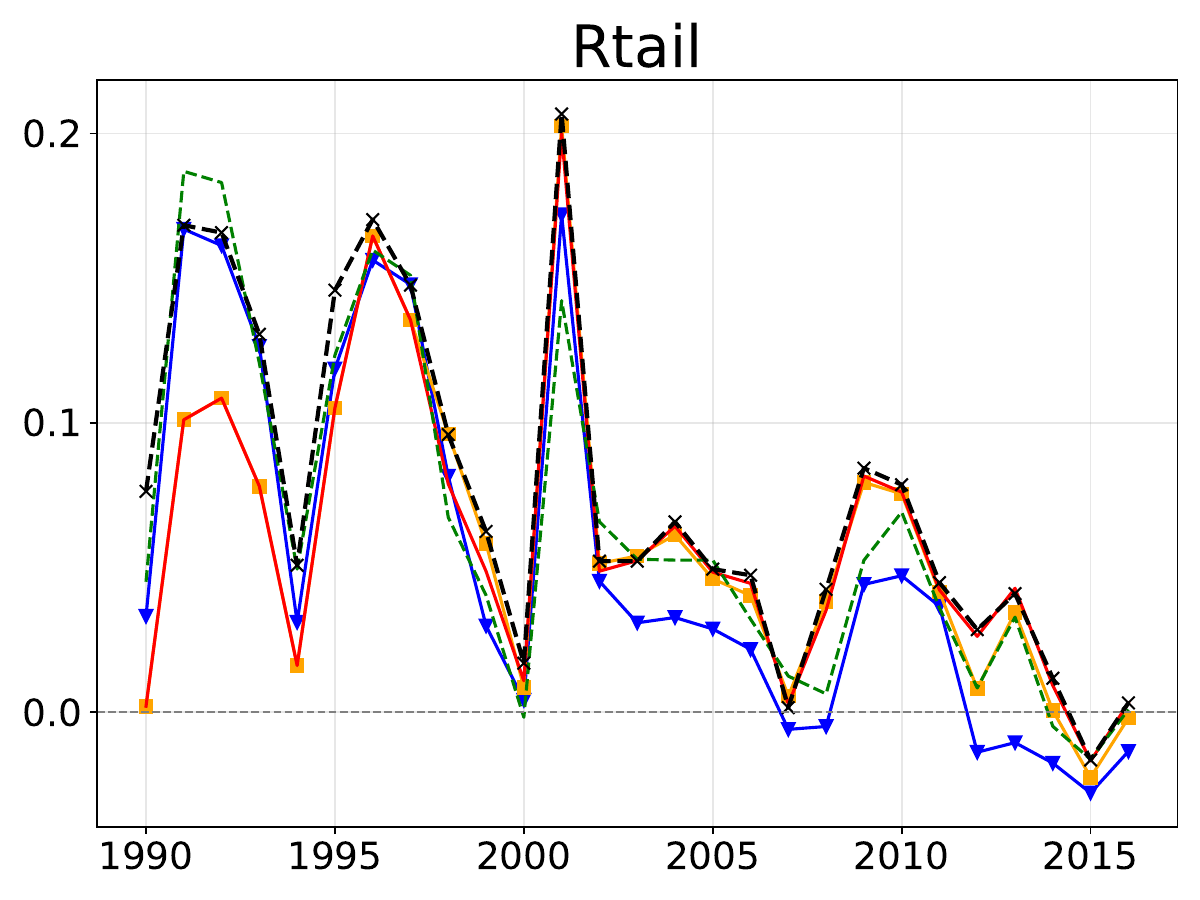}
	\end{subfigure}
    \begin{subfigure}{0.24\textwidth}
    	\centering
        \includegraphics[width=\linewidth]{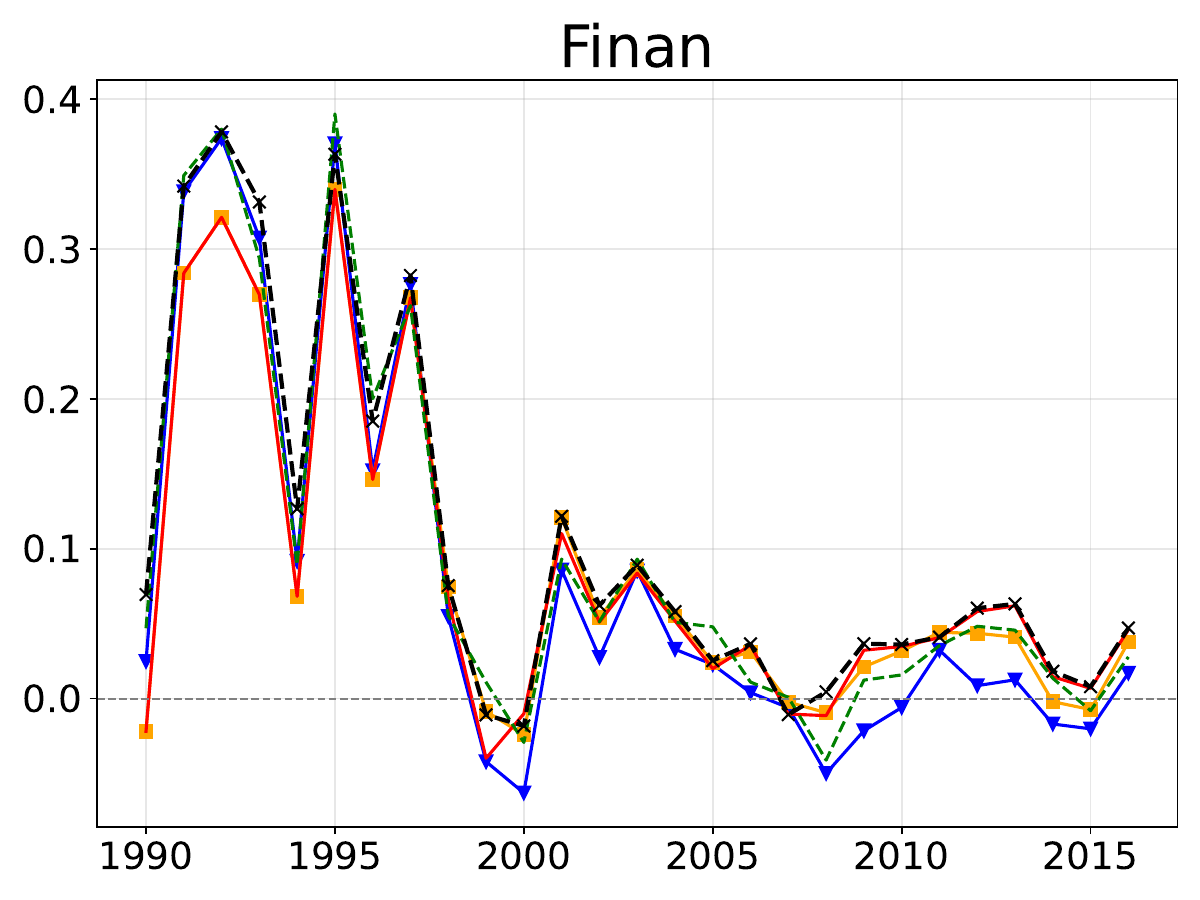}
	\end{subfigure}

    \begin{subfigure}{0.24\textwidth}
        \centering
        \includegraphics[width=\linewidth]{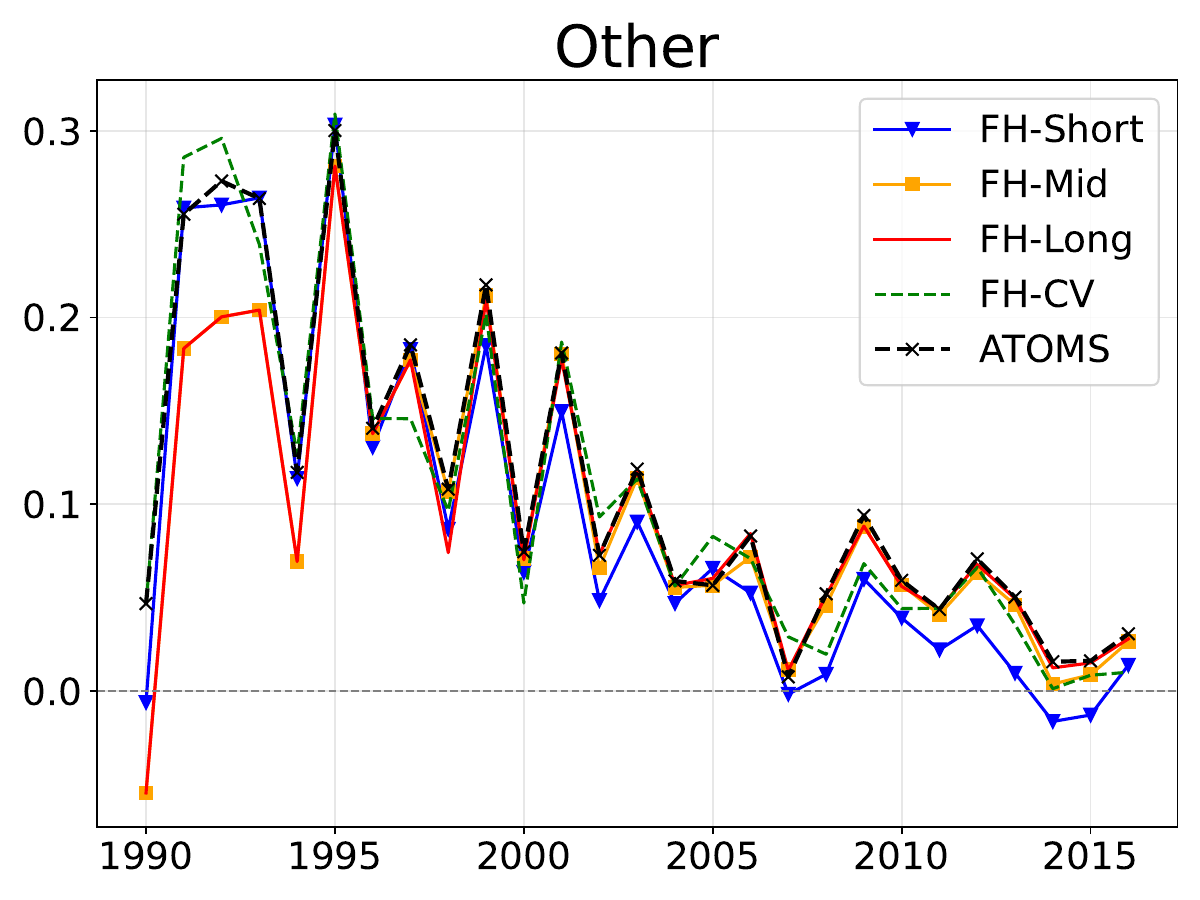}
	\end{subfigure}
      \bnotefig{This figure reports the annual out-of-sample $R^2$ of our adaptive model selection method $\adaptive$ (black dashed line with $\times$'s), as well as the fixed-window baselines $\fwshort$ (blue $\blacktriangledown$'s), $\fwmed$ (orange $\blacksquare$'s), and $\fwlong$ (red). The $\fh$ benchmarked methods use fixed windows as described in Table \ref{tab-fh-benchmarks}. The title in each subfigure is Kenneth French's acronym for each industry. For the full names of these industries, please refer to Table \ref{tab-industry-name-mapping}.}
\end{figure}

In more details, \myCref{fig-industry-yearly} plots the annual out-of-sample $R^2$ of $\adaptive$ and the benchmarks in each industry. We observe that $\adaptive$ maintains its advantage across diverse sectors including cyclical industries (Durbl, Cars, Trans) and defensive sectors (Food, Utils, Cnsum). This cross-sectional consistency further suggests that the observed performance reflects genuine adaptivity to changing market conditions rather than industry-specific anomalies.

Notably, $\adaptive$ shows particular strength in industries most sensitive to business cycle fluctuations, such as durable goods (Durbl), consumer discretionary (Rtail), and financial services (Finan). This pattern aligns with economic intuition, as these sectors experience the most dramatic shifts in risk premia during economic transitions. Our method's ability to capture these dynamics more effectively than benchmarks suggests it successfully identifies the changing risk-return tradeoffs that characterize different phases of the business cycle.

\paragraph{Adaptation versus Best-Window Identification.}
Perhaps counter-intuitively, our experiments show that short validation windows may not be optimal during recession periods. In non-stationary environments, the bias-variance tradeoff involved in choosing the window size can be subtle. For example, crisis data is often highly volatile, so a longer window may stabilize model comparison. A longer window may also include data from previous crisis periods whose distribution is closer to the current crisis than samples immediately preceding it. Consequently, several window choices may achieve statistically and economically similar performance, and there need not be a universal ordering of window sizes in terms of performance, even during recessions. This is reflected in the fixed-horizon benchmarks in Table~\ref{tab:oos_r2_industry_time}: among $\fwshort$, $\fwmed$, and $\fwlong$, the best-performing fixed window varies across recession periods, with $\fwshort$ performing best during the Gulf War recession, $\fwmed$ during the 2001 recession, and $\fwlong$ during the 2008 Financial Crisis.
Therefore, it is generally not feasible to adapt to the non-stationarity through best-window identification. We do not aim to infer a uniquely optimal validation window at each point in time, especially when several window choices may yield similar performance. Instead, $\adaptive$ is designed to select a window size that gives predictive performance close to the best fixed window in hindsight.

\subsection{Portfolio Performance and Investment Gains}

We now evaluate whether the conditional return forecasts generated by $\adaptive$ translate into economic value in an implementable allocation problem. We construct two portfolio rules over the 17 industry portfolios and compare risk-adjusted performance, turnover, and cumulative wealth for $\adaptive$ and the benchmarks. All portfolio results are strictly out-of-sample, with the trading period running from January 1990 (month $t=1$) through November 2016.

\paragraph{Portfolio Construction and Timing.} At the beginning of each month $t$, each method estimates $N=17$ conditional mean functions $\widehat{f}_t^{(n)}$, $n=1,\ldots,N$, one for each industry portfolio return, using only information available up to $t-1$. On each trading day $s\in\batch_t$ of month $t$, given predictors $\bx_s$, the vector of conditional return forecasts is
\[
\widehat{\bbf}_t(\bx_s) = \big(\widehat{f}_t^{(1)}(\bx_s),...,\widehat{f}_t^{(N)}(\bx_s)\big)^\top.
\]
These forecasts are mapped into daily cross-sectional portfolio weights $\bw_s=\big(w_s^{(1)},...,w_s^{(N)}\big)^\top$. We consider two allocation rules.
\begin{enumerate}
    \item Mean-variance efficient portfolio: We form the mean-variance efficient direction
\[
    \widetilde{\bw}_s^{\MV}
    =  \widehat{\bSigma}_s^{-1}\widehat{\bbf}_t(\bx_s),
\]
where $\widehat{\bSigma}_s$ is the sample covariance matrix of daily industry returns estimated with a one-year rolling window. Portfolio weights are then normalized so that their absolute values sum to $1$:
\[
    \bw_s^{\MV} = \frac{\widetilde{\bw}_s^{\MV}}{\big\| \widetilde{\bw}_s^{\MV} \big\|_1}.
\]
\item Signed portfolio: We assign equal absolute weight $1/N = 1/17$ to each industry, and take long positions for industries with positive predicted returns and short positions for industries with negative predicted returns:
\[
\bw_s^{\EWsign} = \frac{1}{N} \cdot \sign\big( \widehat{\bbf}_t (\bx_s) \big),
\]
where $\sign(\cdot)$ is applied componentwise.
We also compare against the equally weighted portfolio ($\EWlong$) which longs every industry with weight $1/N$: $\bw^{\EW}_s = (\frac{1}{N} ,\ldots,\frac{1}{N})^\top$.
\end{enumerate}

\paragraph{Performance and Implementation Metrics.} For portfolio weights $\bw_s$, the gross daily portfolio return is $r^{\texttt{g}}_s = \bw_s^\top \by_s$. Daily turnover is measured by the $\ell_1$ rebalancing norm, $\turnover_s = \| \bw_s - \bw_{s-1} \|_1$, where $s-1$ denotes the previous trading day. The net return is $r_s = r^{\texttt{g}}_s - c\cdot\turnover_s$, where $c = 4\times 10^{-4} / 252$\footnote{We consider the annualized trading cost of 4bps that are on the same scale reported by \cite{harvey2025unintended}, and treat each year as 252 trading days.} is the trading cost, and $c\cdot \turnover_s$ takes into account implementation frictions.

We aggregate daily results to the monthly frequency. For each month $t$, we compute the monthly Sharpe ratio
\[
\SR_t = \frac{\widehat{\EE}_t[r]}{\sqrt{\widehat{\var}_t(r)}},
\]
where the sample mean $\widehat{\EE}_t[r]$ and sample variance $\widehat{\var}_t(r)$ are computed from daily net returns within month $t$:
\[
\widehat{\EE}_t[r] = \frac{1}{\batch_t} \sum_{s\in\batch_t} r_s
\quad\text{and}\quad
\widehat{\var}_t(r) = \frac{1}{\batch_t-1} \sum_{s\in\batch_t}\Big( r_s - \widehat{\EE}_t[r]  \Big)^2.
\]

We also assess the cumulative wealth across time. Starting with initial wealth $W_0=1$, the cumulative wealth evolves according to $
W_t = W_{t-1}\prod_{s\in \batch_t} \left( 1 + r_s \right) $ and the terminal wealth $W_T$ at time $T$ is then $
W_T = \prod_{t=1}^T \prod_{s\in \batch_t} \left( 1 + r_s \right)$.

To capture risk characteristics, we note that path-dependent downside risk is better captured by maximum drawdown (MDD) than by average return
or Sharpe ratio alone.
We measure the maximum drawdown of the wealth curve as the largest peak-to-trough decline
over the investment horizon:
\[
  \drawdown = \max_{1\le t\le T} \bigg( 1 - \frac{W_t}{\max\limits_{1\le\tau\le t} W_{\tau}} \bigg).
\]
For MDD, a large value indicates that the strategy's cumulative value fell sharply from its running historical peak at some point along the path---a loss event that is invisible to statistics that average across time.

To isolate stress-period behavior, we define the \emph{conditional drawdown} over a subperiod $[t_0,t_1]$ with a lookback reference window of length $S$:
\[
  \drawdown\big|_{t_0,\,t_1}
  = \max_{t_0\le t\le t_1}
        \bigg( 1 - \frac{W_t}{\max\limits_{t_0-S \le \tau \le\, t} W_{\tau}} \bigg).
\]
This metric restricts attention to the worst trough within $[t_0,t_1]$, measured relative
to the running peak computed over the immediately preceding $S$-length window.
For example, setting $S = 3$~months and $[t_0,t_1]$ equal to the 2001 recession,
$\drawdown|_{t_0,t_1}$ captures the deepest portfolio decline from any peak attained
since the 3 months before the contraction began, through the end of the recession.

Smaller drawdowns have direct economic consequences as they reduce the probability of investor
capital withdrawals, margin calls, and forced deleveraging after adverse return streaks.
We therefore treat drawdown as a complementary implementability criterion because among strategies
with similar risk-adjusted returns, the one with the shallower trough will remain viable in real time.
Drawdown also functions as a consistency check on the cumulative wealth results where a strategy
whose wealth gains concentrate in a handful of extreme observations will show sharp
troughs during recessions. We would thus reveal fragility in any lack of risk diversification with the $\drawdown|_{t_0,t_1}$ metric.

\paragraph{Results: Mean-Variance Portfolio.} The mean-variance portfolio provides an empirical proxy for the conditional SDF. By the \citet{HJa91} variance bound, for any valid pricing kernel $m$, $\mathrm{std}(m)/\mathbb{E}(m) \ge \SR_{\max}/(1+R_f)$, where $\SR_{\max}$ is the maximum Sharpe ratio in the traded asset span, and $R_f$ is the risk-free rate. A higher Sharpe ratio tightens this bound, requiring the pricing kernel to vary more to remain consistent with the data.\footnote{The bound applies unconditionally. We interpret improvements in the realized conditional Sharpe ratio as directional evidence of better conditional SDF approximation.} The gains below therefore carry an asset-pricing interpretation beyond investor utility.

\Cref{tab:sharpe_mv_industry_time} shows a clear full-sample ordering: $\adaptive$ delivers the highest out-of-sample mean-variance Sharpe ratio, 0.334, relative to 0.304 for $\fwlong$, 0.300 for $\fixedwindowCV$, 0.296 for $\fwmed$, 0.241 for $\fwshort$, and 0.232 for $\RS$. These differences correspond to gains of 9.87\%, 11.33\%, 38.59\%, and 44.0\% respectively, indicating economically meaningful improvements in excess return per unit of risk rather than marginal statistical variation.

\begin{table}[h]
\centering
\caption{Average Monthly Sharpe Ratio of Mean-Variance Portfolio over Different Periods}\label{tab:sharpe_mv_industry_time}
\begin{tabular}{@{}lcccccc@{}}
\toprule
\multirow{2}{*}{Method} & \multirow{2}{*}{Full OOS Period} & \multicolumn{3}{c}{Recessions}  \\
\cmidrule{3-5}
& & Gulf War & 2001 Recession & Financial Crisis  \\
\midrule
$\adaptive$ & $0.333$ & $0.187$ & $0.568$ & $0.243$ \\
$\fwshort$ & $0.239$ & $0.148$ & $0.305$ & $0.098$ \\
$\fwmed$ & $0.295$ & $0.145$ & $0.471$ & $0.211$ \\
$\fwlong$ & $0.303$ & $0.145$ & $0.559$ & $0.220$ \\
$\fixedwindowCV$ & $0.299$ & $0.168$ & $0.371$ & $0.053$ \\
$\RS$ & $0.231$ & $0.113$ & $0.399$ & $0.138$ \\
\bottomrule
\end{tabular}
\bnotetab{
    This table reports monthly Sharpe ratio averages for mean-variance portfolios formed from conditional return forecasts for the 17 industry portfolios. Full OOS Period refers to the full out-of-sample period, 01/1990$\sim$11/2016. The $\fh$ benchmarked methods use fixed windows as described in Table \ref{tab-fh-benchmarks}. Columns report Sharpe ratio averages during three recessions, as documented in \href{https://www.nber.org/research/business-cycle-dating}{NBER Business Cycle Dating}: 
    \begin{itemize}
        \item the 1990 Gulf War recession (06/1990$\sim$10/1990);
        \item the 2001 Recession of dot-com bubble burst and the 9/11 attack (05/2001$\sim$10/2001);
        \item the Financial Crisis led by defaults of subprime mortgages (11/2007$\sim$06/2009). 
    \end{itemize}}
\end{table}

The recession subsamples provide a sharper test because they are the episodes in which conditional means and covariance structure move most abruptly. In the Gulf War recession, $\adaptive$ attains 0.188 versus 0.169 for $\fixedwindowCV$, 0.146 for $\fwlong$, and 0.115 for $\RS$ (11.24\%, 28.77\% and 63.48\% improvements). In the 2001 recession, $\adaptive$ reaches 0.569, close to $\fwlong$ at 0.560 (1.61\%), but far above $\RS$ at 0.400 (42.25\%) and $\fixedwindowCV$ at 0.371 (53.37\%), consistent with substantial value from adaptive window choice under rapid repricing. In the Financial Crisis, $\adaptive$ records 0.244 versus 0.221 for $\fwlong$, 0.140 for $\RS$, and 0.054 for $\fixedwindowCV$ (10.41\%, 47.29\% and 351.85\% improvements), suggesting that adaptive updating remains effective even in a prolonged downturn.

For asset-pricing interpretation, these results matter beyond investor utility. Within the span of traded industry returns, the mean-variance efficient portfolio is tied to the mean-variance efficient pricing kernel. Accordingly, superior out-of-sample Sharpe ratios for this portfolio imply a tighter empirical approximation to the conditional SDF, or equivalently, smaller pricing-kernel misspecification in the traded cross section. Taken together, the evidence supports a state-dependent environment in which adaptive estimation improves both implementable portfolio efficiency and SDF approximation.

These Sharpe ratio improvements are not generated by mechanically higher trading intensity. \Cref{tab:turnover_mv_industry_time} shows that $\adaptive$ has lower turnover than $\fwshort$, $\fwmed$, $\fwlong$ and $\RS$ in the full sample (11.94 versus 16.39, 13.87, 12.45 and 20.40), and comparable turnover in recession windows. Relative to $\fixedwindowCV$, $\adaptive$ trades somewhat more but delivers materially higher Sharpe ratios, implying a more favorable return-cost trade-off.

\begin{table}[h]
\centering
\caption{Average Monthly Turnover of Mean-Variance Portfolio over Different Periods}\label{tab:turnover_mv_industry_time}
\begin{tabular}{@{}lcccccc@{}}
\toprule
\multirow{2}{*}{Method} & \multirow{2}{*}{Full OOS Period} & \multicolumn{3}{c}{Recessions}  \\
\cmidrule{3-5}
& & Gulf War & 2001 Recession & Financial Crisis  \\
\midrule
$\adaptive$ & $11.94$ & $16.85$ & $13.71$ & $13.62$ \\
$\fwshort$ & $16.39$ & $17.43$ & $16.91$ & $17.86$ \\
$\fwmed$ & $13.87$ & $18.63$ & $15.07$ & $15.12$ \\
$\fwlong$ & $12.45$ & $18.63$ & $14.06$ & $13.34$ \\
$\fixedwindowCV$ & $11.31$ & $13.07$ & $10.92$ & $10.06$ \\
$\RS$ & $20.40$ & $22.83$ & $19.06$ & $20.62$ \\
\bottomrule
\end{tabular}
\bnotetab{
    This table reports monthly turnover averages for mean-variance portfolios formed from conditional return forecasts. Turnover is computed as the average number of portfolio weight changes per month across all $17$ industry portfolios, i.e. the $\ell_1$ norm of the difference in portfolio weights between consecutive months. The $\fh$ benchmarked methods use fixed windows as described in Table \ref{tab-fh-benchmarks}.}
\end{table}

\begin{figure}[h]
\centering
\caption{Cumulative Wealth of Mean-Variance Portfolio\label{fig-wealth-mv}}
\includegraphics[scale=0.45]{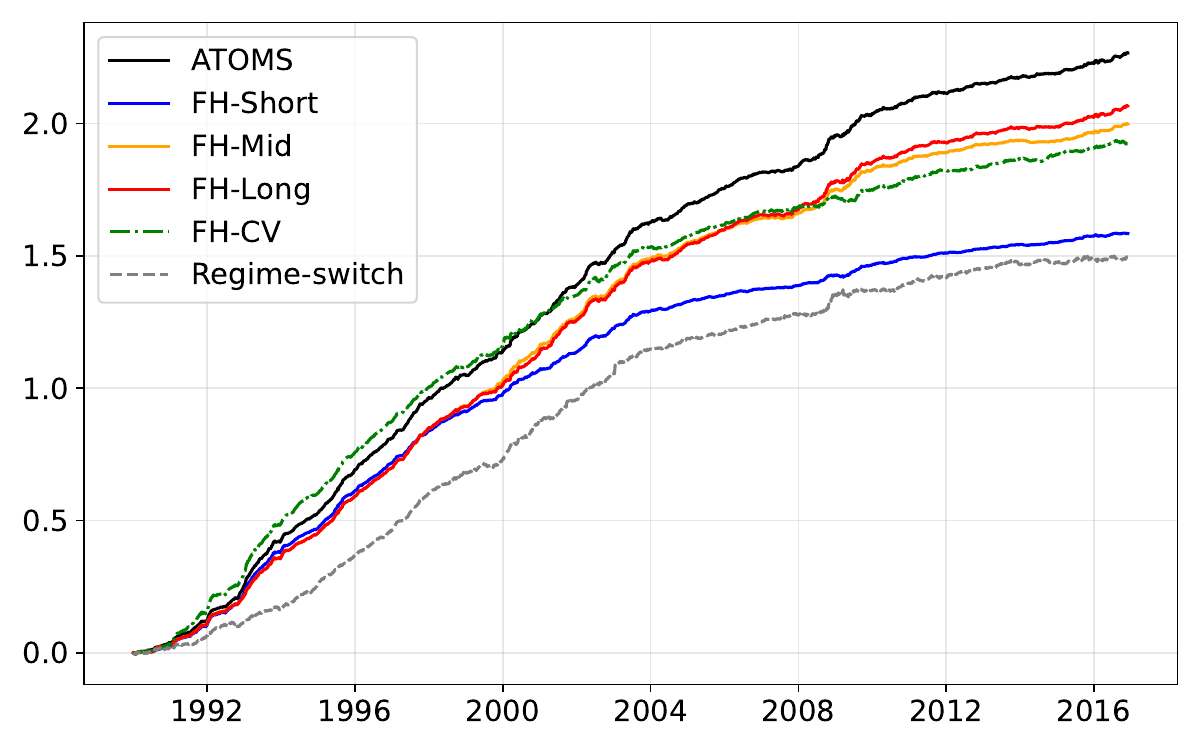}
\bnotefig{This figure plots cumulative wealth (in logarithmic scale) for the mean-variance portfolio formed from conditional return forecasts under $\adaptive$ and the benchmark methods. Starting with initial wealth $W_0=1$, the wealth in each month $t$ evolves according to $W_t = W_{t-1}\prod_{s\in \batch_t} \left( 1 + r^{\texttt{g}}_s - c\cdot \turnover_s \right)$, where $\batch_t$ collects all days in month $t$, $r^{\texttt{g}}_s$ is the return of the mean-variance portfolio on day $s$, $c = 4\times 10^{-4} / 252$ is the trading cost, and $\turnover_s$ is the daily turnover.}
\end{figure}

\Cref{fig-wealth-mv} plots log cumulative wealth. $\adaptive$ dominates the fixed-window alternatives and the regime-switching model over most of the out-of-sample period and separates notably from $\fixedwindowCV$ after 2001. The wealth paths confirm that the Sharpe-ratio gains are economically meaningful after implementation costs.

In \Cref{tab-drawdown-mv}, we report the maximum drawdown of the mean-variance portfolio; for the recession periods, we compute the conditional maximum drawdown with a three-month look-back window. $\adaptive$ has a full-sample drawdown of $3.0\%$, below $\fwshort$ ($4.3\%$) and $\fixedwindowCV$ ($4.4\%$), and close to $\fwlong$ ($3.1\%$). During recessions, $\adaptive$ also maintains a low downside risk: it has the smallest drawdown during the Gulf War recession ($0.7\%$), ties the best drawdown during the 2001 recession ($0.6\%$), and matches $\fwmed$ during the Financial Crisis ($2.4\%$). Although $\RS$ has a slightly smaller full-sample drawdown, it also has materially lower Sharpe ratios and much higher turnover. These results indicate that the mean-variance portfolio based on $\adaptive$ improves risk-adjusted performance without generating larger peak-to-trough losses. In \myCref{sec-additional-drawdown-results}, we report additional drawdown results using alternative look-back windows of $S=1,2,6$ months.

\begin{table}[h]
\centering
\caption{Maximum Drawdown of Mean-Variance Portfolio}\label{tab-drawdown-mv}
\begin{tabular}{@{}lcccccc@{}}
\toprule
\multirow{2}{*}{Method} & \multirow{2}{*}{Full OOS Period} & \multicolumn{3}{c}{Recessions}  \\
\cmidrule{3-5}
& & Gulf War & 2001 Recession & Financial Crisis  \\
\midrule
$\adaptive$ & $3.0\%$ & $0.7\%$ & $0.6\%$ & $2.4\%$ \\
$\fwshort$ & $4.3\%$ & $1.0\%$ & $0.9\%$ & $3.5\%$ \\
$\fwmed$ & $3.3\%$ & $1.0\%$ & $0.7\%$ & $2.4\%$ \\
$\fwlong$ & $3.1\%$ & $1.0\%$ & $0.6\%$ & $2.7\%$ \\
$\fixedwindowCV$ & $4.4\%$ & $0.9\%$ & $0.9\%$ & $4.2\%$ \\
$\RS$ & $2.7\%$ & $1.2\%$ & $0.8\%$ & $2.7\%$ \\
\bottomrule
\end{tabular}
\bnotetab{This table reports maximum drawdowns for mean-variance portfolios formed from conditional return forecasts of $\adaptive$ and benchmark methods. Full OOS Period refers to the full out-of-sample period, 01/1990$\sim$11/2016. Recession columns report conditional drawdowns computed within each recession window using a three-month look-back reference window ($S=3$ months). The $\fh$ benchmarked methods use fixed windows as described in Table \ref{tab-fh-benchmarks}. Smaller values indicate smaller peak-to-trough losses.}
\end{table}

\paragraph{Results: Signed Portfolio.} The signed portfolio isolates the directional information in the predicted returns: it holds each industry with the same absolute weight, but uses the model forecast to determine whether the position is long or short. \Cref{tab:sharpe_ew_industry_time} shows that $\adaptive$ again achieves the highest full-sample Sharpe ratio, 0.359, outperforming all the benchmark methods. We observe that the best fixed-horizon benchmark among $\fwshort$, $\fwmed$ and $\fwlong$ varies across recessions: $\fwshort$ performs the best during the Gulf War recession, while $\fwmed$ performs the best during the 2001 recession and marginally leads during the Financial Crisis. $\adaptive$ keeps up with the best fixed-horizon benchmark in hindsight, and outperforms all three fixed-horizon benchmarks in the Gulf War and 2001 recessions while remaining close to the best fixed-horizon benchmark during the Financial Crisis.

The comparison with the equally weighted benchmark $\EWlong$ is more pronounced. $\EWlong$ attains a full-sample Sharpe ratio of 0.239, so $\adaptive$ improves the Sharpe ratio by 50.21\%. The same pattern holds in recession subsamples: $\adaptive$ delivers Sharpe ratios of 0.397, 0.430, and 0.215 during the Gulf War recession, the 2001 recession, and the Financial Crisis, compared with $-0.123$, 0.097, and 0.071 for $\EWlong$. This shows that even a simple equal-weight trading strategy benefits substantially from conditioning positions on the signs of the predicted returns. The additional gains of $\adaptive$ over the other benchmarks show that ATOMS further improves the directional information in the predicted returns.

\begin{table}[h]
\centering
\caption{Average Monthly Sharpe Ratio of Signed Portfolio over Different Periods}\label{tab:sharpe_ew_industry_time}
\begin{tabular}{@{}lcccccc@{}}
\toprule
\multirow{2}{*}{Method} & \multirow{2}{*}{Full OOS Period} & \multicolumn{3}{c}{Recessions}  \\
\cmidrule{3-5}
& & Gulf War & 2001 Recession & Financial Crisis  \\
\midrule
$\adaptive$ & $0.359$ & $0.397$ & $0.430$ & $0.215$ \\
$\fwshort$ & $0.354$ & $0.329$ & $0.408$ & $0.202$ \\
$\fwmed$ & $0.352$ & $0.293$ & $0.425$ & $0.223$ \\
$\fwlong$ & $0.351$ & $0.293$ & $0.414$ & $0.216$ \\
$\fixedwindowCV$ & $0.352$ & $0.283$ & $0.426$ & $0.172$ \\
$\RS$ & $0.311$ & $0.187$ & $0.384$ & $0.203$ \\
$\EWlong$ & $0.239$ & $-0.123$ & $0.097$ & $0.071$ \\
\bottomrule
\end{tabular}
\bnotetab{
    This table reports monthly Sharpe ratio averages for signed portfolios formed from conditional return forecasts for the 17 industry portfolios. Full OOS Period refers to the full out-of-sample period, 01/1990$\sim$11/2016. The $\fh$ benchmarked methods use fixed windows as described in Table \ref{tab-fh-benchmarks}. Columns report Sharpe ratio averages during three recessions, as documented in \href{https://www.nber.org/research/business-cycle-dating}{NBER Business Cycle Dating}: 
    \begin{itemize}
        \item the 1990 Gulf War recession (06/1990$\sim$10/1990);
        \item the 2001 Recession of dot-com bubble burst and the 9/11 attack (05/2001$\sim$10/2001);
        \item the Financial Crisis led by defaults of subprime mortgages (11/2007$\sim$06/2009). 
    \end{itemize}}
\end{table}

In terms of turnover, \Cref{tab:turnover_ew_industry_time} shows that $\adaptive$ has lower turnover than $\fwshort$, $\fwmed$, $\fwlong$, and $\RS$ in the full sample (3.22 versus 5.46, 4.09, 3.53, and 10.62), though slightly more than $\fixedwindowCV$ (3.01). The long-only equally weighted benchmark $\EWlong$ has zero turnover by construction because it maintains constant equal weights. Relative to this zero-turnover benchmark, the prediction-based methods, including $\adaptive$ and the fixed-horizon benchmarks, introduce only modest trading: $\adaptive$ changes only 3.22 units of $\ell_1$ weight per month on average, while attaining a substantially higher Sharpe ratio. During recession periods, $\adaptive$ generally remains below or comparable to the fixed-horizon and regime-switching benchmarks, while delivering stronger risk-adjusted performance.

\begin{table}[h]
\centering
\caption{Average Monthly Turnover of Signed Portfolio over Different Periods}\label{tab:turnover_ew_industry_time}
\begin{tabular}{@{}lcccccc@{}}
\toprule
\multirow{2}{*}{Method} & \multirow{2}{*}{Full OOS Period} & \multicolumn{3}{c}{Recessions}  \\
\cmidrule{3-5}
& & Gulf War & 2001 Recession & Financial Crisis  \\
\midrule
$\adaptive$ & $3.22$ & $5.15$ & $5.30$ & $3.68$ \\
$\fwshort$ & $5.46$ & $6.56$ & $6.61$ & $7.49$ \\
$\fwmed$ & $4.09$ & $8.16$ & $5.66$ & $4.31$ \\
$\fwlong$ & $3.53$ & $8.16$ & $5.16$ & $3.58$ \\
$\fixedwindowCV$ & $3.01$ & $4.40$ & $2.48$ & $3.90$ \\
$\RS$ & $10.62$ & $14.38$ & $10.94$ & $11.06$ \\
$\EWlong$ & $0$ & $0$ & $0$ & $0$ \\
\bottomrule
\end{tabular}
\bnotetab{
    This table reports monthly turnover averages for signed portfolios formed from conditional return forecasts. Turnover is computed as the average number of portfolio weight changes per month across all 17 industry portfolios, i.e. the $\ell_1$-norm of the difference in portfolio weights between consecutive months. The $\fh$ benchmarked methods use fixed windows as described in Table \ref{tab-fh-benchmarks}.} 
\end{table}

\begin{figure}[h!]
\centering
\caption{Cumulative Wealth of Signed Portfolio\label{fig-wealth-ew}}
\includegraphics[scale=0.45]{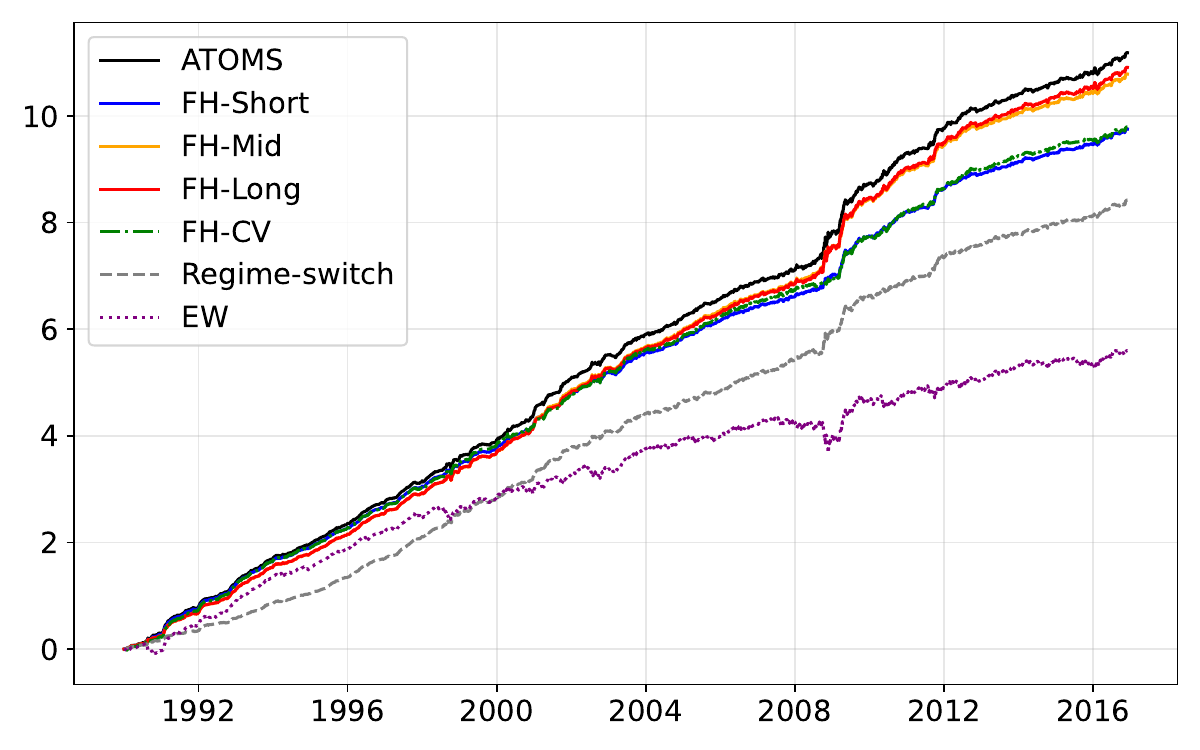}
\bnotefig{This figure plots cumulative wealth (in logarithmic scale) for the equally weighted portfolio formed from conditional return forecasts under $\adaptive$ and the benchmark approaches. Starting with initial wealth $W_0=1$, the wealth in each month $t$ evolves according to $W_t = W_{t-1}\prod_{s\in \batch_t} \left( 1 + r^{\texttt{g}}_s - c\cdot \turnover_s \right)$, where $\batch_t$ collects all days in month $t$, $r_s^{\texttt{g}}$ is the return of the mean-variance portfolio on day $s$, $c = 4\times 10^{-4} / 252$ is the trading cost, and $\turnover_s$ is the daily turnover.}
\end{figure}

\Cref{fig-wealth-ew} reports the log cumulative wealth for the signed portfolio. $\adaptive$ again dominates all benchmarks for most of the sample, and substantially outperforms $\fixedwindowCV$ and $\RS$ after 2000, indicating persistent economic gains from adaptive portfolio formation. Taken together, the mean-variance and equally weighted evidence supports a conditional asset-pricing view in which expected returns vary with the economic state and are better captured by adaptive training windows.

In \Cref{tab-drawdown-ew}, we report maximum drawdowns for the signed portfolio using the same three-month look-back window as the mean-variance portfolio. The prediction-based strategies have much smaller drawdowns than the long-only equal-weight benchmark: $\adaptive$ has a full-sample drawdown of $16.3\%$, compared with $48.8\%$ for $\EWlong$. The difference is especially pronounced in recessions, where $\EWlong$ has drawdowns of $17.8\%$, $14.2\%$, and $46.9\%$ during the Gulf War recession, the 2001 recession, and the Financial Crisis, respectively, while $\adaptive$ has drawdowns of only $1.9\%$, $2.4\%$, and $16.3\%$. Although $\fwshort$ and $\RS$ have smaller full-sample drawdowns than $\adaptive$, they also have lower Sharpe ratios. Overall, the adaptive prediction-based strategy delivers stronger risk-adjusted performance while avoiding the large downside exposure of the long-only equal-weight benchmark. In \myCref{sec-additional-drawdown-results}, we report additional drawdown results using alternative look-back windows of $S=1,2,6$ months.

\begin{table}[h]
\centering
\caption{Maximum Drawdown of Signed Portfolio\label{tab-drawdown-ew}}
\begin{tabular}{@{}lcccccc@{}}
\toprule
\multirow{2}{*}{Method} & \multirow{2}{*}{Full OOS Period} & \multicolumn{3}{c}{Recessions}  \\
\cmidrule{3-5}
& & Gulf War & 2001 Recession & Financial Crisis  \\
\midrule
$\adaptive$ & $16.3\%$ & $1.9\%$ & $2.4\%$ & $16.3\%$ \\
$\fwshort$ & $13.6\%$ & $2.1\%$ & $2.1\%$ & $12.8\%$ \\
$\fwmed$ & $15.2\%$ & $2.7\%$ & $2.4\%$ & $15.1\%$ \\
$\fwlong$ & $16.5\%$ & $2.7\%$ & $2.3\%$ & $16.5\%$ \\
$\fixedwindowCV$ & $18.4\%$ & $2.4\%$ & $1.9\%$ & $18.0\%$ \\
$\RS$ & $13.0\%$ & $1.2\%$ & $1.6\%$ & $13.0\%$ \\
$\EWlong$ & $48.8\%$ & $17.8\%$ & $14.2\%$ & $46.9\%$ \\
\bottomrule
\end{tabular}
\bnotetab{This table reports maximum drawdowns for signed portfolios. Full OOS Period refers to the full out-of-sample period, 01/1990$\sim$11/2016. Recession columns report conditional drawdowns computed within each recession window using a three-month look-back reference window ($S=3$ months). The $\fh$ benchmarked methods use fixed windows as described in Table \ref{tab-fh-benchmarks}. Smaller values indicate smaller peak-to-trough losses.}
\end{table}

\section{Conclusions}\label{sec-discussions}

Our empirical results demonstrate the practical value of our adaptive framework across multiple dimensions. Most notably, during periods of heightened economic stress, our adaptive method $\adaptive$ exhibits superior performance compared to fixed-window approaches. Our approach is well motivated by the documented facts in \myCref{sec-tradeoff-empirics} that, during recession periods including the 1990 Gulf War recession, the 2001 dot-com bubble burst and 9/11 attack, and the 2007--2009 Financial Crisis, simpler models trained on shorter windows consistently outperformed more complex models trained on longer windows. This empirical evidence validates our theory of the nonstationarity-complexity tradeoff.

The adaptive algorithm's performance during economic downturns is worth highlighting. As shown in \myCref{tab:oos_r2_industry_time}, $\adaptive$ achieves an out-of-sample $R^2$ of $0.027$ during the brief but severe 1990 Gulf War recession, while the best fixed-window benchmark $\fwlong$ produces a negative $R^2$ of $-0.031$. During the 2001 recession, $\adaptive$ attains an impressive $R^2$ of $0.125$, outperforming $\fwlong$ by 6.8\% (0.117). Even during the prolonged Global Financial Crisis of 2007-2009, $\adaptive$ maintains its advantage with an $R^2$ of $0.041$ compared to $\fwlong$'s $0.039$. Beyond statistical performance metrics, the economic significance of our approach is demonstrated through trading strategy analysis. Averaged across the industries, our model generates 31\% higher cumulative wealth net of trading costs over the full out-of-sample period, as shown in \Cref{fig-wealth-mv}.

Several future directions are worth exploring. First, our adaptive model selection framework relies on the assumption that data is independent across time even though the distribution can change arbitrarily. While numerical experiments show that our method is robust against temporal dependence in real-world financial time series, it would be interesting and important to extend the framework to handle dependent data in a principled way. Second, our framework of joint model and training window selection requires training a large number of candidate models, which can be computationally intensive. A valuable future direction is to reduce these training costs. For example, one might utilize the trained models from previous periods as warm starts for subsequent training.

{
\bibliographystyle{ims}
\bibliography{bib}

\begin{thebibliography}{50}
\expandafter\ifx\csname natexlab\endcsname\relax\def\natexlab#1{#1}\fi
\expandafter\ifx\csname url\endcsname\relax
  \def\url#1{\texttt{#1}}\fi
\expandafter\ifx\csname urlprefix\endcsname\relax\def\urlprefix{URL }\fi

\bibitem[{Adrian and Shin(2014)}]{AdrShi14}
\textsc{Adrian, T.} and \textsc{Shin, H.~S.} (2014).
\newblock Procyclical leverage and value-at-risk.
\newblock \textit{The Review of Financial Studies} \textbf{27} 373--403.

\bibitem[{Andrews(1993)}]{And93}
\textsc{Andrews, D. W.~K.} (1993).
\newblock Tests for parameter instability and structural change with unknown
  change point.
\newblock \textit{Econometrica} \textbf{61} 821--856.

\bibitem[{Bai and Perron(1998)}]{BPe98}
\textsc{Bai, J.} and \textsc{Perron, P.} (1998).
\newblock Estimating and testing linear models with multiple structural
  changes.
\newblock \textit{Econometrica} \textbf{66} 47--78.

\bibitem[{Banerjee and Urga(2005)}]{BUr05}
\textsc{Banerjee, A.} and \textsc{Urga, G.} (2005).
\newblock Modelling structural breaks, long memory and stock market volatility:
  an overview.
\newblock \textit{Journal of Econometrics} \textbf{129} 1--34.
\newblock Modelling structural breaks.

\bibitem[{Bartlett et~al.(2005)Bartlett, Bousquet and Mendelson}]{BBM05}
\textsc{Bartlett, P.~L.}, \textsc{Bousquet, O.} and \textsc{Mendelson, S.}
  (2005).
\newblock {Local Rademacher complexities}.
\newblock \textit{The Annals of Statistics} \textbf{33} 1497 -- 1537.

\bibitem[{Boucheron et~al.(2013)Boucheron, Lugosi and Massart}]{BLM13}
\textsc{Boucheron, S.}, \textsc{Lugosi, G.} and \textsc{Massart, P.} (2013).
\newblock \textit{Concentration Inequalities: A Nonasymptotic Theory of
  Independence}.
\newblock Oxford University Press.

\bibitem[{Chen et~al.(2024)Chen, Pelger and Zhu}]{CPZ24}
\textsc{Chen, L.}, \textsc{Pelger, M.} and \textsc{Zhu, J.} (2024).
\newblock Deep learning in asset pricing.
\newblock \textit{Management Science} \textbf{70} 714--750.

\bibitem[{Chib(1998)}]{Chi98}
\textsc{Chib, S.} (1998).
\newblock Estimation and comparison of multiple change-point models.
\newblock \textit{Journal of Econometrics} \textbf{86} 221--241.

\bibitem[{Chinco et~al.(2019)Chinco, Clark-Joseph and Ye}]{CCY19}
\textsc{Chinco, A.}, \textsc{Clark-Joseph, A.~D.} and \textsc{Ye, M.} (2019).
\newblock Sparse signals in the cross-section of returns.
\newblock \textit{The Journal of Finance} \textbf{74} 449--492.

\bibitem[{Choi et~al.(2025)Choi, Jiang and Zhang}]{ChoiJiangZhang2025}
\textsc{Choi, D.}, \textsc{Jiang, W.} and \textsc{Zhang, C.} (2025).
\newblock Alpha go everywhere: Machine learning and international stock
  returns.
\newblock \textit{The Review of Asset Pricing Studies} \textbf{15} 288--331.

\bibitem[{Chow(1960)}]{Cho60}
\textsc{Chow, G.~C.} (1960).
\newblock Tests of equality between sets of coefficients in two linear
  regressions.
\newblock \textit{Econometrica} \textbf{28} 591--605.

\bibitem[{Cormen et~al.(2022)Cormen, Leiserson, Rivest and Stein}]{CLRS22}
\textsc{Cormen, T.~H.}, \textsc{Leiserson, C.~E.}, \textsc{Rivest, R.~L.} and
  \textsc{Stein, C.} (2022).
\newblock \textit{Introduction to algorithms}.
\newblock MIT press.

\bibitem[{Didisheim et~al.(2024)Didisheim, Ke, Kelly and Malamud}]{DKK24}
\textsc{Didisheim, A.}, \textsc{Ke, S.~B.}, \textsc{Kelly, B.~T.} and
  \textsc{Malamud, S.} (2024).
\newblock {APT} or {``AIPT''}? {T}he surprising dominance of large factor
  models.
\newblock Tech. rep., National Bureau of Economic Research.

\bibitem[{Fama and French(1993)}]{FFr93}
\textsc{Fama, E.~F.} and \textsc{French, K.~R.} (1993).
\newblock Common risk factors in the returns on stocks and bonds.
\newblock \textit{Journal of Financial Economics} \textbf{33} 3--56.

\bibitem[{Fama and French(1996)}]{fama1996multifactor}
\textsc{Fama, E.~F.} and \textsc{French, K.~R.} (1996).
\newblock Multifactor explanations of asset pricing anomalies.
\newblock \textit{The Journal of Finance} \textbf{51} 55--84.

\bibitem[{Fama and French(2015)}]{FFr15}
\textsc{Fama, E.~F.} and \textsc{French, K.~R.} (2015).
\newblock A five-factor asset pricing model.
\newblock \textit{Journal of Financial Economics} \textbf{116} 1--22.

\bibitem[{Freyberger et~al.(2020)Freyberger, Neuhierl and Weber}]{FNW20}
\textsc{Freyberger, J.}, \textsc{Neuhierl, A.} and \textsc{Weber, M.} (2020).
\newblock Dissecting characteristics nonparametrically.
\newblock \textit{The Review of Financial Studies} \textbf{33} 2326--2377.

\bibitem[{Goldenshluger and Lepski(2008)}]{GLe08}
\textsc{Goldenshluger, A.} and \textsc{Lepski, O.} (2008).
\newblock {Universal pointwise selection rule in multivariate function
  estimation}.
\newblock \textit{Bernoulli} \textbf{14} 1150 -- 1190.

\bibitem[{Gu et~al.(2020)Gu, Kelly and Xiu}]{GKX20}
\textsc{Gu, S.}, \textsc{Kelly, B.} and \textsc{Xiu, D.} (2020).
\newblock Empirical asset pricing via machine learning.
\newblock \textit{The Review of Financial Studies} \textbf{33} 2223--2273.

\bibitem[{Hamilton(1989)}]{Ham89}
\textsc{Hamilton, J.~D.} (1989).
\newblock A new approach to the economic analysis of nonstationary time series
  and the business cycle.
\newblock \textit{Econometrica} \textbf{57} 357--384.

\bibitem[{Han et~al.(2024)Han, Huang and Wang}]{HHW24}
\textsc{Han, E.}, \textsc{Huang, C.} and \textsc{Wang, K.} (2024).
\newblock Model assessment and selection under temporal distribution shift.
\newblock In \textit{Proceedings of the 41st International Conference on
  Machine Learning}, vol. 235 of \textit{Proceedings of Machine Learning
  Research}. PMLR.

\bibitem[{Hansen and Jagannathan(1991)}]{HJa91}
\textsc{Hansen, L.~P.} and \textsc{Jagannathan, R.} (1991).
\newblock Implications of security market data for models of dynamic economies.
\newblock \textit{Journal of Political Economy} \textbf{99} 225--262.

\bibitem[{Harvey et~al.(2025)Harvey, Mazzoleni and
  Melone}]{harvey2025unintended}
\textsc{Harvey, C.~R.}, \textsc{Mazzoleni, M.~G.} and \textsc{Melone, A.}
  (2025).
\newblock The unintended consequences of rebalancing.
\newblock Tech. rep., National Bureau of Economic Research.

\bibitem[{Heckman(1979)}]{Hec79}
\textsc{Heckman, J.~J.} (1979).
\newblock Sample selection bias as a specification error.
\newblock \textit{Econometrica} \textbf{47} 153--161.

\bibitem[{Homm and Breitung(2012)}]{HBr12}
\textsc{Homm, U.} and \textsc{Breitung, J.} (2012).
\newblock Testing for speculative bubbles in stock markets: A comparison of
  alternative methods.
\newblock \textit{Journal of Financial Econometrics} \textbf{10} 198--231.

\bibitem[{Inoue et~al.(2017)Inoue, Jin and Rossi}]{IJR17}
\textsc{Inoue, A.}, \textsc{Jin, L.} and \textsc{Rossi, B.} (2017).
\newblock Rolling window selection for out-of-sample forecasting with
  time-varying parameters.
\newblock \textit{Journal of Econometrics} \textbf{196} 55--67.

\bibitem[{Jacot et~al.(2018)Jacot, Gabriel and Hongler}]{JGH18}
\textsc{Jacot, A.}, \textsc{Gabriel, F.} and \textsc{Hongler, C.} (2018).
\newblock Neural tangent kernel: Convergence and generalization in neural
  networks.
\newblock \textit{Advances in neural information processing systems}
  \textbf{31}.

\bibitem[{Jones and Kaul(1996)}]{JonKau96}
\textsc{Jones, C.~M.} and \textsc{Kaul, G.} (1996).
\newblock Oil and the stock markets.
\newblock \textit{The Journal of Finance} \textbf{51} 463--491.

\bibitem[{Kelly et~al.(2024{\natexlab{a}})Kelly, Kuznetsov, Malamud, Xu and
  Zhang}]{kelly2024large}
\textsc{Kelly, B.}, \textsc{Kuznetsov, B.}, \textsc{Malamud, S.}, \textsc{Xu,
  T.~A.} and \textsc{Zhang, Y.} (2024{\natexlab{a}}).
\newblock {Large and Deep Factor Models}.

\bibitem[{Kelly et~al.(2024{\natexlab{b}})Kelly, Malamud and Zhou}]{KMZ24}
\textsc{Kelly, B.}, \textsc{Malamud, S.} and \textsc{Zhou, K.}
  (2024{\natexlab{b}}).
\newblock The virtue of complexity in return prediction.
\newblock \textit{The Journal of Finance} \textbf{79} 459--503.

\bibitem[{Kelly and Xiu(2023)}]{KXi23}
\textsc{Kelly, B.} and \textsc{Xiu, D.} (2023).
\newblock Financial machine learning.
\newblock \textit{Foundations and Trends{\textregistered} in Finance}
  \textbf{13} 205--363.

\bibitem[{Kelly and Malamud(2025)}]{KellyMalamud2025Understanding}
\textsc{Kelly, B.~T.} and \textsc{Malamud, S.} (2025).
\newblock Understanding the virtue of complexity.
\newblock Tech. Rep. 25-96, Swiss Finance Institute.

\bibitem[{Kelly et~al.(2022)Kelly, Malamud and Zhou}]{Kelly2022Virtue}
\textsc{Kelly, B.~T.}, \textsc{Malamud, S.} and \textsc{Zhou, K.} (2022).
\newblock The virtue of complexity everywhere.
\newblock Tech. Rep. 22-57, Swiss Finance Institute.

\bibitem[{Koh et~al.(2021)Koh, Sagawa, Marklund, Xie, Zhang, Balsubramani, Hu,
  Yasunaga, Phillips, Gao, Lee, David, Stavness, Guo, Earnshaw, Haque, Beery,
  Leskovec, Kundaje, Pierson, Levine, Finn and Liang}]{KSM21}
\textsc{Koh, P.~W.}, \textsc{Sagawa, S.}, \textsc{Marklund, H.}, \textsc{Xie,
  S.~M.}, \textsc{Zhang, M.}, \textsc{Balsubramani, A.}, \textsc{Hu, W.},
  \textsc{Yasunaga, M.}, \textsc{Phillips, R.~L.}, \textsc{Gao, I.},
  \textsc{Lee, T.}, \textsc{David, E.}, \textsc{Stavness, I.}, \textsc{Guo,
  W.}, \textsc{Earnshaw, B.}, \textsc{Haque, I.}, \textsc{Beery, S.~M.},
  \textsc{Leskovec, J.}, \textsc{Kundaje, A.}, \textsc{Pierson, E.},
  \textsc{Levine, S.}, \textsc{Finn, C.} and \textsc{Liang, P.} (2021).
\newblock {WILDS}: A benchmark of in-the-wild distribution shifts.
\newblock In \textit{Proceedings of the 38th International Conference on
  Machine Learning}, vol. 139 of \textit{Proceedings of Machine Learning
  Research}. PMLR.

\bibitem[{Kozak et~al.(2020)Kozak, Nagel and Santosh}]{KNS20}
\textsc{Kozak, S.}, \textsc{Nagel, S.} and \textsc{Santosh, S.} (2020).
\newblock Shrinking the cross-section.
\newblock \textit{Journal of Financial Economics} \textbf{135} 271--292.

\bibitem[{Kpotufe and Martinet(2021)}]{KMa21}
\textsc{Kpotufe, S.} and \textsc{Martinet, G.} (2021).
\newblock {Marginal singularity and the benefits of labels in covariate-shift}.
\newblock \textit{The Annals of Statistics} \textbf{49} 3299 -- 3323.

\bibitem[{Liu and Maheu(2007)}]{LMa07}
\textsc{Liu, C.} and \textsc{Maheu, J.~M.} (2007).
\newblock Are there structural breaks in realized volatility?
\newblock \textit{Journal of Financial Econometrics} \textbf{6} 326--360.

\bibitem[{Ma et~al.(2023)Ma, Pathak and Wainwright}]{MPW23}
\textsc{Ma, C.}, \textsc{Pathak, R.} and \textsc{Wainwright, M.~J.} (2023).
\newblock {Optimally tackling covariate shift in RKHS-based nonparametric
  regression}.
\newblock \textit{The Annals of Statistics} \textbf{51} 738 -- 761.

\bibitem[{Massart(2000)}]{Mas00}
\textsc{Massart, P.} (2000).
\newblock {About the constants in Talagrand's concentration inequalities for
  empirical processes}.
\newblock \textit{The Annals of Probability} \textbf{28} 863 -- 884.

\bibitem[{McLean and Pontiff(2016)}]{MPo16}
\textsc{McLean, R.~D.} and \textsc{Pontiff, J.} (2016).
\newblock Does academic research destroy stock return predictability?
\newblock \textit{The Journal of Finance} \textbf{71} 5--32.

\bibitem[{Mendelson(2002)}]{Men02}
\textsc{Mendelson, S.} (2002).
\newblock Geometric parameters of kernel machines.
\newblock In \textit{Computational Learning Theory}. Springer, Berlin,
  Heidelberg.

\bibitem[{P{\'a}stor and Stambaugh(2003)}]{PasSta03}
\textsc{P{\'a}stor, {\v{L}}.} and \textsc{Stambaugh, R.~F.} (2003).
\newblock Liquidity risk and expected stock returns.
\newblock \textit{Journal of Political Economy} \textbf{111} 642--685.

\bibitem[{Patil et~al.(2024)Patil, Du and Tibshirani}]{PDT24}
\textsc{Patil, P.}, \textsc{Du, J.-H.} and \textsc{Tibshirani, R.} (2024).
\newblock Optimal ridge regularization for out-of-distribution prediction.
\newblock In \textit{Proceedings of the 41st International Conference on
  Machine Learning}, vol. 235 of \textit{Proceedings of Machine Learning
  Research}. PMLR.

\bibitem[{Pesaran and Pick(2011)}]{PPi11}
\textsc{Pesaran, M.~H.} and \textsc{Pick, A.} (2011).
\newblock Forecast combination across estimation windows.
\newblock \textit{Journal of Business \& Economic Statistics} \textbf{29}
  307--318.

\bibitem[{Pesaran and Timmermann(2007)}]{PTi07}
\textsc{Pesaran, M.~H.} and \textsc{Timmermann, A.} (2007).
\newblock Selection of estimation window in the presence of breaks.
\newblock \textit{Journal of Econometrics} \textbf{137} 134--161.

\bibitem[{Sugiyama and Kawanabe(2012)}]{SKa12}
\textsc{Sugiyama, M.} and \textsc{Kawanabe, M.} (2012).
\newblock \textit{Machine learning in non-stationary environments: Introduction
  to covariate shift adaptation}.
\newblock MIT press.

\bibitem[{Tsybakov(2008)}]{Tsy08}
\textsc{Tsybakov, A.~B.} (2008).
\newblock \textit{Introduction to Nonparametric Estimation}.
\newblock 1st ed. Springer Publishing Company, Incorporated.

\bibitem[{Wahba(1990)}]{Wah90}
\textsc{Wahba, G.} (1990).
\newblock \textit{Spline models for observational data}.
\newblock SIAM.

\bibitem[{Wainwright(2019)}]{Wai19}
\textsc{Wainwright, M.~J.} (2019).
\newblock \textit{High-Dimensional Statistics: A Non-Asymptotic Viewpoint}.
\newblock Cambridge Series in Statistical and Probabilistic Mathematics,
  Cambridge University Press.

\bibitem[{Wang(2026)}]{Wan26}
\textsc{Wang, K.} (2026).
\newblock {Pseudo-Labeling for kernel ridge regression under covariate shift}.
\newblock \textit{The Annals of Statistics} \textbf{54} 252 -- 276.

\end{thebibliography}
}

\newpage

\appendix

\vspace{.5cm}
\begin{center}
    \huge\textbf{Appendix}
\end{center}

\section{Additional Experiment Details for \myCref{sec-experiments}}\label{sec-experiment-details}

\subsection{Data}\label{sec-dataapdx}
We use the $17$ industry portfolio returns from Kenneth French's website\footnote{\url{https://mba.tuck.dartmouth.edu/pages/faculty/ken.french/Data_Library/det_17_ind_port.html}}. We first describe our rich set of covariates, which combines daily and monthly sources to construct a comprehensive time series combining macroeconomic and cross-sectional signals. The final daily-frequency sample spans the time period from September 1987 to November 2016.

\paragraph{Common Factors.} To capture the underlying macroeconomic and systematic risk structure in our response variables, we include 15 common factors from the monthly dataset of \citet{CPZ24}, who develop a deep learning framework for estimating the stochastic discount factor (SDF) consistent with the no-arbitrage condition.\footnote{Their model integrates three neural network components: a feedforward network to approximate the nonlinear functional form of the SDF, a recurrent Long-Short Term Memory (LSTM) network to extract macroeconomic state variables, and a generative adversarial network (GAN) that constructs the most informative test assets for pricing. This architecture yields an empirically estimated SDF that represents the conditional mean-variance efficient portfolio of all U.S. equities.}

The \citet{CPZ24} dataset provides three distinct elements relevant to our analysis:  
(i) the estimated stochastic discount factor (SDF), capturing the aggregate price of risk implied by their no-arbitrage model;  
(ii) the returns of ten equal-weighted decile portfolios sorted by firms' exposure to the SDF (beta-sorted), which serve as cross-sectional test assets; and  
(iii) four macroeconomic hidden state variables, derived from the GAN model, that summarize the joint dynamics of 178 macroeconomic time series into a small set of nonlinear factors reflecting business cycle conditions and systemic risk.\footnote{These components jointly span the main drivers of systematic variation in expected returns. The SDF and beta-sorted deciles capture the cross-sectional risk-return trade-offs, while the macroeconomic hidden states track time variation in the pricing kernel associated with expansions, recessions, and crisis periods. The dataset covers the period 1967-2016; to align frequencies with our daily response data, each monthly observation is assigned to all trading days within the corresponding month.}

In addition to these 15 factors, we incorporate the daily Fama-French three factors (FF3) from \citet{FFr93}, which include the excess market return (MKT), the size factor (SMB), and the value factor (HML). These factors provide a benchmark set of linear risk exposures that have been shown to explain broad cross-sectional patterns in stock returns. Each factor is lagged appropriately to avoid look-ahead bias.

 This unified set of common factors allows us to incorporate both the traditional linear risk structure captured by FF3 and the nonlinear, macroeconomically conditioned dynamics extracted from \citet{CPZ24}.

\paragraph{Firm-Level Returns and Characteristics.} We augment the dataset with daily firm-level stock returns obtained from the CRSP database. Excess returns are computed relative to the daily risk-free rate from Kenneth French's data library.\footnote{To mitigate the impact of outliers that can distort cross-sectional averages, we Winsorize the daily cross-section of excess returns by removing the top and bottom 1\% of firms within each day based on their excess return values. This procedure ensures that the resulting portfolio and factor constructions are not unduly influenced by extreme realizations.} To characterize the cross-sectional heterogeneity in firm fundamentals and trading behavior, we incorporate the comprehensive panel of firm-level characteristics from \citet{GKX20}. Their dataset provides 94 standardized monthly characteristics for U.S. equities, constructed to ensure comparability across firms and over time. 

Specifically, \cite{GKX20} identify three dominant groups of predictive signals: (i) price trend variables: measures of short- and long-term momentum, industry momentum, and short-term reversal that capture the persistence of returns and investor underreaction; (ii) liquidity and size variables: market capitalization, trading volume, turnover, and bid-ask spread, which reflect trading frictions and the limits to arbitrage; (iii) volatility and risk variables: total and idiosyncratic volatility, market beta, and higher-order terms such as beta squared, which proxy for systematic and residual risk exposure.\footnote{As the \cite{GKX20} characteristics are available only at a monthly frequency, we assign each monthly vector of characteristics to all trading days within that month for the corresponding firm. This alignment ensures consistency between the daily CRSP return data and the lower-frequency fundamental information.}

\subsubsection{Size Filter}

We apply a size-based filter to focus the analysis on the subset of firms that are most representative of the small-cap universe. Specifically, within each day $t$, we retain only the bottom 25\% of firms by market equity.\footnote{We measure market equity with the (monthly) ``mvel1'' size covariate from \citet{GKX20}. It represents the firm's market capitalization at the end of month $m$, standardized for comparability across firms and time.}

Formally, let $Z_{i,t}$ denote the size of firm $i$ in month $t$. For all trading days in month $t$, we retain only those firms satisfying $Z_{i,t} \le Q_{0.25}\big( \{ Z_{j,t} \}_{j \in \cI_t} \big)$, where $Q_{0.25}(\cdot)$ denotes the 25th percentile of the cross-sectional size distribution, and $\cI_t$ is the set of all firms available in month $t$. Restricting the sample in this way allows us to study the predictive role of firm characteristics and macroeconomic factors within a homogeneous segment of the equity market, mitigating heterogeneity arising from scale effects in firm size.

\subsubsection{Long-Short Decile Portfolios from Characteristics}

To transform the firm-level characteristics into tradable, interpretable covariates, we construct daily long-short portfolios sorted on each of the \citet{GKX20} characteristics. For each of the 94 characteristics, denoted by $c \in [94]$, and for each day $t$, we sort firms into ten equal-weighted portfolios based on the value of characteristic $c$.

Formally, let $X_{i,c,t}$ denote the value of characteristic $c$ for firm $i$ in month $t$, which is assigned to all trading days within that month, and let $R_{i,s}^{\text{ex}}$ denote the daily excess return of firm $i$ on day $s$. We compute monthly breakpoints for characteristic $c$ using cross-sectional quantiles $b_{c,k,t} = Q_{k/10} \big( \{ X_{i,c,t} \}_{i \in \cI_s} \big)$, $k\in[10]$, where $Q_{p}(\cdot)$ denotes the $p$-th quantile, and $\cI_s$ is the set of all firms observed on day $s$. 

We then compute the daily return of the $k$-th decile portfolio for characteristic $c$ as:
\[
D_{c,k,s} = \frac{1}{N_{c,k,s}} \sum_{i:\, b_{c,k-1,t} < X_{i,c,t} \le b_{c,k,t}} R_{i,s}^{\text{ex}},
\]
where $N_{c,k,s}$ is the number of firms assigned to the $k$-th decile on day $s$. Each decile is equal-weighted to ensure that portfolio performance reflects cross-sectional variation in firm characteristics rather than differences in market capitalization.

The corresponding long-short portfolio return for characteristic $c$ is defined as $\text{LS}_{c,s} = D_{c,10,s} - D_{c,1,s}$, representing the daily return to a strategy that longs the highest-decile firms (with the largest $X_{i,c,t}$) and shorts the lowest-decile firms (with the smallest $X_{i,c,t}$).

This construction yields a balanced set of long-short factor returns that isolate the pricing implications of each firm characteristic. Repeating this procedure for all $c \in [94]$ produces a time series matrix of 94 characteristic-sorted long-short portfolio returns at daily frequency.

\subsubsection{Final Covariate Set}

Our final set of predictors integrates macroeconomic, factor-based, and cross-sectional sources of variation to form a unified time series of covariates for forecasting the French 17 industry portfolio returns. The resulting dataset combines information from three complementary dimensions of the asset pricing literature:

\begin{enumerate}
    \item \textbf{Macroeconomic and systematic factors:} the 10 equal-weighted beta-sorted decile portfolios, the 4 macroeconomic hidden states, and the estimated stochastic discount factor (SDF) from \citet{CPZ24}. 
    
    \item \textbf{Benchmark risk factors:} the daily Fama-French 3 factors (FF3) from \citet{FFr93}, consisting of the market excess return (MKT), the size factor (SMB), and the value factor (HML). 
    
    \item \textbf{Cross-sectional characteristic factors:} the 94 daily long-short characteristic-sorted portfolio returns, from the \citet{GKX20} characteristics.

    \item \textbf{Lagging features:} we augment the feature set with one-day lagged returns. Specifically, we use the full vector of the $17$ industry portfolio lagged returns.
\end{enumerate}

By combining the first three components, our dataset links the macroeconomic and cross-sectional perspectives on asset pricing. The \citet{CPZ24} factors embed the nonlinear, time-varying structure of the stochastic discount factor, while the \citet{GKX20} characteristic-sorted long-short portfolios summarize the cross-sectional distribution of risk premia across firms. Incorporating the Fama-French 3 factors provides a benchmark for evaluating whether these modern, high-dimensional representations offer predictive power beyond the traditional linear framework.

\subsection{Name Mapping for the 17 Industries}
\begin{table}[h]
    \centering
    \caption{Name Mapping for the 17 Industries}
    \label{tab-industry-name-mapping}
    \begin{tabular}{@{}ll@{}} 
        \toprule
        Industry Acronym & Full Industry Name \\
        \midrule
        Food  & Food \\
        Mines & Mining and Minerals \\
        Oil   & Oil and Petroleum Products \\
        Clths & Textiles, Apparel \& Footwear \\
        Durbl & Consumer Durables \\
        Chems & Chemicals \\
        Cnsum & Drugs, Soap, Perfumes, Tobacco \\
        Cnstr & Construction and Construction Materials \\
        Steel & Steel Works Etc \\
        FabPr & Fabricated Products \\
        Machn & Machinery and Business Equipment \\
        Cars  & Automobiles \\
        Trans & Transportation \\
        Utils & Utilities \\
        Rtail & Retail Stores \\
        Finan & Banks, Insurance Companies, and Other Financials \\
        Other & Other \\
        \bottomrule
    \end{tabular}
\end{table}

\subsection{Machine Learning Model Implementation Details}\label{sec-hyperparameters}


\subsubsection{Model Hyperparameter Grids}

For each model class considered in our analysis, we systematically explore a comprehensive grid of hyperparameter values for model selection. The hyperparameter grids are designed to balance computational efficiency with thorough exploration of the model space.

\paragraph{Linear Models with Regularization.}

For the ridge regression ($\ridge$), LASSO ($\lasso$), and elastic net ($\enet$) models, we consider the following hyperparameter specifications:

\begin{enumerate}
    \item For ridge regression, we consider values of the regularization parameter $\alpha$ on a logarithmic scale: $\alpha \in \{10^{-3}, 10^{-1.5}, 1, 10^{1.5}, 10^3\}$. This range allows for both strong regularization (small $\alpha$) and weak regularization (large $\alpha$), accommodating different levels of multicollinearity in our high-dimensional covariate space.
    \item For LASSO, we consider values of the regularization parameter $\alpha$ on a logarithmic scale: $\alpha \in \{ 10^{-5}, 10^{-3.5}, 10^{-2}, 10^{-0.5}, 10 \}$. The $\ell_1$ penalty in LASSO facilitates feature selection, which is particularly valuable given our large set of covariates.
    \item For the elastic net, we consider the following combinations of the regularization parameter $\alpha$ and the mixing parameter $r$: $\alpha \in \{ 10^{-3}, 1, 10^3 \}$ and $r\in \{0.01, 0.05, 0.1 \}$. This grid explores the balance between feature selection and coefficient shrinkage.
\end{enumerate}

\paragraph{Random Forest.} For the random forest models, we consider the following combinations of the number of trees $n_{\texttt{tree}}$ and the maximum tree depth $d_{\max}$: $n_{\texttt{tree}}\in \{10, 100, 200\}$ and $d_{\max} \in \{ 3, 5, 10 \}$. Increasing the number of trees generally improves model stability and reduces variance. Trees with shallower depths provide stronger regularization, while deeper trees can capture more complex nonlinear relationships.

\subsubsection{Hyperparameters for $\adaptive$}

In our implementation of $\adaptive$, we set $\delta'=0.1$ and $M'=10^{-3}$. \Cref{sec-hyperparam-ATOMS-ablation} provides further details on the choice of $M'$ and reports ablation studies showing that the performance of $\adaptive$ remains stable when $M'$ is doubled or halved.

\newpage

\vspace{.5cm}
\begin{center}
    \huge\textbf{Internet Appendix}
\end{center}

\section{Theoretical Analysis of the Nonstationarity-Complexity Tradeoff}\label{sec-tradeoff-theory-appendix}

In this section, we provide more details on \myCref{thm-tradeoff} and its proof. 

We first set up some mathematical notation. Recall that $f_t^*(\cdot) = \EE_{(\bx,y)\sim\distP_t}[y \mid \bx=\cdot ]$ is the Bayes optimal least squares estimator, which minimizes $L_t(f)$ over all measurable $f:\featurespace\to\RR$. Let
\[
\bar{f}_t = \argmin_{f\in\modelclass} L_t(f),
\]
which minimizes $L_t(f)$ over all $f\in\modelclass$. For each $t$, let $\distP_{t,\bx}$ be the marginal distribution of $\distP_t$ with respect to the covariates $\bx$. The distribution $\distP_{t,\bx}$ induces an inner product $\langle \cdot,\cdot\rangle_t$ and norm $\|\cdot\|_t$, given by
\[
\langle f, g \rangle_t = \EE_{\bx\sim\distP_{t,\bx}}\left[ f(\bx)g(\bx) \right]
\qquad\text{and}\qquad
\| f \|_t = 
\sqrt{\langle f,f\rangle_t}
=
\sqrt{\EE_{\bx\sim\distP_{t,\bx}}\left[ f(\bx)^2 \right]}.
\]
It can be easily shown that $\excessrisk_t(f) = \|f - f_t^* \|_t^2$ for all $f:\cX\to\RR$.

\subsection{A Non-Stationary Local Rademacher Complexity} 

We now formally define $r_{t,k}(\modelclass)$ through a non-stationary version of the \emph{local Rademacher complexity} \citep{BBM05}. We first define the \emph{Rademacher complexity}, which reflects the richness of a function class with respect to certain data samples.

\begin{definition}[Rademacher complexity]
Let $\{\bz_i\}_{i=1}^n$ be independent random variables in $\featurespace$. Let $\cG$ be a class of functions from $\featurespace$ to $\RR$. The \textbf{Rademacher complexity} of $\cG$ associated with $\{\bz_i\}_{i=1}^n$ is defined by
\[
\Rademacher(\cG;\{\bz_i\}_{i=1}^n) = \EE \left[ \sup_{g\in\cG} \frac{1}{n}  \sum_{i=1}^n\varepsilon_ig(\bz_i) \right],
\]
where $\varepsilon_1,...,\varepsilon_n$ are i.i.d.~random variables following the Rademacher distribution $\PP(\varepsilon_1=1)=\PP(\varepsilon_1=-1)=1/2$, and are independent of $\{z_i\}_{i=1}^n$.
\end{definition}

The local Rademacher complexity is the Rademacher complexity of some local function class centered at $\bar{f}_t$. Let $\ntrain_j = |\batchtr_j|$.

\begin{definition}[Local function class]
For every $r\ge 0$, define the local function class
\[
\modelclass_{t,k}(r) = \bigg\{ f\in\modelclass : \frac{1}{\ntrain_{t,k}} \sum_{j=t-k}^{t-1} \ntrain_j \left\| f - \bar{f}_t \right\|_j^2 \le r \bigg\}.
\]
\end{definition}

In words, $\modelclass_{t,k}(r)$ consists of functions $f\in\modelclass$ that are close to $\bar{f}_t$ with respect to the distributions $\{\distP_{j,\bx}\}_{j=t-k}^{t-1}$ on average. We are now ready to define the quantity $r_{t,k}(\modelclass)$, which is more formally known as the \emph{critical radius} of the function
\[
\Rademacher_{t,k}(r;\modelclass) = \Rademacher \left( \modelclass_{t,k}(r) ; \dataset_{t,k}^{\tr} \right). 
\]

\begin{definition}[Subroot function]
A function $\psi:\RR_+\to\RR_+$ is \textbf{subroot} if it is increasing and $r\mapsto\psi(r)/\sqrt{r}$ is decreasing on $(0,\infty)$.
\end{definition}

\begin{definition}[Critical radius]
The \textbf{critical radius} of $\Rademacher_{t,k}(r;\modelclass)$ is defined by
\[
r_{t,k}(\modelclass) = \inf\left\{ r\ge 0 : \exists\,\text{subroot } \psi \text{ such that } \psi(r)=r, \text{ and } \psi(s) \ge \Rademacher_{t,k}(s;\modelclass) \ \forall s \ge r \right\}.
\]
\end{definition}

\subsection{Proof of \myCref{thm-tradeoff}}\label{sec-thm-tradeoff-proof}

We now prove \myCref{thm-tradeoff}. We will prove a more refined result that decomposes the non-stationarity into conditional shift and covariate shift, where
\begin{itemize}
    \item the conditional shift between a past time period $\tau$ and the current time $t$ is measured by $\| f_{\tau}^* - f_{t}^* \|_t$;
    \item the covariate shift between time $\tau$ and $t$ is characterized by the following integral probability metric:
    \begin{equation}\label{eqn-IPM-covariate-shift}
        D_t\left(\distP_{\tau,\bx},\distP_{t,\bx} \right) 
    =
    \sup_{h\in(\partial\modelclass_t)^2}\left| \EE_{\bx\sim\distP_{\tau,\bx}} [h (\bx)] - \EE_{\bx\sim\distP_{t,\bx}} [h (\bx)] \right|,
    \end{equation}
    with $(\partial\modelclass_t)^2 = \left\{(f-f_{\tau}^*)^2:f\in\modelclass, \, \tau\in[t] \right\}$.
\end{itemize}

We now state the more refined prediction error bound.

\begin{theorem}[Refined prediction error bound]\label{thm-tradeoff-refined}
Let Assumptions \ref{assumption-independence} and \ref{assumption-bounded} hold, and fix $\delta\in(0,1)$. With probability at least $1-\delta$, the model $\widehat{f}$ defined by \eqref{eqn-ERM} satisfies
\[
\excessrisk_t(\widehat{f})
\lesssim 
\min_{f\in\modelclass}\excessrisk_t(f)
+
M^2\left(
r_{t,k}(\modelclass)
+
\frac{\log(1 / \delta)}{\ntrain_{t,k}}
\right)
+
\max_{t-k\le {\tau}\le t-1} \left\{ \| f_{\tau}^* - f_{t}^* \|_t^2 + D_t\left(\distP_{\tau,\bx},\distP_{t,\bx} \right) \right\}.
\]
Here $\lesssim$ hides a universal constant, and $D_t\left(\distP_{\tau,\bx},\distP_{t,\bx} \right)$ is defined by \eqref{eqn-IPM-covariate-shift}.
\end{theorem}

From \Cref{thm-tradeoff-refined}, we can prove \Cref{thm-tradeoff} by showing that both the conditional shift and covariate shift terms are $O\left(M^2\cdot\TV(\distP_{\tau},\distP_t)\right)$. To bound the conditional shift term, for each $\tau$,
\begin{align*}
\| f_{\tau}^* - f_{t}^* \|_t^2
&=
L_t(f_{\tau}^*) - L_t(f_t^*) \\[4pt]
&=
\big[ L_t(f_{\tau}^*) - L_{\tau}(f_{\tau}^*) \big]
+
\big[ L_{\tau}(f_{\tau}^*) - L_{\tau}(f_t^*) \big]
+
\big[ L_{\tau}(f_t^*) - L_t(f_t^*) \big] \\[4pt]
&\le 
\big[ L_t(f_{\tau}^*) - L_{\tau}(f_{\tau}^*) \big]
+ 0 +
\big[ L_{\tau}(f_t^*) - L_t(f_t^*) \big].
\end{align*}
For all $f:\cX\to[-M,M]$, we have
\[
|L_t(f) - L_{\tau}(f)|
=
\left| \EE_{(\bx,y)\sim\distP_t}\left[ (f(\bx)-y)^2 \right] - \EE_{(\bx,y)\sim\distP_{\tau}}\left[ (f(\bx)-y)^2 \right] \right|
\le 4M^2\cdot\TV(\distP_t,\distP_{\tau}).
\]
Thus,
\begin{equation}\label{eqn-thm-tradeoff-proof-TV-1}
  \| f_{\tau}^* - f_{t}^* \|_t^2 \le 8M^2 \cdot \TV(\distP_t,\distP_{\tau}).  
\end{equation}

To bound the covariate shift term, since every $h\in(\partial\modelclass_t)^2$ satisfies $|h(\bx)|\le 4M^2$ for all $\bx\in\cX$, then
\begin{align}
    D_t\left(\distP_{\tau,\bx},\distP_{t,\bx} \right) 
    =
    \sup_{h\in(\partial\modelclass_t)^2}\left| \EE_{\bx\sim\distP_{\tau,\bx}} [h (\bx)] - \EE_{\bx\sim\distP_{t,\bx}} [h (\bx)] \right|
    &\le 
    4M^2\cdot \TV(\distP_{\tau,\bx}, \distP_{t,\bx}) \notag \\[4pt]
    &\le 
    4M^2\cdot \TV(\distP_{\tau}, \distP_{t}). \label{eqn-thm-tradeoff-proof-TV-2}
\end{align}
Combining \eqref{eqn-thm-tradeoff-proof-TV-1} and \eqref{eqn-thm-tradeoff-proof-TV-2}, we obtain
\[
\max_{t-k\le {\tau}\le t-1} \left\{ \| f_{\tau}^* - f_{t}^* \|_t^2 + D_t\left(\distP_{\tau,\bx},\distP_{t,\bx} \right) \right\}
\le 
12M^2\cdot \max_{t-k\le {\tau}\le t-1}\TV(\distP_{\tau},\distP_t).
\]

In the next section, we prove \Cref{thm-tradeoff-refined}.

\subsection{Proof of \Cref{thm-tradeoff-refined}}
In the proof, we will write $\widehat{f}=\widehat{f}_{t,k}$ to emphasize its dependence on the time $t$ and the training window size $k$. The key of the proof is the following lemma, which is a non-stationary version of Theorem 3.3 in \cite{BBM05}. 

\begin{lemma}[Localized uniform concentration]\label{lem-local-uniform-concentration}
Let $\bz_1,...,\bz_n$ be independent random variables taking values in a space $\cZ\subseteq\RR^{d+1}$. Let $\cG$ be a collection of functions from $\cZ$ to $[a,b]$. Suppose that there exist $T:\cG\to\RR_+$ and $C,\eta_1,...,\eta_n\ge0$ such that the following noise condition holds:
\begin{equation}\label{eqn-noise-condition}
\frac{1}{n}\sum_{i=1}^n\var\left[g(\bz_i)\right] \le T(g) \le C\cdot \frac{1}{n}\sum_{i=1}^n \big(  \EE \left[g(\bz_i)\right] + \eta_i \big),\qquad\forall g\in\cG.
\end{equation}
Let $\psi$ be a sub-root function with a fixed point $r^*$ satisfying
\[
\psi(r) \ge C\cdot \Rademacher\big(\{g\in\cG:T(g)\le r\};\{\bz_i\}_{i=1}^n\big),\qquad\forall r\ge r^*.
\]
Take $\delta\in(0,1)$. With probability at least $1-\delta$, for all $g\in\cG$ and $K>1$,
\begin{multline*}
    \frac{1}{n}\sum_{i=1}^n \EE \left[ g(\bz_i) \right] \\
\le
\frac{K}{K-1}\cdot\frac{1}{n}\sum_{i=1}^n  g(\bz_i) 
+
c\left[
\frac{K}{C}r^* 
+
\left((b-a)+CK\right)\frac{\log(1/\delta)}{n}
\right] + \frac{1}{K-1}\cdot\frac{1}{n}\sum_{i=1}^n \eta_i.
\end{multline*}
Here $c>0$ is a universal constant.
\end{lemma}

\begin{proof}[Proof of \myCref{lem-local-uniform-concentration}]
See \myCref{sec-proof-lem-local-uniform-concentration}.
\end{proof}

For every $f\in\modelclass$, define $\ell_f:\featurespace\times\RR\to\RR$ by $\ell_f((\bx,y)) = [f(\bx) - y]^2$. We also denote $(\bx,y)$ by $\bz$. In \myCref{lem-local-uniform-concentration}, take $\{\bz_i\}_{i=1}^n = \cB_{t,k}^{\tr}$ and $\cG = \{\ell_f-\ell_{\bar{f}_t}:f\in\modelclass\}$. The following lemma suggests a choice of $g_i$ and $T$ for which \eqref{eqn-noise-condition} holds. 

\begin{lemma}[Noise condition]\label{lem-noise-condition-regression}
Let Assumption \myref{assumption-bounded} hold. For all $f,\bar{f}\in\modelclass$,
\begin{align*}
&\frac{1}{\ntrain_{t,k}}\sum_{j=t-k}^{t-1}\ntrain_j\var_{\bz\sim\distP_j}\left[\ell_f(\bz) - \ell_{\bar{f}}(\bz)\right] \\[4pt]
&\qquad\le 
16M^2 \cdot \frac{1}{\ntrain_{t,k}} \sum_{j=t-k}^{t-1} \ntrain_j \|f - \bar{f}\|_j^2 \\[4pt]
&\qquad\le 
32M^2 \cdot \frac{1}{\ntrain_{t,k}}  \sum_{j=t-k}^{t-1} \ntrain_j \left\{ \EE_{\bz\sim\distP_j} \left[ \ell_f(\bz) - \ell_{\bar{f}}(\bz) \right] + 2\excessrisk_j(\bar{f}) \right\}.
\end{align*}
\end{lemma}

\begin{proof}[Proof of \myCref{lem-noise-condition-regression}]
See \myCref{sec-proof-lem-noise-condition-regression}.
\end{proof}

Define
\[
T(\ell_f-\ell_{\bar{f}_t}) = 16M^2 \cdot \frac{1}{\ntrain_{t,k}} \sum_{j=t-k}^{t-1} \ntrain_j \left\| f - \bar{f}_t \right\|_j^2, \qquad\forall f\in\modelclass.
\]
By \myCref{lem-noise-condition-regression}, the noise condition \eqref{eqn-noise-condition} holds with $C=32M^2$, and \myCref{lem-local-uniform-concentration} is applicable. Moreover, for all $r \ge 0$,
\begin{align*}
&\Rademacher\left( \left\{ \ell_f - \ell_{\bar{f}_t} : f\in\modelclass, \, T(\ell_f-\ell_{\bar{f}_t}) \le r \right\} ; \dataset^{\tr}_{t,k} \right) \\[4pt]
&\qquad =
\EE\sup\left\{ \frac{1}{\ntrain_{t,k}}  \sum_{j=t-k}^{t-1}\sum_{s\in\batchtr_j}\varepsilon_s \left[f(\bx_s)-y_s\right]^2  : f\in\modelclass, \, T(\ell_f-\ell_{\bar{f}_t}) \le r \right\} \notag \\[4pt]
&\qquad \le
4M \cdot \EE\sup\left\{  \frac{1}{\ntrain_{t,k}}  \sum_{j=t-k}^{t-1} \sum_{s\in\batchtr_j}\varepsilon_s \left[ f(\bx_s) - y_s\right]  : f\in\modelclass, \, T(\ell_f-\ell_{\bar{f}_t}) \le r \right\} \notag \\[4pt]
&\qquad =
4M \cdot \EE\sup\left\{  \frac{1}{\ntrain_{t,k}}  \sum_{j=t-k}^{t-1} \sum_{s\in\batchtr_j}\varepsilon_s f(\bx_s)  : f\in\modelclass, \, T(\ell_f-\ell_{\bar{f}_t}) \le r \right\} \notag \\[4pt]
&\qquad =
4M\cdot \Rademacher\left(
\left\{f\in\cF : T(\ell_f-\ell_{\bar{f}_t}) \le r \right\};\dataset^{\tr}_{t,k}
\right),
\end{align*}
where $\{\varepsilon_s\}$ are i.i.d.~Rademacher random variables, and the inequality is due to Theorem A.6 in \cite{BBM05}. 

Define
\[
L_{t,k}^{\tr}(f) = \frac{1}{\ntrain_{t,k}} \sum_{j=t-k}^{t-1} \ntrain_j L_j(f).
\] 
Applying \myCref{lem-local-uniform-concentration} with $g=\widehat{f}_{t,k}-\bar{f}_t$, we obtain that if $\widetilde{\psi}:\RR_+\to\RR_+$ is a sub-root function with fixed point $\widetilde{r}$ satisfying
\begin{equation}\label{eqn-subroot-requirement}
\widetilde{\psi}(r) \ge 128M^3\cdot \Rademacher\left(
\left\{f\in\cF : T(\ell_f-\ell_{\bar{f}_t}) \le r \right\};\dataset^{\tr}_{t,k}
\right), \qquad \forall r \ge \widetilde{r},
\end{equation} 
then with probability at least $1-\delta$,
\begin{align}
L^{\tr}_{t,k}(\widehat{f}_{t,k}) - L^{\tr}_{t,k}(\bar{f}_t)
&\lesssim 
\left(\widehat{L}_{t,k}^{\tr}(\widehat{f}_{t,k}) - \widehat{L}_{t,k}^{\tr}(\bar{f}_t)\right)
+
\left[
\frac{\widetilde{r}}{M^2}
+
\frac{M^2\log(1/\delta)}{\ntrain_{t,k}}
\right]
+
\max_{t-k\le j\le t-1} \excessrisk_j(\bar{f}_t) \notag \\[4pt]
&\lesssim
\left[
\frac{\widetilde{r}}{M^2}
+
\frac{M^2\log(1/\delta)}{\ntrain_{t,k}}
\right]
+
\max_{t-k\le j\le t-1} \excessrisk_j(\bar{f}_t), \label{eqn-excess-risk-bar-bound-2}
\end{align}
where $\lesssim$ hides a universal constant, and the second inequality is due to $\widehat{L}^{\tr}_{t,k}(\widehat{f}_{t,k}) \le \widehat{L}^{\tr}_{t,k}(\bar{f}_t)$.

It remains to express $\widetilde{r}$ in terms of $r_{t,k}$, and convert \eqref{eqn-excess-risk-bar-bound-2} into a bound for $L_t(\widehat{f}_{t,k})$. We work on $\widetilde{r}$ first. Take a subroot function $\psi:\RR_+\to\RR_+$ with fixed point $r_{t,k}$ such that
\[
\psi(r) \ge \Rademacher\bigg( \bigg\{ f\in\modelclass : \frac{1}{n_{t,k}} \sum_{j=t-k}^{t-1} n_j \left\| f - \bar{f}_t \right\|_j^2 \le r \bigg\} \bigg), \quad \forall r \ge r_{t,k}.
\]
We now show that $\widetilde{\psi}(r) = 128 M^3 \psi\left( \frac{r}{16M^2} \right)$ satisfies \eqref{eqn-subroot-requirement}. By \myCref{lem-property-subroot}, the fixed point $\widetilde{r}$ of $\widetilde{\psi}$ satisfies
\begin{equation}\label{eqn-fixed-point-rescaled}
\min\{1,8M \}^2 16M ^2r_{t,k} \le \widetilde{r} \le \max\{1,8M \}^2 16M ^2r_{t,k}.
\end{equation}
For all $r\ge \widetilde{r}$, since $r \ge r_{t,k}$, then
\[
\widetilde{\psi}(r) \ge 128M ^3\cdot \Rademacher\bigg( \bigg\{ f\in\modelclass : 16M^2 \cdot \frac{1}{n_{t,k}} \sum_{j=t-k}^{t-1} n_j \left\| f - \bar{f}_t \right\|_j^2 \le r \bigg\} \bigg),
\]
so \eqref{eqn-subroot-requirement} holds for this choice of $\widetilde{\psi}$, and \eqref{eqn-excess-risk-bar-bound-2} becomes
\begin{equation}\label{eqn-excess-risk-bar-bound-3}
L^{\tr}_{t,k}(\widehat{f}_{t,k}) - L^{\tr}_{t,k}(\bar{f}_t)
\le
M^2\left[
r_{t,k}
+
\frac{\log(1/ \delta)}{\ntrain_{t,k}}
\right]
+
\max_{t-k\le j\le t-1} \excessrisk_j(\bar{f}_t).
\end{equation}

To convert \eqref{eqn-excess-risk-bar-bound-3} into a bound for $\excessrisk_t(\widehat{f}_{t,k})$, we note that
\begin{align}
    \excessrisk_t(\widehat{f}_{t,k}) - \excessrisk_t(\bar{f}_t)
    &\le
    \frac{1}{\ntrain_{t,k}} \sum_{j=t-k}^{t-1} \ntrain_j \excessrisk_j(\widehat{f}_{t,k}) - \frac{1}{\ntrain_{t,k}} \sum_{j=t-k}^{t-1} \ntrain_j \excessrisk_j(\bar{f}_{t}) \notag \\[4pt]
    &\qquad+
    \max_{t-k\le j \le t-1} |\excessrisk_j(\widehat{f}_{t,k}) - \excessrisk_t(\widehat{f}_{t,k})|
    +
    \max_{t-k\le j \le t-1} |\excessrisk_j(\bar{f}_{t}) - \excessrisk_t(\bar{f}_{t})| \notag \\[6pt]
    &=
    L^{\tr}_{t,k}(\widehat{f}_{t,k}) - L^{\tr}_{t,k}(\bar{f}_t) \notag \\[4pt]
    &\qquad+
    \max_{t-k\le j \le t-1} |\excessrisk_j(\widehat{f}_{t,k}) - \excessrisk_t(\widehat{f}_{t,k})|
    +
    \max_{t-k\le j \le t-1} |\excessrisk_j(\bar{f}_{t}) - \excessrisk_t(\bar{f}_{t})| \notag \\[6pt]
    &\le 
    C\left\{M^2\left[
    r_{t,k}
    +
    \frac{\log(1/ \delta)}{\ntrain_{t,k}}
    \right]
    +
    \max_{t-k\le j\le t-1} \excessrisk_j(\bar{f}_t) \right\} \notag \\[4pt]
    &\qquad +
    \max_{t-k\le j \le t-1} |\excessrisk_j(\widehat{f}_{t,k}) - \excessrisk_t(\widehat{f}_{t,k})|
    +
    \max_{t-k\le j \le t-1} |\excessrisk_j(\bar{f}_{t}) - \excessrisk_t(\bar{f}_{t})| \notag \\[4pt] 
    &\le 
    C\left\{M^2\left[
    r_{t,k}
    +
    \frac{\log(1/ \delta)}{\ntrain_{t,k}}
    \right]
    +\excessrisk_t(\bar{f}_t) \right\} \notag \\[4pt]
    &\qquad +
    \max_{t-k\le j \le t-1} |\excessrisk_j(\widehat{f}_{t,k}) - \excessrisk_t(\widehat{f}_{t,k})|
    +
    (C+1)\max_{t-k\le j \le t-1} |\excessrisk_j(\bar{f}_{t}) - \excessrisk_t(\bar{f}_{t})|
    \label{eqn-excess-risk-bound-1}
\end{align}
for some universal constant $C>0$. We now control the last two terms in \eqref{eqn-excess-risk-bound-1}. For all $j\in\ZZ_+$ and $f$, we have
\begin{align}
    |\excessrisk_j(f) - \excessrisk_t(f)|
    &=
    \big| \| f - f_j^* \|_j^2 - \| f - f_t^* \|_t^2 \big| \notag \\[4pt]
    &\le 
    \big| \| f - f_j^* \|_j^2 - \| f - f_j^* \|_t^2 \big|
    +
    \big| \| f - f_j^* \|_t^2 - \| f - f_t^* \|_t^2 \big|, \label{eqn-excess-risk-bound-complete-square-1}
\end{align}
where the first term on the right hand side of \eqref{eqn-excess-risk-bound-complete-square-1} involves covariate shift and can be bounded by the integral probability metric \eqref{eqn-IPM-covariate-shift} via
\begin{equation}\label{eqn-excess-risk-bound-complete-square-2}
    \| f - f_j^* \|_j^2 - \| f - f_j^* \|_t^2 
\le 
D_t(\distP_{j,\bx}, \distP_{t,\bx}),
\end{equation}
and the second term involves conditional shift and can be bounded via
\begin{align}
    \big| \| f - f_j^* \|_t^2 - \| f - f_t^* \|_t^2 \big|
    &=
    \big| \langle f_j^*-f_t^*, \, 2f - f_j^* - f_t^* \rangle_t \big| \notag \\[4pt]
    &=
    \big| \langle f_j^*-f_t^*, \, 2(f - f_t^*) - (f_j^* - f_t^*) \rangle_t \big| \notag \\[4pt]
    &\le 
    2\|f_j^*-f_t^*\|_t\|f-f_t^*\|_t + \|f_j^*-f_t^*\|_t^2. \label{eqn-excess-risk-bound-complete-square-3}
\end{align}
Substituting \eqref{eqn-excess-risk-bound-complete-square-1}, \eqref{eqn-excess-risk-bound-complete-square-2}, and \eqref{eqn-excess-risk-bound-complete-square-3} into \eqref{eqn-excess-risk-bound-1} yields
\begin{align*}
    \|\widehat{f}_{t,k}-f_t^*\|_t^2 - \|\bar{f}_t - f_t^*\|_t^2
    &\le 
    C\left\{M^2\left[
    r_{t,k}
    +
    \frac{\log(1/ \delta)}{\ntrain_{t,k}}
    \right]
    +
    \|\bar{f}_t-f_t^*\|_t^2 \right\} \\[4pt]
    &\qquad +
    C'\Big[2\max_{t-k\le j\le t-1} \|f_j^*-f_t^*\|_t \|\widehat{f}_{t,k}-f_t^*\|_t \\[4pt]
    &\qquad\qquad + 
    2\max_{t-k\le j\le t-1} \|f_j^*-f_t^*\|_t \|\bar{f}_t-f_t^*\|_t \\[4pt]
    &\qquad\qquad + 
    2\max_{t-k\le j \le t-1}\|f_j^*-f_t^*\|_t^2 + 2\max_{t-k\le j \le t-1}D_t(\distP_{j,\bx}, \distP_{t,\bx})\Big]
\end{align*}
for some universal constant $C'>0$. Completing the squares for $\|\widehat{f}_{t,k}-f_t^*\|_t$ and $\|\bar{f}_t-f_t^*\|_t$ yields
\begin{align*}
    &\left( \|\widehat{f}_{t,k}-f_t^*\|_t - C'\max_{t-k\le j\le t-1} \|f_j^*-f_t^*\|_t \right)^2 \\
    &\qquad\le 
    \left( \|\bar{f}_t-f_t^*\|_t + C'  \max_{t-k\le j\le t-1} \|f_j^*-f_t^*\|_t \right)^2 \\[4pt]
    &\qquad\quad + C\left\{M^2\left[
    r_{t,k}
    +
    \frac{\log(1/ \delta)}{\ntrain_{t,k}}
    \right]
    +
    \|\bar{f}_t-f_t^*\|_t^2 \right\} \\[4pt]
    &\qquad\quad + C' \left[ 2\max_{t-k\le j \le t-1}\|f_j^*-f_t^*\|_t^2 + 2\max_{t-k\le j \le t-1}D_t(\distP_{j,\bx}, \distP_{t,\bx}) \right].
\end{align*}
This implies
\begin{align*}
    \|\widehat{f}_{t,k}-f_t^*\|_t &\lesssim
    \|\bar{f}_t-f_t^*\|_t 
    +  
    M\left[
    r_{t,k}
    +
    \frac{\log(1/ \delta)}{\ntrain_{t,k}}
    \right]^{1/2} \\[4pt]
    &\qquad +
    \max_{t-k\le j\le t-1} \left\{ \|f_j^*-f_t^*\|_t 
    +
    D_t(\distP_{j,\bx}, \distP_{t,\bx})^{1/2} \right\}.
\end{align*}
Squaring both sides yields
\begin{align*}
    \excessrisk_t(\widehat{f}_{t,k})
    &\lesssim
    \excessrisk_t(\bar{f}_t) + M^2\left[
    r_{t,k}
    +
    \frac{\log(1/ \delta)}{\ntrain_{t,k}}
    \right] 
    +
    \max_{t-k\le j\le t-1} \left\{ \|f_j^*-f_t^*\|_t^2 
    +
    D_t(\distP_{j,\bx}, \distP_{t,\bx}) \right\}.
\end{align*}
This finishes the proof.

\subsection{Proof of \myCref{lem-local-uniform-concentration}}\label{sec-proof-lem-local-uniform-concentration}

The core techniques are the same as those of Theorem 3.3 in \cite{BBM05}, with small changes in the quantities to bound. For $r,\lambda>0$, let
\[
w(g) = \min\{r\lambda^k:k\in\{0\}\cup\ZZ_+,\,r\lambda^k\ge T(g) \},
\qquad
\cG_r = \left\{\frac{r}{w(g)}g:g\in\cG\right\},
\]
and
\[
V_r^+ = \sup_{g\in\cG} \left\{ \frac{r}{w(g)} \cdot \frac{1}{n}\sum_{i=1}^n \big(\EE [ g(\bz_i) ] -  g (\bz_i) \big) \right\}.
\]
Similar to Lemma 3.8 of \cite{BBM05}, for every $K>1$ and $g\in\cG$,
\[
V_r^+ \le \frac{r}{\lambda CK}
\quad \text{implies} \quad
\frac{1}{n}\sum_{i=1}^n\EE[g (\bz_i)]
\le 
\frac{K}{K-1}\cdot\frac{1}{n}\sum_{i=1}^n g(\bz_i)
+
\frac{r}{\lambda CK} + \frac{1}{K-1}\cdot\frac{1}{n}\sum_{i=1}^n \eta_i.
\]

We now invoke a uniform convergence result to give a bound for $V_r^+$. It is a non-stationary version of Theorem 2.1 in \cite{BBM05}.

\begin{lemma}[Uniform concentration]\label{lem-uniform-concentration}
Consider the setting of \myCref{lem-local-uniform-concentration}. Define 
\[
v = \frac{1}{n}\sup_{g\in\cG}\sum_{i=1}^n\var[g(\bz_i)].
\] 
Let $\delta\in(0,1)$. With probability at least $1-\delta$,
\begin{align*}
&\sup_{g\in\cG}\frac{1}{n} \sum_{i=1}^n \big(\EE [ g(\bz_i) ] -  g (\bz_i) \big) \\
&\qquad \lesssim
\inf_{\alpha>0}
\left\{
(1+\alpha)\Rademacher(\cG;\{\bz_i\}_{i=1}^n)
+
\sqrt{\frac{v\log(1/\delta)}{n}}
+
(1 + \alpha^{-1})\frac{(b-a)\log(1/\delta)}{n}
\right\},
\end{align*}
where $\lesssim$ hides a universal constant.
\end{lemma}

\begin{proof}[Proof of \myCref{lem-uniform-concentration}] Let $Z = \sup_{g\in\cG} \frac{1}{n} \sum_{i=1}^n \left[ g(\bz_i) - \EE g (\bz_i)\right] $. Adapting the proof of Theorem 4 in \cite{Mas00}, with probability at least $1-\delta$,
\[
Z - \EE Z
\lesssim
\sqrt{\frac{\log(1/\delta)}{n} 
\left(v + (b-a)\EE Z\right)}
+
\frac{(b-a)\log(1/\delta)}{n},
\]
where $\lesssim$ hides a universal constant. By Young's inequality, for every $\alpha>0$,
\[
\sqrt{\frac{\log(1/\delta)}{n}
(b-a)\EE Z}
\le 
\alpha \EE Z
+
\alpha^{-1}\frac{(b-a)\log(1/\delta)}{n}.
\] 
By Rademacher symmetrization (e.g., Lemma A.5 in \cite{BBM05}), for i.i.d.~Rademacher random variables $\varepsilon_1,...,\varepsilon_n$ independent of $\bz_1,...,\bz_n$, it holds that
\[
\EE Z \le 
2\EE\left[
\sup_{g\in\cG}
\frac{1}{n} \sum_{i=1}^n \varepsilon_i g(\bz_i)
\right]
=
2\Rademacher(\cG;\{\bz_i\}_{i=1}^n).
\]
This completes the proof.
\end{proof}

By \myCref{lem-uniform-concentration}, each of the following inequalities holds with probability at least $1-\delta$:
\[
V_r^+ \lesssim \inf_{\alpha>0} \left\{ (1+\alpha)\Rademacher(\cG_r;\{\bz_i\}_{i=1}^n) + \sqrt{\frac{v\log(1/\delta)}{n}} + (1+\alpha^{-1})\frac{(b-a)\log(1/\delta)}{n} \right\}, \]
The rest of the proof follows that in Section 3.2 of \cite{BBM05}.

\subsection{Proof of \myCref{lem-noise-condition-regression}}\label{sec-proof-lem-noise-condition-regression}

Recall that for $\bz=(\bx,y)$, we define $\ell_f(\bz) = [f(\bx) - y]^2$. Then
\begin{align*}
&\frac{1}{\ntrain_{t,k}}\sum_{j=t-k}^{t-1}\ntrain_j\var_{\bz\sim\distP_j}\left[\ell_f(\bz) - \ell_{\bar{f}}(\bz)\right] \\[4pt]
&\qquad =
\frac{1}{\ntrain_{t,k}}\sum_{j=t-k}^{t-1}\ntrain_j\var_{(\bx,y)\sim\distP_j}\left[\big(f(\bx)-y\big)^2 - \big(\bar{f}(\bx)-y\big)^2\right] \\[4pt]
&\qquad \le 
\frac{1}{\ntrain_{t,k}}\sum_{j=t-k}^{t-1}\ntrain_j\EE_{(\bx,y)\sim\distP_j} \left\{\left[\big(f(\bx)-y\big)^2 - \big(\bar{f}(\bx)-y\big)^2\right]^2 \right\} \\[4pt]
&\qquad = 
\frac{1}{\ntrain_{t,k}}\sum_{j=t-k}^{t-1}\ntrain_j\EE_{(\bx,y)\sim\distP_j}\left[\big( f(\bx) - \bar{f}(\bx) \big)^2 \big(f(\bx) + \bar{f}(\bx) - 2y \big)^2\right] \\[4pt]
&\qquad \le 
16M^2\cdot \frac{1}{\ntrain_{t,k}}\sum_{j=t-k}^{t-1}\ntrain_j\EE_{(\bx,y)\sim\distP_j}\left[\left(f(\bx)-\bar{f}(\bx)\right)^2\right]
=
16M^2\cdot \frac{1}{\ntrain_{t,k}}\sum_{j=t-k}^{t-1} \ntrain_j \| f - \bar{f} \|_j^2 \\[4pt]
&\qquad \le
32M^2\cdot \frac{1}{\ntrain_{t,k}}\sum_{j=t-k}^{t-1} \ntrain_j \left( \| f - f_j^* \|_j^2 + \|\bar{f} - f_j^* \|_j^2 \right) \\[4pt]
&\qquad =
32M^2 \cdot \frac{1}{\ntrain_{t,k}}  \sum_{j=t-k}^{t-1} \ntrain_j \left\{ \EE_{\bz\sim\distP_j} \left[ \ell_f(\bz) - \ell_{\bar{f}}(\bz) \right] + 2\excessrisk_j(\bar{f}) \right\}.
\end{align*}
This finishes the proof.

\subsection{Unavoidable Cost of Non-stationarity}\label{sec-prop-tv-lower-bound}

\myCref{sec-tradeoff-theory} argues informally that the non-stationarity term in \Cref{thm-tradeoff} cannot be removed by a cleverer estimator. We state and prove this claim formally here.

\begin{prop}[Unavoidable cost of non-stationarity]
\label{prop-tv-lower-bound}
Fix $t \ge 2$ and a window $k\in[t-1]$. Let $|\cX|\ge 2$. Suppose for simplicity that $M=1$ and $|\dataset_{\tau}^{\tr}|=B$ for all $\tau\in[t-1]$. For $V\in[0,1]$, define a class of non-stationary data distributions over $\cX\times [-1,1]$:
\begin{multline*}
    \mathscr{Q}(V) = \bigg\{ (\distP_1,...,\distP_t): \supp(\distP_{\tau})\subseteq\cX\times [-1,1]\ \forall \tau\in[t],\ \max_{t-k\le \tau \le t-1} \TV(\distP_{\tau},\distP_t) \le V \bigg\}.
\end{multline*}
Then any estimator $\widehat{f}_t\in\modelclass$ based on the data $\{\dataset_{\tau}^{\tr}\}_{\tau=t-k}^{t-1}$ must have a worst-case out-of-sample prediction error of at least
\[
\sup_{(\distP_1,...,\distP_t)\in\mathscr{Q}(V)}
\mathbb{E} \left[\excessrisk_t(\widehat{f}_t)\right]
\gtrsim
\min_{f\in\modelclass} \excessrisk_t(f)
+
\frac{1}{Bk} + V.
\]
\end{prop}

\Cref{prop-tv-lower-bound} considers all non-stationary data-generating processes for which the current distribution $\distP_t$ is at most $V$ away, in total variation, from each historical data distribution $\distP_{\tau}$ used for training. When $V$ is small, the last $k$ periods are close to the current environment; when $V$ is large, the training window may contain observations from highly different regimes.

The lower bound decomposes the unavoidable error into the same three components as \myCref{thm-tradeoff}, and by cross-checking, establishes that our upper bound in \myCref{thm-tradeoff} is indeed tight. The term $\min_{f\in\modelclass}\excessrisk_t(f)$ is the approximation error from using the chosen model class at the current distribution. The term $1/(Bk)$ is the statistical uncertainty due to having only $Bk$ observations in the training window. The term $V$ is the irreducible cost of training on historical distributions that differ from the current distribution. Thus, even if the model class is correctly chosen and the estimator is optimal, non-stationarity alone can generate an error of order $V$. This establishes that the upper bound in \myCref{thm-tradeoff} has the correct dependence on the non-stationarity.

\subsection{Proof of \Cref{prop-tv-lower-bound}}\label{sec-prop-tv-lower-bound-proof}
Let
\[
\cL = \sup_{(\distP_1,...,\distP_t)\in\mathscr{Q}(V)}
\mathbb{E} \left[\excessrisk_t(\widehat{f}_t)\right] 
\]
It suffices to prove the following three lower bounds separately:
\[
\cL
\ge 
\min_{f\in\modelclass} \excessrisk_t(f),
\qquad
\cL
\gtrsim 
\frac{1}{Bk},
\qquad
\cL
\gtrsim 
V.
\]
The first one is immediate since $\widehat{f}_t\in\modelclass$ always. The second one holds regardless of the non-stationarity, and follows from standard minimax lower bound results \citep{Tsy08}. We now prove the third bound.

Let $p = \min\{V,1/4\}$. Take two distinct elements $\bx_1,\bx_2\in\cX$. We define three distributions $\distQ$, $\distQ^+$ and $\distQ^-$ as follows.
\begin{itemize}
    \item A sample $(\bx,y)\sim\distQ$ satisfies $\PP\big((\bx,y)=(\bx_1,0)\big) = 1-p$ and $\PP\big((\bx,y)=(\bx_2,0)\big) = p$.
    \item A sample $(\bx,y)\sim\distQ^+$ satisfies $\PP\big((\bx,y)=(\bx_1,0)\big) = 1-p$ and $\PP\big((\bx,y)=(\bx_2,1)\big) = p$. For $Q^+$, the Bayes least squares optimal estimator $f^*_+(\cdot) = \EE_{(\bx,y)\sim\distQ^+}[y\mid\bx=\cdot]$ is given by $f^*_+(\bx_1)=0$ and $f^*_-(\bx_2)=1$.
    \item A sample $(\bx,y)\sim\distQ^-$ satisfies $\PP\big((\bx,y)=(\bx_1,0)\big) = 1-p$ and $\PP\big((\bx,y)=(\bx_2,-1)\big) = p$. For $Q^-$, the Bayes least squares optimal estimator $f^*_-(\cdot) = \EE_{(\bx,y)\sim\distQ^-}[y\mid\bx=\cdot]$ is given by $f^*_+(\bx_1)=0$ and $f^*_-(\bx_2)=-1$.
\end{itemize}
We have $\TV(Q,Q^+)=\TV(Q,Q^-)= p \in[V/4, V]$.

Define two non-stationary environments $\cP^+ = \big(\distP_1^{+},..., \distP_t^{+}\big)$ and $\cP^- = \big(\distP_1^{-},..., \distP_t^{-}\big)$ by
\begin{itemize}
    \item $\distP_1^+=\cdots=\distP_{t-1}^+=\distQ$ and $\distP_t^+=\distQ^+$,
    \item $\distP_1^-=\cdots=\distP_{t-1}^-=\distQ$ and $\distP_t^-=\distQ^-$.
\end{itemize}
Clearly $\cP^+,\cP^-\in\mathscr{Q}(V)$. Consider any estimator $\widehat{f}_t$. Under both environments, the training data have exactly the same distribution, because all training sample are drawn from $\distQ$. Thus, $\widehat{f}_t$ has the same distribution under $\cP^+$ and $\cP^-$.

Under $\cP^+$,
\[
\excessrisk_t(\widehat{f}_t) = \EE_{(\bx,y)\sim\distQ^+} \left[ \big( \widehat{f}_t(\bx) - f_+^*(\bx) \big)^2 \right]
\ge 
p\big( \widehat{f}_t(\bx_2) - f_+^*(\bx_2) \big)^2
=
p\big( \widehat{f}_t(\bx_2) - 1 \big)^2.
\]
Under $\cP^-$,
\[
\excessrisk_t(\widehat{f}_t) = \EE_{(\bx,y)\sim\distQ^+} \left[ \big( \widehat{f}_t(\bx) - f_-^*(\bx) \big)^2 \right]
\ge 
p\big( \widehat{f}_t(\bx_2) - f_-^*(\bx_2) \big)^2
=
p\big( \widehat{f}_t(\bx_2) + 1 \big)^2.
\]
Therefore,
\[
\sup_{(\distP_1,...,\distP_t)\in\mathscr{Q}(V)}
\mathbb{E} \left[\excessrisk_t(\widehat{f}_t)\right]
\ge 
\EE\left[ \max \left\{ p\big( \widehat{f}_t(\bx_2) - 1 \big)^2, ~ p\big( \widehat{f}_t(\bx_2) + 1 \big)^2\right\} \right]
\ge 
p \ge \frac{V}{4}.
\]
This completes the proof.

\subsection{Proofs for \myCref{eg-linear} and \myCref{eg-kernel}}\label{sec-proof-eg-classes}




To obtain the results in \myCref{eg-linear} and \myCref{eg-kernel}, we invoke a useful lemma. It is an extension of Theorem 41 in \cite{Men02} to the non-i.i.d.~case. The proof is omitted.

\begin{lemma}\label{lem-Rademacher-linear}
Take $\bar{\btheta} \in B(\bm{0},\sqrt{M})\subseteq \RR^d$. Suppose $\bx_i\sim Q_i$, $i\in[n]$ are independent random vectors in $\RR^d$. Let $\varepsilon_1,...,\varepsilon_n$ be i.i.d.~Rademacher random variables independent of $\{\bx_i\}_{i=1}^n$. For $r\ge 0$, define
\[
\cR(r) = \EE \sup\left\{ \frac{1}{n}  \sum_{i=1}^n \varepsilon_i \bx_i^\top\btheta  : \frac{1}{n} \sum_{i=1}^n\EE\left[ \big(\bx_i^\top (\btheta - \bar{\btheta})\big)^2 \right] \le r,~ \|\btheta\|_2^2 \le M  \right\}.
\]
Let $\bSigma = \frac{1}{n} \sum_{i=1}^n\EE(\bx_i \bx_i^{\top})$, and denote by $\{ \lambda_i \}_{i=1}^d$ its eigenvalues sorted in descending order. Then
\[
\cR(r) \le
\sqrt{\frac{c}{n}\sum_{i=1}^d \min\{r,M\lambda_i\}}.
\]
\end{lemma}

The above lemma leads to $\cR(r) \le
\sqrt{ c r d / n}$.
The right-hand side is subroot and has a unique fixed point $r = cd/n$. This verifies \myCref{eg-linear}.

To get \myCref{eg-kernel}, we apply \Cref{lem-Rademacher-linear} to the transformed features $\phi(\bx_{j, i})$ rather than the raw ones. Correspondingly, the feature space becomes $\HH$, and the matrix $\bSigma$ in \myCref{eg-kernel} becomes
$\frac{1}{\ntrain_{t, k}}
\sum_{\tau=t-k}^{t-1}
\sum_{s\in\batch_{\tau}}\EE[\phi(\bx_{s}) \otimes \phi (\bx_{s}) ]$. Here $\otimes$ denotes the tensor product. 

In \myCref{eg-kernel}, the assumption regarding the trace-class operator $\bS$ forces $\bSigma \preceq \bS$.
Then, \Cref{lem-Rademacher-linear} yields
\[
\cR(r) \lesssim
\sqrt{\frac{1}{\ntrain_{t, k}}\sum_{i=1}^{\infty} \min\{r, \mu_i\}}
\leq  \frac{1}{\sqrt{\ntrain_{t, k}}} \min_{j \geq 1} \sqrt{
j r +
\sum_{i=j+1}^{\infty} \mu_i },
\]
where $\lesssim$ hides a constant factor.

\begin{itemize}
\item If there are constants $c_1,c_2>0$ such that $\mu_{j} \leq c_1 e^{- c_2 j }$ holds for all $j$, then 
$\sum_{i=j+1}^{\infty} \mu_i \lesssim e^{-c_2 j }$. Taking $j = \lceil c_2^{-1} \log (1 / r) \rceil$, we get
\[
\cR(r) \lesssim
\frac{1}{\sqrt{\ntrain_{t, k}}} \sqrt{
r \lceil c_2^{-1} \log (1 / r) + 1 \rceil
}
.
\]
The right-hand side is sub-root. Elementary calculation yields 
$r_{t,k}(\modelclass) \lesssim (\log \ntrain_{t, k} ) / \ntrain_{t, k}$.

\item If there are constants $c>0$ and $\alpha \ge 1$ such that $\mu_j \le c j^{-2 \alpha}$ for all $j$, then 
$\sum_{i=j+1}^{\infty} \mu_i \lesssim j^{1-2\alpha}$. Taking $j = \lceil r^{-1/(2\alpha)} \rceil$, we get
\[
\cR(r) \lesssim 
\sqrt{
\frac{
r^{1 - 1 / (2 \alpha)}
}{\ntrain_{t,k}}
}.
\]
The right-hand side is sub-root. Then, we can easily get
$r_{t,k}(\modelclass) \lesssim \big(\ntrain_{t, k}\big)^{-\frac{2 \alpha}{2 \alpha + 1}}$.
\end{itemize}

\section{Proofs for \myCref{sec-select} and \myCref{sec-select-R2}}

\subsection{Algorithmic Details}\label{sec-algorithmic-details}

We record here the full pseudocode for the tournament procedure and the pairwise comparison subroutine described in prose in \myCref{sec-select}. The proofs below refer directly to objects defined in this pseudocode, such as the pivot model and the surviving set $\survivorsnew$.

\begin{algorithm}[h]
	\begin{algorithmic}
		\STATE {\bf Input:} Candidate models $\{f_{\modelidx}\}_{\modelidx=1}^{\nmodel}$, validation data $\{ \dataset^{\va}_{\tau} \}_{\tau=1}^{t-1}$, pairwise model comparison subroutine $\alg$.
		\STATE Initialize $S = \{f_{\modelidx} \}_{\modelidx=1}^{\nmodel}$. \hfill \texttt{// collection of remaining models}\\
		\WHILE{$|\survivors| > 1$}
			\STATE Choose a pivot model $f\in \survivors$ uniformly at random.
			\STATE Initialize $\survivorsnew \leftarrow \emptyset$. \hfill \texttt{// collection of models in $\survivors$ that outperform $f$}
			\FOR{$f' \in \survivors \backslash \{ f \}$}
			\STATE Run $\alg$ to compare $\{f,f'\}$ to obtain $\alg(f,f')$.
			\STATE If $\alg(f,f') = f'$, set $\survivorsnew \leftarrow \survivorsnew\cup\{f'\}$.
			\ENDFOR
			\IF{$S'  = \emptyset$}
			\RETURN $\widehat{f} = f$. \hfill \texttt{// if no model outperforms $f$, output $f$}
			\ELSE
			\STATE Set $\survivors \leftarrow \survivorsnew$.
			\ENDIF
		\ENDWHILE
		\RETURN the only model $\widehat{f} \in \survivors$.
	\caption{Adaptive Tournament Model Selection ($\adaptive$)}
	\label{alg-tournament}
	\end{algorithmic}
\end{algorithm}

\begin{algorithm}[t]
	\begin{algorithmic}
	\STATE {\bf Input:} Candidate models $\{f_1,f_2\}$, validation data $\{ \dataset^{\va}_{\tau} \}_{\tau=1}^{t-1}$, hyperparameters $\delta'$ and $M'$.\\
	\FOR{$\wval = 1,\cdots, t-1$}
	\STATE Compute $\widehat{\gap}_{t,\wval}$, $\widehat{\psi}(t,\wval,\delta')$ and $\widehat\phi(t,\wval,\delta')$ according to \eqref{eqn-performance-gap-hat}, \eqref{eqn-psi-hat} and \eqref{eqn-phi-hat}.
	\ENDFOR
	\STATE Choose window size
	\begin{equation}\label{GL-k-hat}
	\widehat{\wval} \in \argmin_{ \wval \in [t-1] }  \Big\{ \widehat\phi (t, \wval, \delta')  + \widehat{\psi} (t , \wval , \delta')  \Big\}.
	\end{equation}
	\STATE Select $\widehat{\modelidx} = 1$ if $\widehat{\gap}_{t,\widehat{\wval}} \le 0$, and $\widehat{\modelidx} = 2$ otherwise. \\
	\RETURN $\widehat{f} = f_{\widehat{\modelidx}}$.
	\caption{Adaptive Rolling Window for Model Comparison}
	\label{alg-compare}
	\end{algorithmic}
\end{algorithm}

\subsection{Proof of \myCref{lem-complexity-tournament}}\label{sec-lem-complexity-tournament-proof}

Given two models $f,f'\in\modelclass$, denote the output of $\alg$ by $\alg(f,f') \in \{f,f'\}$. Let $T(\nmodel)$ be the maximum expected number of times \myCref{alg-tournament} can call $\alg$ after there are $\nmodel$ remaining models $\{f_{\modelidx}\}_{\modelidx=1}^{\nmodel}$ at the end of a while loop, where the maximum is taken over all possible choices of $\{f_{\modelidx}\}_{\modelidx=1}^{\nmodel}$. Then $T(\nmodel)$ is increasing in $\nmodel$, and $T(\nmodel)\le \nmodel^2/2$. Let $N$ denote the number of remaining models at the end of the next while loop. Since that while loop calls $\alg$ at most $\nmodel-1$ times, then
\begin{equation}\label{eqn-complexity-tournament-1}
T(\nmodel) \le (\nmodel-1) + \EE\left[T(N)\right].
\end{equation}
For each $\modelidx\in[\nmodel]$, let $n_{\modelidx} = |\{\modelidx'\in[\nmodel]\backslash\{\modelidx\}: \alg(f_\modelidx,f_{\modelidx'}) = f_{\modelidx'}\}|$ be the number of remaining models that would beat $f_\modelidx$ if they were paired. Since \myCref{alg-tournament} chooses each $f_{\modelidx}$ as the pivot model uniformly at random, then
\begin{equation}\label{eqn-complexity-tournament-2}
\EE\left[ T(N) \right] =  \EE\left[ \frac{1}{\nmodel}\sum_{\modelidx=1}^\nmodel T(n_\modelidx) \right].
\end{equation}
Since $\sum_{\modelidx=1}^{\nmodel} n_{\modelidx}$ counts exactly one of $(f_{\modelidx},f_{\modelidx'})$ and $(f_{\modelidx'},f_{\modelidx})$ for all $\modelidx \neq \modelidx'$, then $\sum_{\modelidx=1}^{\nmodel} n_{\modelidx} = \nmodel(\nmodel-1)/2$. Let $n_{(1)}\le\cdots\le n_{(\nmodel)}$ be the order statistics of $n_1,...,n_\nmodel$. Then for all $i = 1,...,\lceil \nmodel/3 \rceil$,
\[
n_{(i)} \le n_{(\lceil \nmodel/3 \rceil)} \le \frac{1}{\nmodel - \lceil \nmodel/3 \rceil + 1} \sum_{i = \lceil \nmodel/3 \rceil}^\nmodel n_i \le \frac{3}{2\nmodel}\cdot\frac{\nmodel(\nmodel-1)}{2} = \frac{3(\nmodel-1)}{4}.
\]
By the monotonicity of $T$,
\begin{align*}
\frac{1}{\nmodel}\sum_{\modelidx=1}^\nmodel T(n_\modelidx) = \frac{1}{\nmodel}\sum_{i=1}^\nmodel T(n_{(i)}) 
&\le
\frac{1}{\nmodel} \bigg( \lfloor \nmodel/3 \rfloor T( \lceil 3\nmodel/4 \rceil ) + \left( \nmodel - \lfloor \nmodel/3 \rfloor \right) T(\nmodel-1) \bigg] \\[4pt]
&\le
\frac{1}{3} T( \lceil 3\nmodel/4 \rceil ) + \left(\frac{2}{3} + \frac{1}{\nmodel}\right) T(\nmodel) \\[4pt]
&\le 
\frac{1}{3} T( \lceil 3\nmodel/4 \rceil ) + \frac{2}{3} T(\nmodel) + \nmodel.
\end{align*}
Substituting this into \eqref{eqn-complexity-tournament-2} and \eqref{eqn-complexity-tournament-1} gives
\[
T(\nmodel) \le 6\nmodel + T( \lceil 3\nmodel/4 \rceil ).
\]
By the Master Theorem (Theorem 4.1) in \cite{CLRS22}, we conclude that $T(\nmodel) = \Theta(\nmodel)$.

\subsection{Proof of \myCref{thm-model-comparison}}\label{sec-thm-model-comparison-proof}

We will prove the following stronger result: with probability at least $1-\delta$,
\begin{equation}\label{eqn-model-comparison-fast-root}
\sqrt{\excessrisk_t(\widehat{f})}
-
\sqrt{\excessrisk_t(f_2)}
\lesssim
M\sqrt{\log(t/\delta)} \cdot \left( \max_{t-\wval\le j\le t-1} \TV(\distP_j,\distP_t) + \frac{1}{\nval_{t,\wval}} \right)^{1/2}.
\end{equation}
The bound in \myCref{thm-model-comparison} is obtained by squaring both sides of \eqref{eqn-model-comparison-fast-root}.

Without loss of generality, assume $L_t(f_1) \ge L_t(f_2)$, so $\min_{\modelidx\in[2]}L_t(f_{\modelidx}) = L_t(f_2)$. By Theorem 4.2 in \cite{HHW24}, with probability at least $1-\delta$, \myCref{alg-compare} outputs a model $\widehat{f}$ satisfying
\begin{align*}
L_t(\widehat{f}) - L_t(f_2) 
& \le
| \widehat{\gap}_{t, \widehat{\wval} } - \gap_{t} |  \\[4pt]
& \lesssim
\min_{\wval \in [t-1]} \left\{
\sqrt{ \log ( t / \delta )  } \cdot 
\max_{t-\wval\le j\le t-1} | \gap_{j} - \gap_{t} |
+
\sd_{t, \wval} \sqrt{ \frac{ \log( t / \delta) }{ \nval_{t,\wval} } } + \frac{ M^2 \log ( t / \delta) }{ \nval_{t, \wval} }
\right\},
\end{align*}
where $\lesssim$ hides a universal constant. When this event happens, it holds for all $\wval\in[t-1]$ that
\[
L_t(\widehat{f}) - L_t(f_2)
\lesssim
\sqrt{ \log ( t / \delta )  } \cdot 
\max_{t-\wval\le j\le t-1} | \gap_{j} - \gap_{t} |
+
\sd_{t, \wval} \sqrt{ \frac{ \log( t / \delta) }{ \nval_{t,\wval} } } + \frac{ M^2 \log ( t / \delta) }{ \nval_{t, \wval} }.
\]
When $\widehat{f}=f_2$, \eqref{eqn-model-comparison-fast-root} automatically holds and there is nothing to prove. 

Now consider the case when $\widehat{f}=f_1$. Fix $\wval\in[t-1]$. Define
\[
\excessrisk_{t,\wval}^{\va}(f) = \frac{1}{\nval_{t,\wval}} \sum_{j=t-\wval}^{t-1} \nval_j \big[ L_j(f) - L_j(f_j^*) \big].
\]
Since
\[
\left| \big[\excessrisk_{t,\wval}^{\va}(f_1) - \excessrisk_{t,\wval}^{\va}(f_2) \big] - \big[L_t(f_1) - L_t(f_2) \big]\right|
\le 
\max_{t-\wval\le j\le t-1} |\gap_j - \gap_t|,
\]
then
\begin{align}
\excessrisk_{t,\wval}^{\va}(f_1) - \excessrisk_{t,\wval}^{\va}(f_2) 
&\le 
L_t(f_1) - L_t(f_2) + \max_{t-\wval\le j\le t-1} |\gap_j - \gap_t| \notag \\[4pt]
&\le 
C\left[
\sqrt{ \log ( t / \delta )  } \cdot 
\max_{t-\wval\le j\le t-1} | \gap_{j} - \gap_{t} |
+
\sd_{t, \wval} \sqrt{ \frac{ \log( t / \delta) }{ \nval_{t,\wval} } } + \frac{ M^2 \log ( t / \delta) }{ \nval_{t, \wval} }
\right] \label{eqn-thm-model-comparison-proof-raw}
\end{align}
for some universal constant $C\ge 1$. The variance term $\sd_{t,\wval}^2$ can be bounded by
\begin{align*}
\sd_{t,\wval}^2
&=
\frac{1}{\nval_{t,\wval}}\sum_{j=t-\wval}^{t-1} \nval_j \var_{(\bx,y)\sim\distP_j} \left[ \big( f_1(\bx) - y \big)^2 - \big( f_2(\bx) - y \big)^2 \right] \\[4pt]
&\le 
2\sum_{\modelidx=1}^2 \left[ \frac{1}{\nval_{t,\wval}}\sum_{j=t-\wval}^{t-1} \nval_j \var_{(\bx,y)\sim\distP_j} \left[ \big( f_{\modelidx}(\bx) - y \big)^2 - \big( f_j^*(\bx) - y \big)^2 \right]\right].
\end{align*}
For each $\modelidx\in[2]$ and $j\in\ZZ_+$,
\begin{align*}
\var_{(\bx,y)\sim\distP_j} \left[ \big( f_{\modelidx}(\bx) - y \big)^2 - \big( f_j^*(\bx) - y \big)^2 \right]
&\le 
\EE_{(\bx,y)\sim\distP_j} \left\{ \left[ \big( f_{\modelidx}(\bx) - y \big)^2 - \big( f_j^*(\bx) - y \big)^2 \right]^2 \right\} \\[4pt]
&=
\EE_{(\bx,y)\sim\distP_j} \left[ \big(f_{\modelidx}(\bx) - f_j^*(\bx) \big)^2 \big(f_{\modelidx}+f_j^*(\bx) - 2y \big)^2 \right] \\[4pt]
&\le 
16M^2\EE_{(\bx,y)\sim\distP_j} \left[ \big(f_{\modelidx}(\bx) - f_j^*(\bx) \big)^2 \right] \\[4pt]
&=
16M^2 \excessrisk_j(f_{\modelidx}).
\end{align*}
Thus,
\begin{equation}\label{eqn-thm-model-comparison-proof-variance}
\sigma_{t,\wval}^2
\le 
32M^2\cdot\sum_{\modelidx=1}^{2}\left[ \frac{1}{\nval_{t,\wval}} \sum_{j=t-\wval}^{t-1} \nval_j \excessrisk_j(f_{\modelidx}) \right]
=
32M^2 \big[ \excessrisk_{t,\wval}^{\va}(f_1) + \excessrisk_{t,\wval}^{\va}(f_2) \big].
\end{equation}
Substituting \eqref{eqn-thm-model-comparison-proof-variance} into \eqref{eqn-thm-model-comparison-proof-raw} yields
\[
\excessrisk_{t,\wval}^{\va}(f_1) - \excessrisk_{t,\wval}^{\va}(f_2) 
\le 
2A\left(\sqrt{\excessrisk_{t,\wval}^{\va}(f_1)} + \sqrt{\excessrisk_{t,\wval}^{\va}(f_2)} \, \right)
+
D,
\]
where
\[
A = 2\sqrt{2}CM\sqrt{\frac{\log(t/\delta)}{\nval_{t,\wval}}}
\quad\text{and}\quad
D =
C\left[
\sqrt{ \log ( t / \delta )  } 
\max_{t-\wval\le j\le t-1} | \gap_{j} - \gap_{t} |
+
\frac{ M^2 \log ( t / \delta) }{ \nval_{t, \wval} }
\right]
\]
Completing the squares gives
\[
\left( \sqrt{\excessrisk_{t,\wval}^{\va}(f_1)} - A \right)^2
\le 
\left( \sqrt{\excessrisk_{t,\wval}^{\va}(f_2)} + A \right)^2 + D,
\]
which implies
\begin{align*}
\sqrt{\excessrisk_{t,\wval}^{\va}(f_1)}
-
\sqrt{\excessrisk_{t,\wval}^{\va}(f_2)}
&\le 
2A + \sqrt{D} \\[4pt]
&\lesssim
M\sqrt{\frac{\log(t/\delta)}{\nval_{t,\wval}}} 
+ 
\left[\sqrt{ \log ( t / \delta )  } 
\max_{t-\wval\le j\le t-1} | \gap_{j} - \gap_{t} |
+
\frac{ M^2 \log ( t / \delta) }{ \nval_{t, \wval} } \right]^{1/2} \\[4pt]
&\lesssim
\sqrt{\log(t/\delta)} \cdot \left( \max_{t-\wval\le j\le t-1} | \gap_{j} - \gap_{t} | + \frac{M^2}{\nval_{t,\wval}} \right)^{1/2} \\[4pt]
&\lesssim
M\sqrt{\log(t/\delta)} \cdot \left( \max_{t-\wval\le j\le t-1} \TV(\distP_j,\distP_t) + \frac{1}{\nval_{t,\wval}} \right)^{1/2},
\end{align*}
where the last inequality is due to
\[
| \gap_{j} - \gap_{t} | \lesssim M^2 \cdot \TV(\distP_j,\distP_t).
\]
Finally, for every $f\in\{f_1,f_2\}$,
\begin{align*}
\left| \sqrt{\excessrisk_{t,\wval}^{\va}(f)} - \sqrt{\excessrisk_t(f)} \right|
\le 
\sqrt{\big| \excessrisk_{t,\wval}^{\va}(f) - \excessrisk_t(f) \big| }
&\le 
\max_{t-\wval\le j\le t-1} \sqrt{\big| \excessrisk_j(f) - \excessrisk_t(f) \big|} \\
&\le
2M\max_{t-\wval\le j\le t-1}\sqrt{\TV(\distP_j,\distP_t)},
\end{align*}
so
\begin{align*}
\sqrt{\excessrisk_t(f_1)}
-
\sqrt{\excessrisk_t(f_2)}
&\le
\sqrt{\excessrisk_{t,\wval}^{\va}(f_1)}
-
\sqrt{\excessrisk_{t,\wval}^{\va}(f_2)}
+
4M\max_{t-\wval\le j\le t-1}\sqrt{\TV(\distP_j,\distP_t)} \\[4pt]
&\lesssim
M\sqrt{\log(t/\delta)} \cdot \left( \max_{t-\wval\le j\le t-1} \TV(\distP_j,\distP_t) + \frac{1}{\nval_{t,\wval}} \right)^{1/2}.
\end{align*}
As $\widehat{f}=f_1$, this finishes the proof.

\subsection{Proof of \myCref{thm-select-tournament}}\label{sec-thm-select-tournament-proof}

We first prove the following lemma, which converts any performance guarantee of the subroutine $\alg$ to that of \myCref{alg-tournament}.

\begin{lemma}[From comparison to selection]\label{lem-select-tournament-reduction}
Take a performance metric $\perf:\{f_{\modelidx}\}_{\modelidx=1}^{\nmodel}\to\RR$, and $U:(0,1)\to\RR_+$. Fix $\delta\in(0,1)$. Suppose that the model comparison subroutine $\alg$ in \myCref{alg-tournament} satisfies the following property: given two models $h_1,h_2\in \{f_{\modelidx}\}_{\modelidx=1}^{\nmodel}$, it outputs a model $\widehat{h}\in\{h_1,h_2\}$ satisfying
\[
\PP\left( \perf(\widehat{h}) - \min\left\{\perf(h_1), \perf(h_2) \right\} \le U(\delta) \right) \ge 1 - \delta.
\]
Then the output $\widehat{f}$ of \myCref{alg-tournament} satisfies
\[
\PP\left( 
\perf(\widehat{f})
-
\min_{\modelidx\in[\nmodel]} \perf(f_{\modelidx})
\le
2U(\delta)
\right) \ge 1-\nmodel^2\delta.
\]
\end{lemma}

\begin{proof}[Proof of \myCref{lem-select-tournament-reduction}]

Given two models $f,f'\in\modelclass$, denote the output of $\alg$ by $\alg(f,f') \in \{f,f'\}$. For notational convenience we also set $\alg(f,f) = f$ for every $f\in\modelclass$. By a union bound, with probability at least $1-\nmodel^2\delta$,
\begin{equation}\label{eqn-lem-select-tournaments-proof-compare}
\perf\left( \alg\big( f_{\modelidx'},f_{\modelidx''} \big) \right) - \min_{\modelidx\in\{\modelidx',\modelidx''\}} \perf (f_{\modelidx})
\le 
U(\delta),
\qquad \forall \modelidx',\modelidx''\in [\nmodel].
\end{equation}
From now on suppose that \eqref{eqn-lem-select-tournaments-proof-compare} holds. Take $\bar{f} \in \{f_{\modelidx}\}_{\modelidx=1}^{\nmodel}$ such that $\perf(\bar{f}) = \min_{\modelidx\in[\nmodel]} \perf(f_{\modelidx})$.

If \myCref{alg-tournament} outputs $\widehat{f} = \bar{f}$, then there is nothing to prove. Now assume that $\widehat{f}\neq \bar{f}$. Then, there exists $\wval\in\ZZ_+$ such that at the end of the $K$-th while loop, $\bar{f}$ is not in $\survivorsnew$. Take the smallest such $K$. Let $g_K$ denote the pivot model $f$ during the $K$-th while loop, and let $\survivorsnew_K$ denote the set $\survivorsnew$ at the end of the $K$-th while loop. There are two cases.
\begin{enumerate}
\item If at the end of the $K$-th while loop, $\survivorsnew_K=\emptyset$ and \myCref{alg-tournament} outputs $\widehat{f}$, then during this while loop, a call of \myCref{alg-compare} has yielded $\alg(\{\widehat{f},\bar{f}\}) = \widehat{f}$, so by \eqref{eqn-lem-select-tournaments-proof-compare}, $\perf(\widehat{f}) - \perf(\bar{f}) \le U(\delta)$.
\item Otherwise, at the end of the $K$-th while loop, $\survivorsnew_K\neq\emptyset$. There are two cases.
	\begin{enumerate}
	\item If $g_K = \bar{f}$, then every $f\in\survivorsnew_K$, a call of $\alg$ has yielded $\alg(\{\bar{f},f\}) = f$, so by \eqref{eqn-lem-select-tournaments-proof-compare}, $\perf(f) - \perf(\bar{f}) \le U(\delta)$.
	Since the output $\widehat{f}$ must come from $\survivorsnew_K$, then automatically $\perf(\widehat{f}) - \perf(\bar{f}) \le U(\delta)$.
	\item If $g_K \neq \bar{f}$, then for every $f\in\survivorsnew_K$, a call of $\alg$ has yielded $\alg(\{g_K,f\}) = f$, so by \eqref{eqn-lem-select-tournaments-proof-compare},
	\begin{equation}\label{eqn-lem-select-tournaments-proof-part-2-1}
	\perf(f) - \perf(g_K) \le U(\delta).
	\end{equation}
	Since $\bar{f}\not\in\survivorsnew_K$, then a call of \myCref{alg-compare} has yielded $\alg(\{g_K,\bar{f}\}) = g_K$, so by \eqref{eqn-lem-select-tournaments-proof-compare},
	\begin{equation}\label{eqn-lem-select-tournaments-proof-part-2-2}
	\perf(g_K) - \perf(\bar{f}) \le U(\delta).
	\end{equation}
	Putting together \eqref{eqn-lem-select-tournaments-proof-part-2-1} and \eqref{eqn-lem-select-tournaments-proof-part-2-2} yields that for all $f\in\survivorsnew_K$,
	\[
	\perf(f) \le \perf(g_K) + U(\delta) \le \perf(\bar{f}) + 2U(\delta).
	\]
	Since the output $\widehat{f}$ must come from $\survivorsnew_K$, then automatically $\perf(\widehat{f}) - \perf(\bar{f}) \le 2U(\delta)$.
	\end{enumerate}
\end{enumerate}
In all the cases above, we have $\perf(\widehat{f}) - \perf(\bar{f}) \le 2U(\delta)$.
\end{proof}

We now prove \myCref{thm-select-tournament}. By the stronger bound \eqref{eqn-model-comparison-fast-root} in the proof of \myCref{thm-model-comparison}, we can set in \myCref{lem-select-tournament-reduction} $\perf(f) = \sqrt{\excessrisk_t(f)}$ and
\[
U(\delta)
=
CM \sqrt{ \log(t/\delta) } \cdot 
\min_{\wval\in[t-1]}\left(
\max_{t-\wval\le j\le t-1} \TV(\distP_j, \distP_t)
+
\frac{1}{\nval_{t,\wval}} 
\right)^{1/2},
\]
for some universal constant $C>0$. Then, by \myCref{lem-select-tournament-reduction} and the choice of $\delta'$, with probability $1-\delta$, the output $\widehat{f}$ of \myCref{alg-tournament} satisfies
\[
\sqrt{\excessrisk_t(\widehat{f})} - \min_{\modelidx\in[\nmodel]}\sqrt{\excessrisk_t(f_{\modelidx})}
\lesssim
U(\delta/\nmodel^2)
\lesssim
M \sqrt{\log(\nmodel t/\delta)}  \cdot 
\min_{\wval\in[t-1]}\left(
\max_{t-\wval\le j\le t-1} \TV(\distP_j, \distP_t)
+
\frac{1}{\nval_{t,\wval}} 
\right)^{1/2}.
\]
Here $\lesssim$ only hides universal constants. This finishes the proof.

\subsection{Proof of \myCref{thm-joint-selection}}\label{sec-thm-joint-selection-proof}

Recall that $|\dataset_j^{\tr}| = \ntrain_j$, $|\dataset_j^{\va}| = \nval_j$, $|\dataset_j| = \ntrain_j + \nval_j$, $\ntrain_{t,\wtrain} = \sum_{j=t-\wtrain}^{t-1}\ntrain_j$, $\nval_{t,\wval} = \sum_{j=t-\wval}^{t-1} \nval_j$, and $B_{t,k} = \ntrain_{t,k} + \nval_{t,k}$. Since there are $(t-1)|\metamodelclass|$ candidate models, then by \myCref{thm-select-tournament}, with probability at least $1-\delta/2$, the output $\widehat{f}$ of \myCref{alg-tournament} satisfies
\begin{equation}\label{eqn-joint-selection-proof-val}
\excessrisk_t(\widehat{f})
\lesssim
\min_{\modelclass\in\metamodelclass,\, k\in[t-1]} \excessrisk_t\big( \widehat{h}(\modelclass,k)\big)
+
M^2 \log \big(t^2|\metamodelclass| /\delta\big)  \cdot 
\min_{\wval\in[t-1]}\left\{
\max_{t-\wval\le j\le t-1} \TV(\distP_j, \distP_t)
+
\frac{1}{\nval_{t,\wval}} 
\right\}.
\end{equation}
By \myCref{thm-tradeoff} and a union bound, with probability at least $1-\delta/2$,
\begin{equation}\label{eqn-joint-selection-proof-tr}
\excessrisk_t\big( \widehat{h}(\modelclass,k)\big)
\lesssim 
\min_{f\in\modelclass}\excessrisk_t(f)
+
M^2\left(
r_{t,k}(\modelclass)
+
\frac{\log(t|\metamodelclass| / \delta)}{\ntrain_{t,k}}
\right)
+
M^2 \max_{t-k\le j\le t-1} \TV\left( \distP_j,\distP_t \right).
\end{equation}

Combining \eqref{eqn-joint-selection-proof-val} and \eqref{eqn-joint-selection-proof-tr} yields that, with probability at least $1-\delta$,
\begin{align*}
\excessrisk_t(\widehat{f})
&\lesssim
\min_{\modelclass\in\metamodelclass,\, k\in[t-1]} \left\{
\min_{f\in\modelclass}\excessrisk_t(f)
+
M^2\left(
r_{t,k}(\modelclass)
+
\frac{\log(t|\metamodelclass| / \delta)}{\ntrain_{t,k}}
\right)
+
M^2 \max_{t-k\le j\le t-1} \TV\left( \distP_j,\distP_t \right)
\right\} \\
&\qquad
+ M^2 \log \big(t^2|\metamodelclass| /\delta\big)  \cdot 
\min_{\wval\in[t-1]}\left\{
\max_{t-\wval\le j\le t-1} \TV(\distP_j, \distP_t)
+
\frac{1}{\nval_{t,\wval}} 
\right\} \\[4pt]
&\lesssim
\min_{\modelclass\in\metamodelclass,\, k\in[t-1]} \left\{
\min_{f\in\modelclass}\excessrisk_t(f)
+
M^2\left(
r_{t,k}(\modelclass)
+
\frac{\log(t|\metamodelclass| / \delta)}{\ntrain_{t,k}}
\right)
+
M^2 \max_{t-k\le j\le t-1} \TV\left( \distP_j,\distP_t \right)
\right\} \\
&\qquad
+ M^2 \log \big(t^2|\metamodelclass| /\delta\big)  \cdot 
\min_{k\in[t-1]}\left\{
\max_{t-k\le j\le t-1} \TV(\distP_j, \distP_t)
+
\frac{1}{\ntrain_{t,k}} 
\right\} \tag{by Assumption \myref{assumption-data-split}} \\[4pt]
&\lesssim
\log \big(t|\metamodelclass| /\delta\big)
\cdot \min_{\modelclass\in\metamodelclass,\, k\in[t-1]} \left\{
\min_{f\in\modelclass}\excessrisk_t(f)
+
M^2\left(
r_{t,k}(\modelclass)
+
\frac{1}{\ntrain_{t,k}}
\right)
+
M^2 \max_{t-k\le j\le t-1} \TV\left( \distP_j,\distP_t \right)
\right\}.
\end{align*}
Here the second inequality hides the constant $c$ in Assumption \myref{assumption-data-split}. This completes the proof.

\subsection{Proof of \myCref{lem-bias-variance-decomp-R}}\label{sec-lem-bias-variance-decomp-R-proof}

By the triangle inequality,
\begin{equation}\label{eqn-lem-bias-variance-decomp-proof-decomp-R}
\big| \gapRhat_{t,\wval} - \gapR_t \big|
\le 
\big| \gapRhat_{t,\wval} - \gapR_{t,\wval} \big|
+
\big| \gapR_{t,\wval} - \gapR_t \big|,
\end{equation}
where
\[
\gapR_{t,\wval} = \EE \big[\gapRhat_{t,\wval}\big] = \frac{\gap_{t,\wval}}{V_{t,\wval}}.
\]
By \myCref{lem-max-diff-ratio},
\begin{equation}\label{eqn-lem-bias-variance-decomp-proof-bias-R}
\big| \gapR_{t,\wval} - \gapR_t \big|
=
\left| \frac{\gap_{t,\wval}}{V_{t,\wval}} - \frac{\gap_t}{V_t} \right|
\le 
\max_{t-\wval\le j\le t-1} |\gapR_j - \gapR_t|.
\end{equation}
We have
\[
\gapRhat_{t,\wval} = \frac{1}{\nval_{t,\wval}} \sum_{j=t-\wval}^{t-1} \sum_{i=1}^{\nval_j} \frac{u_{j,i}}{V_{t,\wval}},
\]
where $u_{j,i} = \big[ f_1(\bx_{j,i}^{\va}) - y_{j,i}^{\va} \big]^2 - \big[ f_2(\bx_{j,i}^{\va}) - y_{j,i}^{\va} \big]^2$. By Assumptions \myref{assumption-bounded} and \myref{assumption-positive-variance}, $|u_{j,i}/V_{t,\wval}| \le 8M^2 / v$ for all $j$ and $i$. By Bernstein's concentration inequality (\myCref{lem-Bernstein}), with probability at least $1-\delta$,
\begin{equation}\label{eqn-lem-bias-variance-decomp-proof-variance-R}
\big| \gapRhat_{t,\wval} - \gapR_{t,\wval} \big|
\le 
\sdR_{t,\wval} \sqrt{\frac{2\log(2/\delta)}{\nval_{t,\wval}}} + \frac{16(M^2/v)\log(2/\delta)}{3\nval_{t,\wval}}.
\end{equation}
Substituting \eqref{eqn-lem-bias-variance-decomp-proof-bias-R} and \eqref{eqn-lem-bias-variance-decomp-proof-variance-R} into \eqref{eqn-lem-bias-variance-decomp-proof-decomp-R} completes the proof.

\subsection{Proof of \myCref{thm-select-tournament-R}}\label{sec-thm-select-tournament-R-proof}

We first prove the following theoretical guarantee for the $R^2$-based comparison subroutine \myCref{alg-compare-R}.

\begin{theorem}[Near-optimal model comparison with $R^2$]\label{thm-model-comparison-R}
Let Assumptions \myref{assumption-bounded} and \myref{assumption-positive-variance} hold. Choose $\delta\in(0,1)$ and set $\delta' = 1/(3 t)$ in \myCref{alg-tournament}. With probability at least $1 - \delta $, the output $\widehat{f}$ of \myCref{alg-compare-R} satisfies
\begin{equation}\label{eqn-oracle-model-comparison-R}
\max_{\modelidx\in[2]} \Rpop_t(f_{\modelidx}) - \Rpop_t(\widehat{f})
\lesssim
\log(t/\delta)\cdot\min_{\wval \in [t-1]} \left\{ 
\max_{t-\wval\le j\le t-1} \max_{\modelidx\in[2]} \big| \Rpop_j(f_{\modelidx}) - \Rpop_t(f_{\modelidx}) \big|
+
\frac{M^2 / \vlb}{ \sqrt{ \nval_{t,\wval} } } 
\right\}.
\end{equation} 
Here $\lesssim$ hides a universal constant.
\end{theorem}

\begin{proof}[Proof of \myCref{thm-model-comparison-R}]
Following the same argument as Theorem 4.2 in \cite{HHW24}, we can show that with probability at least $1-\delta$,
\begin{align*}
\max_{\modelidx\in[2]} \Rpop_t(f_{\modelidx}) - \Rpop_t(\widehat{f})
&\le 
\big| \gapRhat_{t,\widehat{\wval}} - \gapR_t \big| \\[4pt]
&\lesssim
\log(t/\delta) \cdot \min_{\wval\in[t-1]} \left\{ \max_{t-\wval\le j\le t-1} \big| \gapR_j - \gapR_t \big|
+ 
\frac{ \totalsdRhat_{t, \wval}  }{ \sqrt{ \nval_{t, \wval} } } + \frac{ M^2 / \vlb }{ \nval_{t, \wval} } \right\}.
\end{align*}
We finish the proof by noting that $\sdR_{t, \wval} \lesssim M^2 / v$ and
\[
\big| \gapR_j - \gapR_t \big| 
=
\left| \Big[ \Rpop_j(f_1) - \Rpop_j(f_2) \Big] - \Big[ \Rpop_t(f_1) - \Rpop_t(f_2) \Big] \right|
\le 
2\max_{\modelidx\in[2]} \left| \Rpop_j(f_{\modelidx}) - \Rpop_j(f_{\modelidx})  \right|.
\]
\end{proof}

We can now use \myCref{lem-select-tournament-reduction} to translate \myCref{thm-model-comparison} to a theoretical guarantee for general model selection. Set $\perf(f) = 1 - \Rpop_t(f)$ and
\[
U(\delta)
=
C\log(t/\delta)\cdot\min_{\wval \in [t-1]} \left\{ 
\max_{t-\wval\le j\le t-1} \max_{\modelidx\in[\nmodel]} \big| \Rpop_j(f_{\modelidx}) - \Rpop_t(f_{\modelidx}) \big|
+
\frac{M^2 / \vlb}{ \sqrt{ \nval_{t,\wval} } } 
\right\},
\]
for a sufficiently large universal constant $C>0$. Then for any $f$,
\[
\perf(f) - \min_{\modelidx\in[\nmodel]} \perf(f_{\modelidx})
=
\max_{\modelidx\in[\nmodel]}\Rpop_t(f_{\modelidx}) - \Rpop_t(f).
\]
By \myCref{lem-select-tournament-reduction} and the choice of $\delta'$, the output $\widehat{f}$ of \myCref{alg-tournament} satisfies that with probability at least $1-\delta$,
\begin{align*}
\max_{\modelidx\in[\nmodel]}\Rpop_t(f_{\modelidx}) - \Rpop_t(\widehat{f})
&\lesssim
U(\delta/\nmodel^2) \\[4pt]
&\lesssim
\log(\nmodel t/\delta)\cdot\min_{\wval \in [t-1]} \left\{ 
\max_{t-\wval\le j\le t-1} \max_{\modelidx\in[\nmodel]} \big| \Rpop_j(f_{\modelidx}) - \Rpop_t(f_{\modelidx}) \big|
+
\frac{M^2 / \vlb}{ \sqrt{ \nval_{t,\wval} } } 
\right\}.
\end{align*}
Here $\lesssim$ only hides universal constants.

\section{Technical Lemmas}

\begin{lemma}\label{lem-property-subroot}
Let $\psi:\RR_+\to\RR_+$ be a sub-root function with fixed point $r^*>0$. For all $A,a>0$, the function $\widetilde{\psi}(r) = A\psi(ar)$ is sub-root and its fixed point $\widetilde{r}$ satisfies 
\[
\frac{\min\{1,Aa\}^2}{a}r^*\le\widetilde{r}\le\frac{\max\{1,Aa\}^2}{a}r^*.
\] 
\end{lemma}

\begin{proof}[Proof of \Cref{lem-property-subroot}] 
It is easy to verify that $\widetilde{\psi}$ is subroot. We now study $\widetilde{r}$. First consider the case $a=1$. Since $A\psi(\widetilde{r}) = \widetilde{r}$, then $\psi(\widetilde{r})/\sqrt{\widetilde{r}} = \sqrt{\widetilde{r}}/A$. There are two cases.
\begin{itemize}
\item If $\widetilde{r}\ge r^*$, then $\psi(\widetilde{r})/\sqrt{\widetilde{r}} \le \psi(r^*)/\sqrt{r^*} = \sqrt{r^*}$, so $\widetilde{r} \le A^2r^*$.
\item If $\widetilde{r}\le r^*$, then $\psi(\widetilde{r})/\sqrt{\widetilde{r}} \ge \psi(r^*)/\sqrt{r^*} = \sqrt{r^*}$, so $\widetilde{r} \ge A^2r^*$.
\end{itemize}
Therefore, if $A<1$, then $A^2r^*\le \widetilde{r}< r^*$. If $A> 1$, then $r^*< \widetilde{r} \le A^2r^*$. This shows that 
\[
\min\{1,A\}^2r^*\le\widetilde{r}\le\max\{1,A\}^2r^*.
\] 

In the general case of $a>0$, the function $r\mapsto a^{-1}\psi(ar)$ is sub-root and has fixed point $a^{-1}r^*$. The proof is finished by noting $\widetilde{\psi}(r) = (Aa)\cdot a^{-1}\psi(ar)$.
\end{proof}

\begin{lemma}[Bernstein's concentration inequality]\label{lem-Bernstein}
Let $\{ x_i \}_{i=1}^n $ be independent random variables taking values in $[a, b]$ almost surely. Define the average variance $\sd^2 = \frac{1}{n} \sum_{i=1}^n \var (x_i)$. For any $\delta \in (0 , 1 )$, with probability at least $1-\delta$,
\[
\bigg| \frac{1}{n} \sum_{i=1}^{n} ( x_i - \EE x_i ) \bigg| 
\le \sd \sqrt{ \frac{ 2 \log ( 2 / \delta)  }{ n } } + \frac{ 2  (b-a) \log (2 / \delta) }{3 n }.
\]
\end{lemma}

\begin{proof}[Proof of \Cref{lem-Bernstein}]
Inequality (2.10) in \cite{BLM13} implies that for any $t \geq 0$,
\[
\PP \bigg(
\frac{1}{n} \sum_{i=1}^{n} ( x_i - \EE x_i ) > t
\bigg) \le \exp \left(
- \frac{ n t^2 / 2 }{ \sd^2 + (b - a) t / 3 }
\right) .
\]
Fix $\delta \in (0, 1)$. Then,
\begin{align*}
&\quad \exp  \left(
- \frac{ n t^2 / 2 }{ \sd^2 + (b-a) t / 3 }
\right) \le \delta \\[4pt]
 \Leftrightarrow &\quad \frac{n  t^2 }{2} \geq \sd^2 \log (1 / \delta) + \frac{t (b-a) \log ( 1 / \delta ) }{ 3 } \\[4pt]
 \Leftrightarrow &\quad \frac{n}{2} \bigg(
t - \frac{  (b-a) \log (1 / \delta) }{ 3n }
\bigg)^2 \ge \sd^2 \log (1 / \delta)  + \frac{n}{2} \bigg( \frac{  (b-a) \log (1 / \delta) }{3 n } \bigg)^2 \\[4pt]
 \Leftarrow &\quad \bigg(
t - \frac{  (b-a) \log (1 / \delta) }{ 3n }
\bigg)^2 \ge  \bigg( \sd \sqrt{ \frac{ 2 \log (1 / \delta)  }{ n } } + \frac{  (b-a) \log (1 / \delta) }{3 n } \bigg)^2 \\[4pt]
 \Leftarrow &\quad t \ge \sd \sqrt{ \frac{ 2 \log (1 / \delta)  }{ n } } + \frac{ 2  (b-a) \log (1 / \delta) }{ 3n } .
\end{align*}
Hence,
\[
\PP \left(
\frac{1}{n} \sum_{i=1}^{n} ( x_i - \EE x_i ) >
\sd \sqrt{ \frac{ 2 \log (1 / \delta)  }{ n } } + \frac{ 2  (b-a) \log (1 / \delta) }{3 n } 
\right) \le  \delta.
\]
Replacing each $x_i$ by $-x_i$ gives bounds on the lower tail and the absolute deviation.
\end{proof}

\begin{lemma}\label{lem-max-diff-ratio}
For all $a,a_1,...,a_n \ge 0$ and $b,b_1,...,b_n > 0$, it holds that
\[
\left| \frac{a}{b} - \frac{\sum_{i=1}^n a_i}{\sum_{i=1}^n b_i} \right| \le \max_{i\in[n]} \left| \frac{a}{b} - \frac{a_i}{b_i} \right|.
\]
\end{lemma}

\begin{proof}[Proof of \Cref{lem-max-diff-ratio}]
This is due to
\begin{align*}
\left| \frac{a}{b} - \frac{\sum_{i=1}^n a_i}{\sum_{i=1}^n b_i} \right|
=
\left| \frac{a}{b} - \sum_{i=1}^n \frac{b_i}{\sum_{j=1}^n b_j}\cdot \frac{a_i}{b_i} \right|
&=
\left| \sum_{i=1}^n \frac{b_i}{\sum_{j=1}^n b_j}\cdot \left( \frac{a}{b} - \frac{a_i}{b_i} \right) \right| \\[4pt]
&\le 
\sum_{i=1}^n \frac{b_i}{\sum_{j=1}^n b_j}\cdot \left|  \frac{a}{b} - \frac{a_i}{b_i}  \right|
\le 
\max_{i\in[n]} \left| \frac{a}{b} - \frac{a_i}{b_i} \right|.
\end{align*}
This finishes the proof.
\end{proof}

\section{Extension: Model Selection with the $R^2$ Metric}\label{sec-select-R2}

\subsection{Algorithm Design}

In this section, we propose a variant of our model selection method that is tailored to the $R^2$ metric. We consider the following population form:
\begin{equation}\label{eqn-R2-population}
\Rpop_t(f) = 1 - \frac{\EE_{(\bx,y)\sim\distP_t}\left[ \left( f(\bx) - y \right)^2 \right]}{\EE_{(\bx,y)\sim\distP_t}[y^2]},
\end{equation}
Define $\vardenom_t = \EE_{(\bx,y)\sim\distP_t}[y^2]$, then $\Rpop_t(f) = 1 - L_t(f) / V_t$. For simplicity, we assume that at the beginning of each period $t\in\ZZ_+$, we have access to $\{\vardenom_{\tau}\}_{\tau=1}^{t-1}$. In our numerical experiments, we will approximate $V_{\tau}$ by its empirical counterpart computed from the validation data $\dataset_{\tau}^{\va}$. For the population $R^2$ metric to be well defined, we assume for simplicity that $\{\vardenom_t\}_{t=1}^{\infty}$ are bounded away from zero.

\begin{assumption}[Uniformly lower bounded second moments]\label{assumption-positive-variance}
There exists $\vlb>0$ such that $\vardenom_t \ge \vlb$ for all $t\in\ZZ_+$.
\end{assumption}

We first define the model comparison subroutine, which aims to output the better of two given candidate models $f_1$ and $f_2$. Define the $R^2$ performance gap
\[
\gapR_t = \Rpop_t(f_2) - \Rpop_t(f_1) = \frac{L_t(f_1) - L_t(f_2)}{\vardenom_t}.
\]
Then $f_2$ outperforms $f_1$ if and only if $\gapR_t > 0$. For each window size $\wval\in[t-1]$, we can form a rolling-window estimator of $\gapR_t$, given by
\begin{equation}\label{eqn-performance-gap-hat-R}
\gapRhat_{t,\wval}
=
\frac{\widehat{\gap}_{t,\wval}}{V_{t,\wval}}
\end{equation}
where $\widehat{\gap}_{t,\wval}$ is defined by \eqref{eqn-performance-gap-hat}, and $V_{t,\wval} = \frac{1}{\nval_{t,\wval}} \sum_{\tau=t-k}^{t-1} \nval_{\tau} V_{\tau}$. We establish a bias-variance decomposition for the estimation error of $\gapRhat_{t,\wval}$.

\begin{lemma}[Bias-variance decomposition]\label{lem-bias-variance-decomp-R}
Let Assumptions \ref{assumption-independence}, \ref{assumption-bounded} and \ref{assumption-positive-variance} hold. Let $\sdR_{t,\wval} = \sd_{t,\wval} / V_{t,\wval}$, where $\sd_{t,\wval}$ is defined by \eqref{eqn-psi-hat}. For $\delta\in(0,1)$, define $M'=8M^2$,
\begin{align*}
& \phiR (t , \wval ) = \max_{t-\wval\le \tau\le t-1} \big| \gapR_{\tau} - \gapR_t \big| , \\[4pt]
& \psiR (t,\wval,\delta) 
= 
\begin{cases}
M' / \vlb,&\ \text{if } \nval_{t,\wval}=1 \\[6pt]
\displaystyle \sdR_{t, \wval}  \sqrt{ \frac{2 \log( 2 / \delta) }{ \nval_{t, \wval} } } + \frac{ 2 (M'/ \vlb) \log ( 2 / \delta) }{ 3 \nval_{t, \wval} } , &\ \text{if } \nval_{t,\wval}\ge 2
\end{cases} .
\end{align*}
With probability at least $1-\delta$,
\[
\big| \gapRhat_{t , \wval} - \gapR_t \big| \le \phiR (t , \wval ) + \psiR(t, \wval, \delta).
\]
\end{lemma}

\begin{proof}
See \myCref{sec-lem-bias-variance-decomp-R-proof}.
\end{proof}

Based on \myCref{lem-bias-variance-decomp-R} and following the same idea as \eqref{eqn-psi-hat}, we form a data-driven proxy for $\psiR(t,\wval,\delta)$, given by
\begin{equation}\label{eqn-psi-hat-R}
\psiRhat (t, \wval, \delta) 
=
\begin{cases}
M' / \vlb ,&\ \text{if } \nval_{t,\wval}=1 \\[6pt]
\displaystyle \totalsdRhat_{t, \wval}  \sqrt{ \frac{2 \log( 2 / \delta) }{ \nval_{t, \wval} } } + \frac{ 8 (M' / \vlb) \log ( 2 / \delta) }{ 3 ( \nval_{t, \wval} - 1 ) } , &\ \text{if } \nval_{t,\wval}\ge 2
\end{cases},
\end{equation}
where
\begin{equation}\label{eqn-sigma-hat-R} 
\totalsdRhat_{t, \wval} = \frac{\totalsdhat_{t, \wval}}{V_{t,\wval}},
\end{equation}
and $\totalsdhat_{t, \wval}$ is defined in \eqref{eqn-sigma-hat}. Following the same idea as \eqref{eqn-phi-hat}, we also form a data-driven proxy for $\phiR(t,\wval,\delta)$:
\begin{equation}\label{eqn-phi-hat-R}
\phiRhat (t, \wval, \delta)= 
\max_{i \in [\wval]} \bigg( \big| \gapRhat_{t, \wval} - \gapRhat_{t, i} \big|
-  \big[ \psiRhat\left(t, \wval , \delta \right) + \psiRhat\left(t, i , \delta \right)  \big] \bigg)_+.
\end{equation}
This yields \myCref{alg-compare-R} as the model comparison routine. 

\begin{algorithm}[h]
	\begin{algorithmic}
	\STATE {\bf Input:} Candidate models $\{f_1,f_2\}$, validation data $\{ \dataset^{\va}_{\tau} \}_{\tau=1}^{t-1}$, variances $\{\vardenom_{\tau}\}_{\tau=1}^{t-1}$, hyperparameters $\delta'$, $M'$ and $v$. \\
	\FOR{$\wval = 1,\cdots, t-1$}
	\STATE Compute $\gapRhat_{t,\wval}$, $\totalsdRhat_{t,\wval}$, $\psiRhat (t,\wval,\delta')$ and $\phiRhat (t,\wval,\delta')$ according to \eqref{eqn-performance-gap-hat-R}, \eqref{eqn-sigma-hat-R}, \eqref{eqn-psi-hat-R} and \eqref{eqn-phi-hat-R}.
	\ENDFOR
	\STATE Choose window size
	\begin{equation}\label{GL-k-hat-R}
	\widehat{\wval} \in \argmin_{ \wval \in [t-1] }  \Big\{ \phiRhat (t, \wval, \delta')  + \psiRhat (t , \wval , \delta')  \Big\}.
	\end{equation}
	\STATE Select $\widehat{\modelidx}=1$ if $\gapRhat_{t,\widehat{\wval}} \le 0$, and $\widehat{\modelidx}=2$ otherwise. \\
	\RETURN $\widehat{f} = f_{\widehat{\modelidx}}$.
	\caption{Adaptive Rolling Window for Model Comparison ($R^2$ Metric)}
	\label{alg-compare-R}
	\end{algorithmic}
\end{algorithm}

By using \myCref{alg-compare-R} as the pairwise comparison subroutine $\alg$ in \myCref{alg-tournament}, we obtain an $R^2$-based model selection algorithm, which we call $\adaptiveR$. We establish the following guarantee in terms of the $R^2$ metric.

\begin{theorem}[Near-optimal model selection with $R^2$]\label{thm-select-tournament-R}
Let Assumptions \ref{assumption-independence} and \ref{assumption-bounded} hold. Choose $\delta\in(0,1)$ and set $\delta' = 1/(3\nmodel^2 t)$ in \myCref{alg-tournament}. With probability at least $1-\delta$, the output $\widehat{f}$ of $\adaptiveR$ satisfies
\[
\max_{\modelidx\in[\nmodel]} \Rpop_t(f_{\modelidx}) - \Rpop_t(\widehat{f})
\lesssim
\log(\nmodel t/\delta)\cdot\min_{\wval \in [t-1]} \left\{ 
\max_{t-\wval\le \tau\le t-1} \max_{\modelidx\in[\nmodel]} \big| \Rpop_{\tau}(f_{\modelidx}) - \Rpop_t(f_{\modelidx}) \big|
+
\frac{M^2 / \vlb}{ \sqrt{ \nval_{t,\wval} } } 
\right\}.
\]
Here $\lesssim$ hides a universal constant.
\end{theorem}

\begin{proof}[Proof of \myCref{thm-select-tournament-R}]
See \myCref{sec-thm-select-tournament-R-proof}.
\end{proof}

\myCref{thm-select-tournament-R} shares a similar interpretation as \myCref{thm-select-tournament}. The term
\[
\max_{t-\wval\le \tau\le t-1} \max_{\modelidx\in[\nmodel]} \big| \Rpop_{\tau}(f_{\modelidx}) - \Rpop_t(f_{\modelidx}) \big|
\]
quantifies the shift in the $R^2$ metric of the models within the last $\wval$ periods, and $(M^2/v) / \sqrt{\nval_{t,\wval}}$ represents the statistical uncertainty associated with the $\nval_{t,\wval}$ validation data points. Together, they represent the errors that arise when using a fixed validation window $\wval$ to select models. \myCref{thm-select-tournament-R} shows that our $R^2$-based model selection algorithm $\adaptiveR$ is comparable to an oracle that uses the optimal validation window in hindsight to attain the highest $R^2$.

\subsection{Experiment Results for $\adaptiveR$}\label{sec-experiments-adaptive-R2}

We now present experiment results for $\adaptiveR$. We set its hyperparameters $\delta'=0.1$ and $M'=10$. Its performance is similar to the MSE-based approach $\adaptive$. We will report results for the $R^2$ metric that benchmarks against a zero forecast.

\begin{figure}[H]
\centering
\caption{Box Plot of Out-of-Sample $R^2$ of $\adaptiveR$, $\adaptive$ and Benchmarks for $17$ Industry Portfolios.\label{fig-boxplot-supp}}
\includegraphics[scale=0.5]{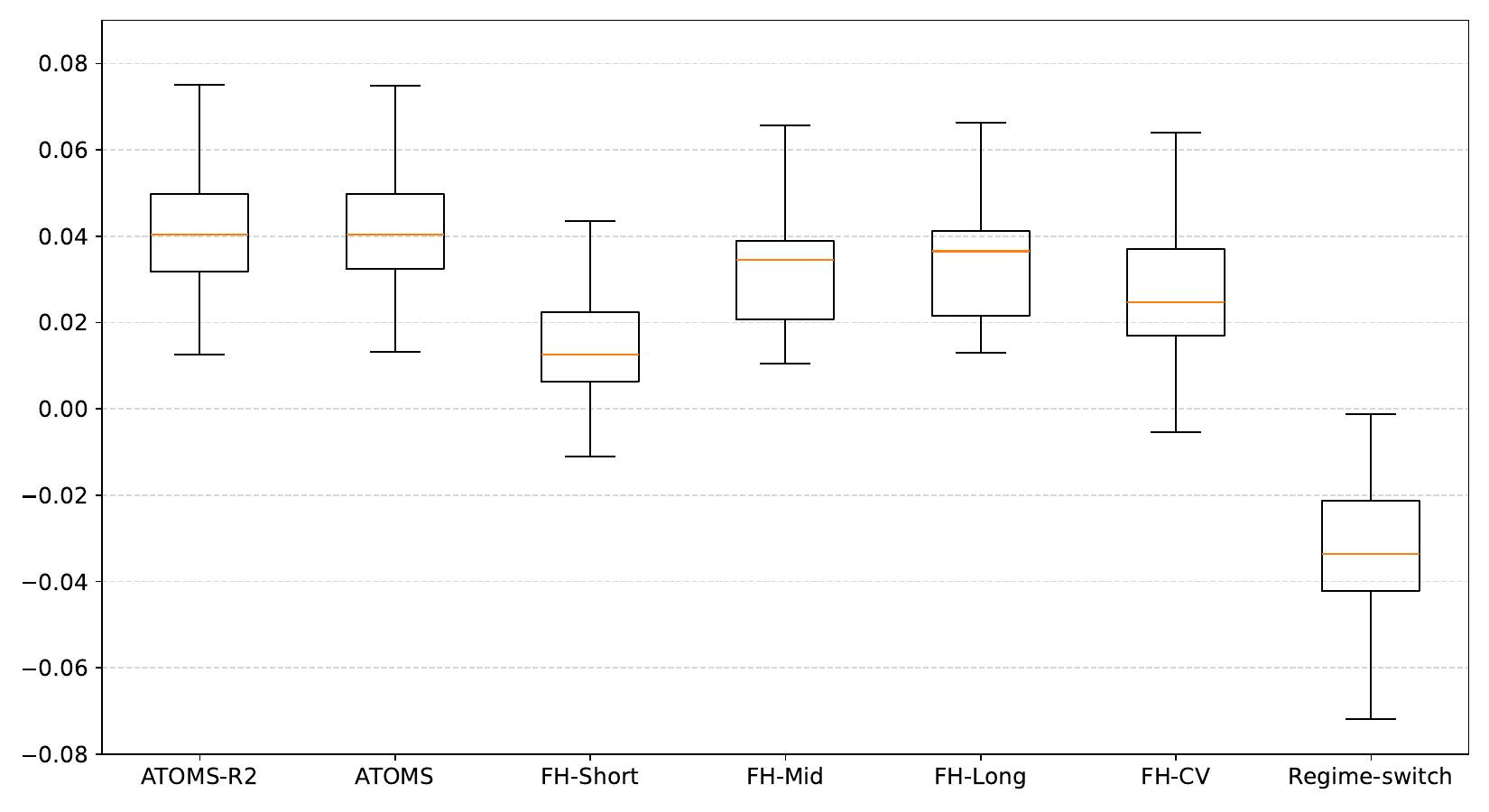}
\bnotefig{This figure describes the distribution of each method's OOS $R^2$. Each box corresponds to all industries and all years in our OOS horizon.}
\end{figure}

\begin{figure}[H]
	\centering
	\caption{Annual Out-of-Sample $R^2$ of $\adaptiveR$ and $\adaptive$ for $17$ Industry Portfolios. \label{fig-industry-yearly-supp}}

    \begin{subfigure}{0.24\textwidth}
    	\centering
        \includegraphics[width=\linewidth]{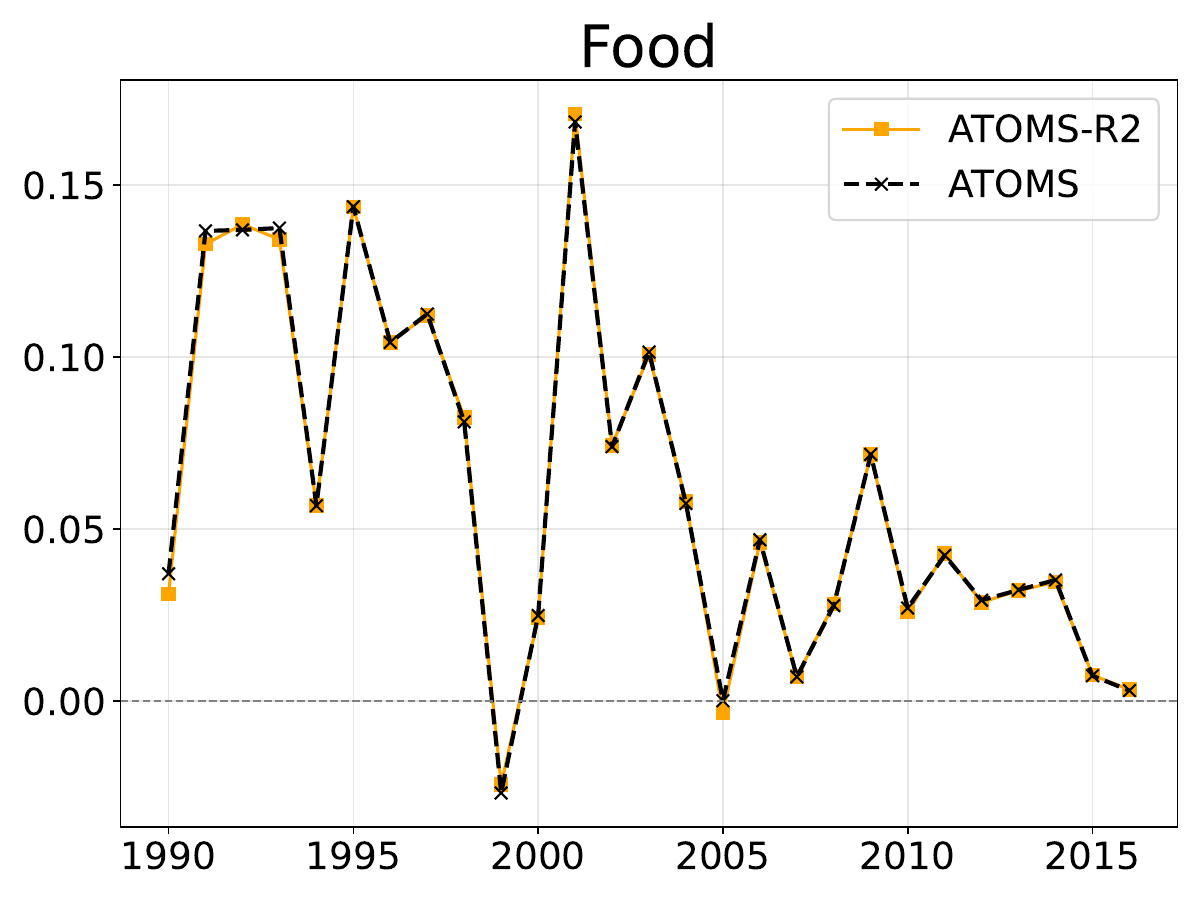}
	\end{subfigure}
    \begin{subfigure}{0.24\textwidth}
        \centering
        \includegraphics[width=\linewidth]{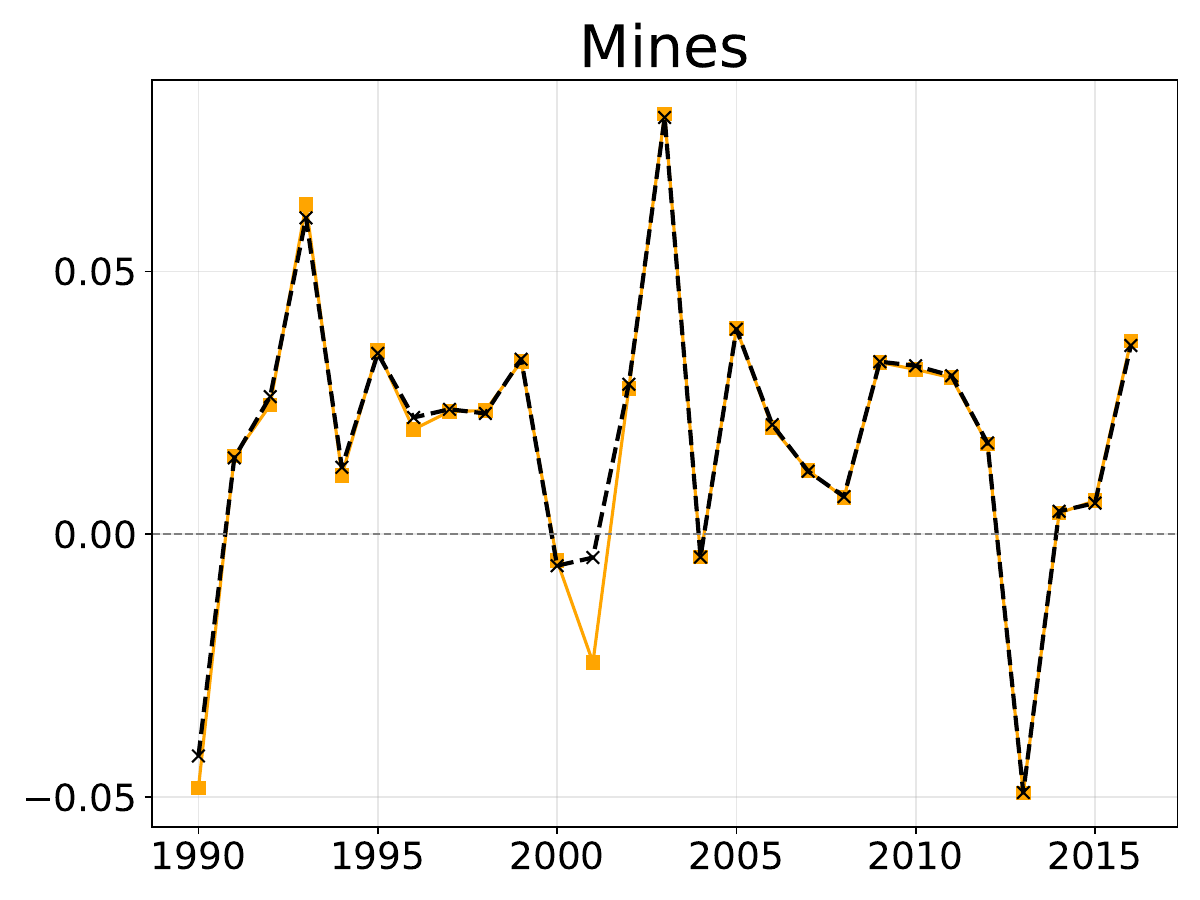}
	\end{subfigure}
    \begin{subfigure}{0.24\textwidth}
        \centering
        \includegraphics[width=\linewidth]{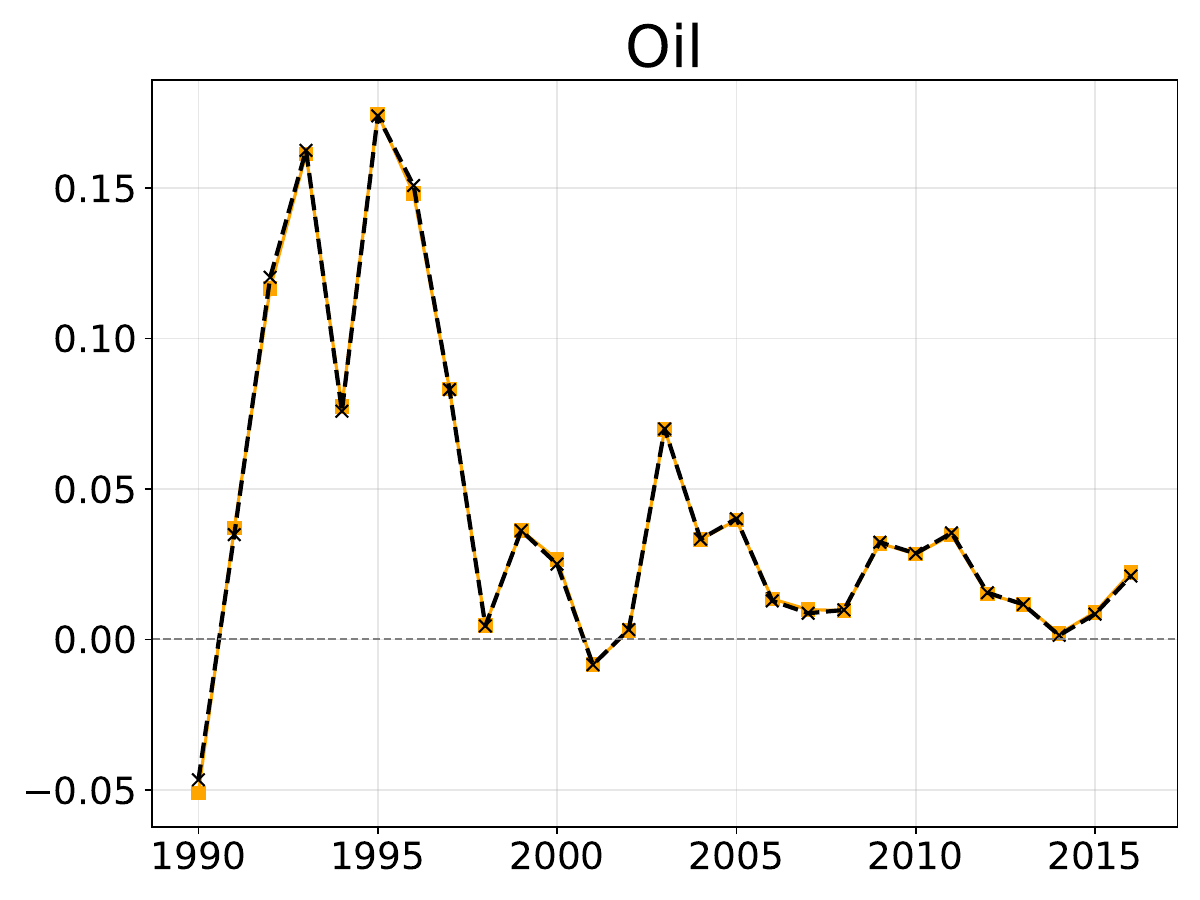}
	\end{subfigure}
    \begin{subfigure}{0.24\textwidth}
    	\centering
        \includegraphics[width=\linewidth]{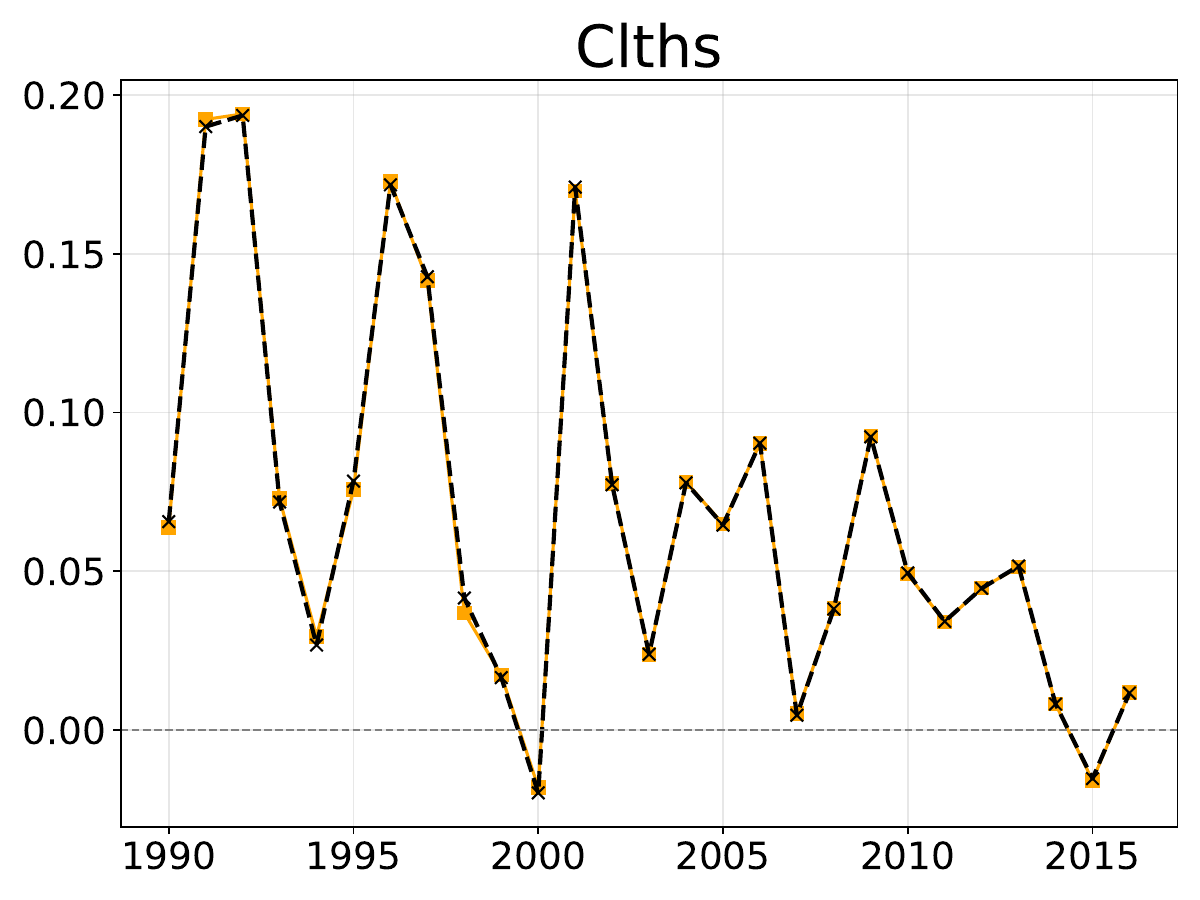}
	\end{subfigure}

    \begin{subfigure}{0.24\textwidth}
        \centering
        \includegraphics[width=\linewidth]{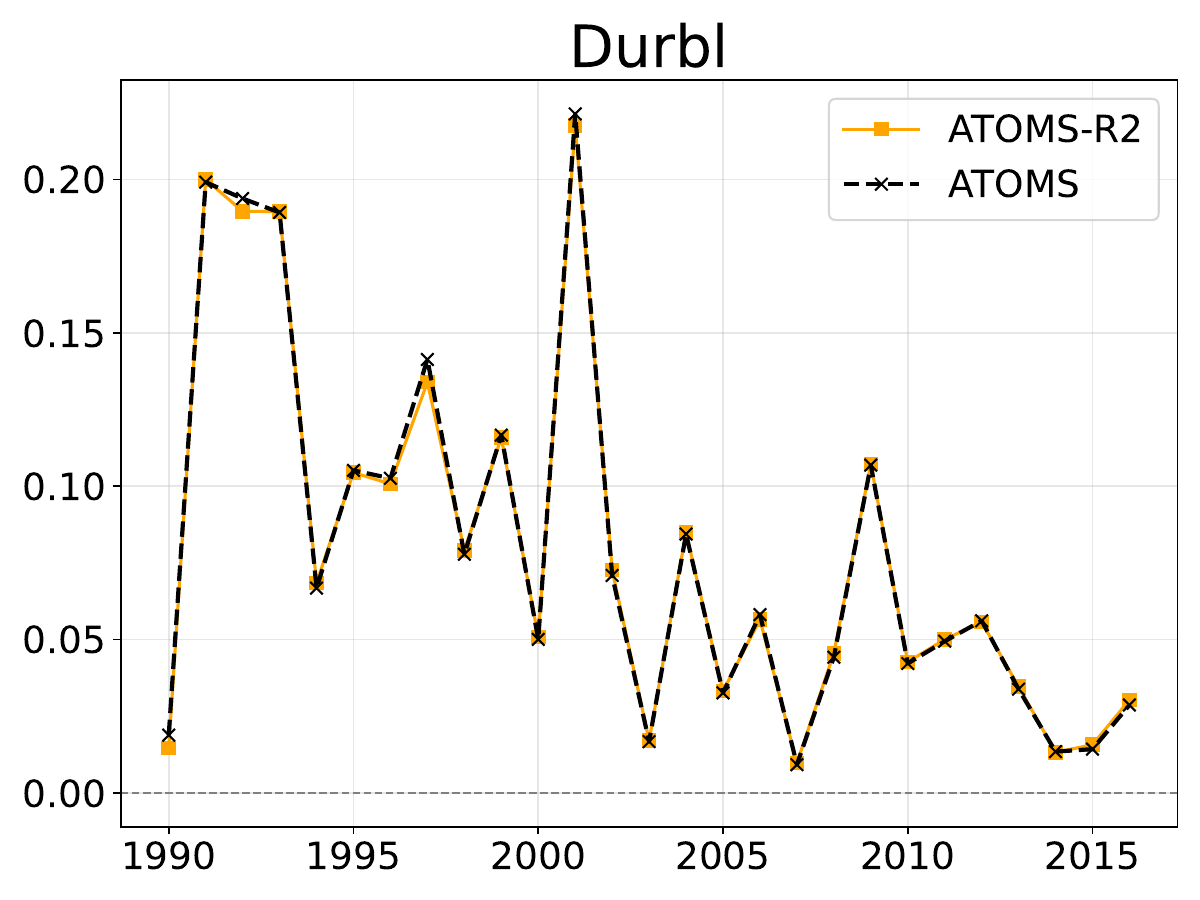}
	\end{subfigure}
    \begin{subfigure}{0.24\textwidth}
        \centering
        \includegraphics[width=\linewidth]{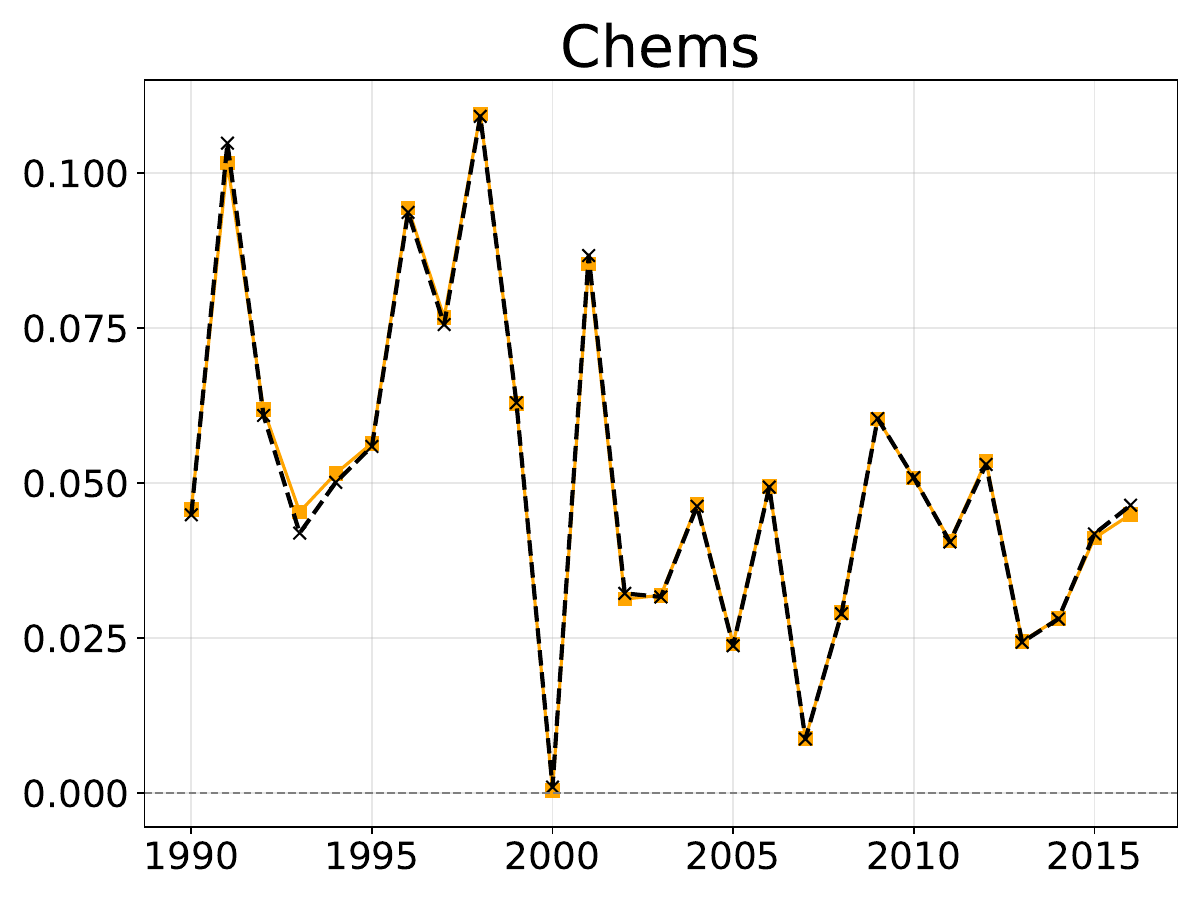}
	\end{subfigure}
    \begin{subfigure}{0.24\textwidth}
    	\centering
        \includegraphics[width=\linewidth]{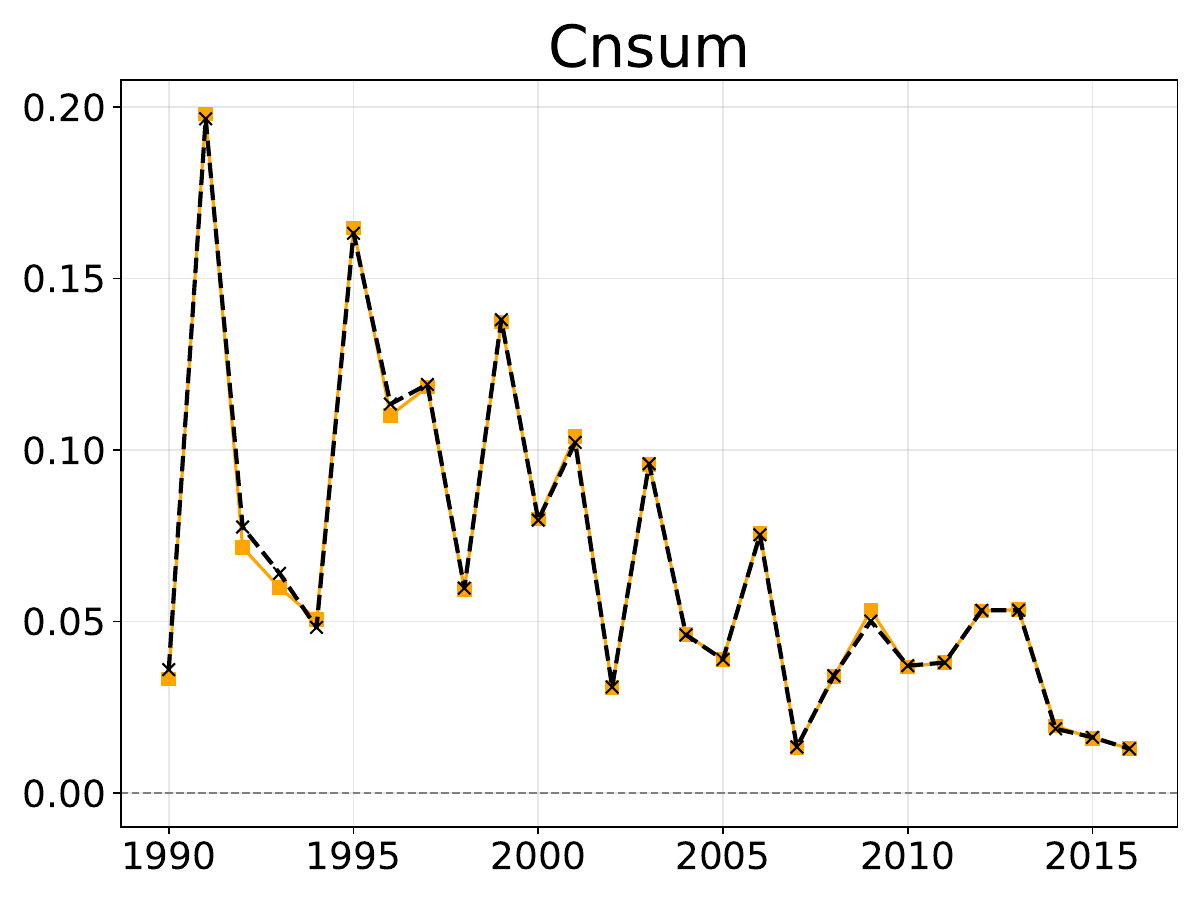}
	\end{subfigure}
    \begin{subfigure}{0.24\textwidth}
        \centering
        \includegraphics[width=\linewidth]{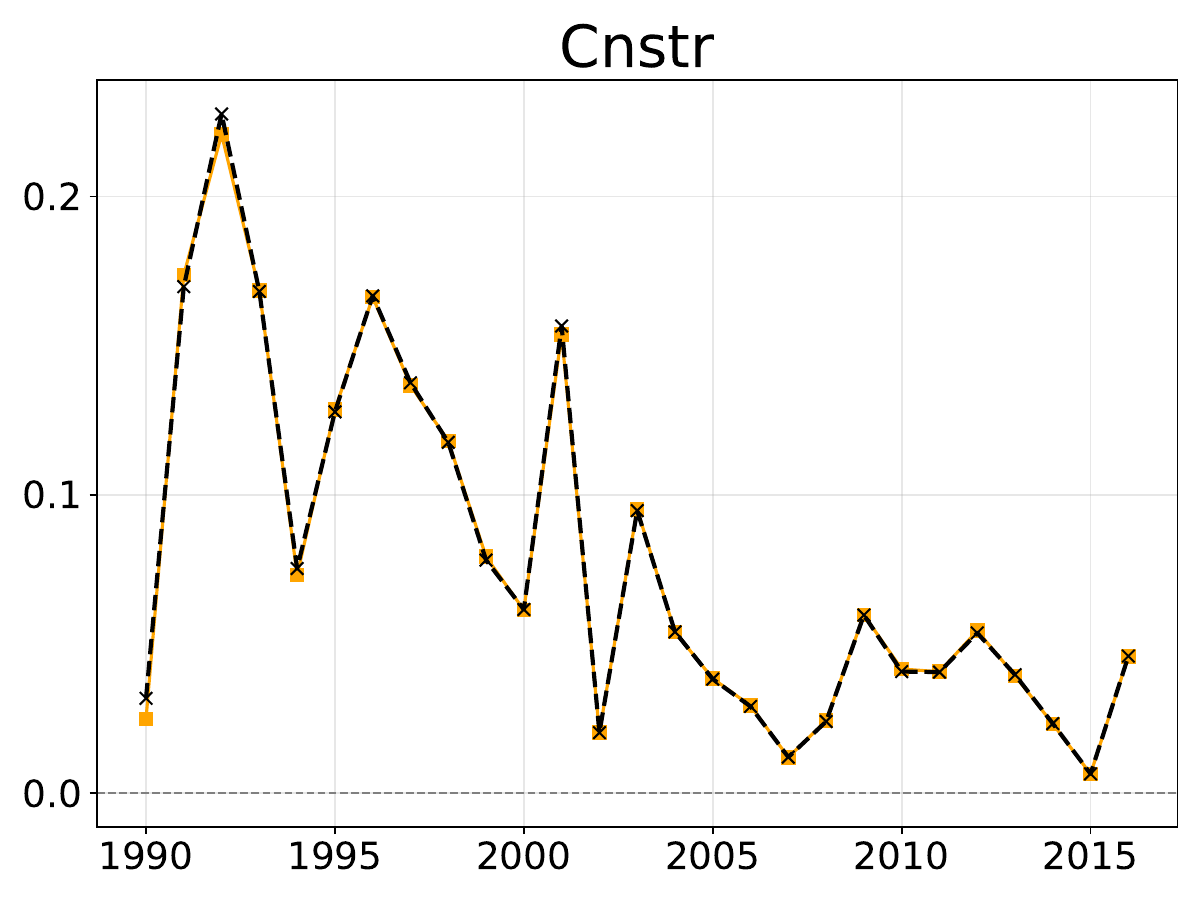}
	\end{subfigure}

    \begin{subfigure}{0.24\textwidth}
        \centering
        \includegraphics[width=\linewidth]{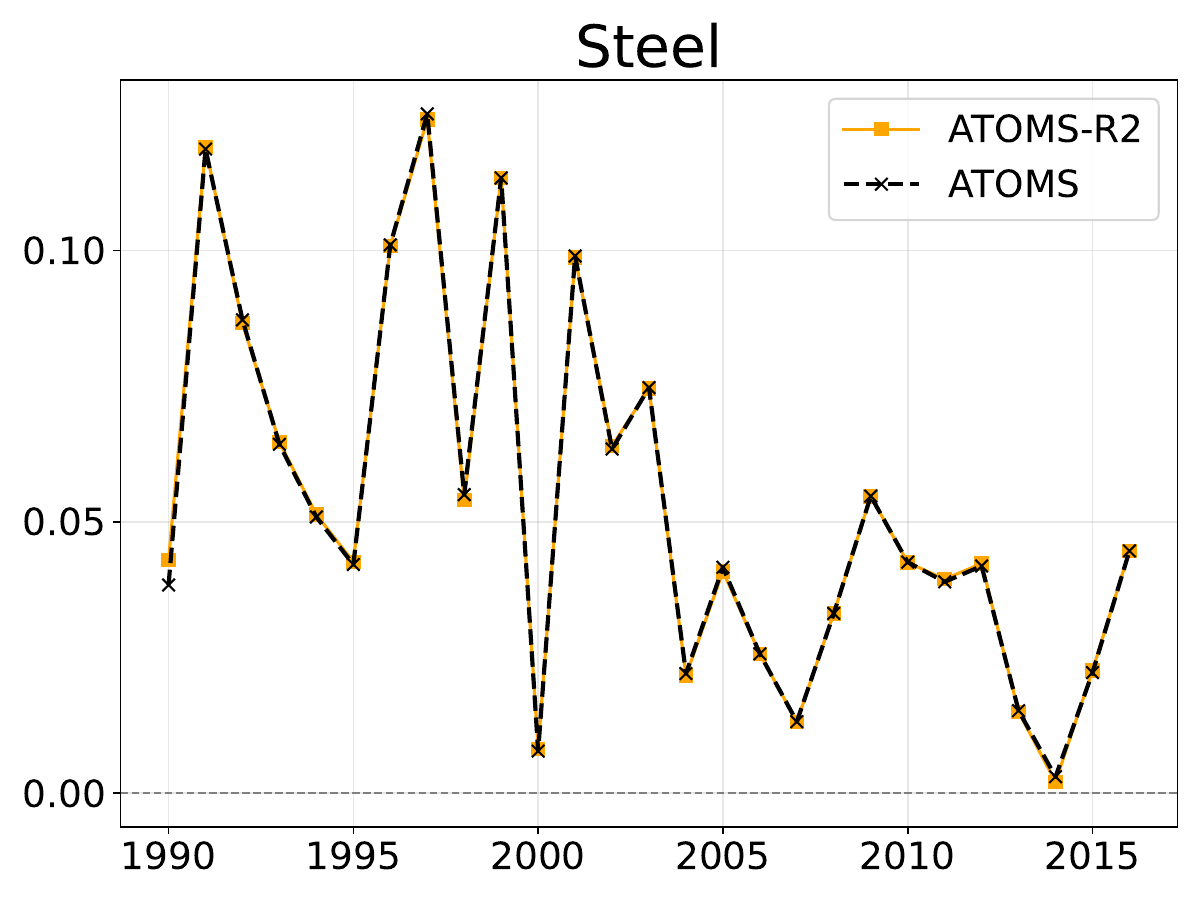}
	\end{subfigure}
    \begin{subfigure}{0.24\textwidth}
    	\centering
        \includegraphics[width=\linewidth]{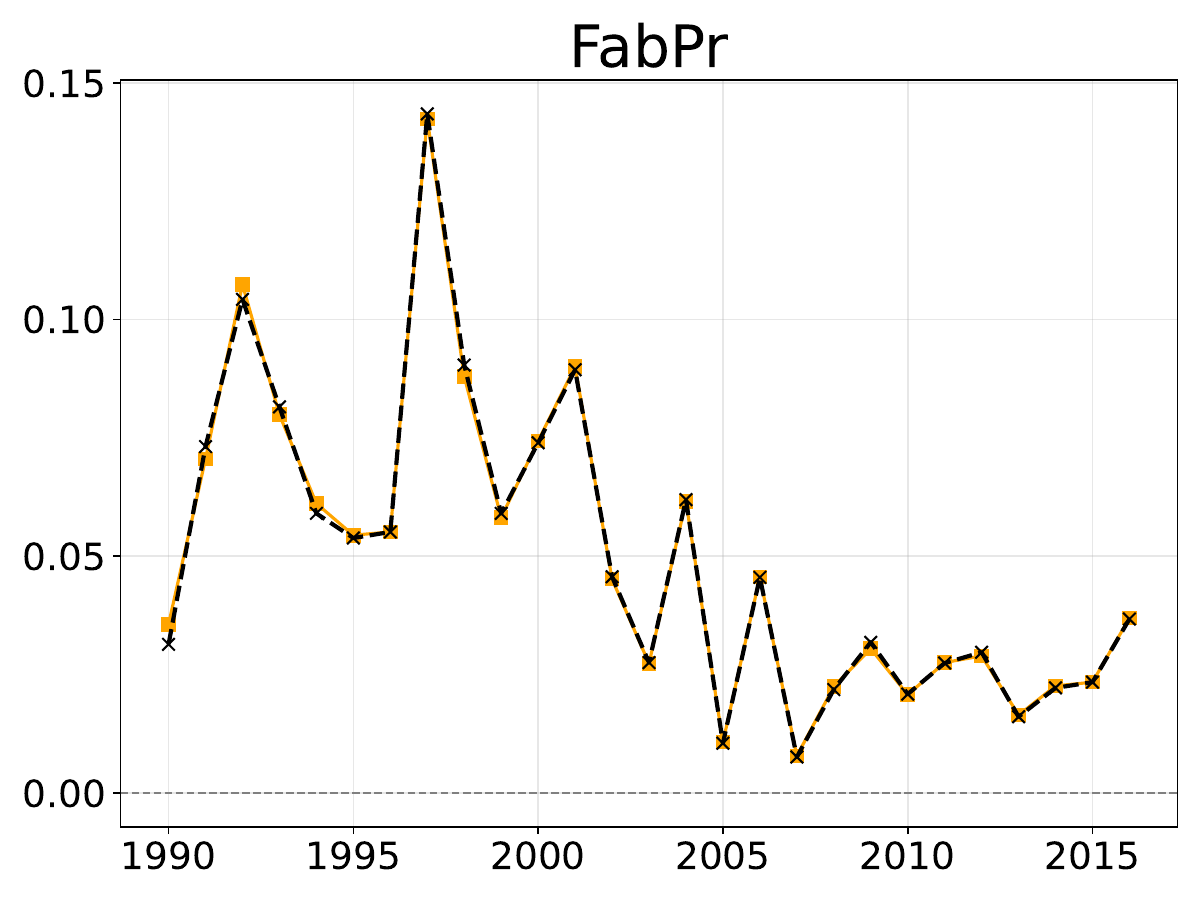}
	\end{subfigure}
    \begin{subfigure}{0.24\textwidth}
        \centering
        \includegraphics[width=\linewidth]{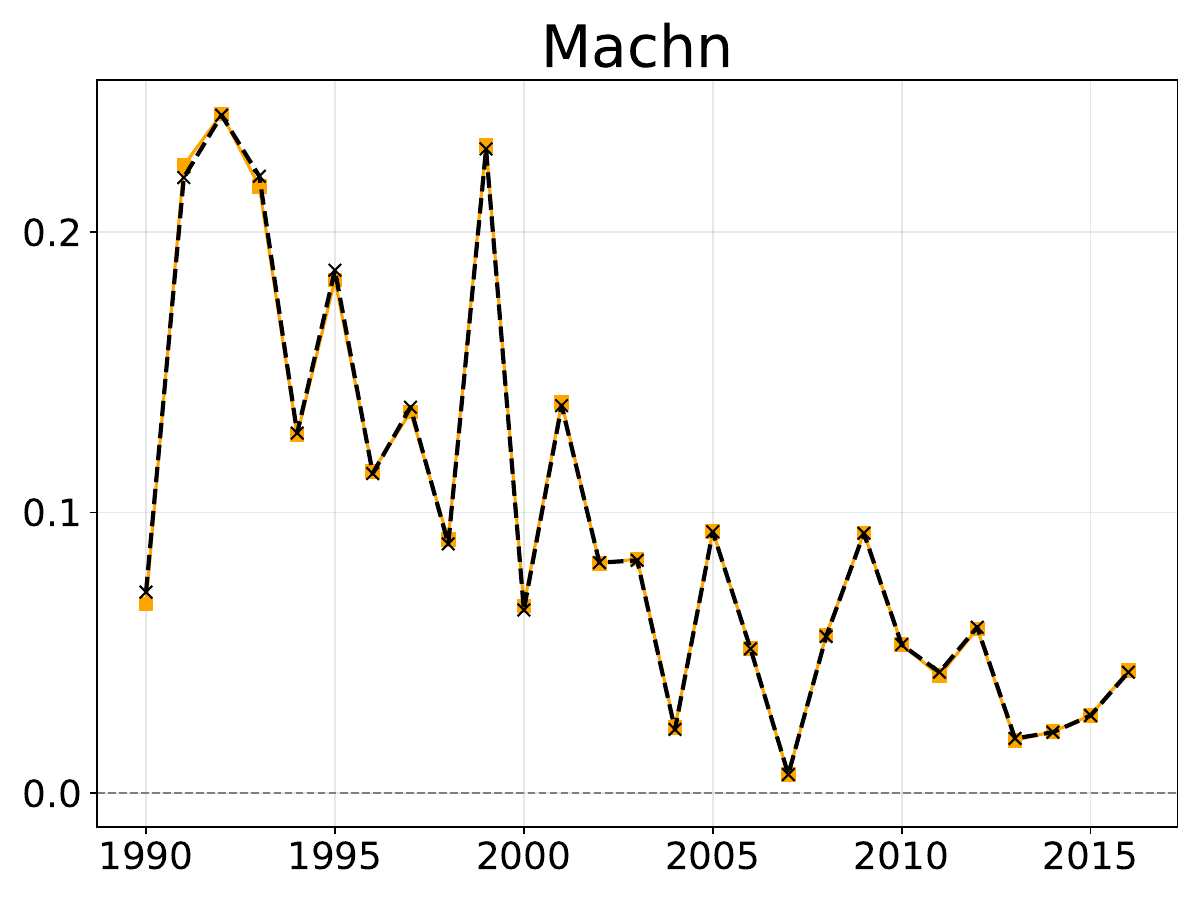}
	\end{subfigure}
    \hfill
    \begin{subfigure}{0.24\textwidth}
        \centering
        \includegraphics[width=\linewidth]{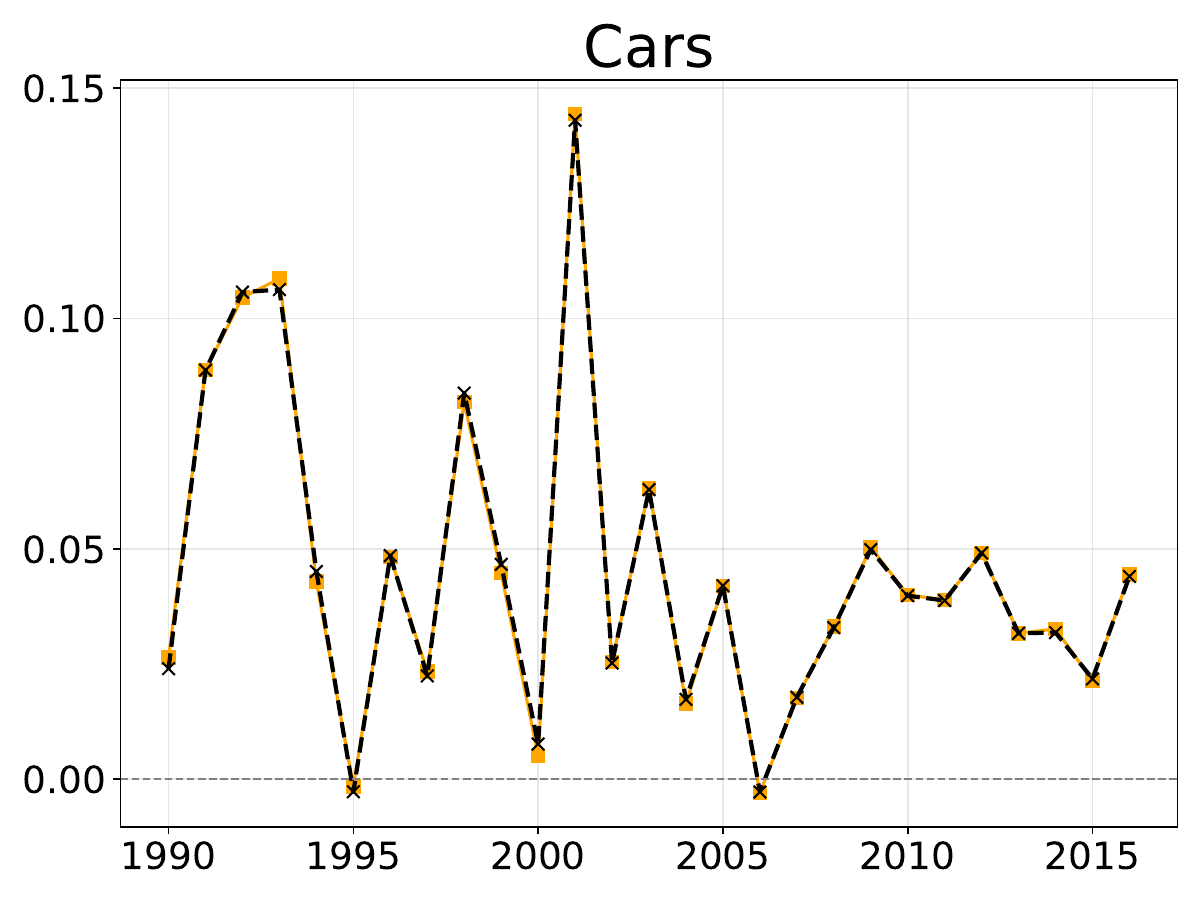}
	\end{subfigure}

    \begin{subfigure}{0.24\textwidth}
    	\centering
        \includegraphics[width=\linewidth]{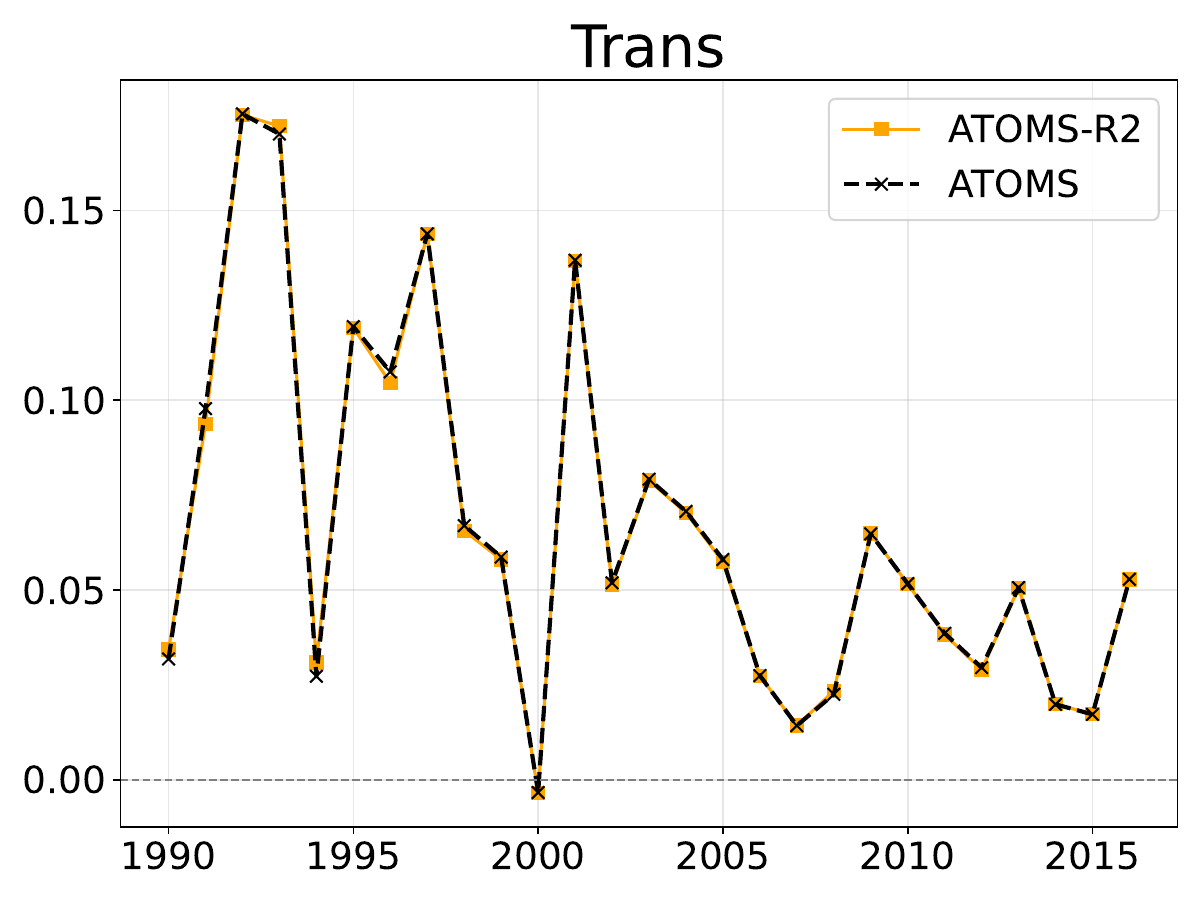}
	\end{subfigure}
    \begin{subfigure}{0.24\textwidth}
        \centering
        \includegraphics[width=\linewidth]{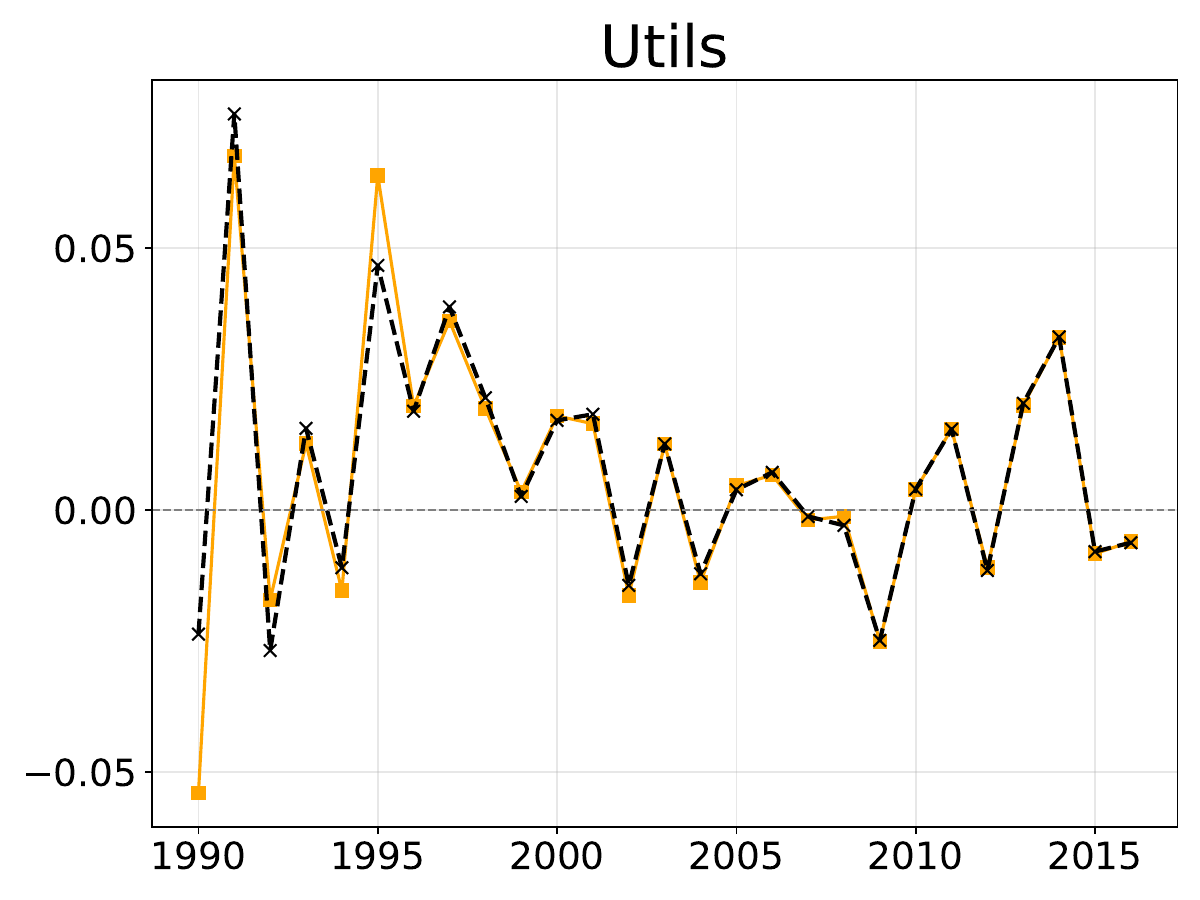}
	\end{subfigure}
    \begin{subfigure}{0.24\textwidth}
        \centering
        \includegraphics[width=\linewidth]{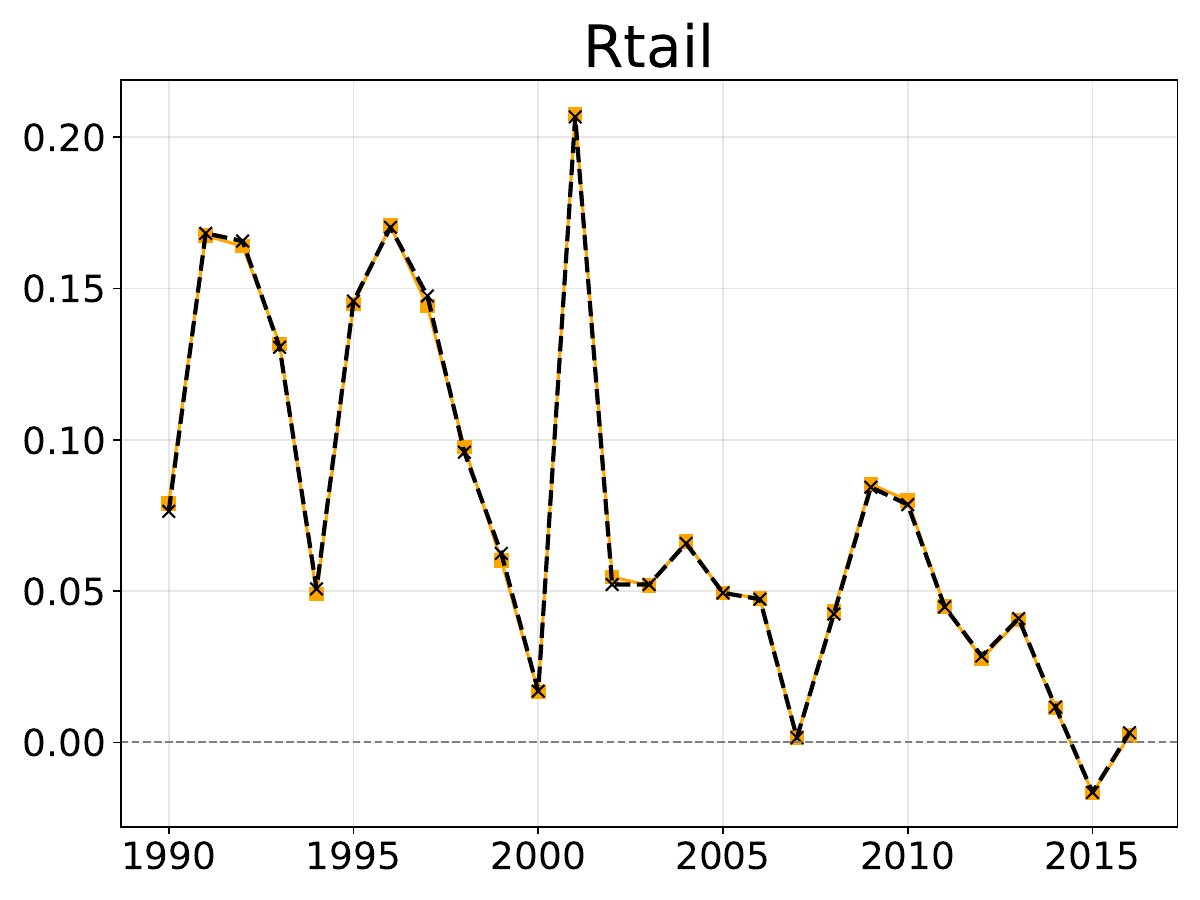}
	\end{subfigure}
    \begin{subfigure}{0.24\textwidth}
    	\centering
        \includegraphics[width=\linewidth]{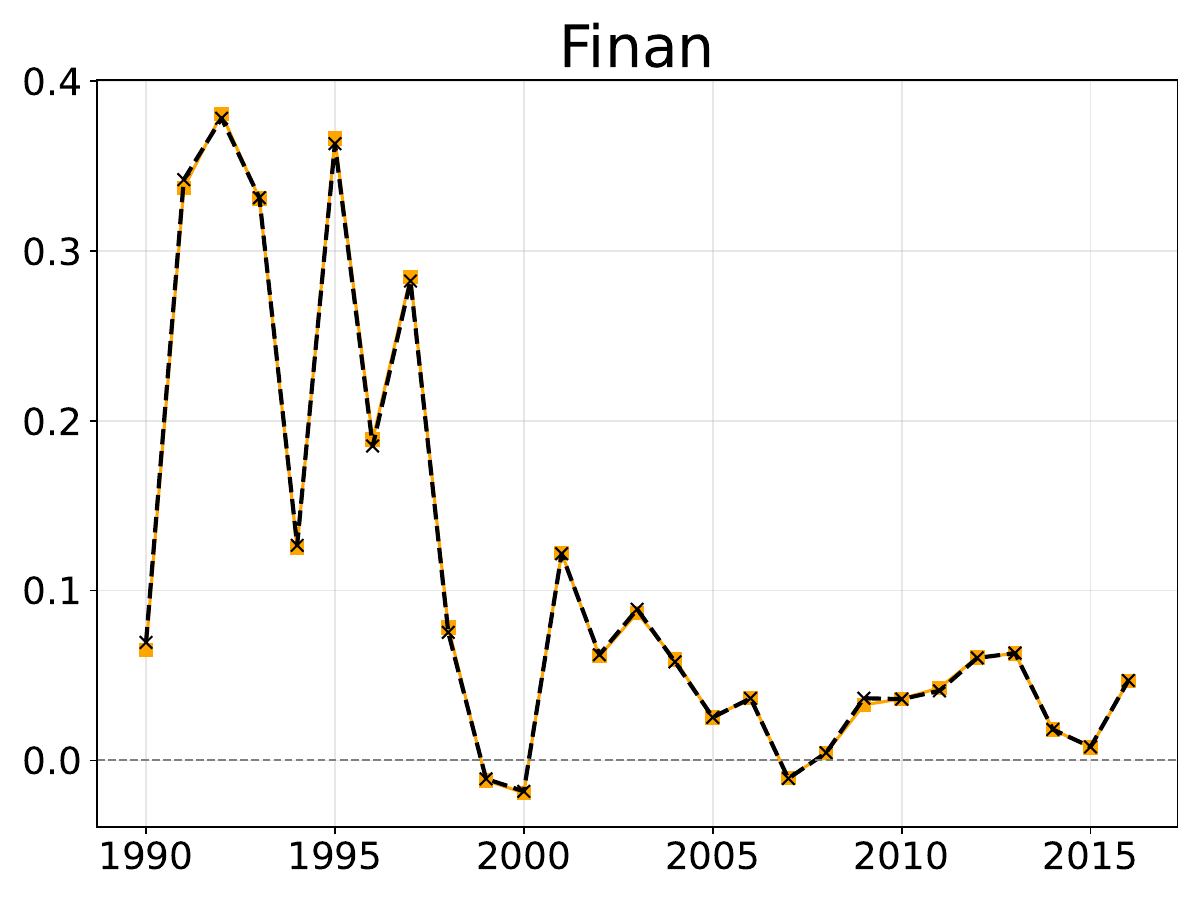}
	\end{subfigure}

    \begin{subfigure}{0.24\textwidth}
        \centering
        \includegraphics[width=\linewidth]{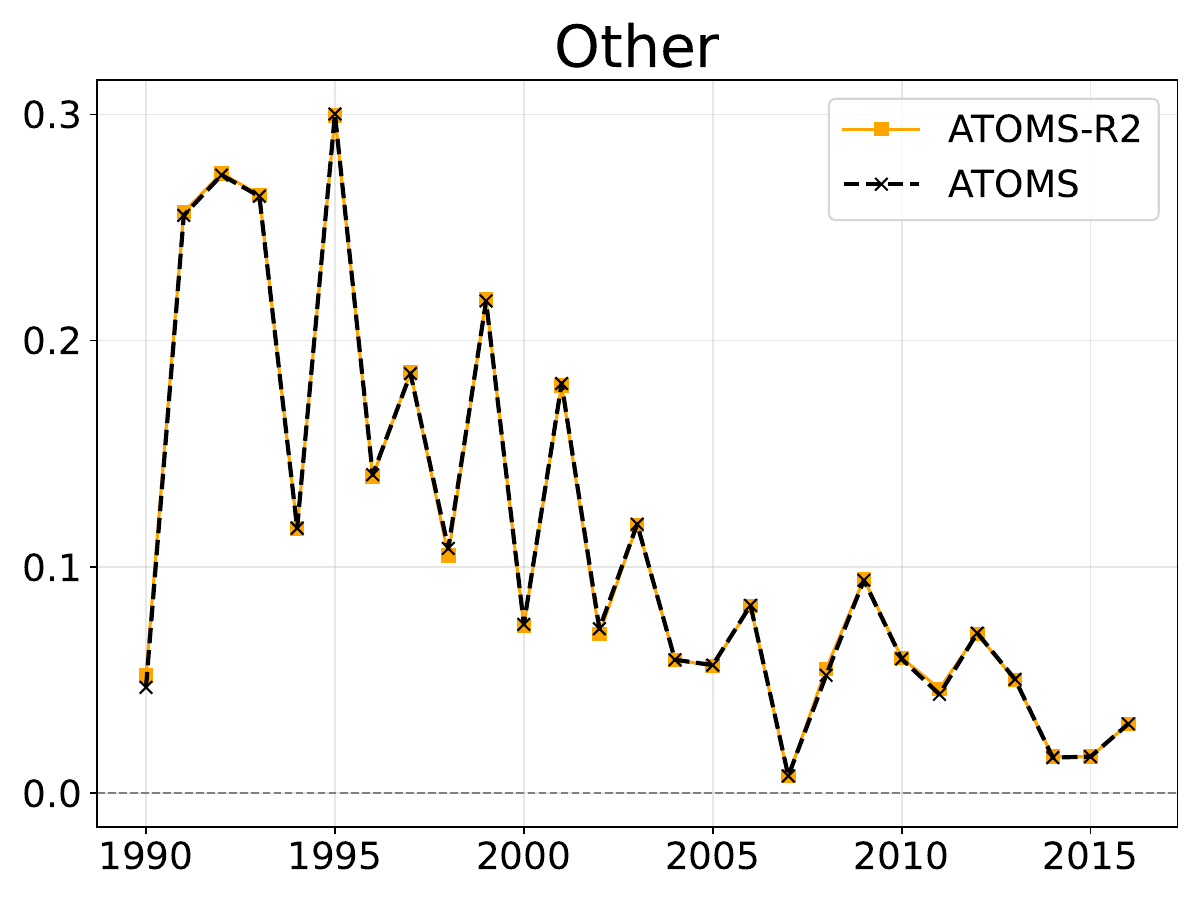}
	\end{subfigure}
      \bnotefig{This figure reports the annual OOS $R^2$ of our adaptive model selection algorithms $\adaptiveR$ (orange line with $\blacksquare$'s) and $\adaptive$ (black dashed line with $\times$'s). The title in each subfigure is Kenneth French's acronym for each industry. For the full names of these industries, please refer to Table \myref{tab-industry-name-mapping}.}
\end{figure}

In \myCref{fig-boxplot-supp}, we give a box plot of the overall out-of-sample $R^2$ of $\adaptiveR$ along with $\adaptive$ and the fixed-window benchmarks across the $17$ industries. In \myCref{fig-industry-yearly-supp}, we compare the annual out-of-sample $R^2$ of $\adaptiveR$ and $\adaptive$ for the $17$ industries. Again, $\adaptiveR$ and $\adaptive$ have similar performance.

\section{Additional Experiment Details}

\subsection{Experiment Details for \myCref{sec-tradeoff-empirics}}\label{sec-tradeoff-empirics-details}

In this section, we given more details on our empirical investigations of the nonstationarity-complexity tradeoff in \myCref{sec-tradeoff-empirics}. 

The three prediction models, along with their hyperparameters, are: (1) a linear model trained by Ridge regression using the most recent $64$ months of data, with $\alpha = 1$, (2) a random forest trained on the most recent $64$ months of data, with $n_{\texttt{tree}}=200$ and $d_{\max}=5$, and (3) a random forest trained on all historical data, with $n_{\texttt{tree}}=200$ and $d_{\max}=5$.

In each month $t$, we construct training data by randomly subsampling $4/5$ of the observations $\dataset_j$ in each previous month $j\in[t-1]$. The process is repeated $20$ times with independent random seeds. We then average the out-of-sample $R^2$ over the $20$ random seeds, which are then used to produce the figures in \myCref{sec-tradeoff-empirics}.

\subsection{Training Window Configurations in \Cref{sec-experiments}}\label{sec-window-details}

To assess the performance of our adaptive model selection algorithm across different data regimes, we train models on windows of different lengths. For each month $t$, we consider training windows of $4^k \wedge (t-1)$ months with $0\le k \le 5$. In particular, since $t\in\{1,...,350\}$, then $k=5$ corresponds to a full training window of $(t-1)$ months. This yields the following window lengths:

\begin{table}[h]
\centering
\caption{Training Window Lengths by Value of $k$}
\begin{tabular}{ccc}
\hline
$k$ & Window Length (months) & Approximate Years \\
\hline
0 & 1 & 0.08 \\
1 & 4 & 0.33 \\
2 & 16 & 1.33 \\
3 & 64 & 5.33 \\
4 & 256 & 21.33 \\
5 & $t-1$ & $(t-1)/12$ \\
\hline
\end{tabular}
\end{table}

This exponential scaling allows us to examine how model performance varies with the amount of historical data available for training. Shorter windows capture recent market dynamics but may be susceptible to noise, while longer windows provide more stable parameter estimates but may miss structural changes in the data-generating process.

\subsection{Hyperparameters for $\adaptive$}\label{sec-hyperparam-ATOMS-ablation}

In our implementation of $\adaptive$, we set $\delta'=0.1$ and $M'=10^{-3}$. The theoretical prescription is $M'=8M^2$, where $M$ is an upper bound on the magnitude of returns $y$. In finite samples, however, using the literal maximum return can make this bound overly conservative because a small number of extreme observations dominate the scale. We therefore calibrate $M'$ to cover typical values of $8y^2$ in the pre-OOS period. From September 1987 to December 1989, before the OOS period begins in January 1990, the $90$th percentile of $8y^2$ is $9\times 10^{-4}$, while the maximum is $0.1$. We therefore set $M'=10^{-3}$.

To assess the sensitivity of $\adaptive$ to the choice of $M'$, \Cref{tab:ablation_oos_r2_industry_time} reports its OOS performance after doubling and halving the baseline choice $M'=10^{-3}$, corresponding to $M'=2\times 10^{-3}$ and $M'=5\times 10^{-4}$,  respectively. The results are stable across these perturbations: the full-period OOS $R^2$ remains between $0.049$ and $0.050$, and the results during the recession periods change only modestly. For all these three values of $M'$, $\adaptive$ continues to outperform the benchmark methods reported in \Cref{tab:oos_r2_industry_time}.

\begin{table}[H]
    \centering
    \caption{Ablation Study of Hyperparameter $M'$ in $\adaptive$: Out-of-Sample $R^2$ Averages Across Industries by Time Period}
    \label{tab:ablation_oos_r2_industry_time}
    \begin{tabular}{@{}lcccccc@{}}  
        \toprule
        \multirow{2}{*}{Method} & \multirow{2}{*}{Full OOS Period} & \multicolumn{3}{c}{Recessions}  \\
        \cmidrule{3-5}
        & & Gulf War & 2001 Recession & Financial Crisis  \\
        \midrule
        $\adaptive$ ($M'=2\times 10^{-3}$)       & $0.050$  & $0.024$  & $0.124$  & $0.041$    \\
        $\adaptive$ ($M'=1\times 10^{-3}$)        & $0.049$  & $0.027$  & $0.125$  & $0.041$    \\
        $\adaptive$ ($M'=5\times 10^{-4}$)         & $0.049$  & $0.028$  & $0.124$  & $0.040$    \\
        \bottomrule
    \end{tabular}\bnotetab{
    This table reports the sensitivity of $\adaptive$'s performance to the choice of $M'$, in terms of the out-of-sample (OOS) $R^2$ averages for return prediction models across all 17 industry portfolios. The baseline choice is $M'=10^{-3}$, which corresponds to the result reported in \Cref{tab:oos_r2_industry_time}. The other two rows double and halve this value, setting $M'=2\times 10^{-3}$ and $M'=5\times 10^{-4}$, respectively. Full OOS Period refers our largest available OOS period covering 01/1990$\sim$11/2016. Columns report OOS $R^2$ averages across all industries and highlight this metric during three recessions, as documented in \href{https://www.nber.org/research/business-cycle-dating}{NBER Business Cycle Dating}: 
    \begin{itemize}
        \item the 1990 Gulf War recession (06/1990$\sim$10/1990);
        \item the 2001 Recession of dot-com bubble burst and the 9/11 attack (05/2001$\sim$10/2001);
        \item the Financial Crisis led by defaults of subprime mortgages (11/2007$\sim$06/2009). 
    \end{itemize}
    That is, the OOS performance in Gulf War column focuses on model performance comparisons exclusively in the out-of-sample period of 06/1990$\sim$10/1990. All values are calculated using monthly return data.}
\end{table}

\subsection{Computational Implementation}

All models are implemented using Python 3.9 with the following libraries:

\begin{itemize}
    \item Linear models: \texttt{scikit-learn} version 1.0.2, specifically \texttt{Ridge}, \texttt{Lasso}, and \texttt{ElasticNet}.
    \item Random forest: \texttt{scikit-learn}'s \texttt{RandomForestRegressor} with \texttt{random\_state=0}.
    \item Regime-switching model: \texttt{statsmodels.tsa.regime\_switching.markov\_regression}.
    \item Data manipulation: \texttt{pandas} version 1.4.2 and \texttt{numpy} version 1.21.5.
\end{itemize}

\subsection{Summary Statistics of the Dataset}

We provide an overview of the long--short firm characteristic covariates used in the analysis, their time-series behavior, and cross-sectional dependence as well as a brief summary of the stochastic discount factor (SDF) and decile portfolios from \citet{CPZ24}. Recall that all of the long-short characteristic portfolios are computed at the daily frequency, for the subsequent summary plots, we have aggregated them into the monthly frequency using within-month averages, in line with the standard practice of aligning signals with monthly returns. The monthly aggregation smooths out day-to-day noise and highlights the economically relevant medium-horizon variations.

\begin{figure}[H]
    \centering
    \caption{Monthly Means of the 12 Most Volatile Covariates.\label{fig:top_most_volatile_monthly}}
    \includegraphics[width=.78\textwidth, trim=0 20 0 20, clip]{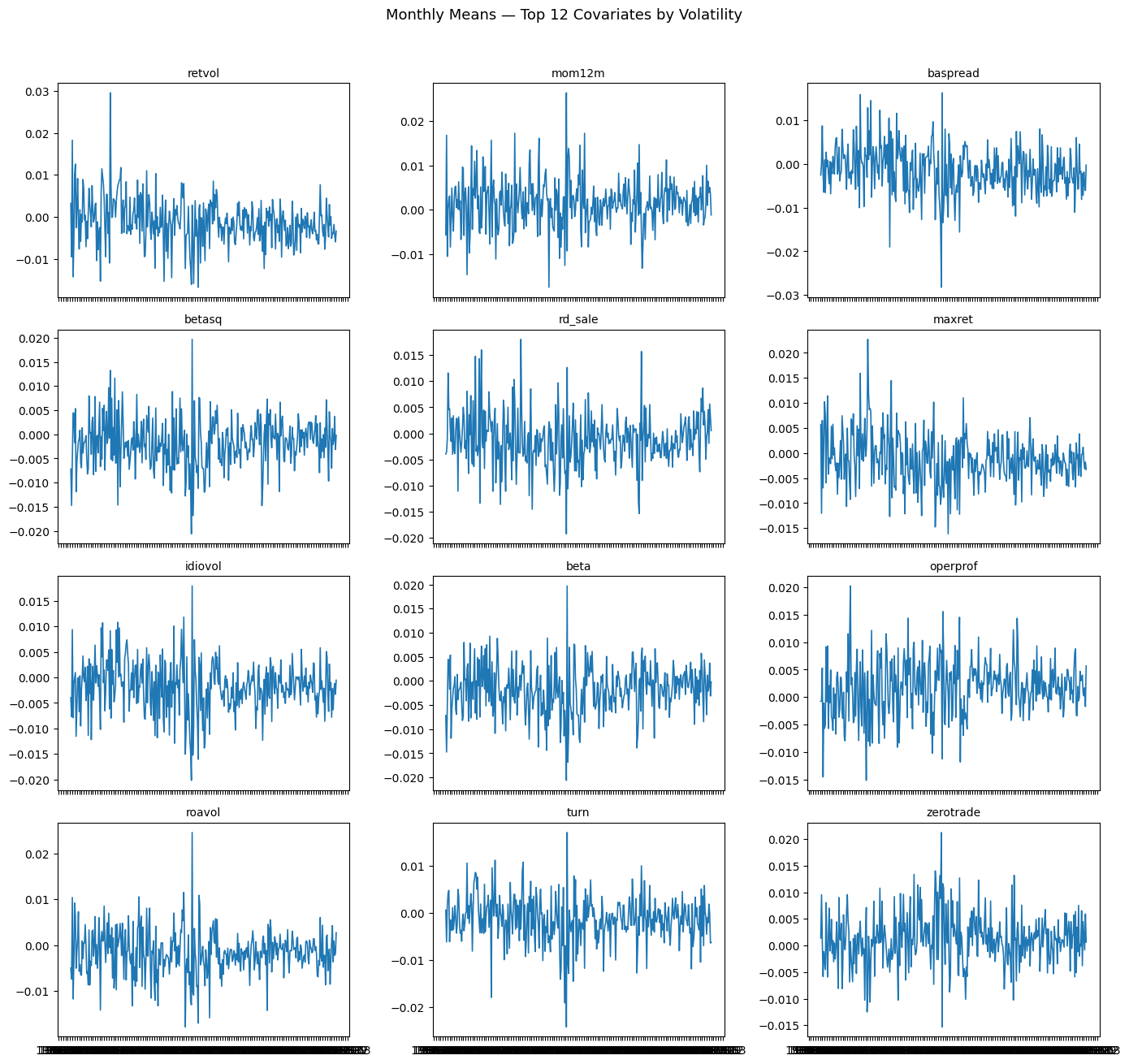}
    \bnotefig{Each panel shows the monthly mean of a long--short firm characteristic. 
    Covariates are ranked by total-sample volatility. 
    The time series reveal which characteristics exhibit the greatest month-to-month variation and long-run persistence.}
\end{figure}

\paragraph{Monthly Evolution of Covariates.}Figure~\ref{fig:top_most_volatile_monthly} displays the time series of monthly mean values for the twelve most volatile covariates, ranked by their total-sample standard deviation. The figure highlights that variables such as \texttt{retvol}, \texttt{mom12m}, and \texttt{baspread} exhibit pronounced month-to-month fluctuations, while others such as \texttt{turn} and \texttt{operprof} remain relatively stable. These series reveal persistent heteroskedasticity and regime shifts over time, particularly during market dislocations such as the early 2000s and 2008 crises.

\begin{figure}[H]
    \centering
    \caption{Distributions (Boxplots) of Standardized Monthly Covariates.\label{fig:boxplots_top_most_volatile_z_scored}}
    \includegraphics[width=0.7\textwidth, height=0.4\textheight, trim=0 20 0 20, clip]{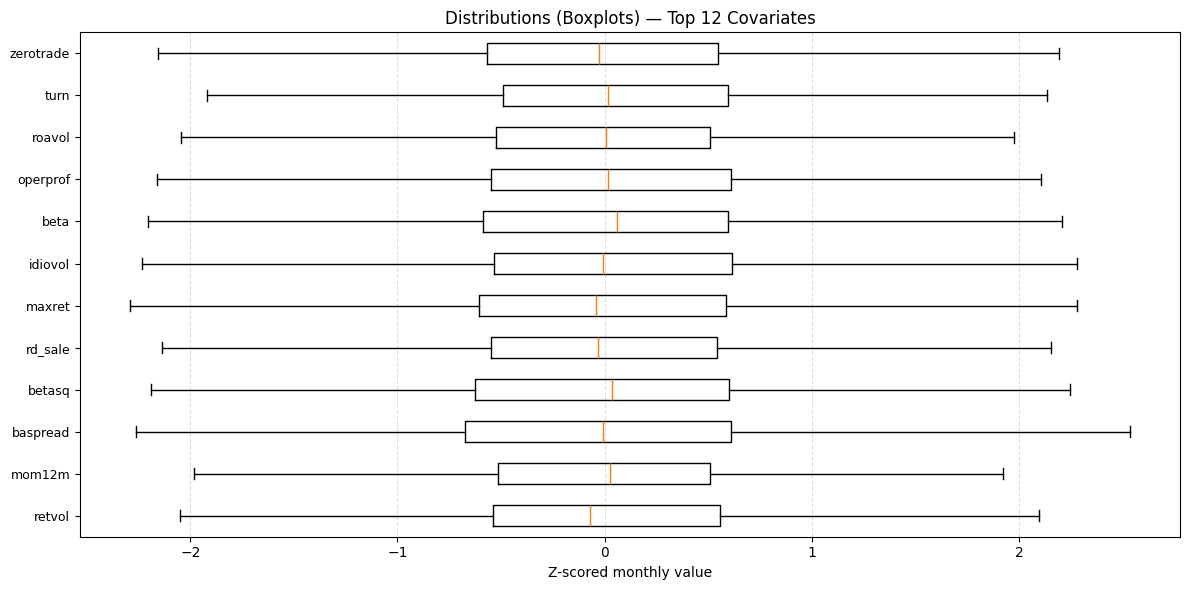}
    \bnotefig{Z-scored monthly series for the twelve most volatile covariates. The figure compares dispersion and tail behavior across characteristics, highlighting differences in amplitude and symmetry.}   
\end{figure}

\paragraph{Distributional and Correlation Structure.} Figure~\ref{fig:boxplots_top_most_volatile_z_scored} summarizes the time-series distributions of the same twelve covariates using standardized (z-scored) monthly values. The median, interquartile range, and whiskers capture the magnitude and symmetry of fluctuations across time. Most variables display near-zero median values but differ in dispersion and tail behavior, consistent with heterogeneous economic mechanisms underlying each characteristic.

For example, variables such as \texttt{retvol}, \texttt{baspread}, and \texttt{mom12m} exhibit wide interquartile ranges and thick tails, suggesting that these signals experience substantial time variation and occasional extreme realizations. In contrast, variables such as \texttt{turn} and \texttt{operprof} have narrower boxes, implying greater stability through time.

 Each distribution is constructed by pooling monthly observations over the entire sample period for that specific covariate. This provides a concise view of the temporal heterogeneity and persistence of each characteristic after accounting for scale differences. The figure thus complements the time-series plots in Figure~\ref{fig:top_most_volatile_monthly} by providing a scale-free summary of long-run variability and skewness in the underlying long--short characteristics.
 
\begin{figure}[H]
    \centering
    \caption{Correlation Heatmap of the 30 Most Volatile Covariates.\label{fig:correlation_heatmap_top_most_volatile}}
    \includegraphics[width=1\textwidth, trim=0 0 0 20, clip]{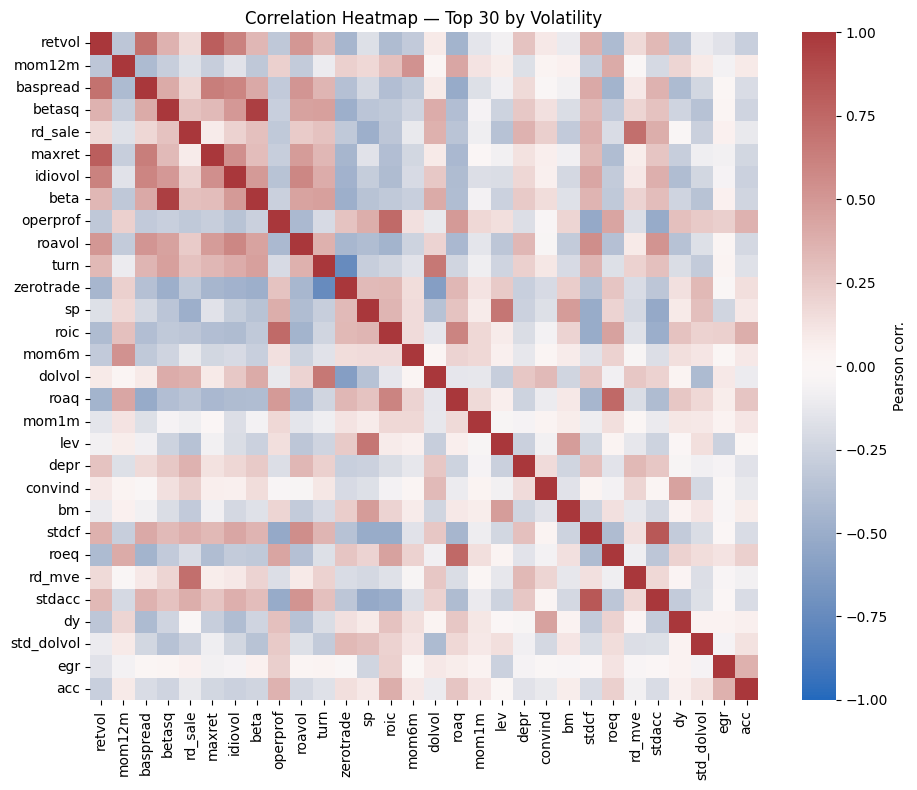}
    \bnotefig{The matrix shows pairwise Pearson correlations between the thirty most volatile monthly covariates. 
    Red indicates positive correlation and blue indicates negative correlation. 
    Distinct blocks suggest clusters of related characteristics.}
\end{figure}

The pairwise dependence structure among the top thirty covariates (in terms of volatility) is visualized in Figure~\ref{fig:correlation_heatmap_top_most_volatile}. The heatmap reveals clusters of strongly correlated signals, such as volatility-related measures (\texttt{retvol}, \texttt{idiovol}, \texttt{roavol}) and liquidity-related variables (\texttt{baspread}, \texttt{zerotrade}, \texttt{turn}). The presence of such correlation blocks indicates there could be shared economic channels.

\begin{figure}[H]
    \centering
    \caption{Twelve-Month Rolling Volatility of the Most Volatile Covariates.\label{fig:rolling_volatility_top_most_volatile}}
    \includegraphics[width=.8\textwidth, trim=0 20 0 20, clip]{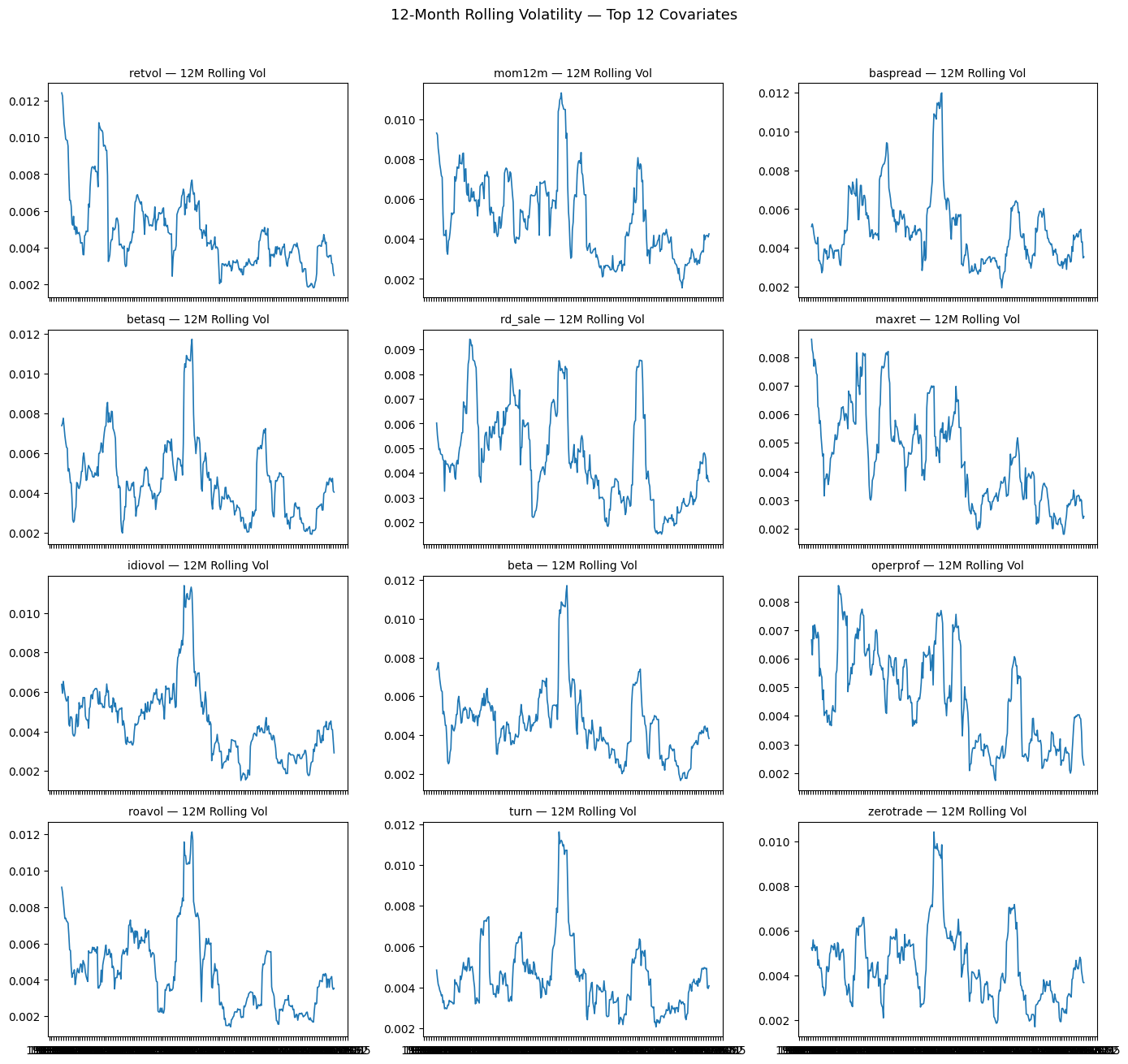}
    \bnotefig{The panels show rolling standard deviations computed using a 12-month moving window for each of the twelve most volatile covariates. 
    This highlights temporal variation in the stability and amplitude of the long-short signals.}
\end{figure}

\paragraph{Time-Varying Volatility of Covariates.} Figure~\ref{fig:rolling_volatility_top_most_volatile} plots the 12-month rolling standard deviation of the twelve most volatile covariates. Unlike Figure~\ref{fig:top_most_volatile_monthly}, which ranks variables by overall volatility, the rolling volatility tracks how the variability of each covariate evolves through time. Periods such as the dot-com bubble and the global financial crisis correspond to distinct spikes in volatility across multiple signals, indicating that the informational strength and instability of certain factors are regime-dependent.

\begin{figure}[H]
    \centering
    \caption{Monthly SDF and Decile Portfolios.\label{fig:pelger_deciles_sdf}}
    \includegraphics[scale=0.45, trim=0 0 0 20, clip]{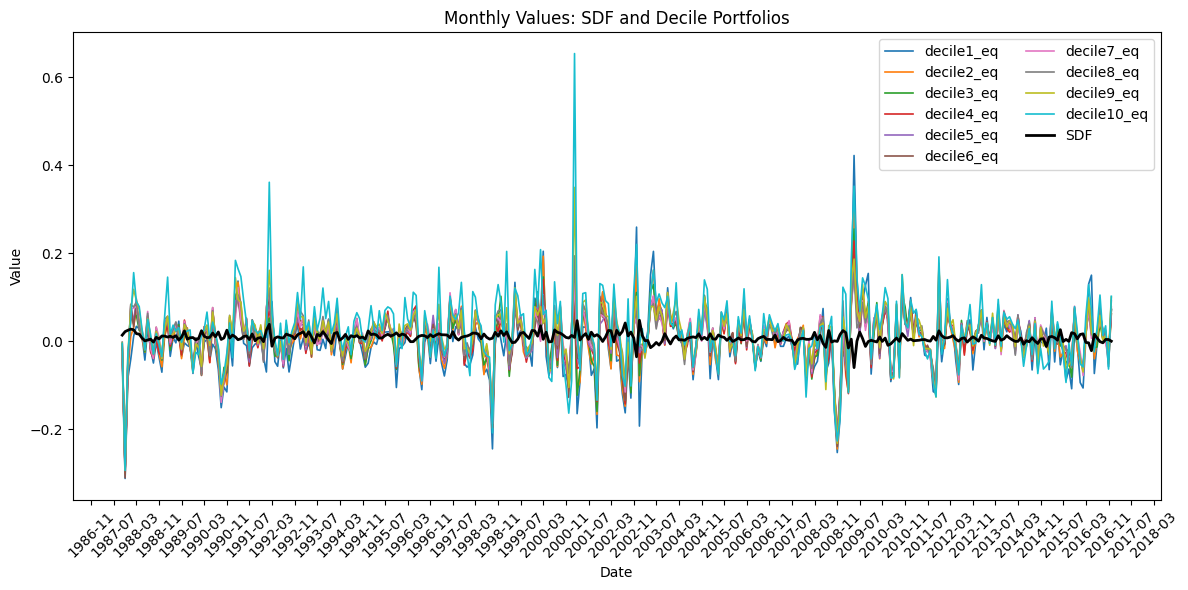}
    \bnotefig{The black line represents the stochastic discount factor (SDF), while the colored lines correspond to ten equal-weighted decile portfolios sorted on firm characteristics. 
    The comovement between the SDF and the characteristic-sorted portfolios provides an initial indication of factor relevance.}
\end{figure}

\paragraph{SDF and Decile Portfolios.} Finally, Figure~\ref{fig:pelger_deciles_sdf} plots the monthly time series of the stochastic discount factor (SDF) alongside the ten equal-weighted decile portfolios sorted by the underlying characteristic. The decile portfolios exhibit substantial comovement, with the SDF (shown in black) fluctuating more smoothly. This figure provides a benchmark for comparing the magnitude and temporal alignment of the SDF with characteristic-sorted portfolio returns, and serves as a diagnostic for whether the constructed SDF captures systematic components of asset pricing variation.

\section{Additional Experiment Results}

\subsection{Additional Results for Regime-Switching Model}\label{sec-RS-addition-results}

In this section, we show that the regimes learned by the regime-switching model capture meaningful market variations. To do this, we will examine the correlation between the learned regimes and CBOE Volatility Index (VIX)\footnote{\url{https://fred.stlouisfed.org/series/VIXCLS}}. 

Specifically, for each industry, for each month $t$, the regime-switching model estimates model parameters, including the regime transition matrix, using data available up to month $(t-1)$. Then, for each day $s$, we recursively compute the daily \emph{filtered regime probability}, which is the estimated regime distribution given the data up to day $s$:
\[
\widehat{p}_s(z) = \widehat{\PP}(z_s = z \mid \cG_s) , \quad \forall z\in\{1,2\},
\]
where $\cG_s$ denote all data points up to and including day $s$. Then, we compute the monthly prevalence of state $2$ in month $t$:
\[
S_t = \frac{1}{\batch_t} \sum_{s\in\batch_t} \ind \left\{ \widehat{p}_s(2) > \widehat{p}_s(1) \right\}.
\]
The variation of $S_t$ over time reflects changes in the learned regimes.

\begin{figure}[H]
\centering
\caption{Box Plot of Pearson Correlation between Learned Industry Regimes and VIX.\label{fig-regime-corr-boxplot}}
\includegraphics[scale=0.6]{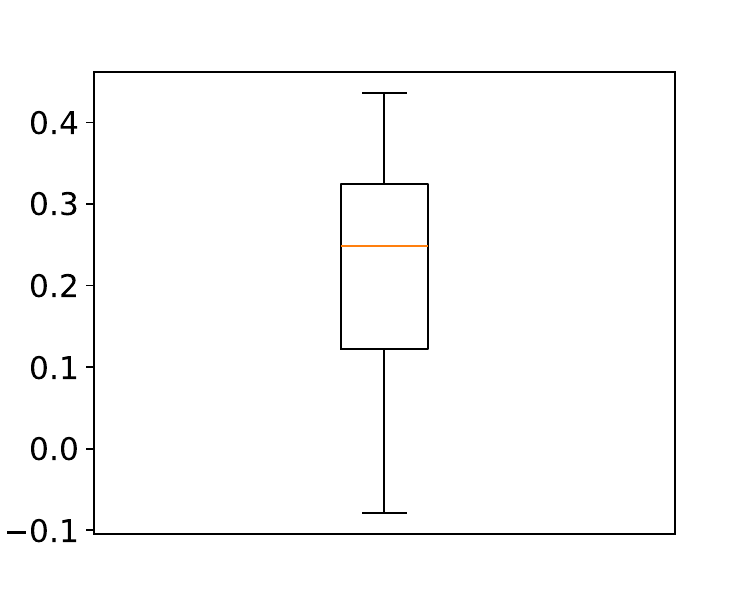}
\bnotefig{This figure describes the distribution of the Pearson correlation between each industry's learned regimes and the VIX. For each of the $17$ industries, we compute the prevalence of state $2$ in each month, and calculate its correlation with the VIX. The box plot is over the $17$ correlations, each from one industry.}
\end{figure}

We compute the Pearson correlation between $\{S_t\}$ with the monthly averages of CBOE Volatility Index (VIX), from January 1990 to November 2016. \Cref{fig-regime-corr-boxplot} shows a box plot across the $17$ industries. In particular, the $25$th and $75$th percentiles of the $17$ correlations are $0.12$ and $0.32$, respectively. Moreover, across the $17$ industries, the raw $p$-values for the correlation between the estimated regimes and the VIX have $70$th percentile equal to $0.0015$. Since $17$ correlations are tested, Bonferroni correction at the $0.05$ level requires $p < 0.05 / 17 \approx 0.0029$. Thus, $70\%$ of the correlations remain statistically significant. This shows that across most industries, there are positive correlations between the learned regimes and the VIX, and thus the regime-switching model captures meaningful market variations.

\subsection{Experiment Results with the Standard $R^2$ Metric}\label{sec-experiments-standard-R2}

In \myCref{sec-experiments}, we evaluated predictive performance using the zero-benchmark $R^2$ to avoid the noise inherent in historical mean estimation. For completeness, this section reports the corresponding results using the standard out-of-sample $R^2$ metric, which benchmarks model performance against the historical sample mean. Qualitatively, the relative performance among models remains consistent with the the observations in \myCref{sec-experiments}: our adaptive algorithm $\adaptive$ continues to outperform fixed-window benchmarks. Quantitatively, we observe that the standard $R^2$ values are generally lower than their zero-benchmark counterparts.

\myCref{tab:oos_r2_industry_time-standardR2} presents out-of-sample standard $R^2$ values of $\adaptive$ and baselines across distinct economic regimes, serving as the counterpart to \myCref{tab:oos_r2_industry_time} in the main text.

\begin{table}[htbp]
    \centering
    \caption{OOS Standard $R^2$ Averages Across Industries by Time Period.}
    \label{tab:oos_r2_industry_time-standardR2}
    \begin{tabular}{@{}lcccccc@{}}  
        \toprule
        \multirow{2}{*}{Method} & \multirow{2}{*}{Full OOS Period} & \multicolumn{3}{c}{Recessions}  \\
        \cmidrule{3-5}
        & & Gulf War & 2001 Recession & Financial Crisis  \\
        \midrule
        $\adaptive$               & $0.041$  & $-0.019$  & $0.115$  & $0.039$    \\
        $\fixedwindow(32)$ & $0.013$  & $-0.038$ & $0.085$  & $-0.003$  \\
        $\fixedwindow(128)$ & $0.032$  & $-0.080$ & $0.113$  & $0.034$  \\
        $\fixedwindow(512)$ & $0.034$  & $-0.080$ & $0.107$  & $0.037$  \\
        $\fixedwindowCV$       &  $0.026$ &  $-0.056$ & $0.060$ & $0.012$   \\
        $\RS$       &  $-0.035$ &  $-0.508$ & $0.070$ & $-0.010$   \\
        \bottomrule
    \end{tabular}\bnotetab{
    This table reports OOS standard $R^2$ averages for return prediction models across all 17 industry portfolios. Full OOS Period refers to OOS period covering 01/1990$\sim$11/2016. Columns report OOS $R^2$ averages across all industries and highlight this metric during three recessions, as documented in \href{https://www.nber.org/research/business-cycle-dating}{NBER Business Cycle Dating}: 
    \begin{itemize}
        \item the 1990 Gulf War recession (06/1990$\sim$10/1990);
        \item the 2001 Recession of dot-com bubble burst and the 9/11 attack (05/2001$\sim$10/2001);
        \item the Financial Crisis led by defaults of subprime mortgages (11/2007$\sim$06/2009). 
    \end{itemize}
    That is, the OOS performance in Gulf War column focuses on model performance comparisons exclusively in the out-of-sample period of 06/1990$\sim$10/1990. All values are calculated using monthly return data.}
\end{table}

\begin{figure}[H]
\centering
\caption{Box Plot of OOS Standard $R^2$ of $\adaptive$ and Benchmarks for $17$ Industry Portfolios.\label{fig-boxplot-standardR2}}
\includegraphics[scale=0.5]{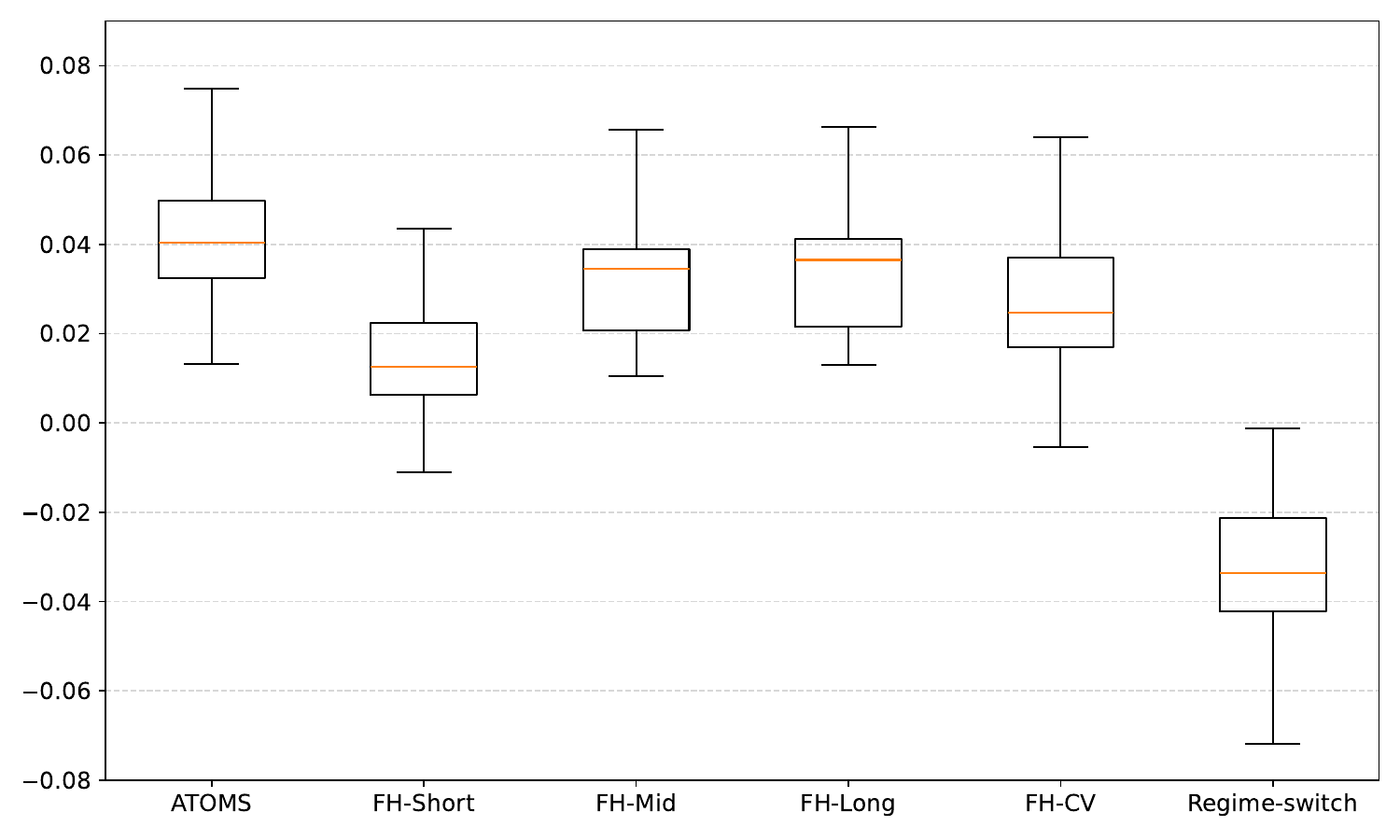}
\bnotefig{This figure describes the distribution of each method's OOS $R^2$. Each box corresponds to all industries and all years in our OOS horizon.}
\end{figure}

\myCref{fig-boxplot-standardR2} gives a box plot of the OOS standard $R^2$ of $\adaptive$ and the fixed-window baselines over the 17 industry portfolios, mirroring \myCref{fig-boxplot}.

\begin{figure}[H]
	\centering
	\caption{Annual OOS Standard $R^2$ of Different Approaches for $17$ Industry Portfolios. \label{fig-industry-yearly-standardR2}}

    \begin{subfigure}{0.24\textwidth}
    	\centering
        \includegraphics[width=\linewidth]{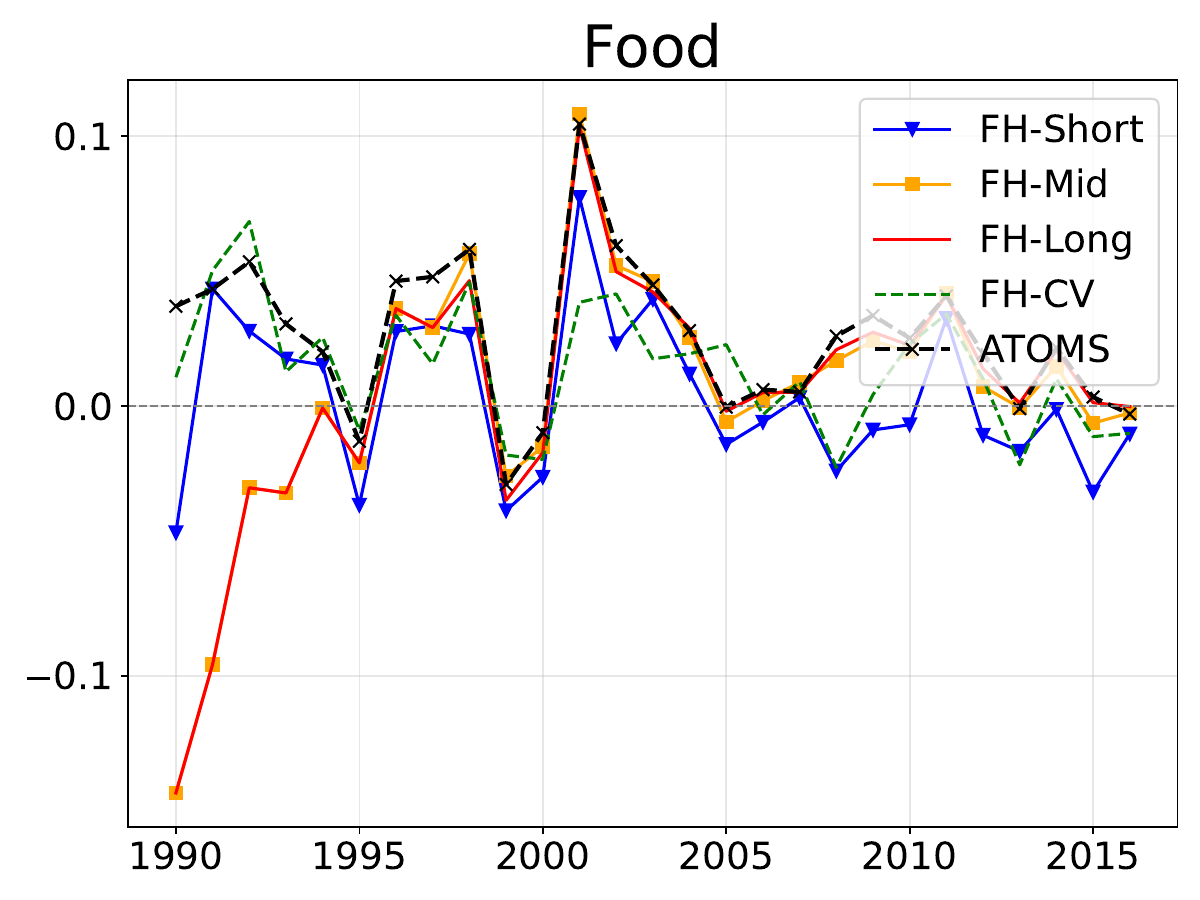}
	\end{subfigure}
    \begin{subfigure}{0.24\textwidth}
        \centering
        \includegraphics[width=\linewidth]{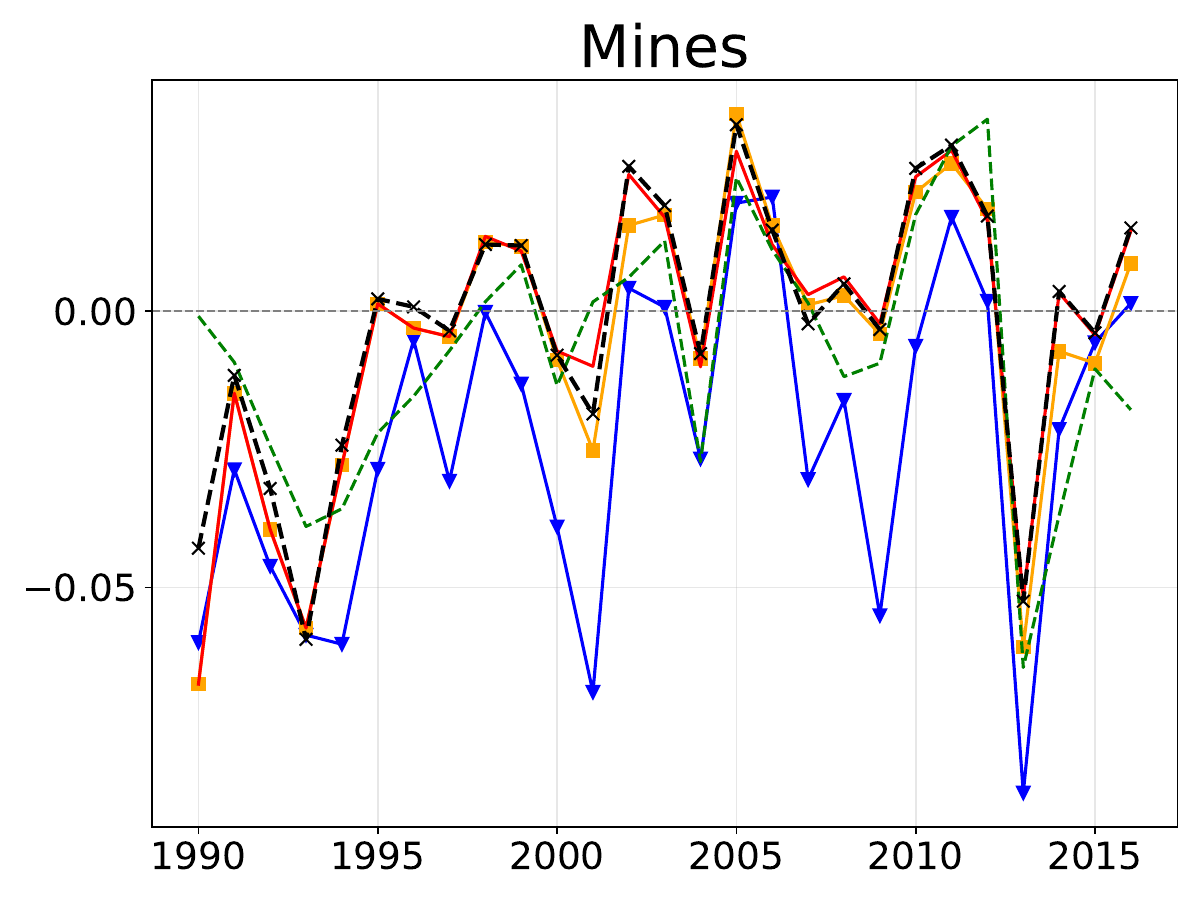}
	\end{subfigure}
    \begin{subfigure}{0.24\textwidth}
        \centering
        \includegraphics[width=\linewidth]{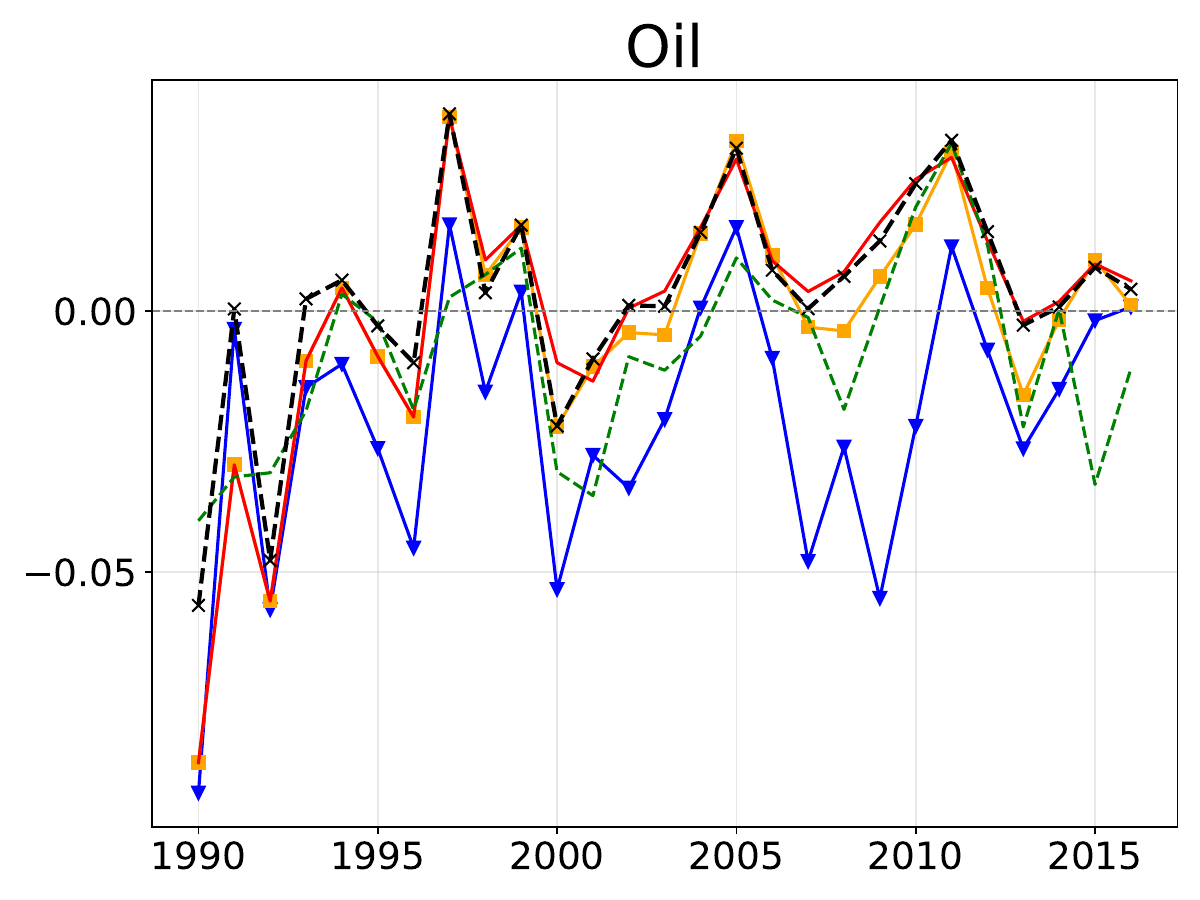}
	\end{subfigure}
    \begin{subfigure}{0.24\textwidth}
    	\centering
        \includegraphics[width=\linewidth]{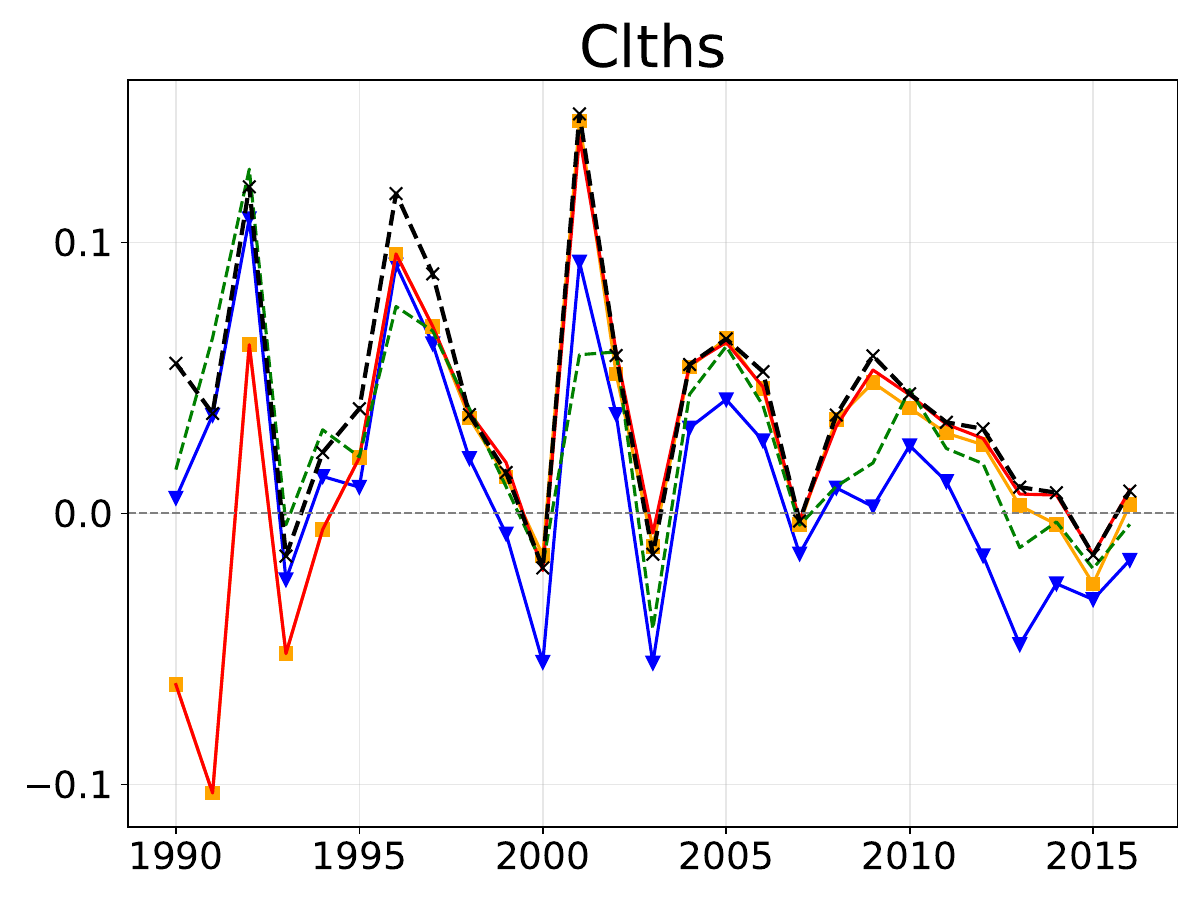}
	\end{subfigure}

    \begin{subfigure}{0.24\textwidth}
        \centering
        \includegraphics[width=\linewidth]{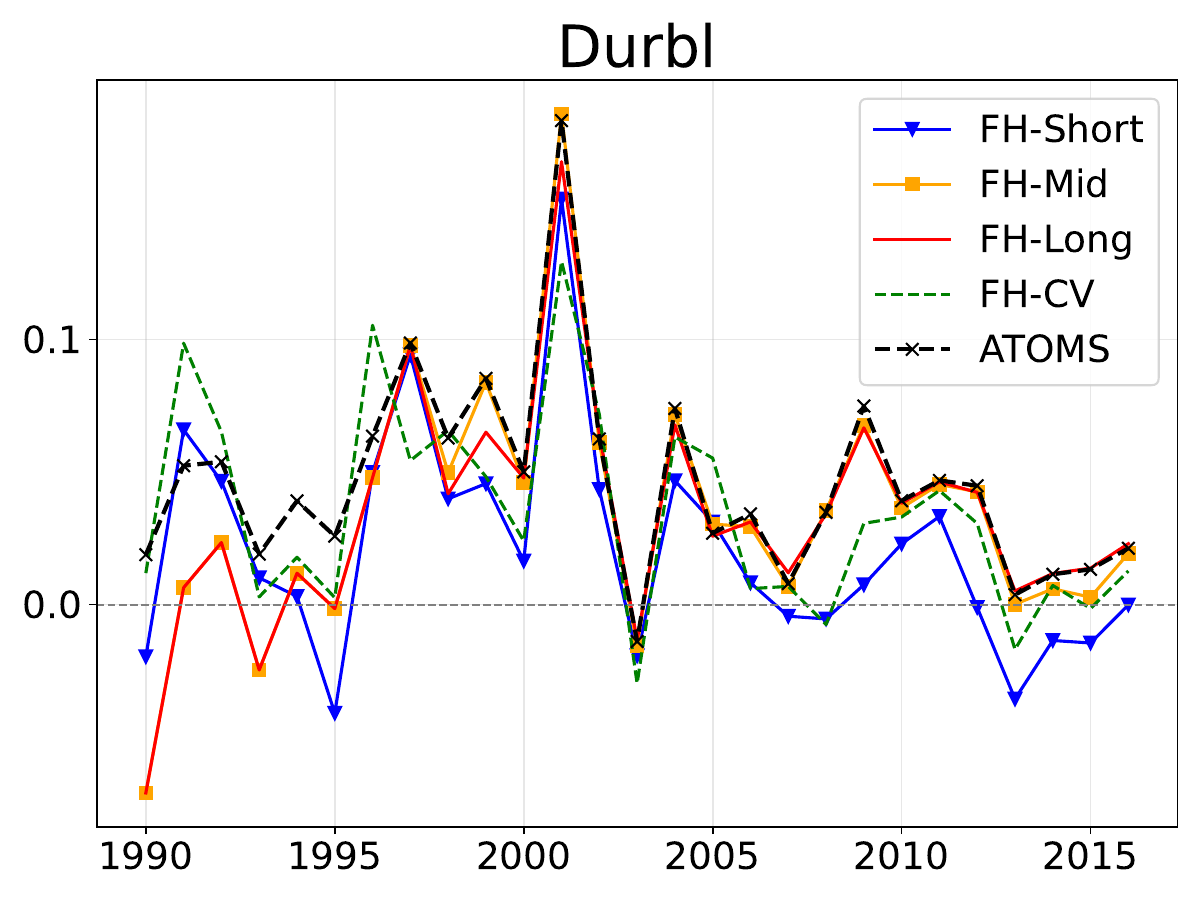}
	\end{subfigure}
    \begin{subfigure}{0.24\textwidth}
        \centering
        \includegraphics[width=\linewidth]{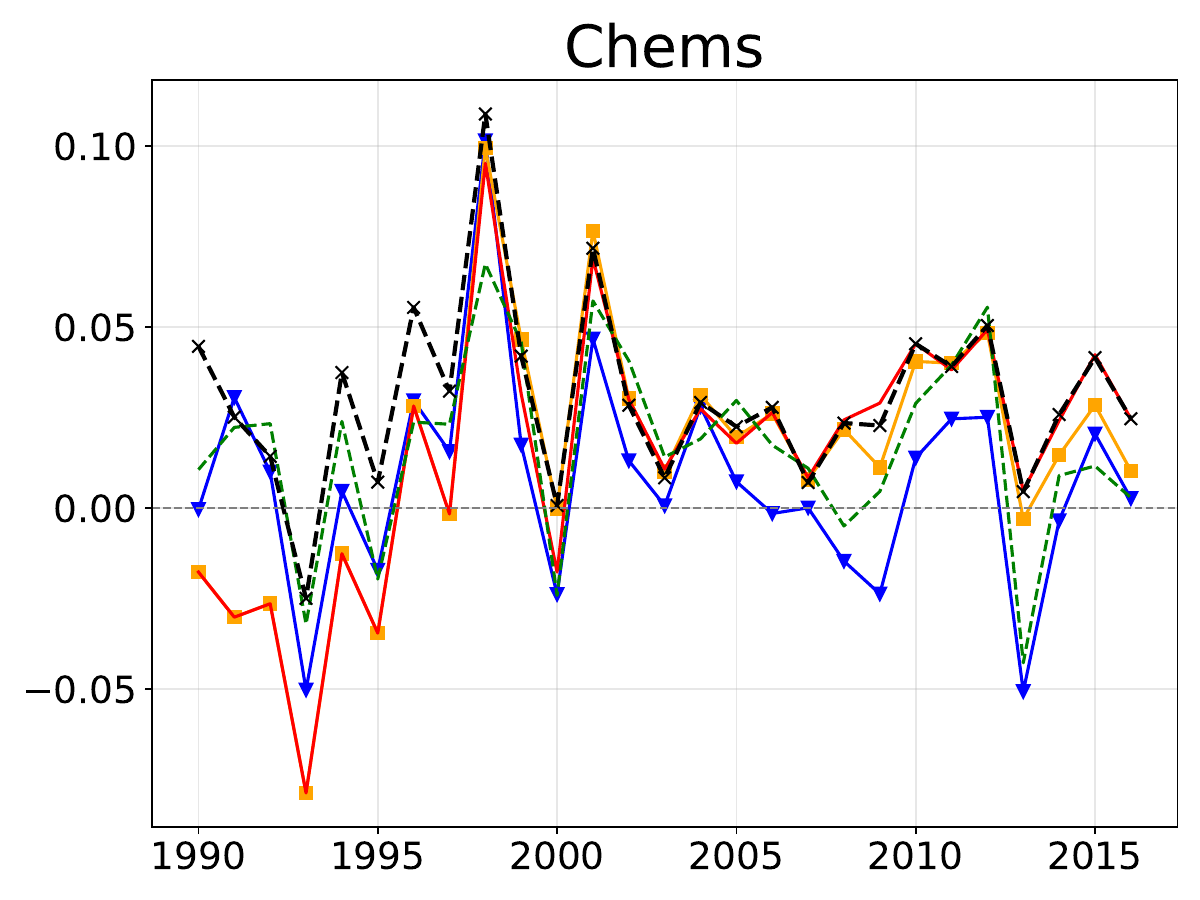}
	\end{subfigure}
    \begin{subfigure}{0.24\textwidth}
    	\centering
        \includegraphics[width=\linewidth]{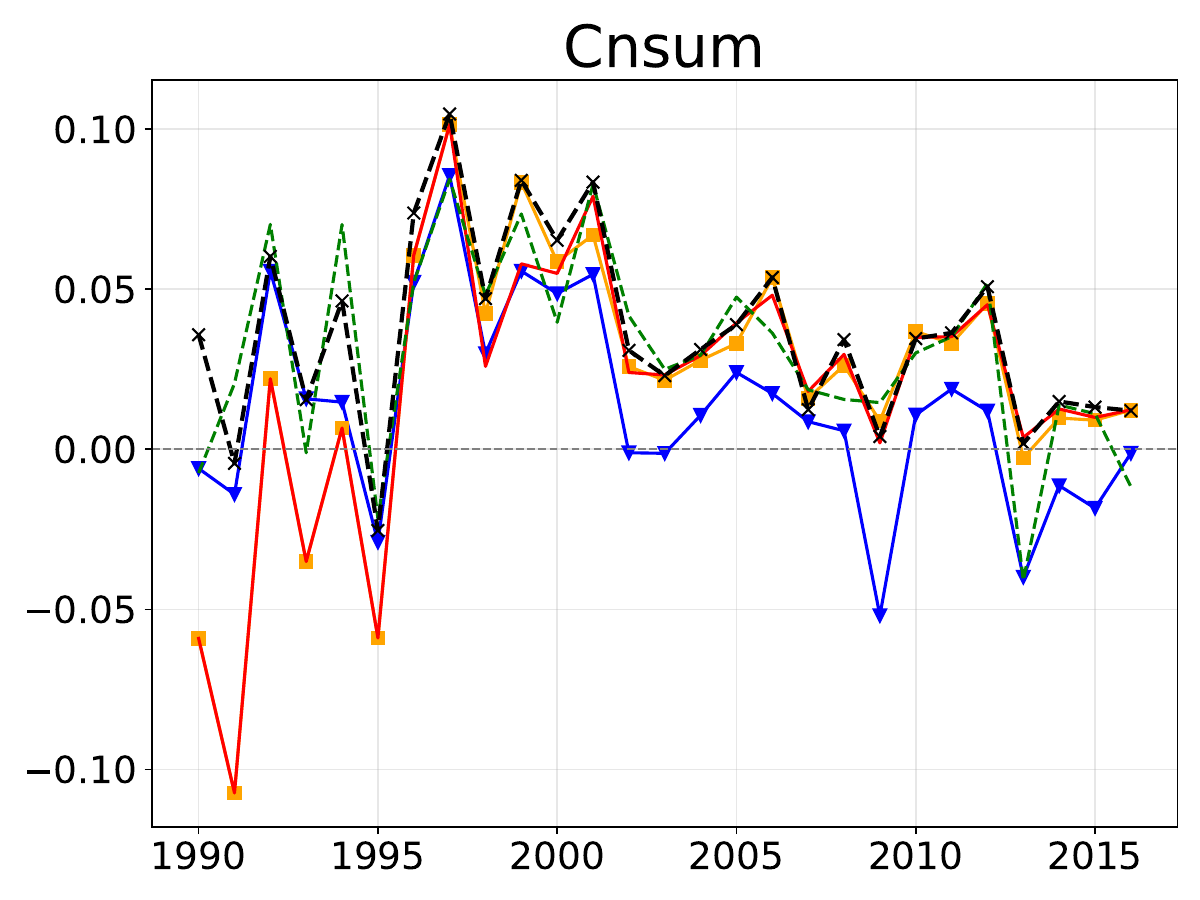}
	\end{subfigure}
    \begin{subfigure}{0.24\textwidth}
        \centering
        \includegraphics[width=\linewidth]{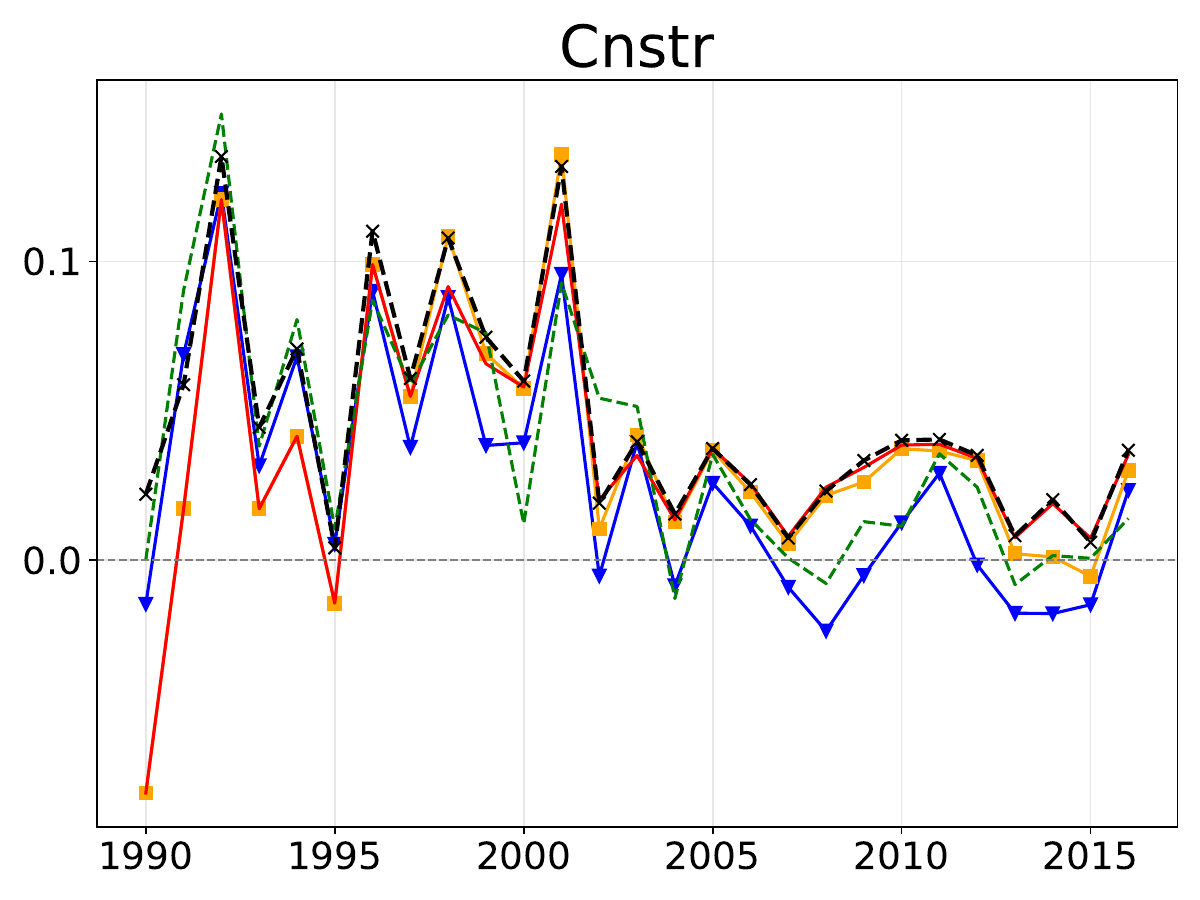}
	\end{subfigure}

    \begin{subfigure}{0.24\textwidth}
        \centering
        \includegraphics[width=\linewidth]{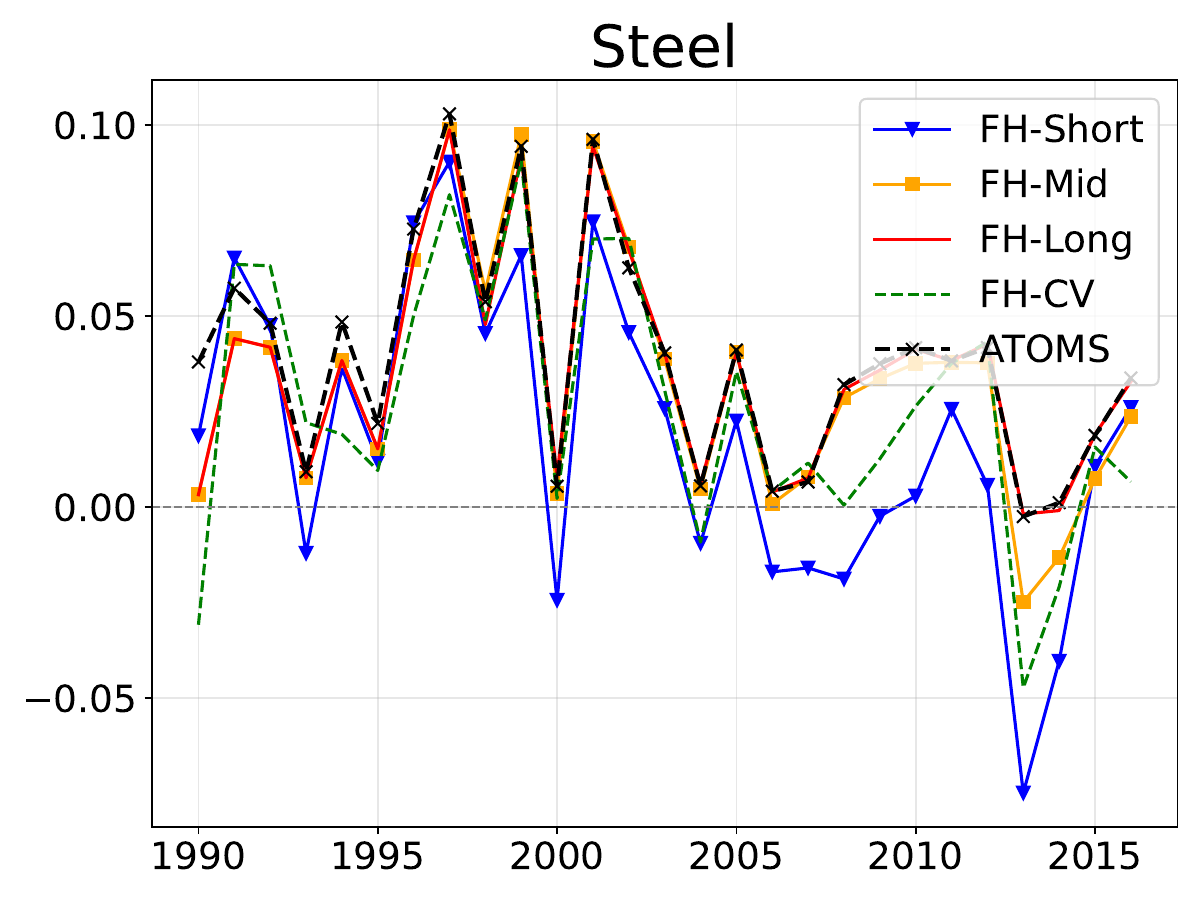}
	\end{subfigure}
    \begin{subfigure}{0.24\textwidth}
    	\centering
        \includegraphics[width=\linewidth]{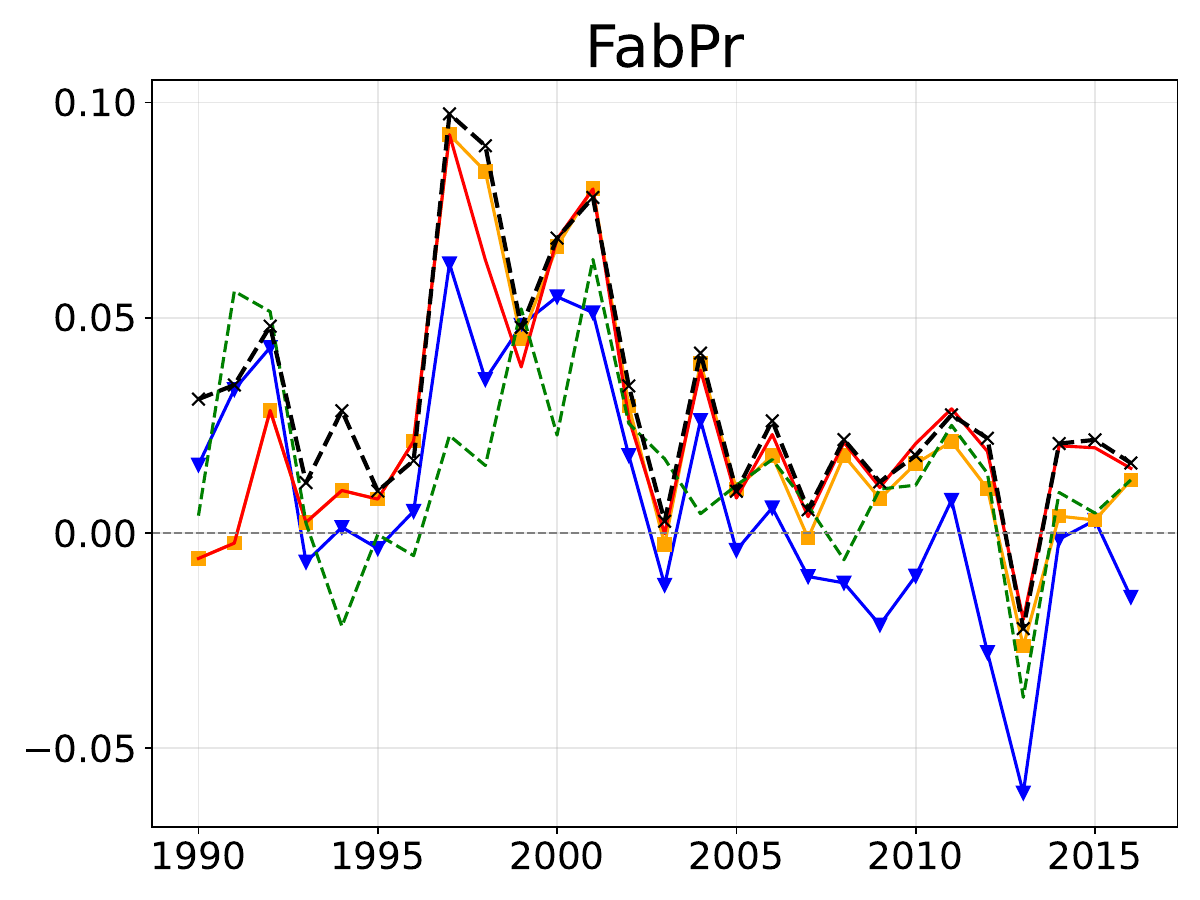}
	\end{subfigure}
    \begin{subfigure}{0.24\textwidth}
        \centering
        \includegraphics[width=\linewidth]{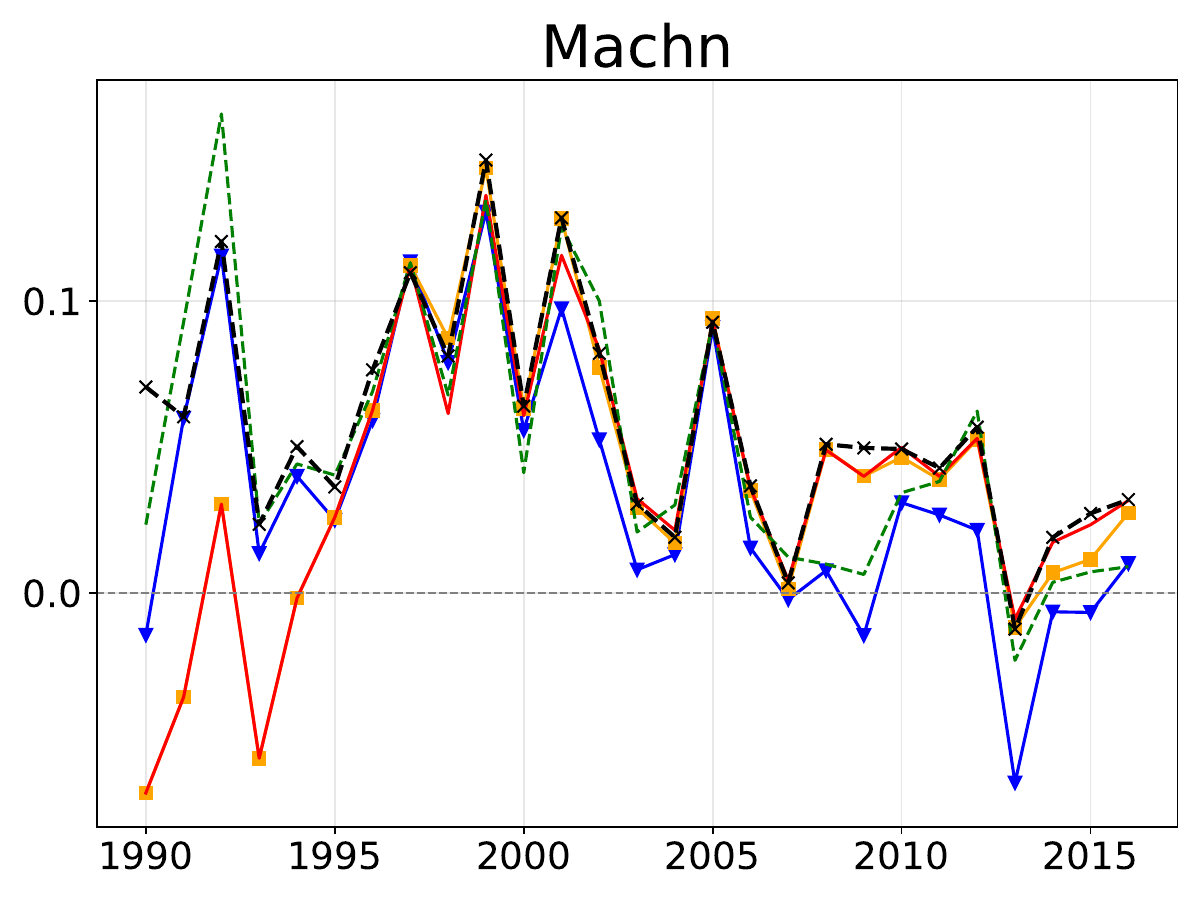}
	\end{subfigure}
    \hfill
    \begin{subfigure}{0.24\textwidth}
        \centering
        \includegraphics[width=\linewidth]{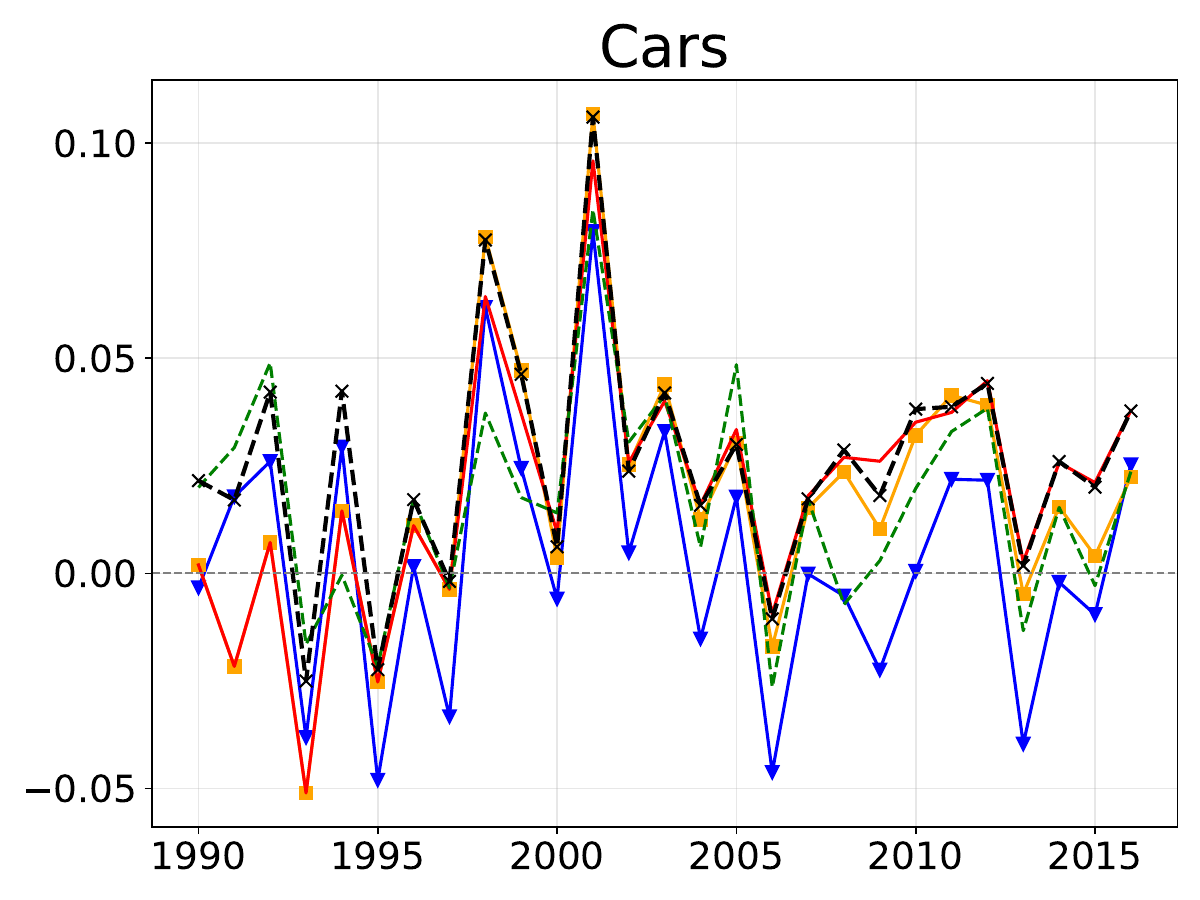}
	\end{subfigure}

    \begin{subfigure}{0.24\textwidth}
    	\centering
        \includegraphics[width=\linewidth]{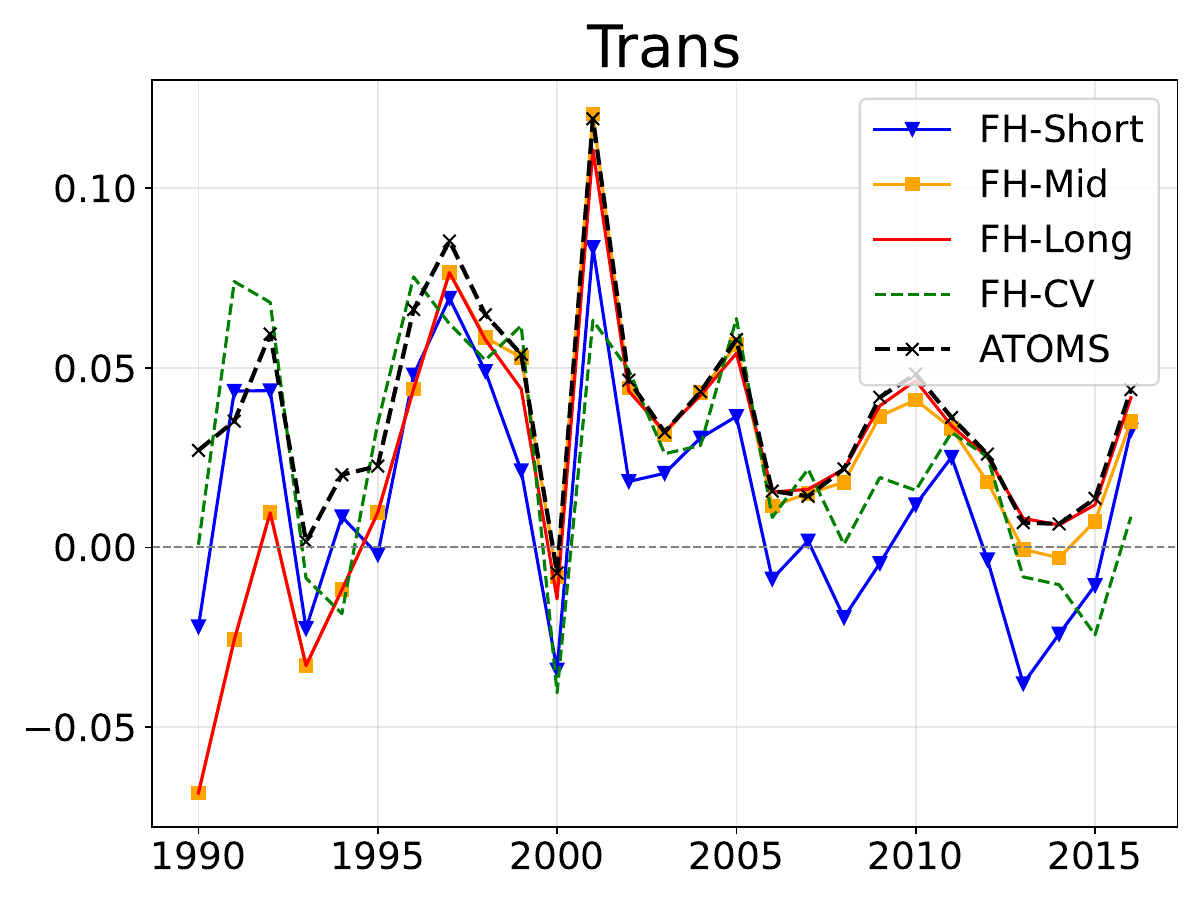}
	\end{subfigure}
    \begin{subfigure}{0.24\textwidth}
        \centering
        \includegraphics[width=\linewidth]{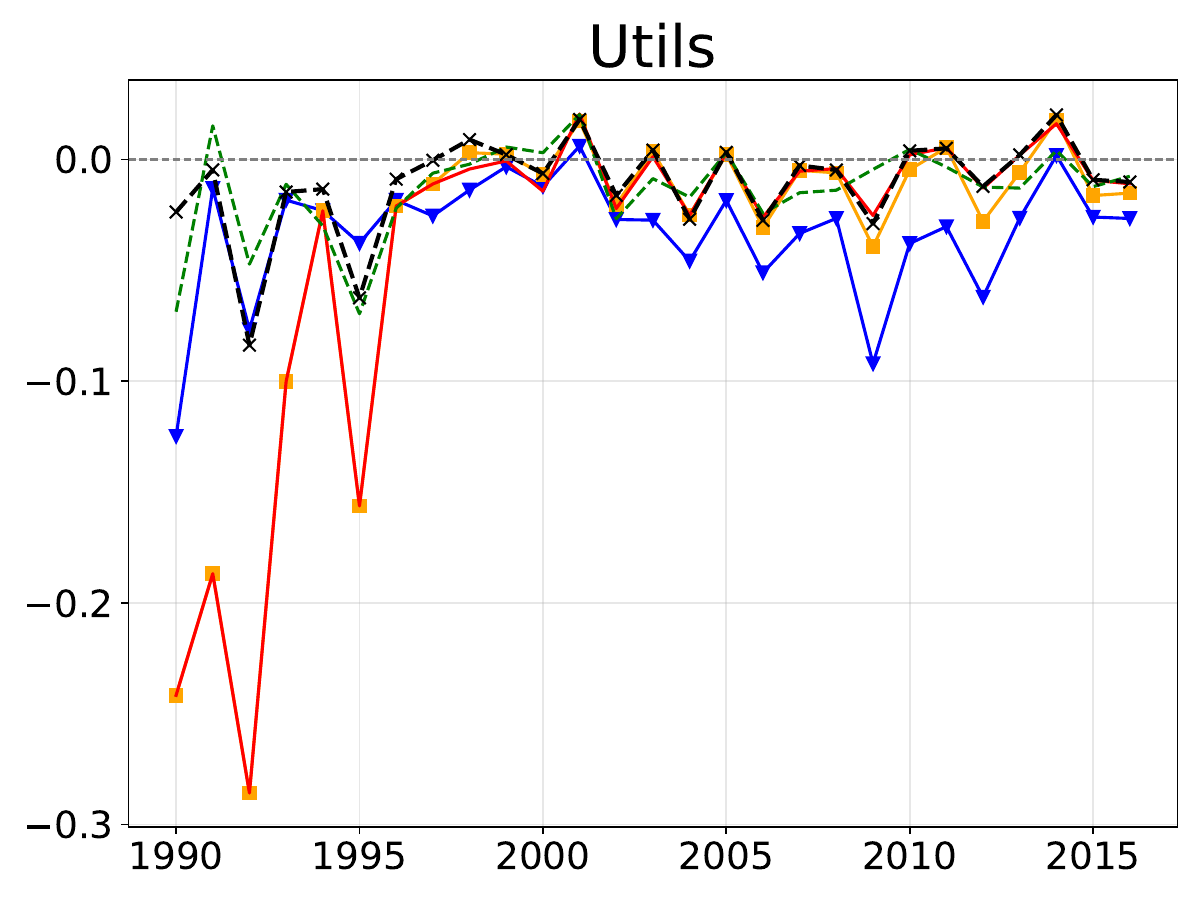}
	\end{subfigure}
    \begin{subfigure}{0.24\textwidth}
        \centering
        \includegraphics[width=\linewidth]{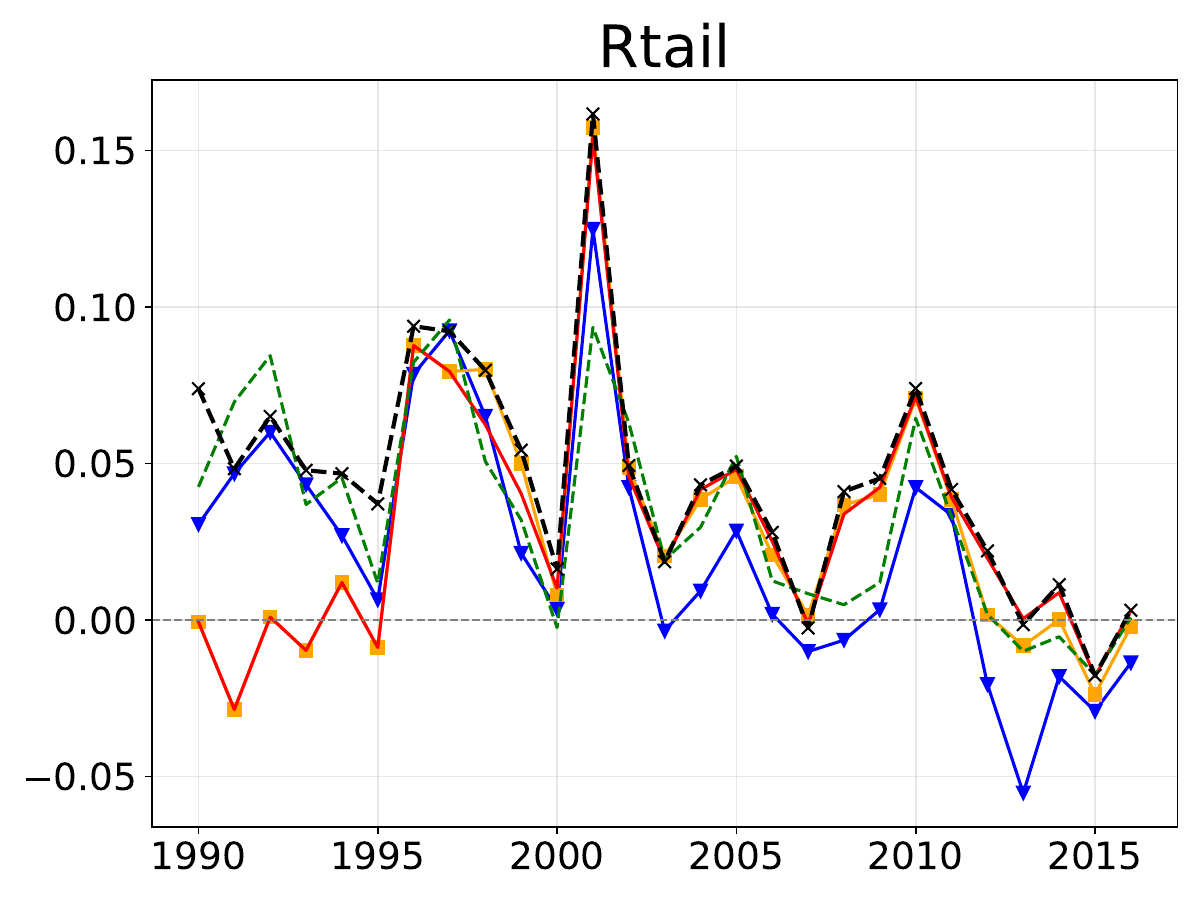}
	\end{subfigure}
    \begin{subfigure}{0.24\textwidth}
    	\centering
        \includegraphics[width=\linewidth]{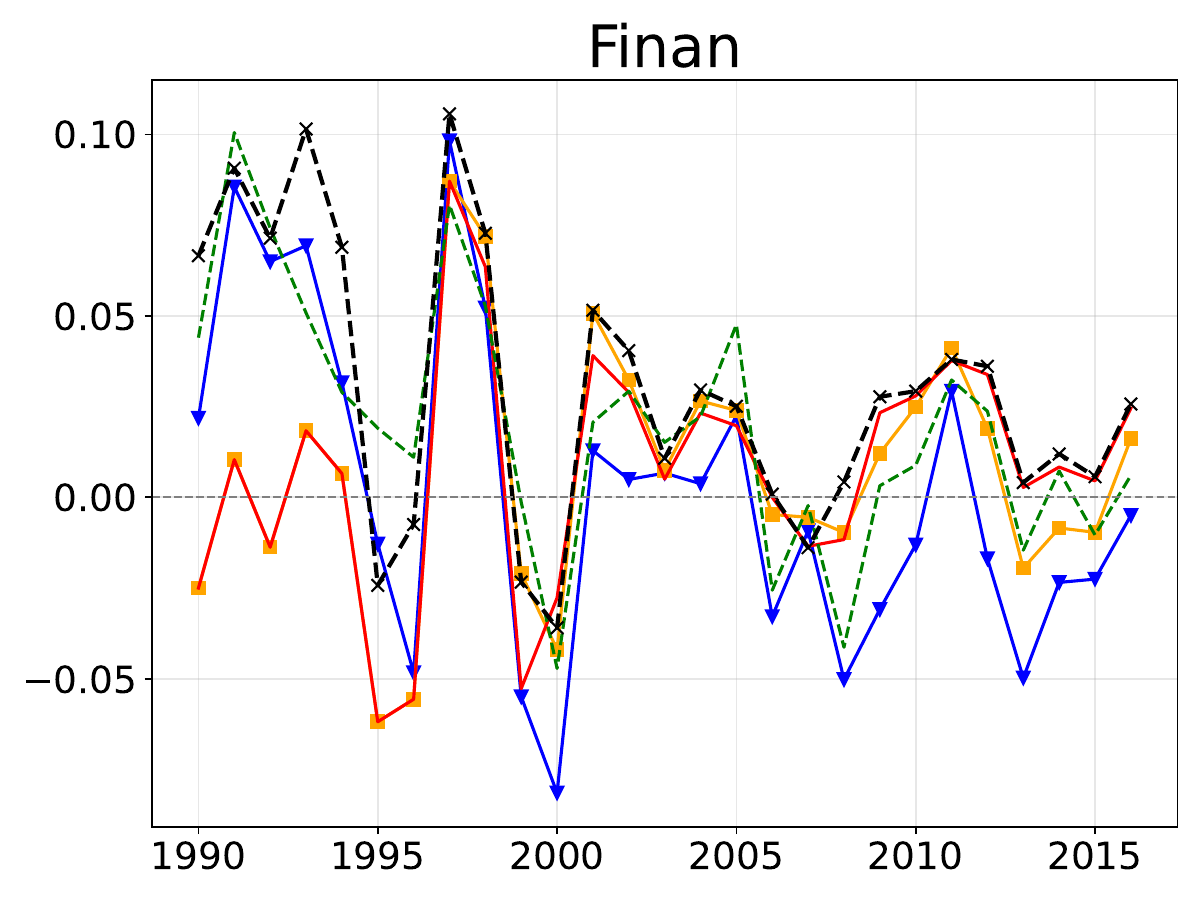}
	\end{subfigure}

    \begin{subfigure}{0.24\textwidth}
        \centering
        \includegraphics[width=\linewidth]{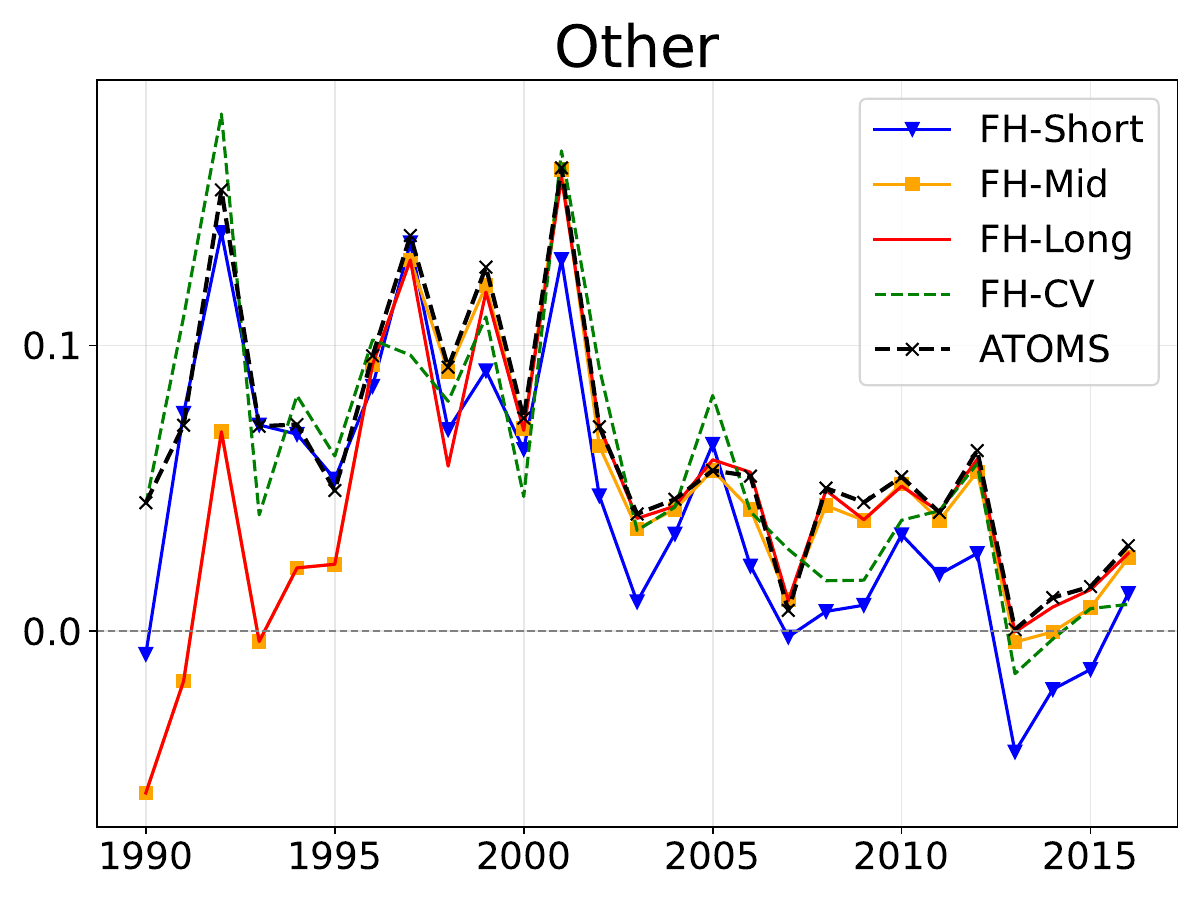}
	\end{subfigure}
      \bnotefig{This figure reports the annual OOS standard $R^2$ of our adaptive model selection algorithm $\adaptive$ (black dashed line with $\times$'s), as well as the fixed-window baselines $\fixedwindow(32)$ (blue $\blacktriangledown$'s), $\fixedwindow(128)$ (orange $\blacksquare$'s), and $\fixedwindow(512)$ (red), which use the last $32$, last $128$ and all months of validation data. The title in each subfigure is Kenneth French's acronym for each industry. For the full names of these industries, please refer to Table \myref{tab-industry-name-mapping}.}
\end{figure}

Finally, \myCref{fig-industry-yearly-standardR2} plots the annual out-of-sample $R^2$ for the $17$ industry portfolios, paralleling \myCref{fig-industry-yearly}.

\subsection{Additional Maximum Drawdown Results}\label{sec-additional-drawdown-results}

This section reports maximum drawdown results for alternative look-back reference windows of $S=1,2,6$ months.

\paragraph{Results: Mean-Variance Portfolio.} \Cref{tab-drawdown-mv-S1,tab-drawdown-mv-S2,tab-drawdown-mv-S6} report the maximum drawdown results for the mean-variance portfolio, with reference windows of $S=1,2,6$ months, respectively. The drawdowns are similar to those computed with a reference window of $S=3$ months, presented in \Cref{tab-drawdown-mv} of the main text.

\begin{table}[h]
\centering
\caption{Maximum Drawdown of Mean-Variance Portfolio with One-Month Reference Window}\label{tab-drawdown-mv-S1}
\begin{tabular}{@{}lcccccc@{}}
\toprule
\multirow{2}{*}{Method} & \multirow{2}{*}{Full OOS Period} & \multicolumn{3}{c}{Recessions}  \\
\cmidrule{3-5}
& & Gulf War & 2001 Recession & Financial Crisis  \\
\midrule
$\adaptive$ & $3.0\%$ & $0.7\%$ & $0.6\%$ & $2.4\%$ \\
$\fwshort$ & $4.3\%$ & $0.8\%$ & $0.9\%$ & $3.5\%$ \\
$\fwmed$ & $3.3\%$ & $0.9\%$ & $0.7\%$ & $2.4\%$ \\
$\fwlong$ & $3.1\%$ & $0.9\%$ & $0.6\%$ & $2.7\%$ \\
$\fixedwindowCV$ & $4.4\%$ & $0.8\%$ & $0.9\%$ & $4.2\%$ \\
$\RS$ & $2.7\%$ & $1.2\%$ & $0.7\%$ & $2.7\%$ \\
\bottomrule
\end{tabular}
\bnotetab{This table reports maximum drawdowns for mean-variance portfolios. Full OOS Period refers to the full out-of-sample period, 01/1990$\sim$11/2016. Recession columns report conditional drawdowns computed within each recession window using a one-month look-back reference window ($S=1$ months). The $\fh$ benchmarked methods use fixed windows as described in Table \ref{tab-fh-benchmarks}. Smaller values indicate smaller peak-to-trough losses.}
\end{table}

\begin{table}[h]
\centering
\caption{Maximum Drawdown of Mean-Variance Portfolio with Two-Month Reference Window}\label{tab-drawdown-mv-S2}
\begin{tabular}{@{}lcccccc@{}}
\toprule
\multirow{2}{*}{Method} & \multirow{2}{*}{Full OOS Period} & \multicolumn{3}{c}{Recessions}  \\
\cmidrule{3-5}
& & Gulf War & 2001 Recession & Financial Crisis  \\
\midrule
$\adaptive$ & $3.0\%$ & $0.7\%$ & $0.6\%$ & $2.4\%$ \\
$\fwshort$ & $4.3\%$ & $0.9\%$ & $0.9\%$ & $3.5\%$ \\
$\fwmed$ & $3.3\%$ & $0.9\%$ & $0.7\%$ & $2.4\%$ \\
$\fwlong$ & $3.1\%$ & $0.9\%$ & $0.6\%$ & $2.7\%$ \\
$\fixedwindowCV$ & $4.4\%$ & $0.8\%$ & $0.9\%$ & $4.2\%$ \\
$\RS$ & $2.7\%$ & $1.2\%$ & $0.8\%$ & $2.7\%$ \\
\bottomrule
\end{tabular}
\bnotetab{This table reports maximum drawdowns for mean-variance portfolios. Full OOS Period refers to the full out-of-sample period, 01/1990$\sim$11/2016. Recession columns report conditional drawdowns computed within each recession window using a two-month look-back reference window ($S=2$ months). The $\fh$ benchmarked methods use fixed windows as described in Table \ref{tab-fh-benchmarks}. Smaller values indicate smaller peak-to-trough losses.}
\end{table}

\begin{table}[h!]
\centering
\caption{Maximum Drawdown of Mean-Variance Portfolio with Six-Month Reference Window}\label{tab-drawdown-mv-S6}
\begin{tabular}{@{}lcccccc@{}}
\toprule
\multirow{2}{*}{Method} & \multirow{2}{*}{Full OOS Period} & \multicolumn{3}{c}{Recessions}  \\
\cmidrule{3-5}
& & Gulf War & 2001 Recession & Financial Crisis  \\
\midrule
$\adaptive$ & $3.0\%$ & $0.7\%$ & $0.6\%$ & $2.4\%$ \\
$\fwshort$ & $4.3\%$ & $1.0\%$ & $1.0\%$ & $3.5\%$ \\
$\fwmed$ & $3.3\%$ & $1.0\%$ & $0.7\%$ & $2.4\%$ \\
$\fwlong$ & $3.1\%$ & $1.0\%$ & $0.6\%$ & $2.7\%$ \\
$\fixedwindowCV$ & $4.4\%$ & $0.9\%$ & $0.9\%$ & $4.2\%$ \\
$\RS$ & $2.7\%$ & $1.2\%$ & $0.8\%$ & $2.7\%$ \\
\bottomrule
\end{tabular}
\bnotetab{This table reports maximum drawdowns for mean-variance portfolios. Full OOS Period refers to the full out-of-sample period, 01/1990$\sim$11/2016. Recession columns report conditional drawdowns computed within each recession window using a six-month look-back reference window ($S=6$ months). The $\fh$ benchmarked methods use fixed windows as described in Table \ref{tab-fh-benchmarks}. Smaller values indicate smaller peak-to-trough losses.}
\end{table}

\paragraph{Results: Signed Portfolio.} \Cref{tab-drawdown-ew-S1,tab-drawdown-ew-S2,tab-drawdown-ew-S6} report the maximum drawdown results for the signed portfolio, with reference windows of $S=1,2,6$ months, respectively. The drawdowns are similar to those computed with a reference window of $S=3$ months, presented in \Cref{tab-drawdown-ew} of the main text.

\begin{table}[h]
\centering
\caption{Maximum Drawdown of Signed Portfolio with One-Month Reference Window}\label{tab-drawdown-ew-S1}
\begin{tabular}{@{}lcccccc@{}}
\toprule
\multirow{2}{*}{Method} & \multirow{2}{*}{Full OOS Period} & \multicolumn{3}{c}{Recessions}  \\
\cmidrule{3-5}
& & Gulf War & 2001 Recession & Financial Crisis  \\
\midrule
$\adaptive$ & $16.3\%$ & $1.9\%$ & $2.4\%$ & $16.3\%$ \\
$\fwshort$ & $13.6\%$ & $2.1\%$ & $2.1\%$ & $12.8\%$ \\
$\fwmed$ & $15.2\%$ & $2.7\%$ & $2.4\%$ & $15.1\%$ \\
$\fwlong$ & $16.5\%$ & $2.7\%$ & $2.3\%$ & $16.5\%$ \\
$\fixedwindowCV$ & $18.4\%$ & $2.4\%$ & $1.9\%$ & $18.0\%$ \\
$\RS$ & $13.0\%$ & $1.2\%$ & $1.6\%$ & $13.0\%$ \\
EW & $48.8\%$ & $17.8\%$ & $14.2\%$ & $46.9\%$ \\
\bottomrule
\end{tabular}
\bnotetab{This table reports maximum drawdowns for signed portfolios. Full OOS Period refers to the full out-of-sample period, 01/1990$\sim$11/2016. Recession columns report conditional drawdowns computed within each recession window using a one-month look-back reference window ($S=1$ months). The $\fh$ benchmarked methods use fixed windows as described in Table \ref{tab-fh-benchmarks}. Smaller values indicate smaller peak-to-trough losses.}
\end{table}

\begin{table}[h]
\centering
\caption{Maximum Drawdown of Signed Portfolio with Two-Month Reference Window}\label{tab-drawdown-ew-S2}
\begin{tabular}{@{}lcccccc@{}}
\toprule
\multirow{2}{*}{Method} & \multirow{2}{*}{Full OOS Period} & \multicolumn{3}{c}{Recessions}  \\
\cmidrule{3-5}
& & Gulf War & 2001 Recession & Financial Crisis  \\
\midrule
$\adaptive$ & $16.3\%$ & $1.9\%$ & $2.4\%$ & $16.3\%$ \\
$\fwshort$ & $13.6\%$ & $2.1\%$ & $2.1\%$ & $12.8\%$ \\
$\fwmed$ & $15.2\%$ & $2.7\%$ & $2.4\%$ & $15.1\%$ \\
$\fwlong$ & $16.5\%$ & $2.7\%$ & $2.3\%$ & $16.5\%$ \\
$\fixedwindowCV$ & $18.4\%$ & $2.4\%$ & $1.9\%$ & $18.0\%$ \\
$\RS$ & $13.0\%$ & $1.2\%$ & $1.6\%$ & $13.0\%$ \\
EW & $48.8\%$ & $17.8\%$ & $14.2\%$ & $46.9\%$ \\
\bottomrule
\end{tabular}
\bnotetab{This table reports maximum drawdowns for signed portfolios. Full OOS Period refers to the full out-of-sample period, 01/1990$\sim$11/2016. Recession columns report conditional drawdowns computed within each recession window using a two-month look-back reference window ($S=2$ months). The $\fh$ benchmarked methods use fixed windows as described in Table \ref{tab-fh-benchmarks}. Smaller values indicate smaller peak-to-trough losses.}
\end{table}

\begin{table}[h]
\centering
\caption{Maximum Drawdown of Signed Portfolio with Six-Month Reference Window}\label{tab-drawdown-ew-S6}
\begin{tabular}{@{}lcccccc@{}}
\toprule
\multirow{2}{*}{Method} & \multirow{2}{*}{Full OOS Period} & \multicolumn{3}{c}{Recessions}  \\
\cmidrule{3-5}
& & Gulf War & 2001 Recession & Financial Crisis  \\
\midrule
$\adaptive$ & $16.3\%$ & $1.9\%$ & $2.4\%$ & $16.3\%$ \\
$\fwshort$ & $13.6\%$ & $2.1\%$ & $2.1\%$ & $12.8\%$ \\
$\fwmed$ & $15.2\%$ & $2.7\%$ & $2.4\%$ & $15.1\%$ \\
$\fwlong$ & $16.5\%$ & $2.7\%$ & $2.3\%$ & $16.5\%$ \\
$\fixedwindowCV$ & $18.4\%$ & $2.4\%$ & $1.9\%$ & $18.0\%$ \\
$\RS$ & $13.0\%$ & $1.2\%$ & $1.6\%$ & $13.0\%$ \\
EW & $48.8\%$ & $17.8\%$ & $14.2\%$ & $48.8\%$ \\
\bottomrule
\end{tabular}
\bnotetab{This table reports maximum drawdowns for signed portfolios. Full OOS Period refers to the full out-of-sample period, 01/1990$\sim$11/2016. Recession columns report conditional drawdowns computed within each recession window using a six-month look-back reference window ($S=6$ months). The $\fh$ benchmarked methods use fixed windows as described in Table \ref{tab-fh-benchmarks}. Smaller values indicate smaller peak-to-trough losses.}
\end{table}

\end{document}